\documentclass[10pt,twocolumn,letterpaper]{article}

\usepackage[pagenumbers]{iccv} 

\definecolor{forestgreen}{rgb}{0.13, 0.55, 0.13}

\renewcommand{\vec}{\mathbf}
\newcommand{\xinone}{\vec{x}_i}
\newcommand{\xione}{\vec{x}_i(1)}

\newcommand{\xitminusdelta}{\vec{x}_i(t - \Delta t)}
\newcommand{\xiz}{\vec{x}_i(0)}
\newcommand{\xio}{\vec{x}_i(1)}

\newcommand{\xit}{\vec{x}_i(t)}

\newcommand{\xlj}{\vec{x}^j}
\newcommand{\xljclip}{\vec{c}^j}
\newcommand{\xljbbox}{\vec{b}^j}
\newcommand{\xljopacity}{\alpha^j}

\newcommand{\xlij}{\vec{x}_i^j}
\newcommand{\xlijo}{\vec{x}_i^j(1)}
\newcommand{\xlijz}{\vec{x}_i^j(0)}
\newcommand{\xlijt}{\vec{x}_i^j(t)}

\newcommand{\xlijoclip}{\vec{c}_i^j(1)}
\newcommand{\xlijclip}{{\vec{c}_i^j}}

\newcommand{\xlijbbox}{{\vec{b}_i^j}}

\newcommand{\xlijoopacity}{\alpha_i^j(1)}
\newcommand{\xlijopacity}{\alpha_i^j}

\newcommand{\yinone}{\vec{y}_i} 
\newcommand{\ylij}{\vec{y}_i^j}
\newcommand{\yit}{\vec{y}_i(t)}

\newcommand{\xlijtbbox}{\vec{b}_i^j(t)}
\newcommand{\xlijtopacity}{\alpha_i^j(t)}

\newcommand{\directionset}{\vec{d}_i}

\newcommand{\driftvector}{\vec{d}_i}

\newcommand{\globalPrompti}{P_i}

\newcommand{\indexedLabel}{\ell}
\newcommand{\indexedLabelPrime}{{\indexedLabel}'}

\newcommand{\pij}{p_i^{\ell}}
\newcommand{\qij}{q_i^{\ell}}
\newcommand{\boundingBoxMetrics}{({\vec{b}_i^{\indexedLabel}})_m} %
\newcommand{\boundingBoxMetricsPrime}{({\vec{b}_i^{\indexedLabelPrime}})_{m^*}}

\newcommand{\KDESingle}{k_i^{\indexedLabel}}
\newcommand{\KDEDouble}{k_i^{{\indexedLabel},{\indexedLabelPrime}}}

\newcommand{\objectNumeracyScore}{O_{\text{Num}}}
\newcommand{\firstOrderPositionalLikelihood}{l_{\text{Pos}}^{(1)}}
\newcommand{\secondOrderPositionalLikelihood}{l_{\text{Pos}}^{(2)}}
\newcommand{\positionalVarianceScore}{\sigma^2_{\text{Pos}}}
\newcommand{\mIoU}{\text{mIoU}}

\newcommand{\method}{SLayR}

\newcommand{\layoutTransformer}{LayoutTransformer}
\newcommand{\layoutDiffusion}{LayoutDM}
\newcommand{\layoutFlow}{LayoutFlow}
\newcommand{\instanceDiffusion}{InstanceDiffusion}
\newcommand{\lmdPlus}{LMD+}
\newcommand{\gligen}{GLIGEN}
\newcommand{\boxdiff}{BoxDiff}
\newcommand{\gptFourO}{GPT4o}
\newcommand{\LLMGroundedDiffusion}{LLM-Grounded Diffusion}

\newcommand{\mainResultsGraphicsWidth}{0.105\textwidth}

\newcommand{\editingResultsGraphicsWidth}{0.3\columnwidth}

\newcommand{\tabitem}{~~\llap{\textbullet}~~}

\newcommand{\introFigureGraphicsWidth}{0.115\textwidth}
\newcommand{\mscocoGraphicsWidth}{0.115\textwidth}

\usepackage{multirow}
\usepackage[mode=buildnew]{standalone}
\usepackage{etoolbox, siunitx, booktabs}
\usepackage[normalem]{ulem}
\usepackage{tikz}
\usetikzlibrary{positioning,shapes.geometric, backgrounds, fit}
\usepackage{pgfplots}
\pgfplotsset{compat=1.18}

\usepackage{tabularx} 
\usepackage{array}    
\usepackage{algorithm}
\usepackage{algorithmic}
\usepackage{longtable}
\usepackage{xcolor}

\definecolor{iccvblue}{rgb}{0.21,0.49,0.74}
\usepackage[pagebackref,breaklinks,colorlinks,allcolors=iccvblue]{hyperref}

\def\paperID{5324} 
\def\confName{ICCV}
\def\confYear{2025}

\title{SLayR: Scene Layout Generation with Rectified Flow}

\author{Cameron Braunstein\textsuperscript{1,3,4} 
\and\hspace{-0.5cm}
Hevra Petekkaya\textsuperscript{1,4}\\
\and\hspace{-0.5cm}
Jan Eric Lenssen\textsuperscript{2}\\
\and\hspace{-0.5cm}
Mariya Toneva\textsuperscript{3}\\
\and\hspace{-0.5cm}
Eddy Ilg\textsuperscript{4}\\
\and \vspace{-1.cm}\\
\vspace{-0.1cm}
\normalsize
\textsuperscript{1}Saarland University, Saarland Informatics Campus
\\
\normalsize
\textsuperscript{2}Max Planck Institute for Informatics, Saarland Informatics Campus
\vspace{-0.1cm}\\
\normalsize
\textsuperscript{3}Max Planck Institute for Software Systems, Saarland Informatics Campus
\vspace{-0.1cm}\\
\normalsize
\textsuperscript{4}Computer Vision and Machine Perception Lab, University of Technology Nuremberg\vspace{-0.1cm} \\\vspace{-0.1cm}
{\tt\small braunstein@cs.uni-saarland.de, eddy.ilg@utn.de}
}

\begin{document}
\twocolumn[{%
\renewcommand\twocolumn[1][]{#1}%
\maketitle

\begin{center}
\centering
\captionsetup{type=figure}
\vspace{-0.8cm}
\includegraphics[scale=0.625]{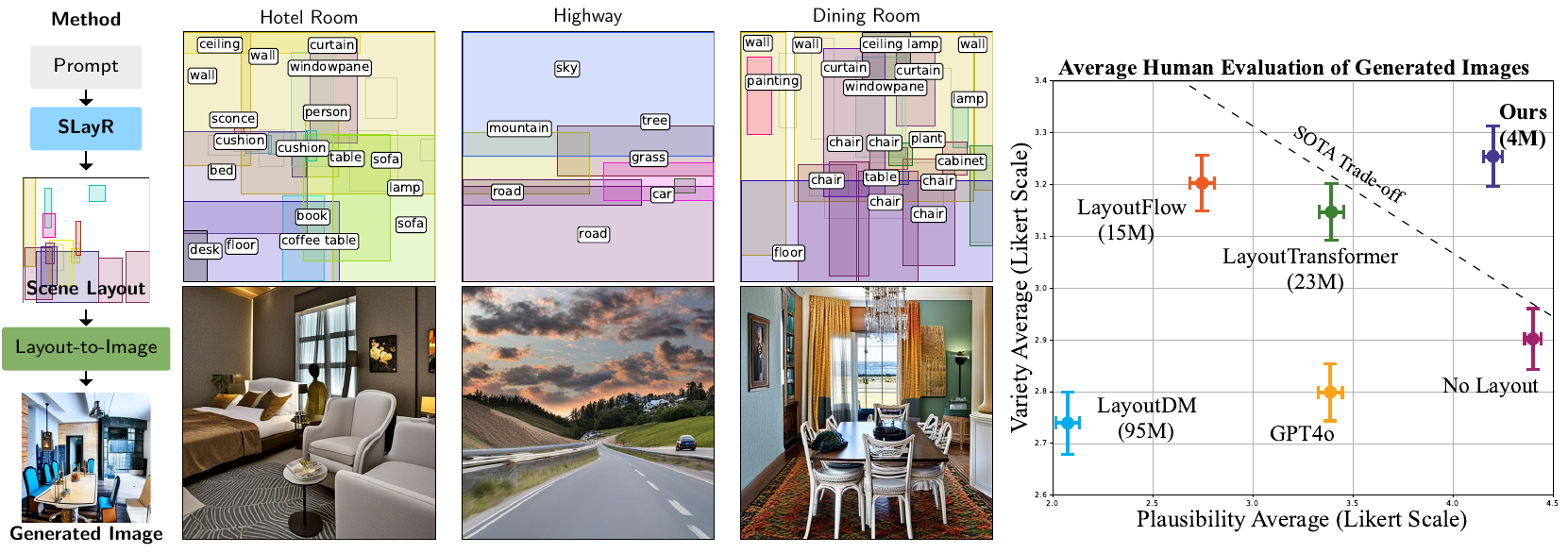}
\vspace{-0.3cm}
\captionof{figure}{
\textbf{Left:} We introduce \textbf{\method}, a method for scene layout generation via rectified flow.
\textbf{Middle:} \method~generates interpretable and editable scene layouts, which can be rendered using a layout-to-image generator. \textbf{Right:} Our method sets a new state of the art in generating more varied and plausible scenes than other methods and LLMs. 
}
\label{fig:teaser}
\end{center}

}]

\begin{abstract}

We introduce \method, \textbf{S}cene \textbf{Lay}out Generation with \textbf{R}ectified flow, a novel transformer-based model for text-to-layout generation which can then be paired with existing layout-to-image models to produce images. {\method} addresses a domain in which current text-to-image pipelines struggle: 
 generating scene layouts that are of significant variety and plausibility, when the given prompt is ambiguous and does not provide constraints on the scene. {\method} surpasses existing baselines including LLMs in unconstrained generation, and can generate layouts from an open caption set. To accurately evaluate the layout generation, we introduce a new benchmark suite, including numerical metrics and a carefully designed repeatable human-evaluation procedure that assesses the plausibility and variety of generated images. We show that our method sets a new state of the art for achieving both at the same time, while being at least 3$\times$ times smaller in the number of parameters. 
\end{abstract}


\vspace{-0.2cm}
\section{Introduction}

\begin{figure}[t!]
    \centering
    \setlength{\tabcolsep}{1pt}
    \begin{tabular}{ cc|cc}

        {\small Ours} & {\small LayoutGPT} & {\small Ours} & {\small LayoutGPT} \\
        \includegraphics[width=\introFigureGraphicsWidth]{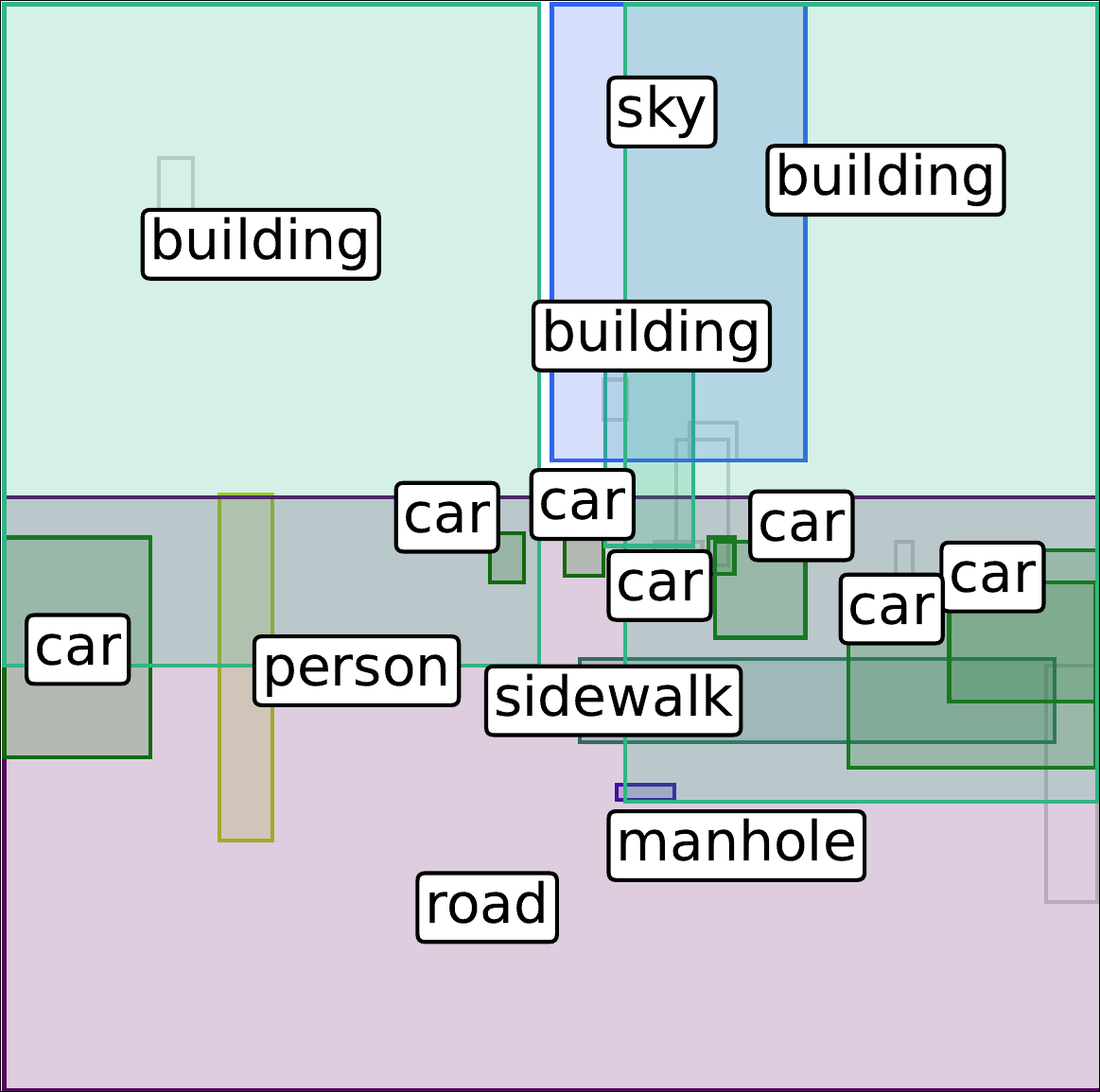} &
        \includegraphics[width=\introFigureGraphicsWidth]{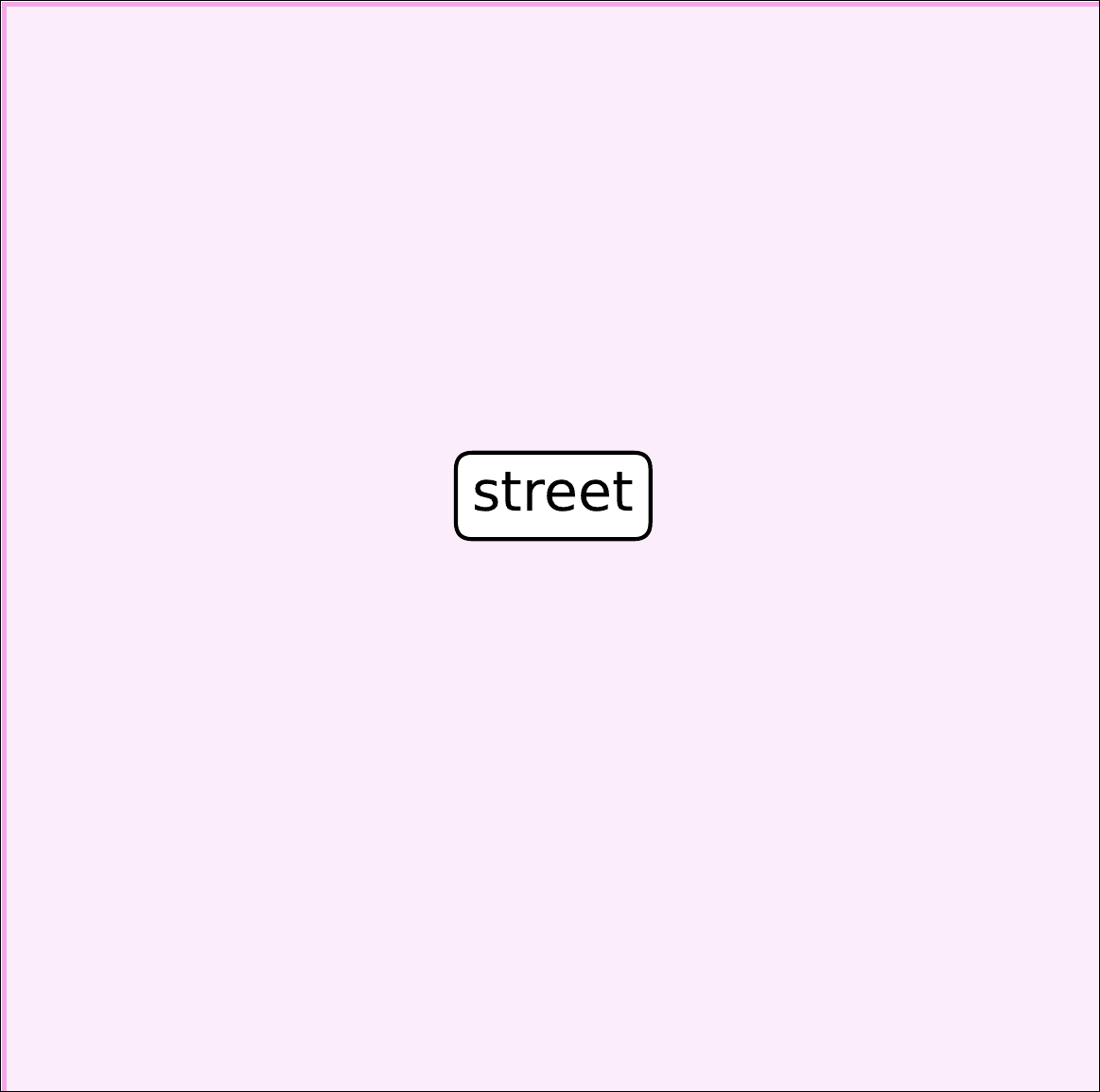} &
        \includegraphics[width=\introFigureGraphicsWidth]{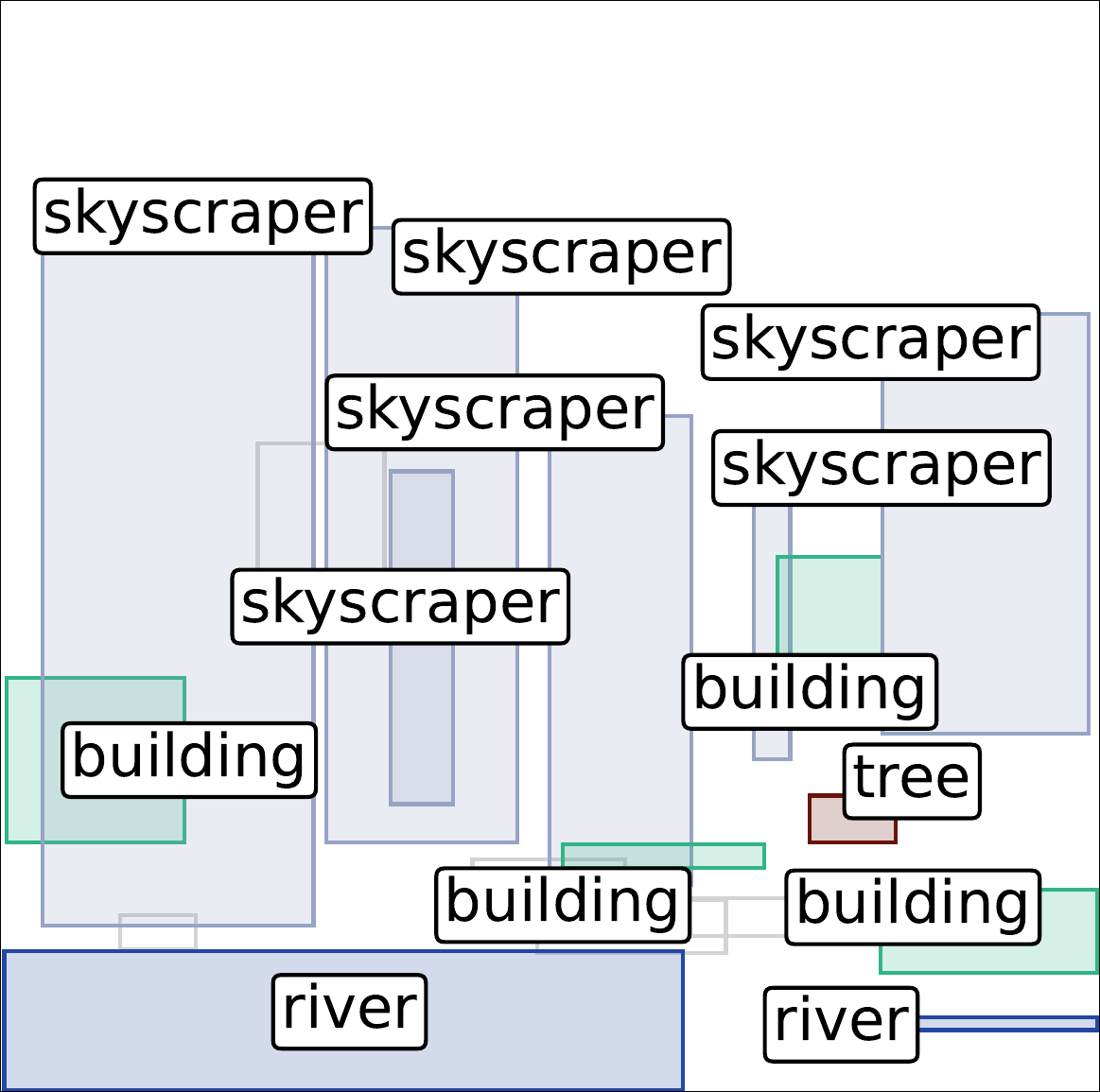} &
        \includegraphics[width=\introFigureGraphicsWidth]{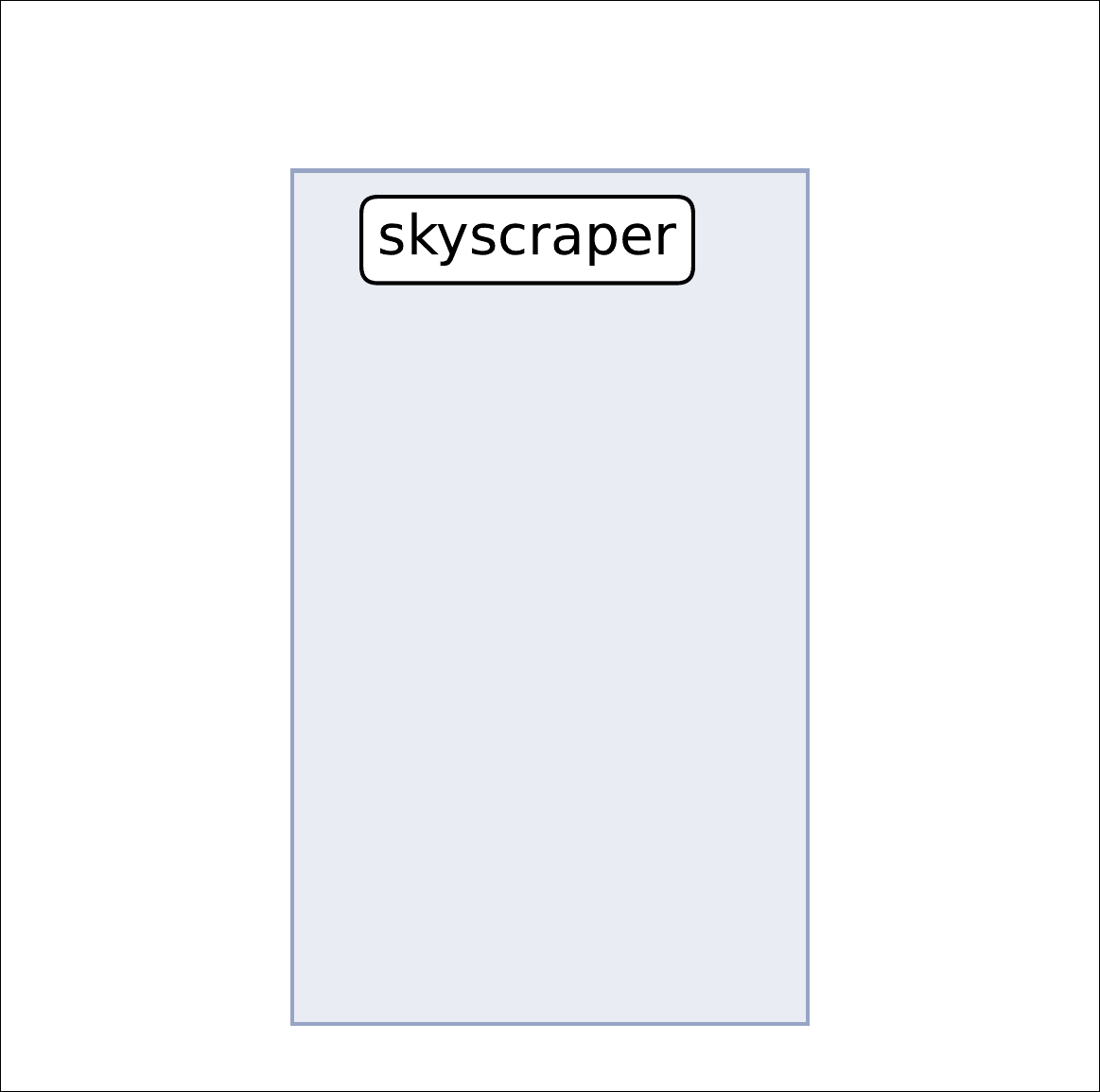} \\

    \end{tabular}
    \caption{Degenerated layouts from unconstrained prompts of a \emph{skyscraper}, and a \emph{street} from LayoutGPT \cite{feng2023layoutgptcompositionalvisualplanning} vs. our layouts. LayoutGPT shows to produce degenerate scenes 2/3rds of the times while our produces detailed plausible scene layouts. 
    }
    \label{fig:layout_gpt_comparison}
    \vspace{-0.3cm}
\end{figure}

Recent advances in text-to-image modeling have focused on training denoising diffusion models ~\cite{sohldickstein2015deepunsupervisedlearningusing, ho2020denoisingdiffusionprobabilisticmodels,song2022denoisingdiffusionimplicitmodels} to generate images from a
prompt encoding and image noise~\cite{ramesh2021zeroshottexttoimagegeneration, rombach2022highresolutionimagesynthesislatent, saharia2022photorealistictexttoimagediffusionmodels,esser2024scalingrectifiedflowtransformers,zhang2023texttoimagediffusionmodelsgenerative,sauer2024fasthighresolutionimagesynthesis}, as well as incorporating finer-grained control modalities \cite{hudson2023sodabottleneckdiffusionmodels, kwon2023diffusionmodelssemanticlatent,park2023understandinglatentspacediffusion,zhang2023addingconditionalcontroltexttoimage, luo2024readoutguidancelearningcontrol,shen2024sgadapterenhancingtexttoimagegeneration,mishra2024imagesynthesisgraphconditioning,Wu2023}. Building upon these advancements, prior works have demonstrated the editability and interpretability advantages of a multistage text-to-layout-to-image model, where the user can view and manipulate an intermediate layout consisting of bounding boxes for object-level scene elements~\cite{lian2024llmgroundeddiffusionenhancingprompt, feng2023layoutgptcompositionalvisualplanning,zhou2024gala3dtextto3dcomplexscene, gao2024graphdreamercompositional3dscene,yuan2024dreamscape3dscenecreation, ocal2024scenetellerlanguageto3dscenegeneration, aguinakang2024openuniverseindoorscenegeneration}. These works use LLMs as text-to-layout generators, and focus on parsing multi-object prompts (e.g. \emph{two dogs next to a cat}). 
However, a closer inspection reveals that these models do not generate high variety (see Figure~\ref{fig:teaser}, right) or collapse entirely (see Figure~\ref{fig:layout_gpt_comparison}), when presented with prompts that have few constraints and leave a high degree of ambiguity.

This motivates us to propose a novel light-weight text-to-layout generation model for expanding unconstrained prompts (e.g. \emph{a park}, \emph{a beach}) into a variety of  plausible scene layouts (see Figure~\ref{fig:teaser}, left and middle). 
The model cannot rely on the prompt to deliver guiding information, and must instead have extensive knowledge about the general distribution of scene structures. Given a CLIP~\cite{radford2021learningtransferablevisualmodels} embedding of a global scene prompt, we generate the layout using rectified flow~\cite{liu2022flowstraightfastlearning}, with a Diffusion Transformer (DiT)~\cite{peebles2023scalablediffusionmodelstransformers}. We test our model's performance across a set of layout-to-image generation models ~\cite{wang2024instancediffusioninstancelevelcontrolimage,li2023gligenopensetgroundedtexttoimage,xie2023boxdifftexttoimagesynthesistrainingfree,lian2024llmgroundeddiffusionenhancingprompt}.
Since the domain of unconstrained text-to-layout generation is largely unexamined for non-LLM approaches, we also evaluate a comprehensive set of baseline architectures adapted from UI/document generation. 

We pair {\method} with several available layout-to-image model generation models and achieve optimal scores in CMMD~\cite{jayasumana2024rethinkingfidbetterevaluation}, FID~\cite{heusel2018gans}, KID~\cite{binkowski2021demystifyingmmdgans}, VQA ~\cite{lin2024evaluatingtexttovisualgenerationimagetotext}, HPSv2 ~\cite{wu2023humanpreferencescorev2}, and ImageReward ~\cite{xu2023imagerewardlearningevaluatinghuman}. As these metrics evaluate only the quality of the images and not the layouts, we furthermore introduce a set of novel metrics to statistically assess the \emph{plausibility} and \emph{variety} of scene layouts.
To validate our findings, we introduce a comprehensive, repeatable human-evaluation study. We show that, across all examined text-to-layout-to-image pipelines, our model yields the best trade-off in generating images that are both diverse and plausible, while being significantly more light-weight than its competitors. Our approach can be conditioned on partial layouts and directional constraints, and enjoys the native editing capabilities of the two-stage pipeline. In summary, our contributions are:
\begin{itemize}
    \item We propose the first general purpose model for text-to-layout generation using rectified flow, which can generate complex scene layouts with varying number and types of objects from a generic text prompt,  supports partial layout conditioning, directional constraints, and can scale to open scene captions.
    \item We establish a comprehensive suite of numerical metrics for generated layout evaluation and    
    a well-designed human-evaluation study that can be used to assess the layout quality and be repeated by others.
    \item We demonstrate that our method can successfully integrate into a text-to-layout-to-image pipeline, yielding the state-of-the-art in achieving both variety and plausibility.
\end{itemize}
Upon publication, the source code and models will be made available to the community.

\section{Related Work}

\noindent\textbf{LLMs in Scene Layout Generation.}
In 2D layout generation, {\LLMGroundedDiffusion} \cite{lian2024llmgroundeddiffusionenhancingprompt} and LayoutGPT \cite{feng2023layoutgptcompositionalvisualplanning} generate scene layouts from multi-object prompts, with a bounding box for each object in the prompt, using prompt templates that facilitate in-context learning. We found that directly plugging in single-word prompts into these templates yields degenerate solutions (layouts with a single bounding box labeled with the prompt) in approximately 2/3 of queries (see Figure~\ref{fig:layout_gpt_comparison}). Therefore, to reasonably measure an LLM's aptitude for our task and to encourage chain-of-thought reasoning \cite{wei2023chainofthoughtpromptingelicitsreasoning} we adapt {\LLMGroundedDiffusion}'s 
template with in-context examples to our domain as a baseline. We note here also that GPT4o is a vastly larger model that was trained on much larger corpora of data than other baselines. 

\noindent\textbf{Adapting UI Generation.} Our task of scene layout generation is distinct from User Interface (UI) generation: scene and object captions are from open sets, whereas UI layouts lack global captions and have labels from a small fixed set. Nevertheless, they can serve as interesting baselines, and we adapt several of these models using their conditional generation capabilities. We use {\layoutTransformer} ~\cite{gupta2021layouttransformerlayoutgenerationcompletion} as a representative for autoregressive transformer approaches, which completes a partial sequence of object bounding boxes to form an image layout. Works such as LayoutFormer++ ~\cite{jiang2023layoutformer} extend {\layoutTransformer} with added conditioning, but this is not the focus of our assessment of adapted UI generation, and these baselines are considered redundant. We also adapt {\layoutDiffusion} ~\cite{inoue2023layoutdmdiscretediffusionmodel} and {\layoutFlow} ~\cite{guerreiro2024layoutflowflowmatchinglayout} as representative baselines for diffusion-based methods for UI generation \cite{zhang2023layoutdiffusionimprovinggraphiclayout,chai2023layoutdmtransformerbaseddiffusionmodel,levi2023dltconditionedlayoutgeneration}. For GAN-based approaches \cite{li2019layoutgangeneratinggraphiclayouts},  while LayoutGAN++ \cite{Kikuchi_2021} supports inter-bounding-box relationships, the Lagrange multiplier constraint formulation cannot be adapted to support global conditioning.  In contrast to our method, UI generation models by design do not extend into the open world scenario.

\begin{figure*}[ht!]
    \centering
    \includegraphics[width=1\linewidth]{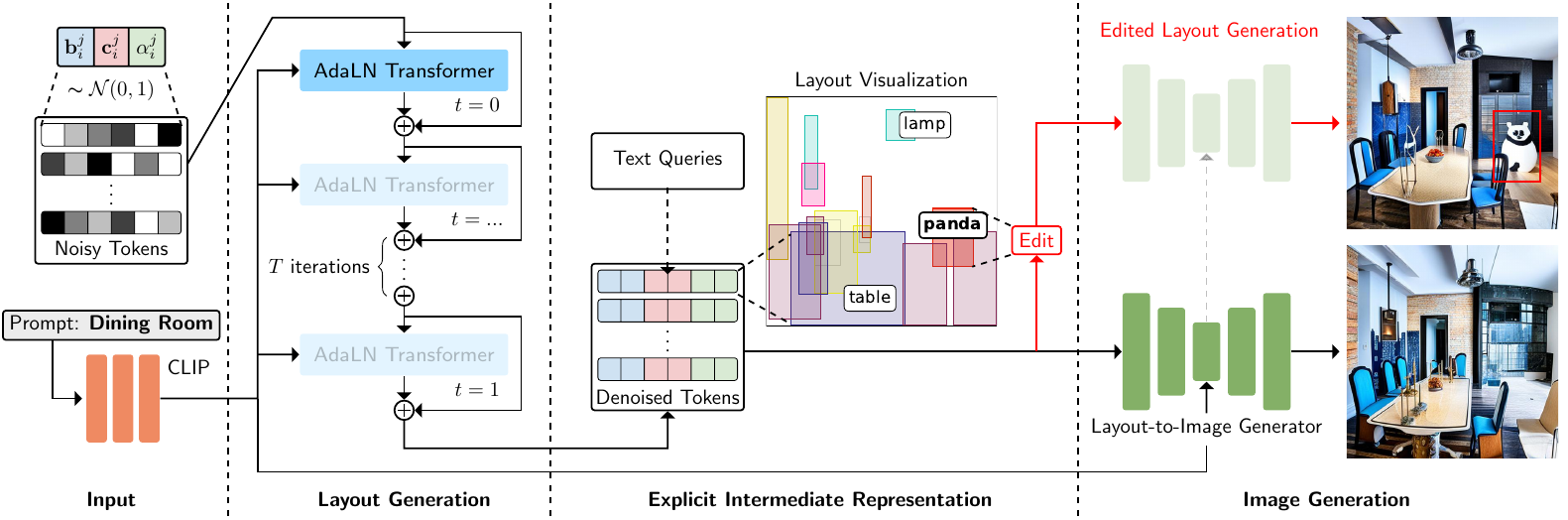}
    \caption{\textbf{Method Overview.}
    Our layout generation model takes a set of noisy tokens and a prompt encoded as a global CLIP embedding as input. The tokens are partitioned into bounding box information $\xlijbbox$, reduced clip embeddings $ \xlijclip$ and opacities $\xlijopacity$, that are subsequently denoised using a transformer. 
    For visualization purposes, the user can query the generated layout with labels and edit boxes by adding, moving or removing them. Finally, the generated layout is passed through an off-the-shelf layout-to-image generator. 
    }
\label{fig:main_method_diagram}
\end{figure*}

\noindent\textbf{Rectified Flow.} Diffusion modeling has inspired numerous variants and improvements, one of which is rectified flow~\cite{liu2022flowstraightfastlearning}. Prior works  on the related text-to-image generation task \cite{liu2024instaflowstephighqualitydiffusionbased,esser2024scalingrectifiedflowtransformers} have successfully leveraged it in their pipelines, with the added benefit of not requiring a noise schedule like DDIM \cite{song2022denoisingdiffusionimplicitmodels} or DDPM \cite{ho2020denoisingdiffusionprobabilisticmodels}-based approaches, motivating us to use it for our task.

\noindent\textbf{Layout-to-Image Generation}.  We demonstrate that {\method} integrates well into downstream conditional diffusion models to form a complete text-to-image pipeline, with the added benefits of an interpretable and controllable intermediate layout phase. 
To control for the effect which the image generator has on the final generated image, we evaluate our layouts across multiple layout-to-image models. Although there are a wide variety of such models, \cite{chen2023trainingfreelayoutcontrolcrossattention,yang2022recoregioncontrolledtexttoimagegeneration,zhao2019imagegenerationlayout,object_centric_generation,bartal2023multidiffusionfusingdiffusionpaths,xiong2024groundingboothgroundingtexttoimagecustomization} we select four which are publically available and have been used successfully with LLM-driven layouts \cite{lian2024llmgroundeddiffusionenhancingprompt,feng2023layoutgptcompositionalvisualplanning} or have shown SOTA performance:  {\instanceDiffusion} ~\cite{wang2024instancediffusioninstancelevelcontrolimage}, {\gligen}~\cite{li2023gligenopensetgroundedtexttoimage}, {\boxdiff} \cite{xie2023boxdifftexttoimagesynthesistrainingfree}, and {\lmdPlus} \cite{lian2024llmgroundeddiffusionenhancingprompt}.

\section{Method}

The central part of our work is the text-to-layout generation module, which we combine with the existing  layout-to-image generators ~\cite{wang2024instancediffusioninstancelevelcontrolimage,li2023gligenopensetgroundedtexttoimage,lian2024llmgroundeddiffusionenhancingprompt,xie2023boxdifftexttoimagesynthesistrainingfree} to form a complete text-to-image pipeline. An overview is provided in~\cref{fig:main_method_diagram}.

\subsection{Rectified Flow Preliminaries}

We briefly recap the basics of rectified flow introduced in ~\cite{liu2022flowstraightfastlearning}. Let $I$ be a set of training sample indices and $\{\xinone \}_{i \in I}$ the ground-truth samples whose distribution we would like to learn using our model $v$. We linearly interpolate between Gaussian noise $\xiz$ and samples 
 $\xio \equiv \xinone$ across timesteps $t \in [0,1]$ as follows:
\begin{equation}\label{eq:x_t_definition}
    \xit = (1-t) \cdot \xiz + t \cdot \xio \, .
\end{equation}
The model $v$ is trained to take $(\xit, t)$ as input and to predict the derivative of the path between $\xiz$ and $\xio$, which according to Equation~\ref{eq:x_t_definition} is $\xio - \xiz$. The model is trained with the following objective: 
\begin{equation}\label{eq:rectified_flow_training_objective}
    \text{min}_v \int_0^1\mathbb{E}_i[ || (\xio - \xiz) - v(\xit,t)||^2]dt
\end{equation}
using stochastic gradient descent. This optimization is carried out across all available samples of the ground-truth distribution. Following~\cite{liu2022flowstraightfastlearning},  noisy values $\xiz$ are resampled at each epoch. The end result is a network $v$ which is effective at predicting the direction from a noisy sample at an intermediate timestep towards the target distribution. Since a single evaluation may be noisy, the inference is performed by integrating over $T$ timesteps:  
\begin{equation} \label{eq:rectified_flow_inference}
    \xio = \xiz +  \sum_{t=1}^T v(\vec{x}_i(\frac{t-1}{T}),\frac{t}{T})\cdot \frac{1}{T}  \, .
\end{equation}

\subsection{Layout Representation}
\label{subsec:layout_representation}

We start with defining a scene representation as the basis for our generative architecture. 
A training sample $(\xinone, \globalPrompti)$ is composed of a global image caption prompt $\globalPrompti$ and a set of $J$ object tokens $\xinone=\{\xlij \in \mathbb{R}^{d+5}\}_{j \in J}$.
The token representation of any single object is composed of
\begin{equation} \label{eq:layout_token_definition}
    \xlij = (\xlijbbox \mathbin\Vert \xlijclip \mathbin\Vert \xlijopacity),
\end{equation}
where \mbox{$\xlijbbox = (x,y,w,h) \in \mathbb{R}^4$} encodes the bounding box coordinates, $\xlijclip \in \mathbb{R}^d$ is a PCA-reduced CLIP~\cite{radford2021learningtransferablevisualmodels} embedding, and $\xlijopacity \in \mathbb{R}$ is an opacity value that defines the existence of a specific bounding box.

\subsection{Our Model Architecture}
\label{subsec:architecture}
Our rectified flow model is built from multihead AdaLN transformer blocks, which can process tokens $\{\xlij\}_{j \in J}$ to iteratively denoise them~\cite{peebles2023scalablediffusionmodelstransformers}.
The timestep $t$, bounding box coordinates $\xlijtbbox$, and opacity values $\xlijtopacity$ are sinusoidally encoded. The timestep $t$ and a linear projection of the global label $\globalPrompti$'s CLIP encoding are passed as conditioning of the adaptive layer normalization of the transformer blocks. The tokens represent the objects in the layout and are processed all at once. 

Inference begins at $t=0$ with the set of tokens $\{\xlijt\}_{j \in J} \equiv \{\xlijo\}_{j \in J}$ initialized from Gaussian noise. Our model then iteratively processes and updates the tokens from $t=0$ to $t=1$ over $T$ iterations using \cref{eq:rectified_flow_inference} based on the global prompt conditioning $\globalPrompti$. As the output of the transformer blocks represents the rate of change of $\xlijt$, we use a linear layer to downproject the output back to the dimension of $\xlijt$ before sinusoidal encoding. 

Using~\cref{eq:rectified_flow_inference} to sum over all timesteps yields the final layout $\{\xlijo\}_{j \in J}$ with PCA-reduced CLIP embeddings, bounding boxes, and opacities. Tokens with $\xlijoopacity < 0.5 $ are considered unused and discarded. For image generation, we unproject each $\xlijoclip$ from the PCA space into the CLIP feature space and pass the embeddings directly into the downstream image generation module. We decode labels into text for visualization purposes only, where the decoding can be achieved through label queries from the user or from a label bank.

\subsection{Partial Conditioning and Constraints}
\label{sec:partcond}
We explain how we adapt the  RePaint~\cite{lugmayr2022repaintinpaintingusingdenoising, schroppel2024neuralpointclouddiffusion} technique for rectified flow to enable \emph{partial layout conditioning} in the supplemental material. As we show in \cref{subsection:additional_model_features}, our model can accommodate partial layouts where only some boxes or  labels are given. Additionally, we can impose inter-bounding box positional constraints by adding a directional drift on the bounding boxes during inference. We provide a brief sketch here and further details in the supplement. Let $\yinone$ be the partial layout, $M$ be a conditioning mask, and $\driftvector$ be a vector specifying all drift conditions, all with the same dimension as $\xit$. During inference, we initialize $\xiz$ from Gaussian noise, and at timestep $t$, we first apply the rectifed flow inference step:
\begin{equation}
    \vec{x}_i(\frac{t}{T}) \leftarrow \vec{x}_i(\frac{t-1}{T}) +  v(\vec{x}_i(\frac{t-1}{T}),\frac{t}{T},\globalPrompti)\cdot \frac{1}{T} \, , 
\end{equation}
then compute time-adjusted partial conditioning:
\begin{equation}
    \vec{y}_i(\frac{t}{T}) \leftarrow  \yinone \cdot \frac{t}{T} + \xiz \cdot \frac{1-t}{T} \, , 
\end{equation}
and finally apply conditioning:
\begin{equation}
    \vec{x}_i(\frac{t}{T}) \leftarrow \vec{x}_i(\frac{t}{T}) \odot (1 - M) + \vec{y}_i(\frac{t}{T}) \odot M + \driftvector \, .
\end{equation}
In addition to handling unconstrained or explicitly constrained prompts, following prior work~\cite{feng2023layoutgptcompositionalvisualplanning}, we integrate the ability to extract constraints from complex text prompts using LLMs by leveraging the LMM to extract the objects and positional constraints from them. 
We show that this integration allows us to perform well in spatial and numerical reasoning tasks in~\cref{subsec:scaling_to_mscoco}. 

\subsection{Training}\label{subsec:training}
To construct a training sample from the ground-truth image layout $i$, we create $\xlijclip$ and $\xlijbbox$ for each bounding box $j$, and initialize $\xlijopacity := 1$. To ensure a consistent amount of tokens, we pad the samples  by adding tokens with $\xlijopacity \xlijbbox := 0$, and $\xlijclip$ to the null string embedding. 

We now treat $\{\xlij\}_{j \in J} \equiv \{\xlijo\}_{j \in J}$, sample $\{\xlijz\}_{j \in J}$ from Gaussian noise, draw $t$ uniformly from $[0,1]$, and compute the set of tokens $\{\xlijt\}_{j \in J}$ by adapting the formula from ~\cref{eq:x_t_definition}, which are then passed to the model as input. We refer to the output of the model as $v(\{\xlijt\}_{j \in J},t,\globalPrompti)$ and compute the training loss derived from \cref{eq:rectified_flow_training_objective}:
\begin{equation}
    \mathcal{L} = \sum_{i \in I, j \in J} || \xlijo - \xlijz  -v(\{\xlijt\}_{j \in J},t,\globalPrompti)_j ||^2 \,.
\end{equation}

\section{Novel Metrics}

\subsection{Human Evaluation of Generated Images}\label{subsubsection:human_evaluation}

To compliment generated image metrics ~\cite{jayasumana2024rethinkingfidbetterevaluation,heusel2018gans,binkowski2021demystifyingmmdgans,lin2024evaluatingtexttovisualgenerationimagetotext,wu2023humanpreferencescorev2,xu2023imagerewardlearningevaluatinghuman} which we measure in ~\cref{subsec:generated_image_metrics_results}, we argue that the true plausibility and variety of images can only be assessed by humans and therefore introduce a human-evaluation study. 

To conduct the evaluation, we survey $60$ participants by presenting them images from text-to-layout-to-image models with different text-to-layout baselines. For each method, each participant rates $10$ collections for their plausibility, and different $10$ collections for their variety, on a Likert scale ($1$ to $5$, where $1$ corresponds to very implausible/very low variance, and $5$ corresponds to very plausible/very high variance). For plausibility, we instructed participants to consider the overall realism of the collection, as well as how effectively it depicts the global text prompt. For variety, we instructed users to consider the spatial arrangement of objects in an image and implied camera angle in addition to overall image appearance. We control for the choice of layout-to-image generator by running our evaluation once for each of our tested layout-to-image generators ({\instanceDiffusion} ~\cite{wang2024instancediffusioninstancelevelcontrolimage}, {\gligen}~\cite{li2023gligenopensetgroundedtexttoimage}, {\boxdiff} \cite{xie2023boxdifftexttoimagesynthesistrainingfree}, and {\lmdPlus} \cite{lian2024llmgroundeddiffusionenhancingprompt}).  We provide the full details of the study in the supplemental material. The study was approved by the Ethics Review Board of our institution, ensuring compliance with ethical standards.  

\subsection{Generated Layout Metrics}\label{subsubsec:numerical_metrics}
In the following, we introduce four metrics to assess the generated scenes layouts' plausibility and variety.

\paragraph{Object Numeracy.}
Our metric $\objectNumeracyScore$ assesses whether generated layouts contain the objects at the expected frequencies. We sample across a collection of global prompts ($\{P_i\}$). The probability distribution for expected occurrences of the object-label $\indexedLabel$ in layouts generated from the global prompt $P_i$ is written $\qij$, and the ground-truth probability distribution is $\pij$. Our metric is the normalized sum of KL-divergence between these two distributions:
\begin{align}
    \objectNumeracyScore :=\frac{\sum_{i,\indexedLabel}KL(\pij||\qij)}{|\{P_i\}|} \,
\end{align}
where lower scores indicate that the model produces layouts with more plausible object numeracy. 

\paragraph{Positional Likelihoods.}

We introduce $\firstOrderPositionalLikelihood$, and $\secondOrderPositionalLikelihood$ to measure how plausible the objects in a generated layout are arranged.
Let $m$ index all bounding boxes of object-label $\indexedLabel$ for prompt $i$. For each object-label $\indexedLabel$, we obtain a distribution $\KDESingle$ with KDE of the object's bounding box $\boundingBoxMetrics$ in all layouts with global prompt $i$. We compute the average likelihood over all objects and all global prompts, to measure the \emph{first-order positional likelihood}:
\begin{equation}
    \firstOrderPositionalLikelihood =     \frac{\sum_{i,\indexedLabel,m}  \KDESingle (\boundingBoxMetrics)}{|\{ \boundingBoxMetrics\}|} \, .
\end{equation}
A higher value for $\firstOrderPositionalLikelihood$ means that object bounding boxes are placed in reasonable locations in the layout.

We also want to measure the likelihood of spatial relationships between objects. Let $m^*$ index all bounding boxes of object-label $\indexedLabelPrime$. For each object-label pair $(\indexedLabel,\indexedLabelPrime)$, we obtain a distribution estimated with KDE $\KDEDouble$ for the difference in the bounding box dimensions. We compute the average likelihood over all objects and all global prompts from our distributions to measure the \emph{second order positional likelihood}: 
\begin{equation}
 \secondOrderPositionalLikelihood=          \frac{\sum_{i,\indexedLabel \neq \indexedLabelPrime,m,m^*} \KDEDouble (\boundingBoxMetrics-\boundingBoxMetricsPrime)}{|\{\boundingBoxMetrics\}|(|\{ \boundingBoxMetrics \}|+1)/2} \,.
\end{equation}
A higher value for $\secondOrderPositionalLikelihood$ means that pairs of objects are plausibly positioned relative to one another. We conduct a grid search across bandwidths with 5-fold cross validation to optimize the KDE bandwidths for both $\firstOrderPositionalLikelihood$ and $\secondOrderPositionalLikelihood$. 

\paragraph{Positional Variance.}
Our metric $\positionalVarianceScore$ measures the variety of bounding boxes. For every bounding box $\boundingBoxMetrics$, we find the bounding box in layouts with global prompt $i$ and object label $\indexedLabel$ that is closest in Euclidean distance to the bounding box. We now redefine $\{m^*\}$ as the set of indices of bounding boxes in other samples which minimize the term $|| \boundingBoxMetrics - \boundingBoxMetricsPrime||$.
We compute all of these Euclidean distances and take the average:
\begin{equation}
    \positionalVarianceScore = \frac{\sum_{i,\indexedLabel,m} \sum_{\{m^*\}}|| \boundingBoxMetrics - \boundingBoxMetricsPrime ||}{\sum_{i,\indexedLabel,m} |\{m^*\}|} \, 
\end{equation}  
If this metric is small, it means that the variance is low.

\paragraph{Adapting UI Generation Metrics.}
Typical UI generation metrics must be reconsidered with shift in domain.  Alignment \cite{lee2020neuraldesignnetworkgraphic} and Overlap \cite{li2019layoutgangeneratinggraphiclayouts} scores are not salient, as real world images often have misaligned or overlapping bounding boxes. The layout-FID \cite{heusel2018gans} metric requires a GAN discriminator to compute, which we do not have in this new domain. We compute a standard max IoU ($\mIoU$) \cite{Kikuchi_2021} averaged across sampled scene categories.

\section{Experiments}

\begin{figure*}[ht]
    \setlength{\tabcolsep}{1pt}

    \centering
        \begin{tabular}{p{0.3cm} ccc | ccc | ccc} 

        &
        \multicolumn{3}{c|}{``Living Room''} &
        \multicolumn{3}{c|}{``Roof Garden''} &
        \multicolumn{3}{c}{``Street''} \\

        \multirow{-5}{*}{\rotatebox[origin=c]{90}{No Layout}}&
        \includegraphics[width=\mainResultsGraphicsWidth]{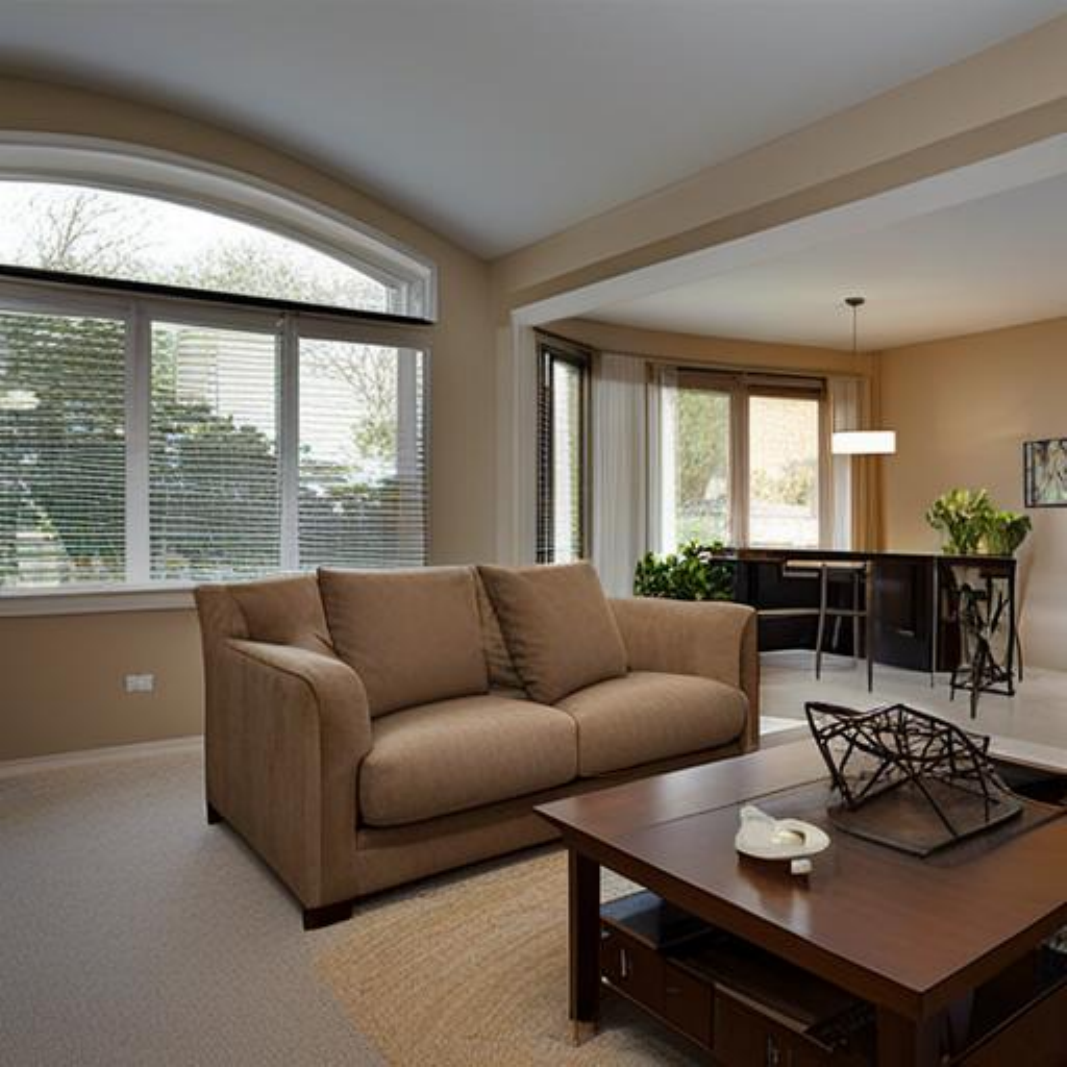} &
        \includegraphics[width=\mainResultsGraphicsWidth]{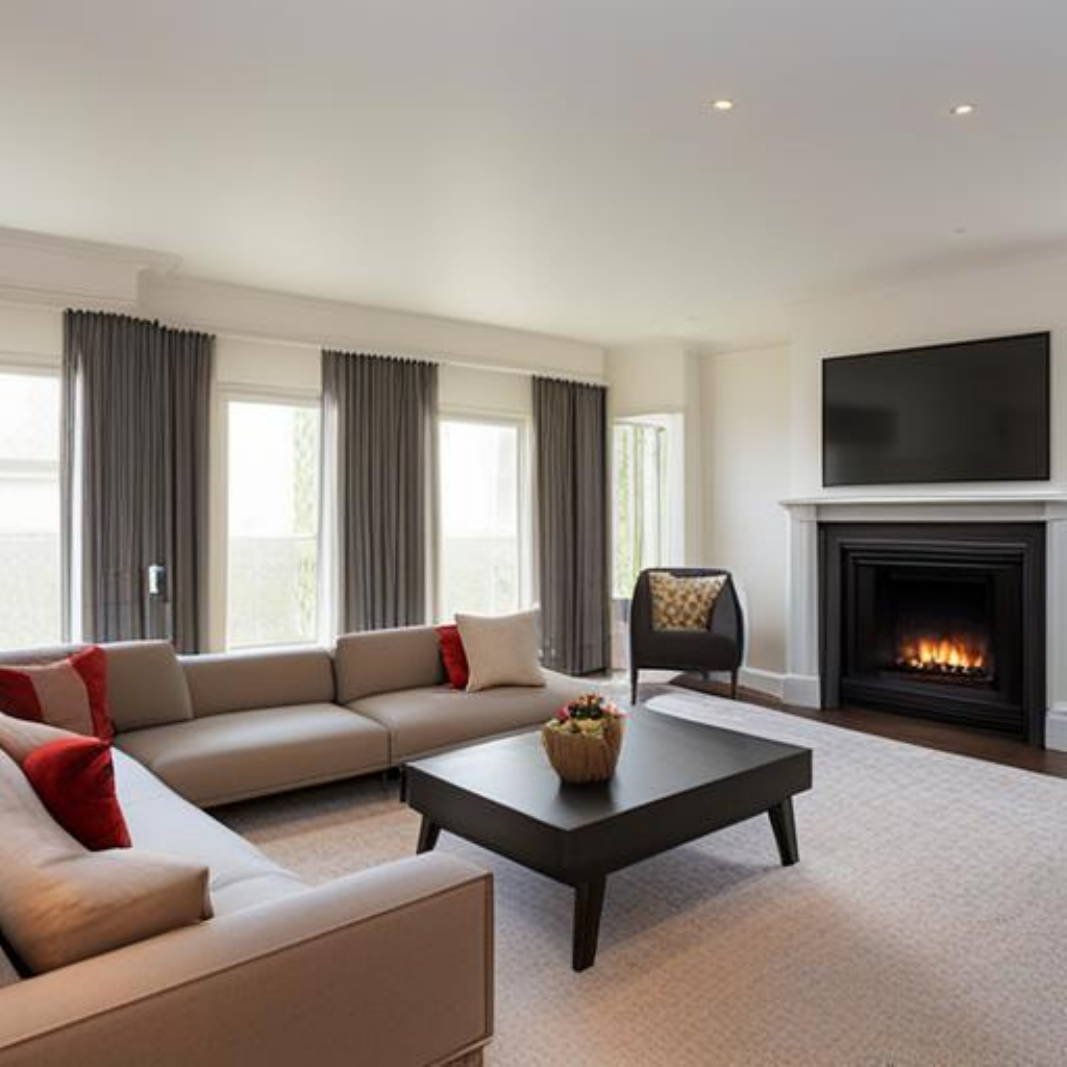} &
        \includegraphics[width=\mainResultsGraphicsWidth]{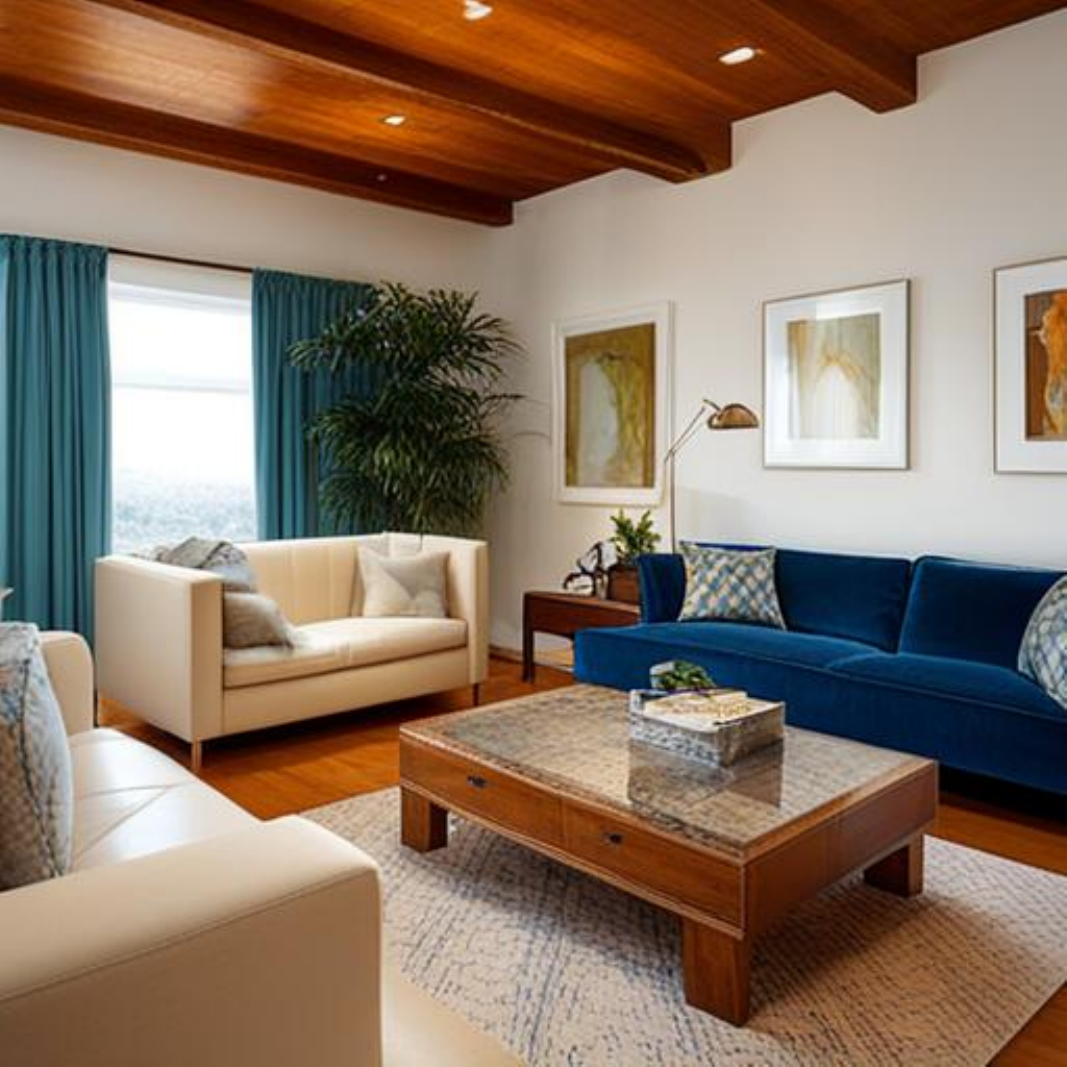} &

        \includegraphics[width=\mainResultsGraphicsWidth]{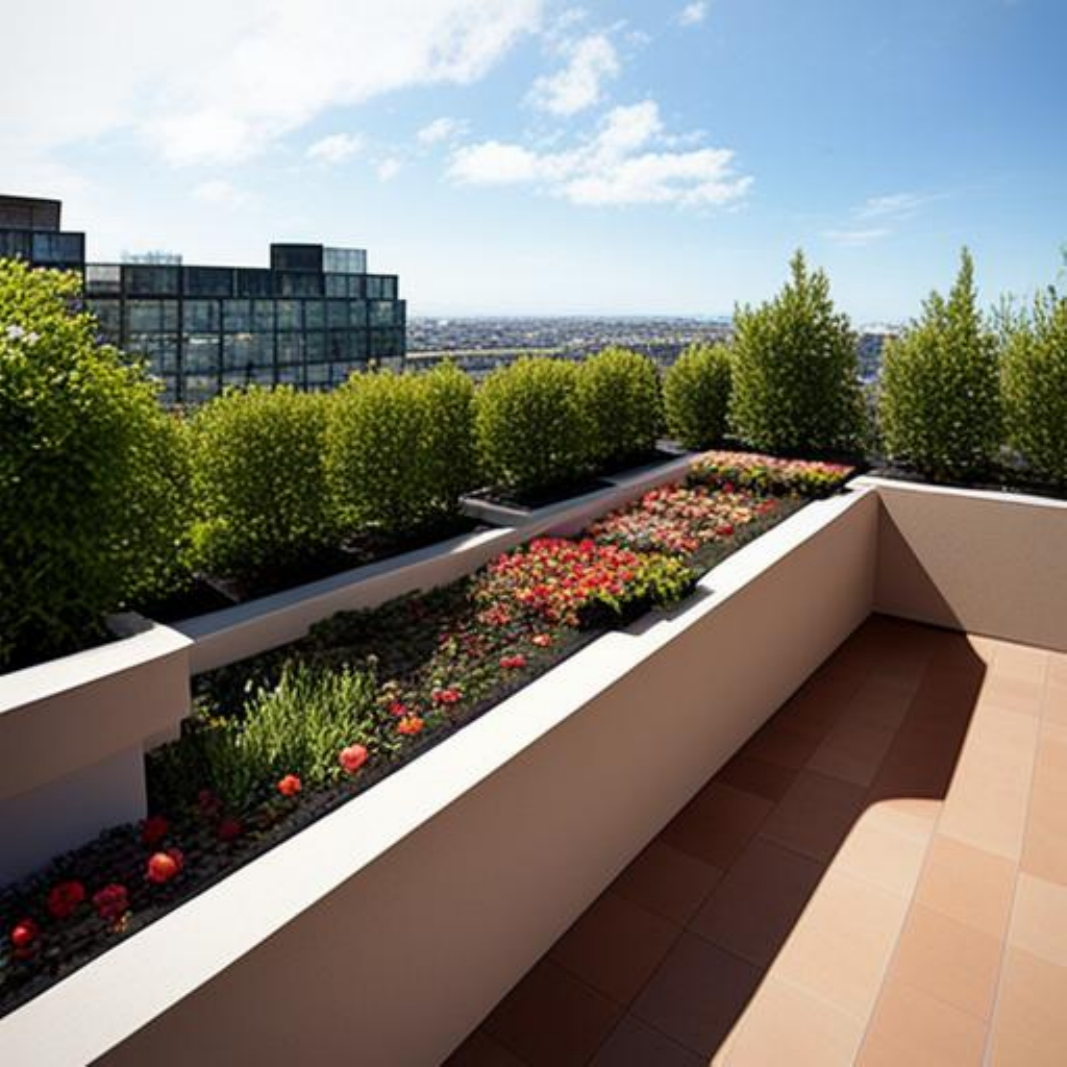} &
        \includegraphics[width=\mainResultsGraphicsWidth]{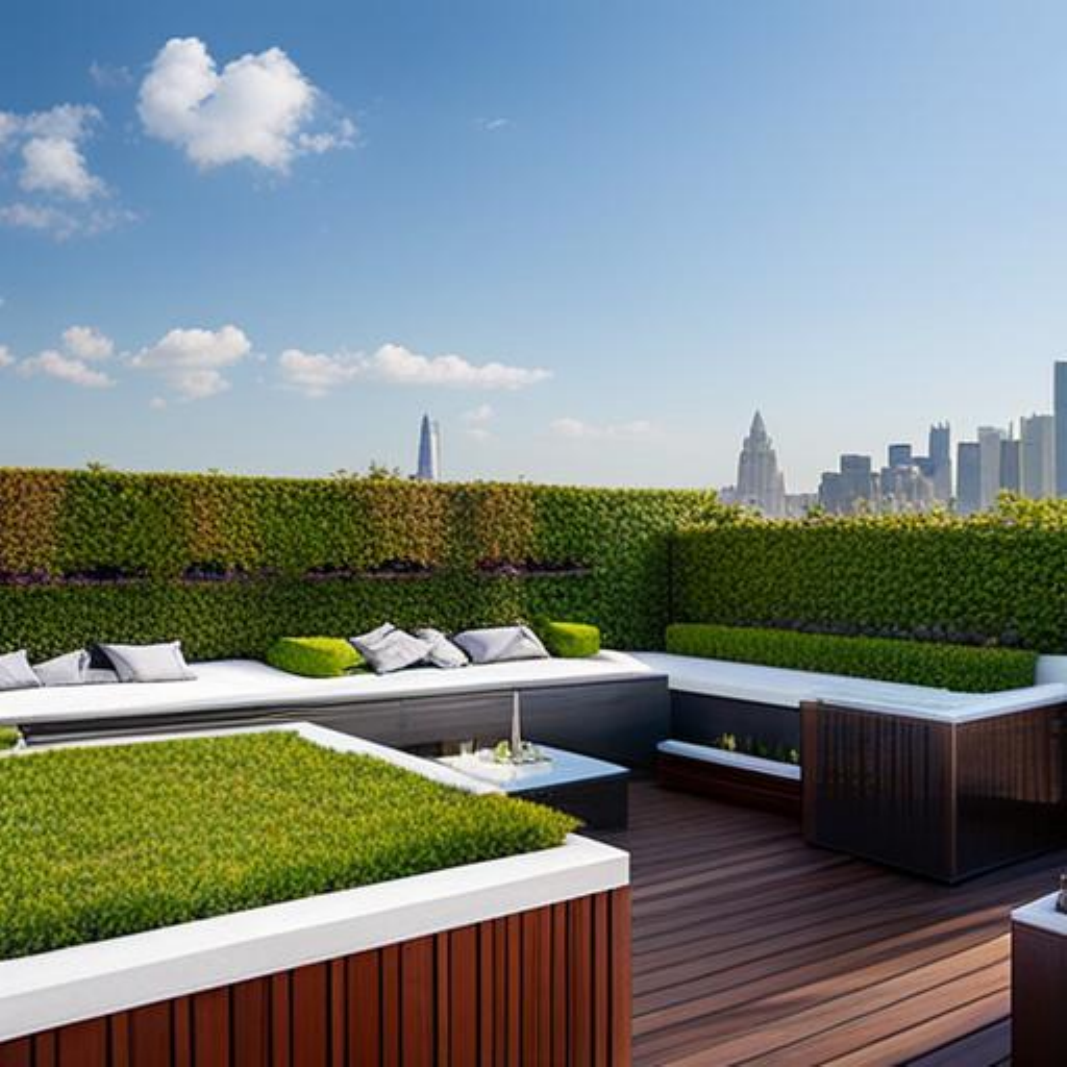} &
        \includegraphics[width=\mainResultsGraphicsWidth]{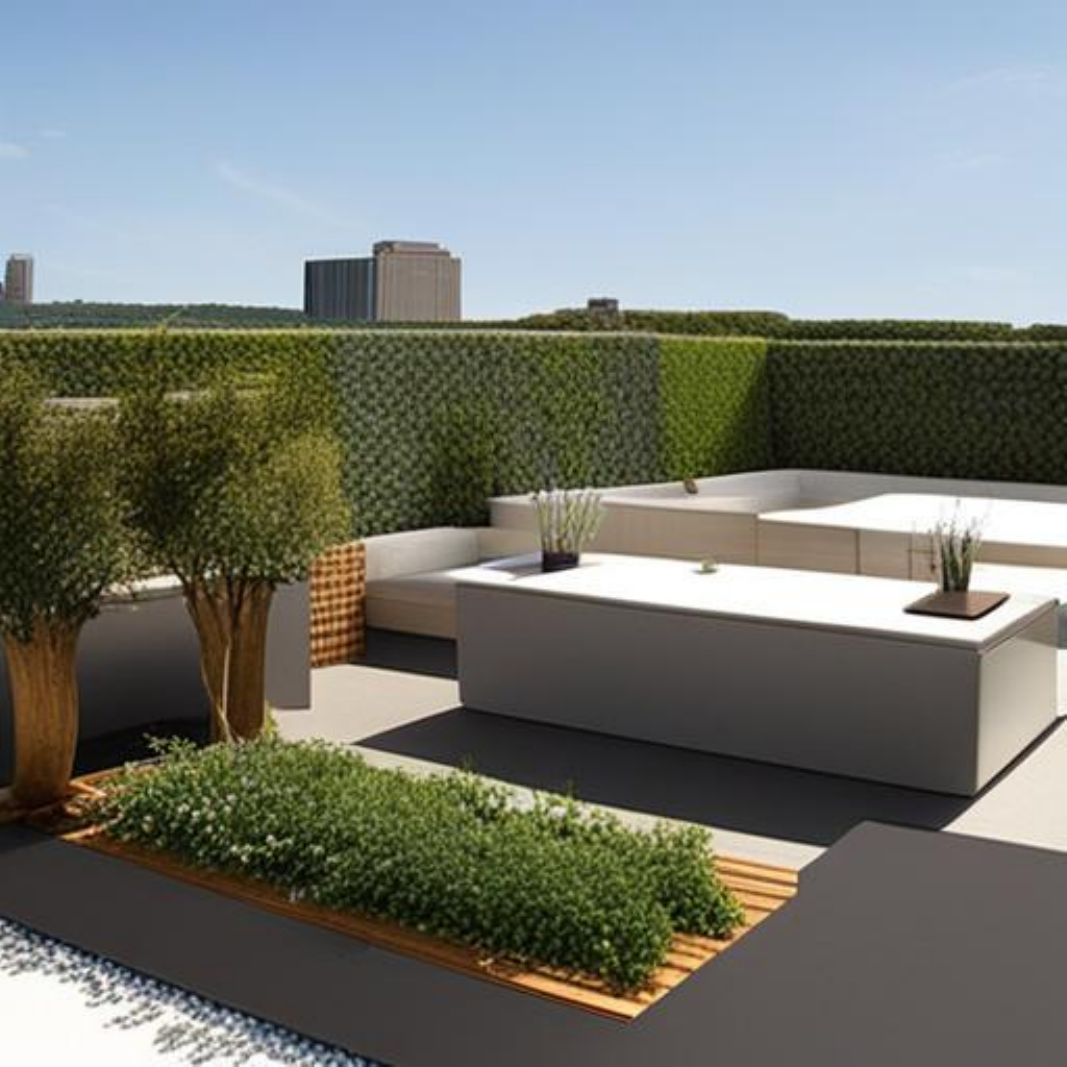} &

        \includegraphics[width=\mainResultsGraphicsWidth]{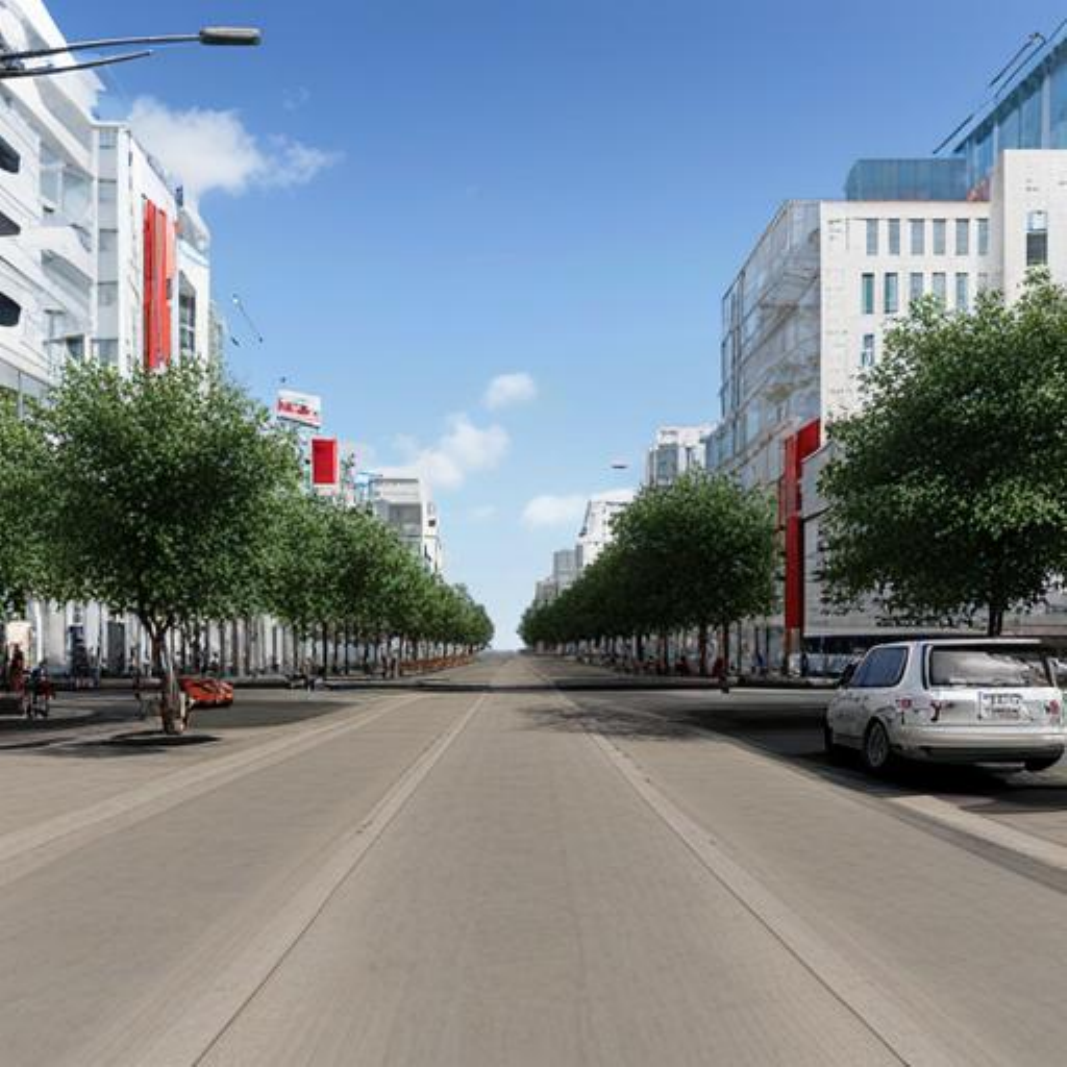} & 
        \includegraphics[width=\mainResultsGraphicsWidth]{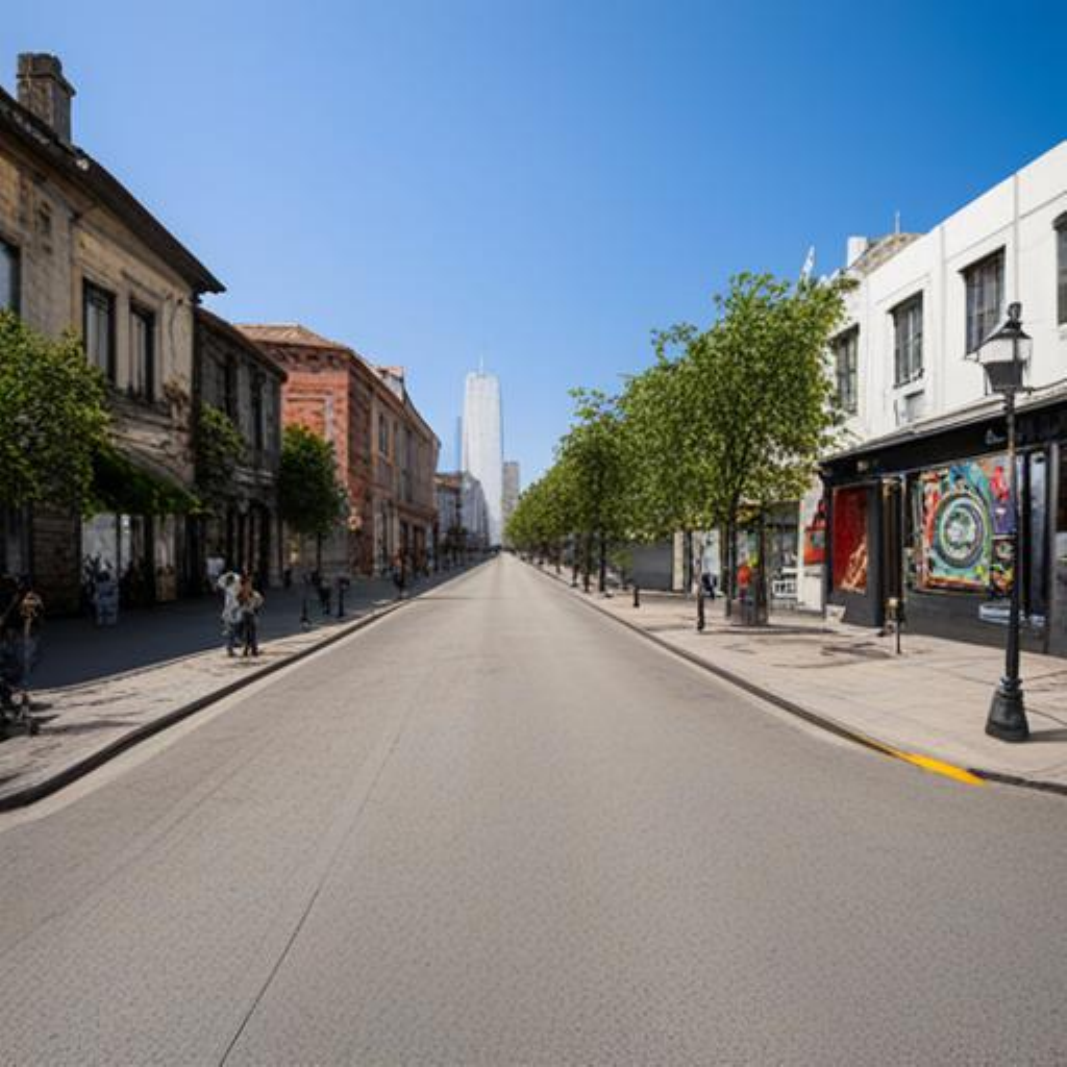} &
        \includegraphics[width=\mainResultsGraphicsWidth]{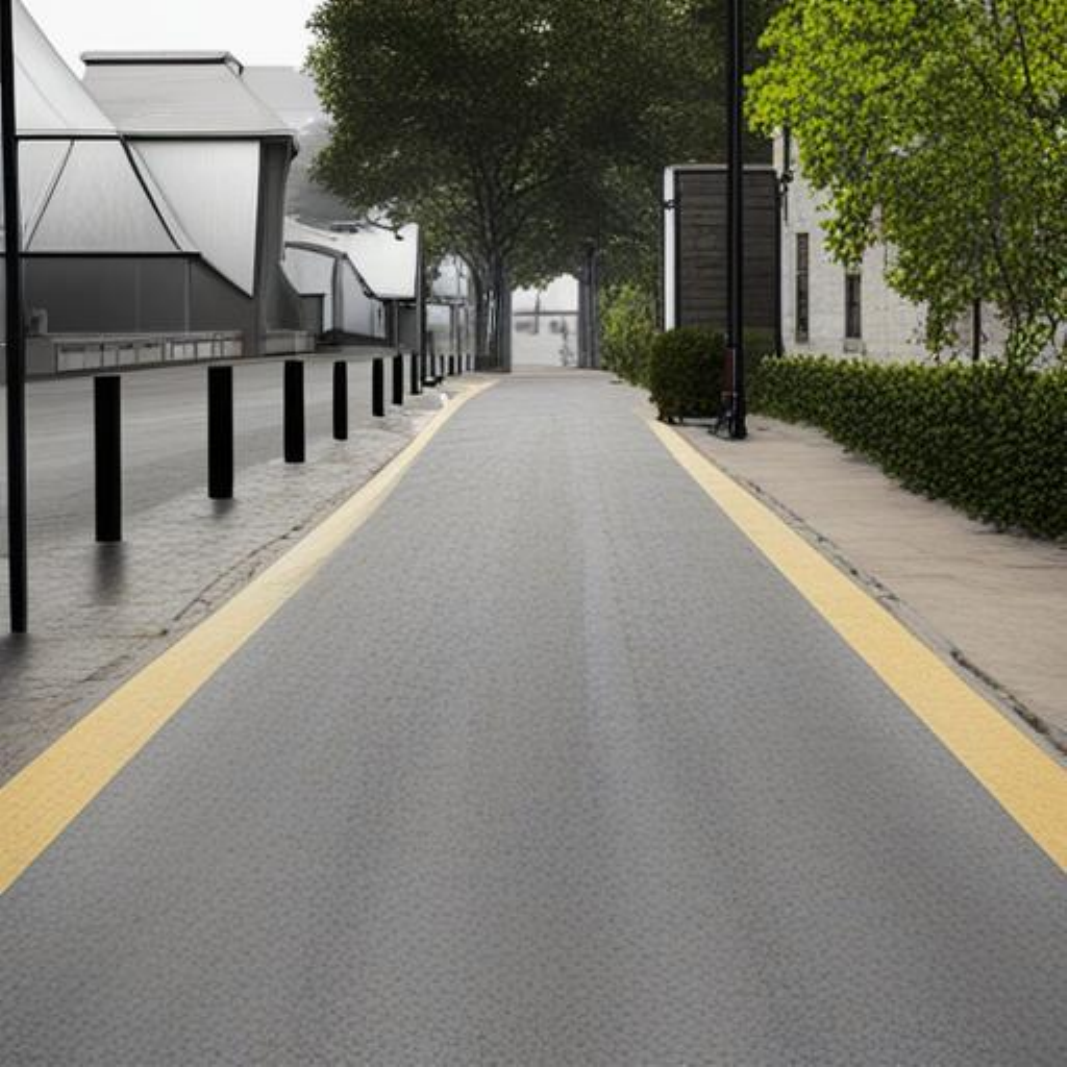} \\
        \midrule


        \multirow{-3.5}{*}{\rotatebox[origin=c]{90}{LayoutDiffusion}}&
        \includegraphics[width={\mainResultsGraphicsWidth}]{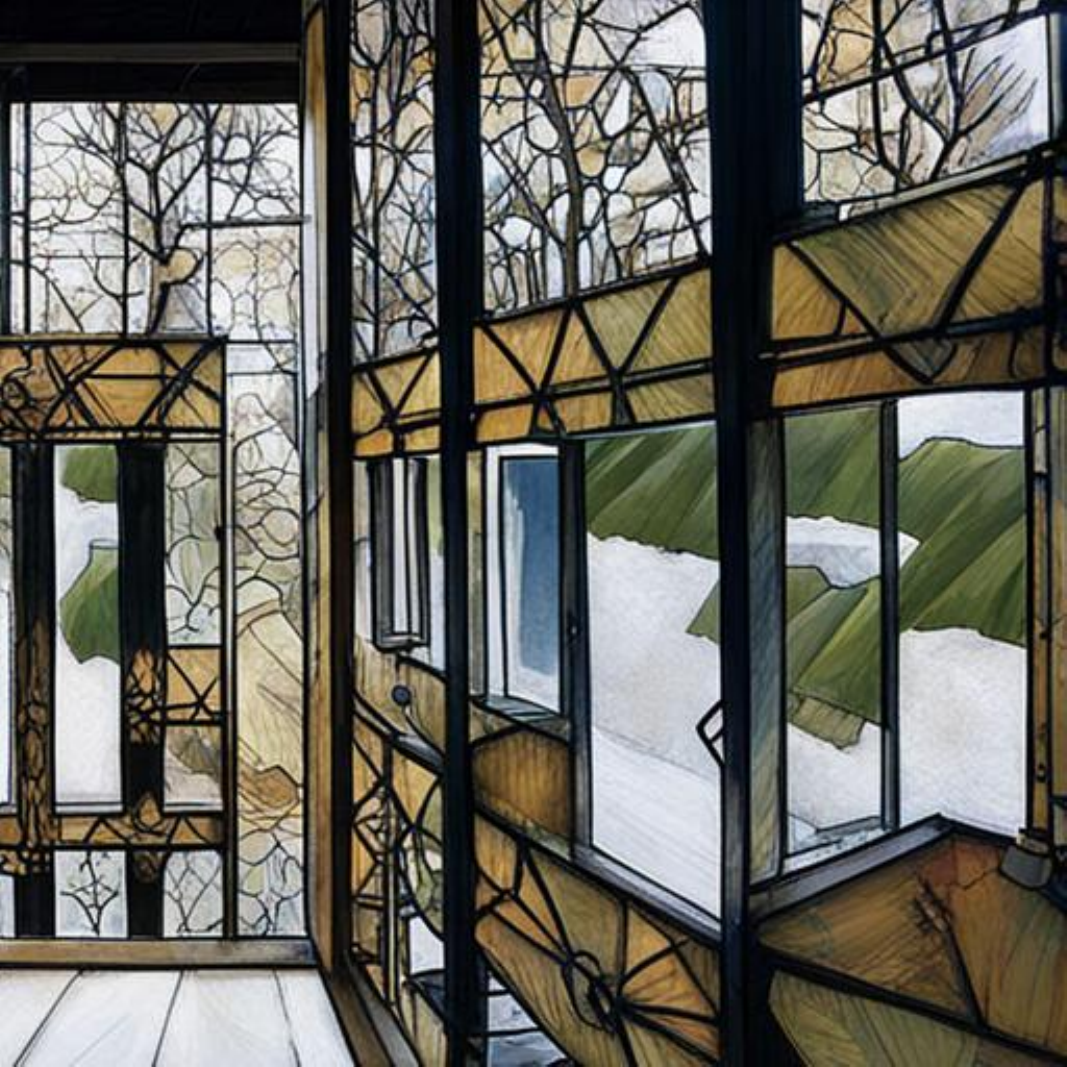} &
        \includegraphics[width={\mainResultsGraphicsWidth}]{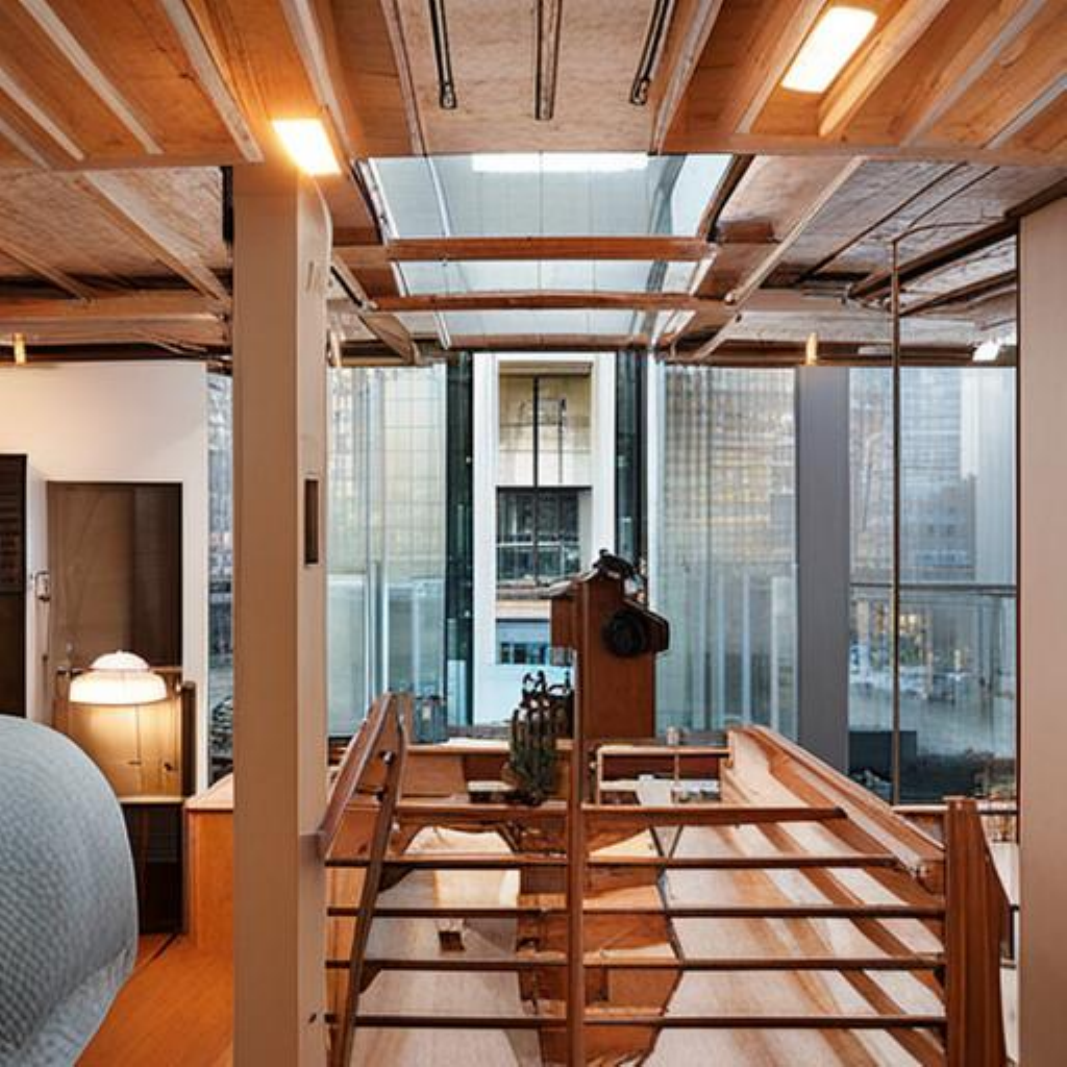} &
        \includegraphics[width={\mainResultsGraphicsWidth}]{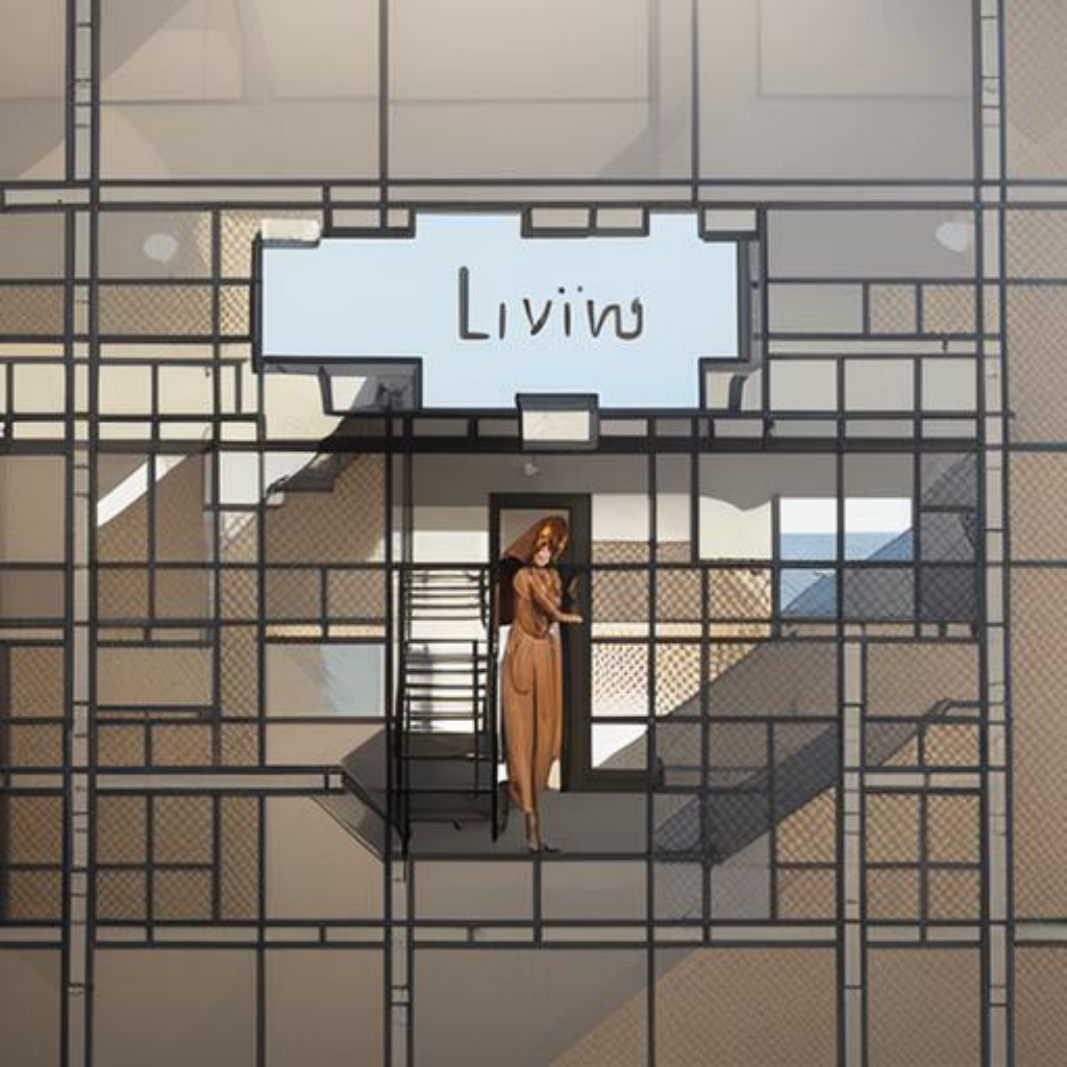} &

        \includegraphics[width={\mainResultsGraphicsWidth}]{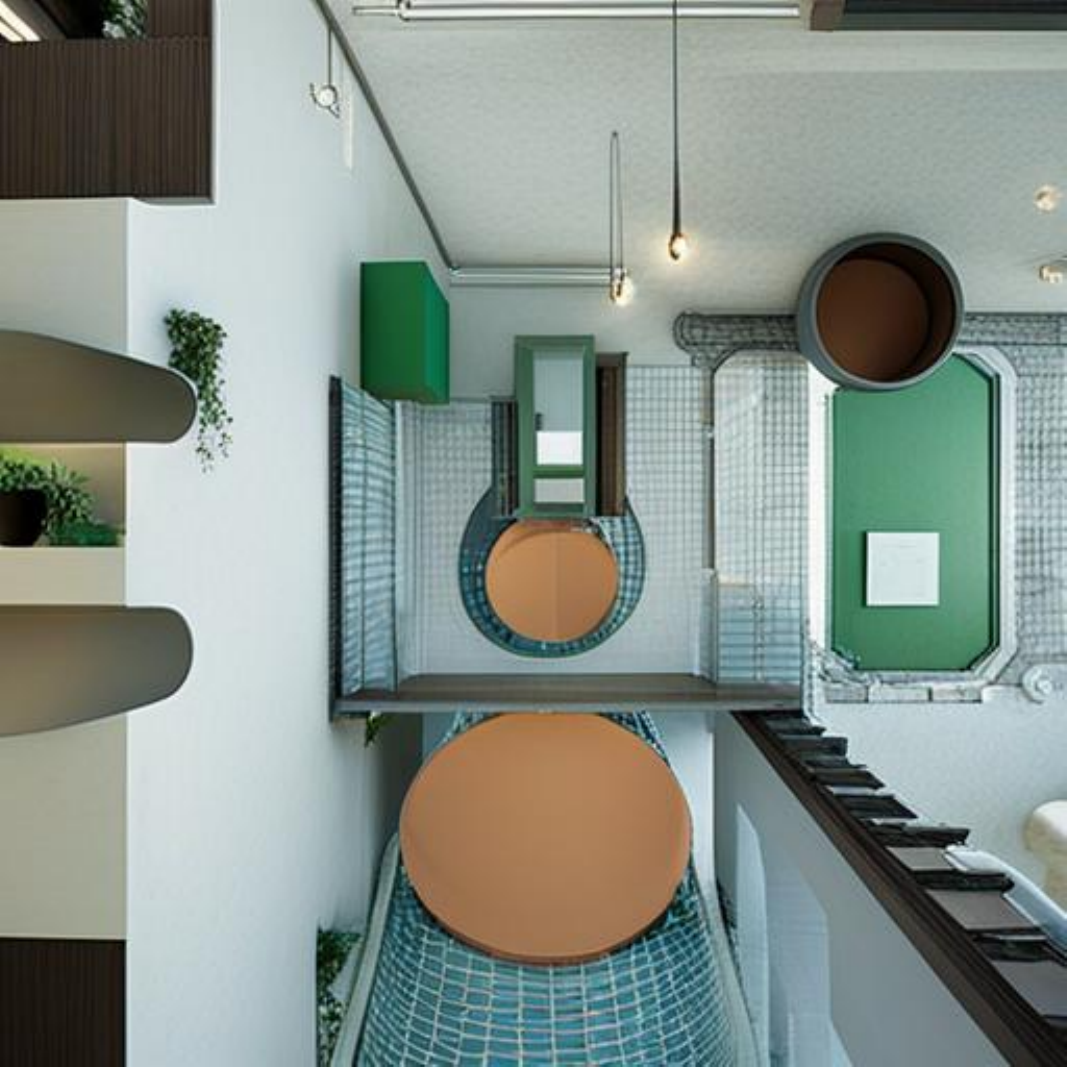} &
        \includegraphics[width={\mainResultsGraphicsWidth}]{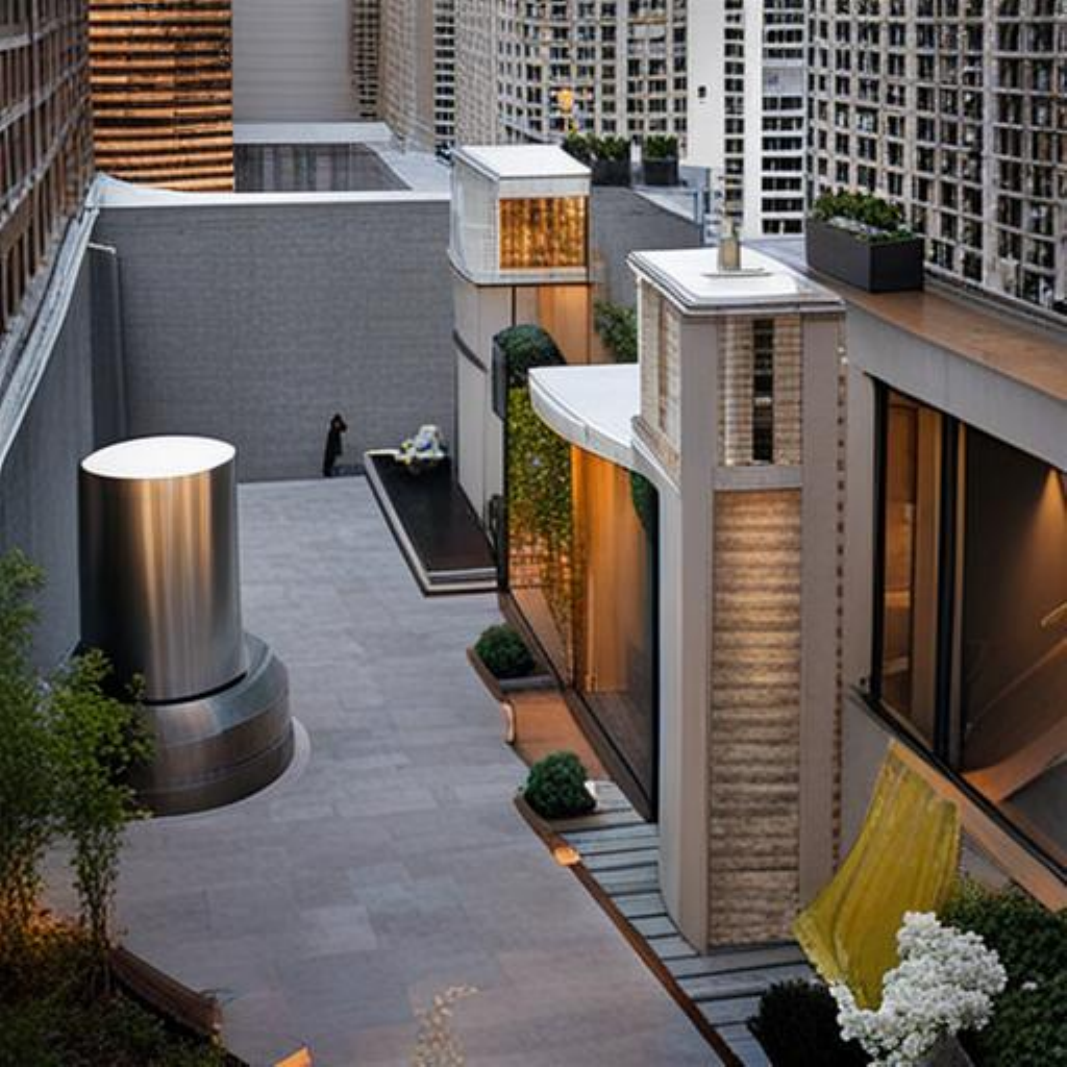} &
        \includegraphics[width={\mainResultsGraphicsWidth}]{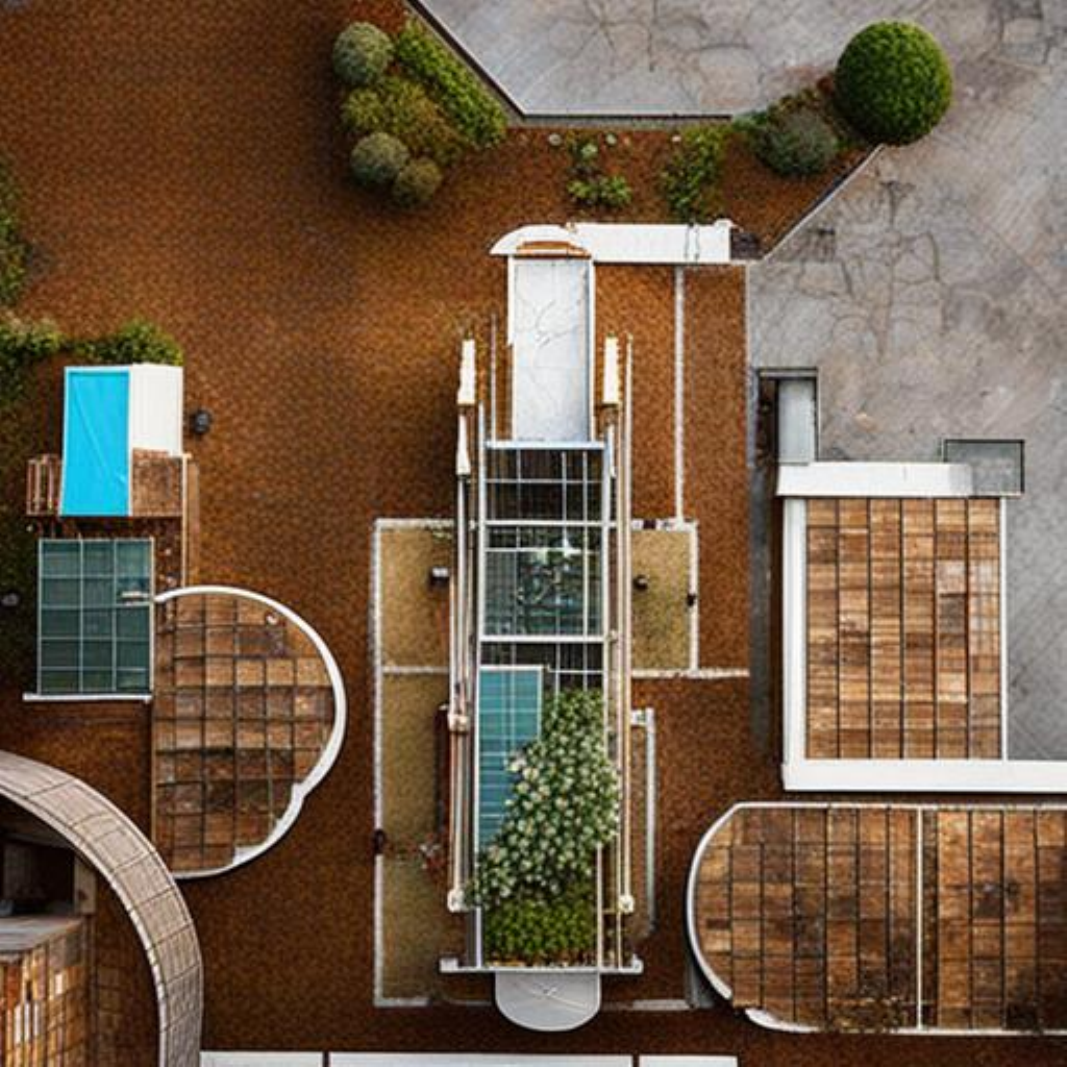} &

        \includegraphics[width={\mainResultsGraphicsWidth}]{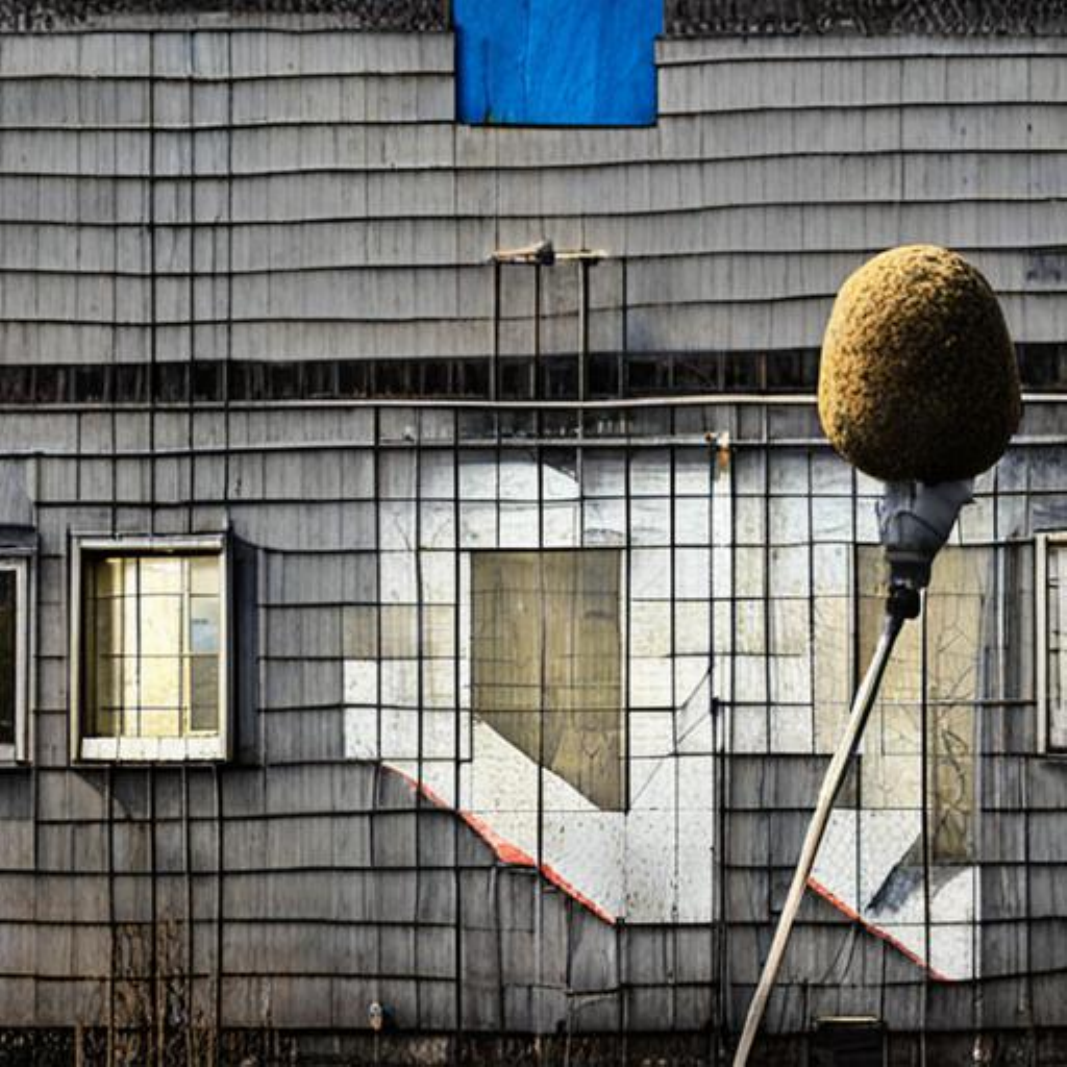} &
        \includegraphics[width={\mainResultsGraphicsWidth}]{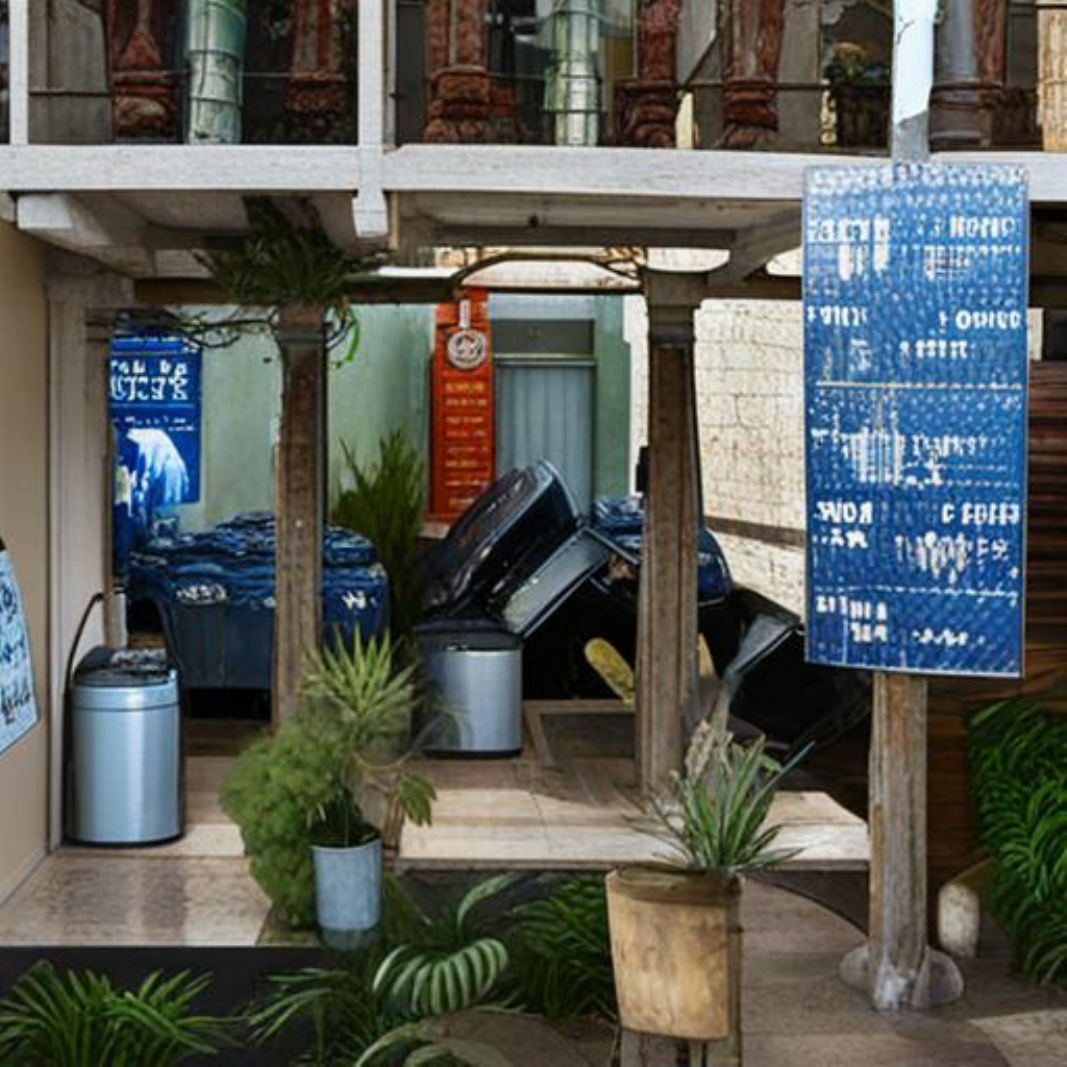} &
        \includegraphics[width={\mainResultsGraphicsWidth}]{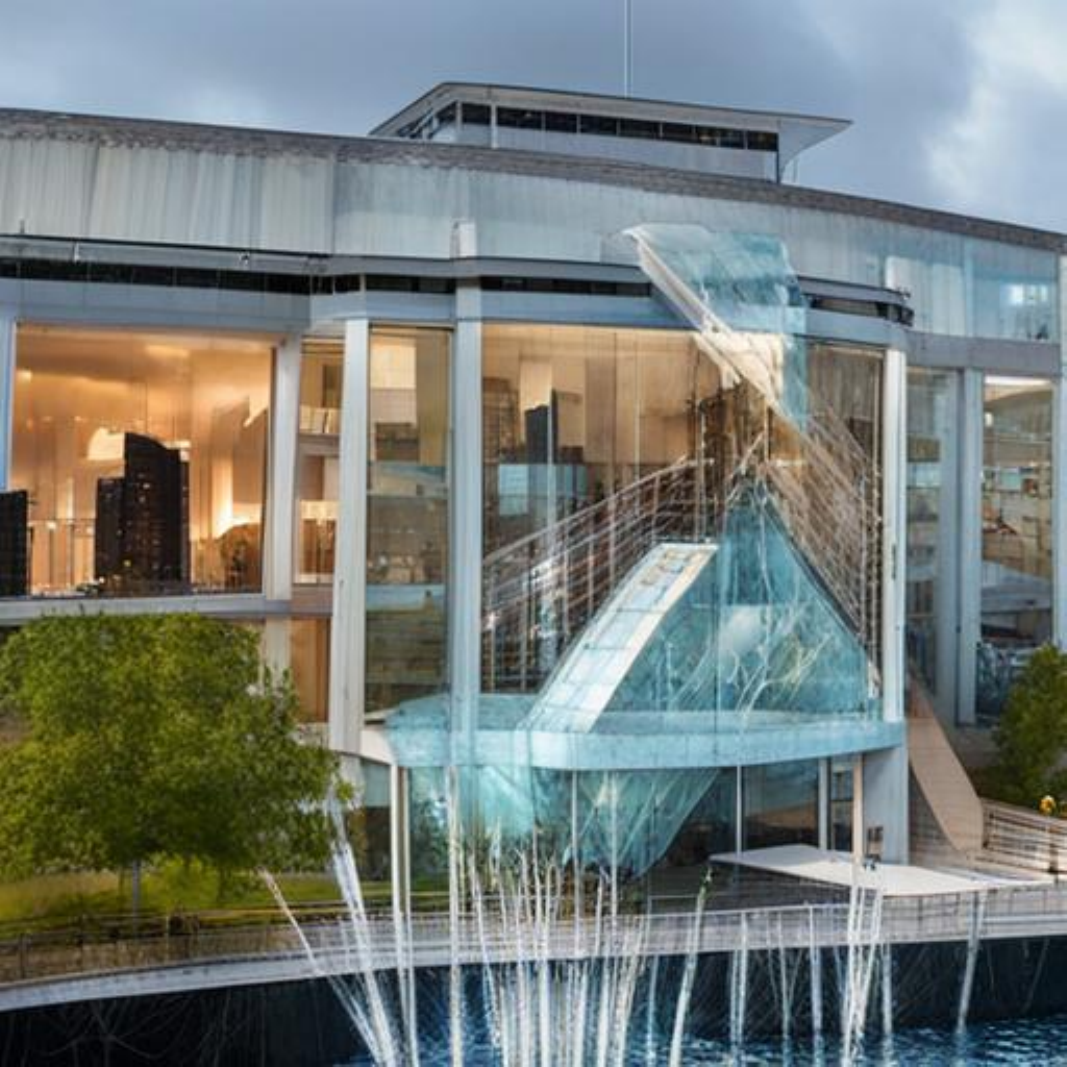} \\

        &
        \includegraphics[width={\mainResultsGraphicsWidth}]{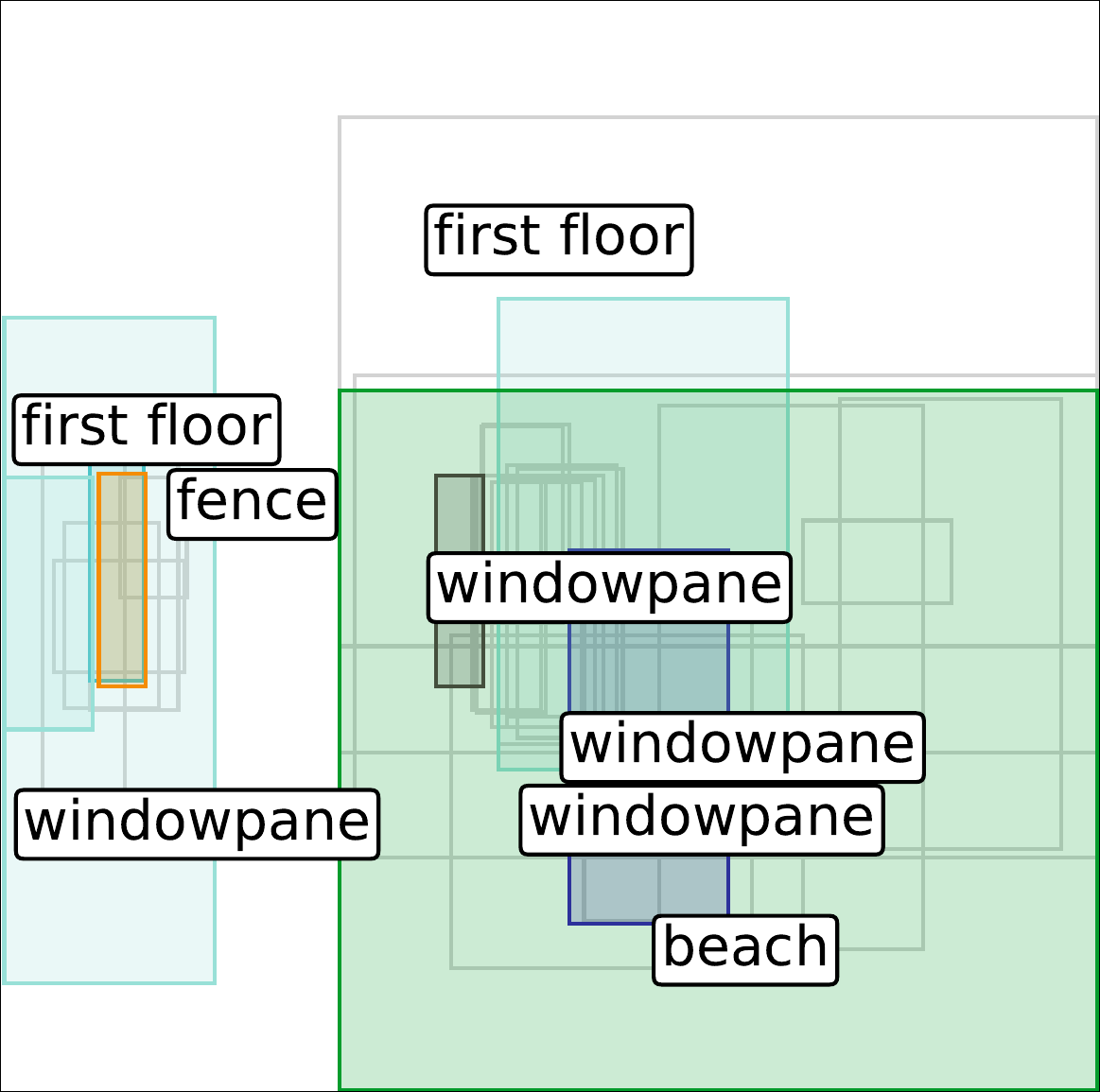} &
        \includegraphics[width={\mainResultsGraphicsWidth}]{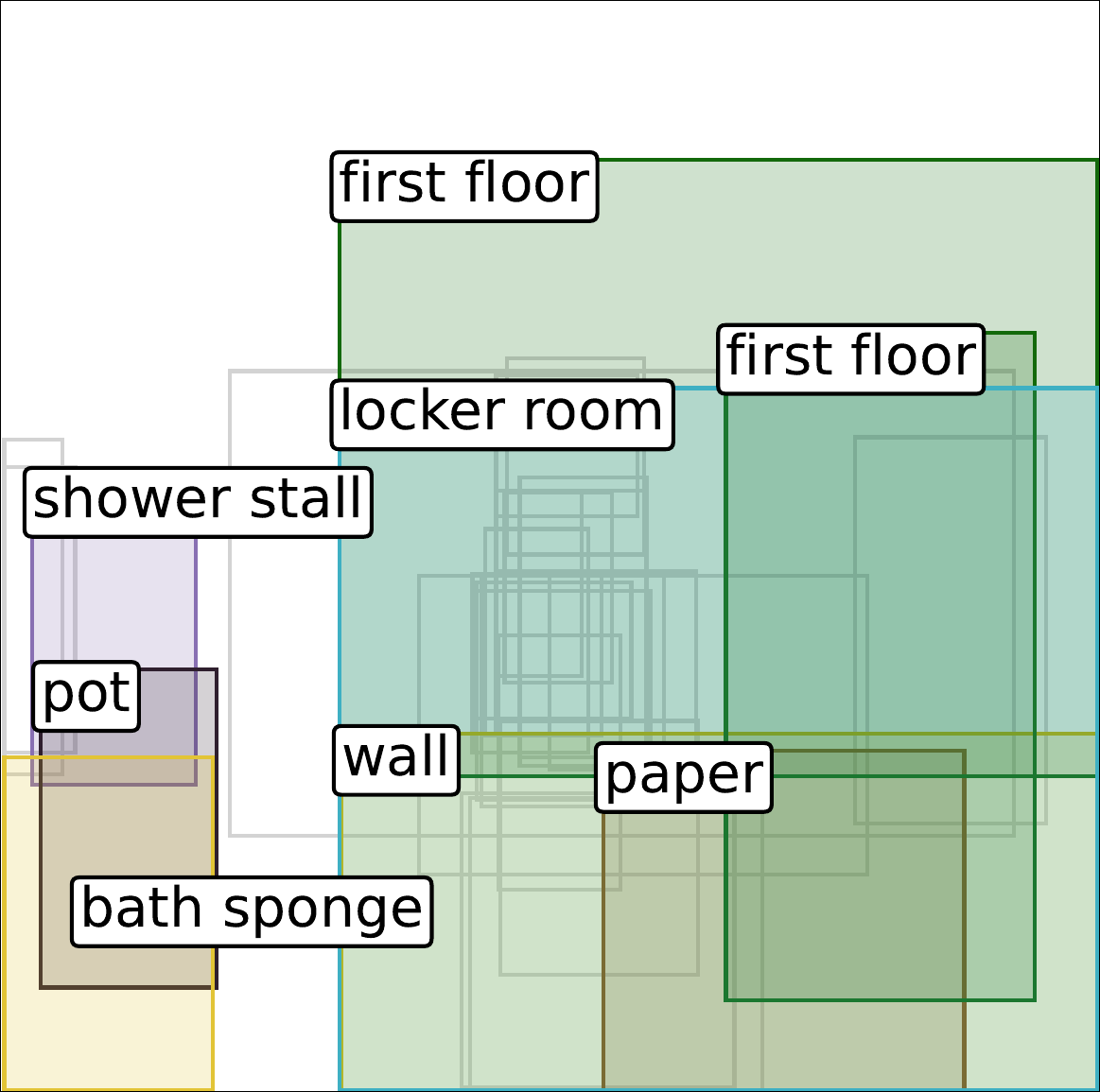} &
        \includegraphics[width={\mainResultsGraphicsWidth}]{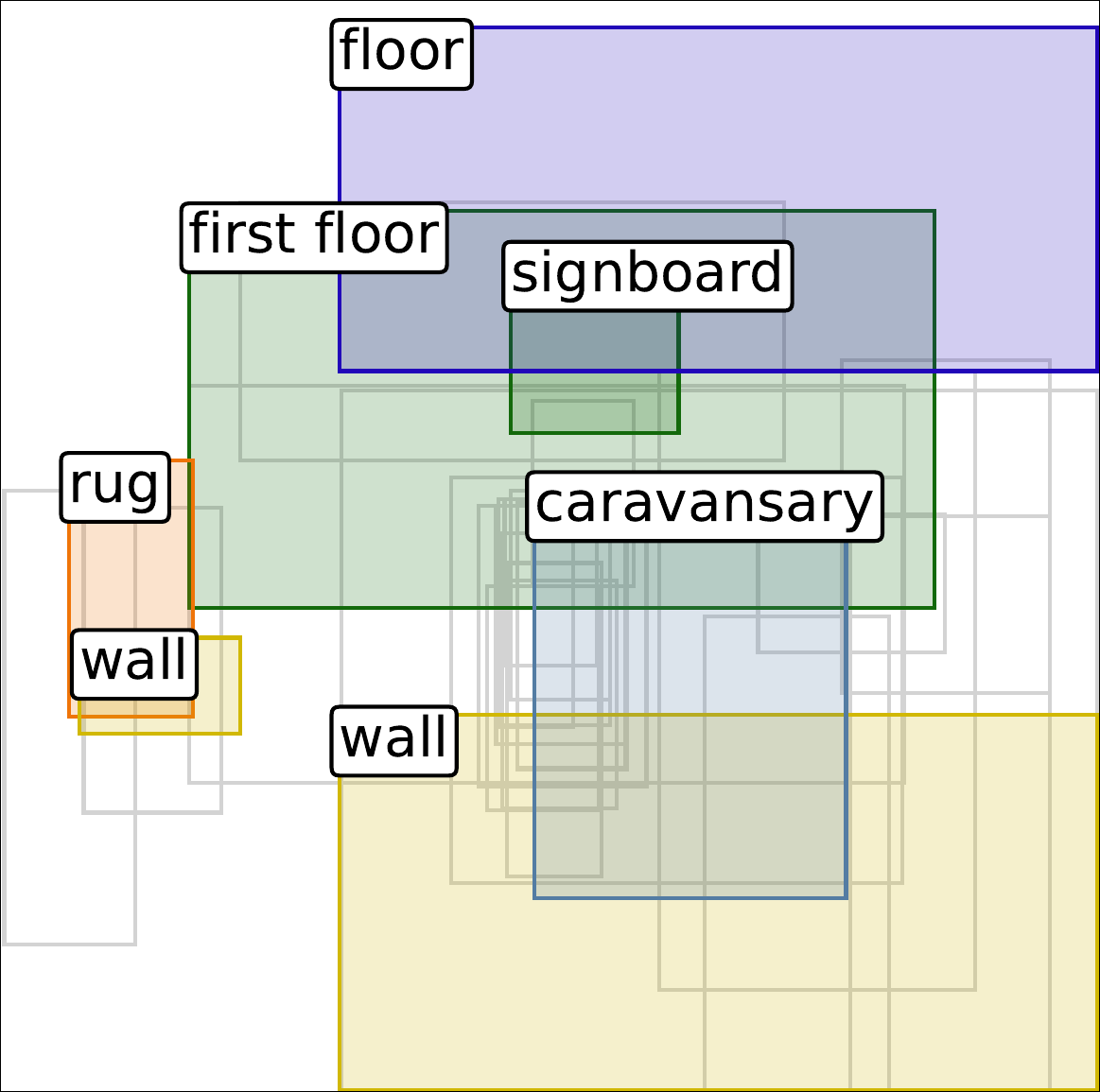} &

        \includegraphics[width={\mainResultsGraphicsWidth}]{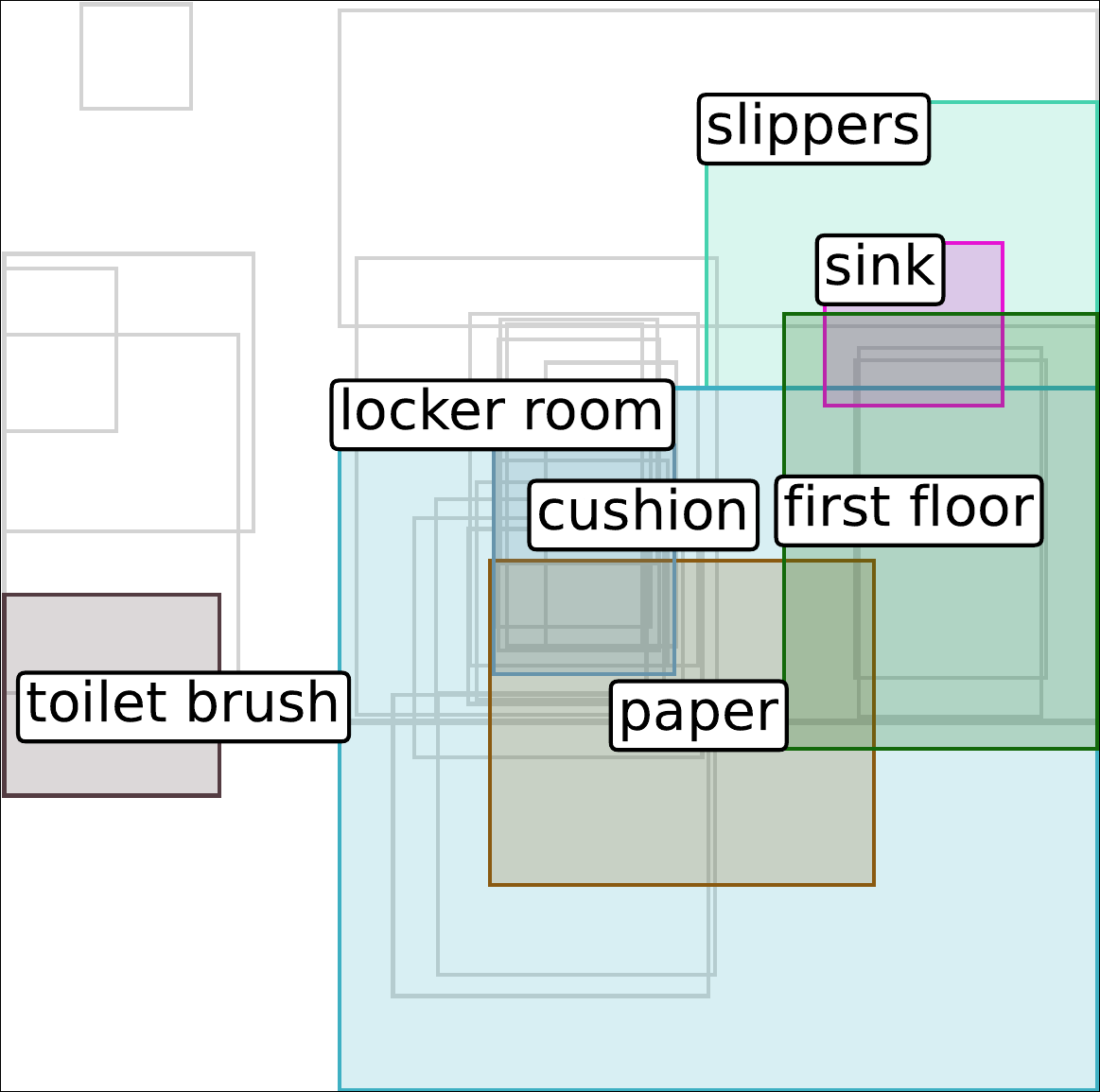} &
        \includegraphics[width={\mainResultsGraphicsWidth}]{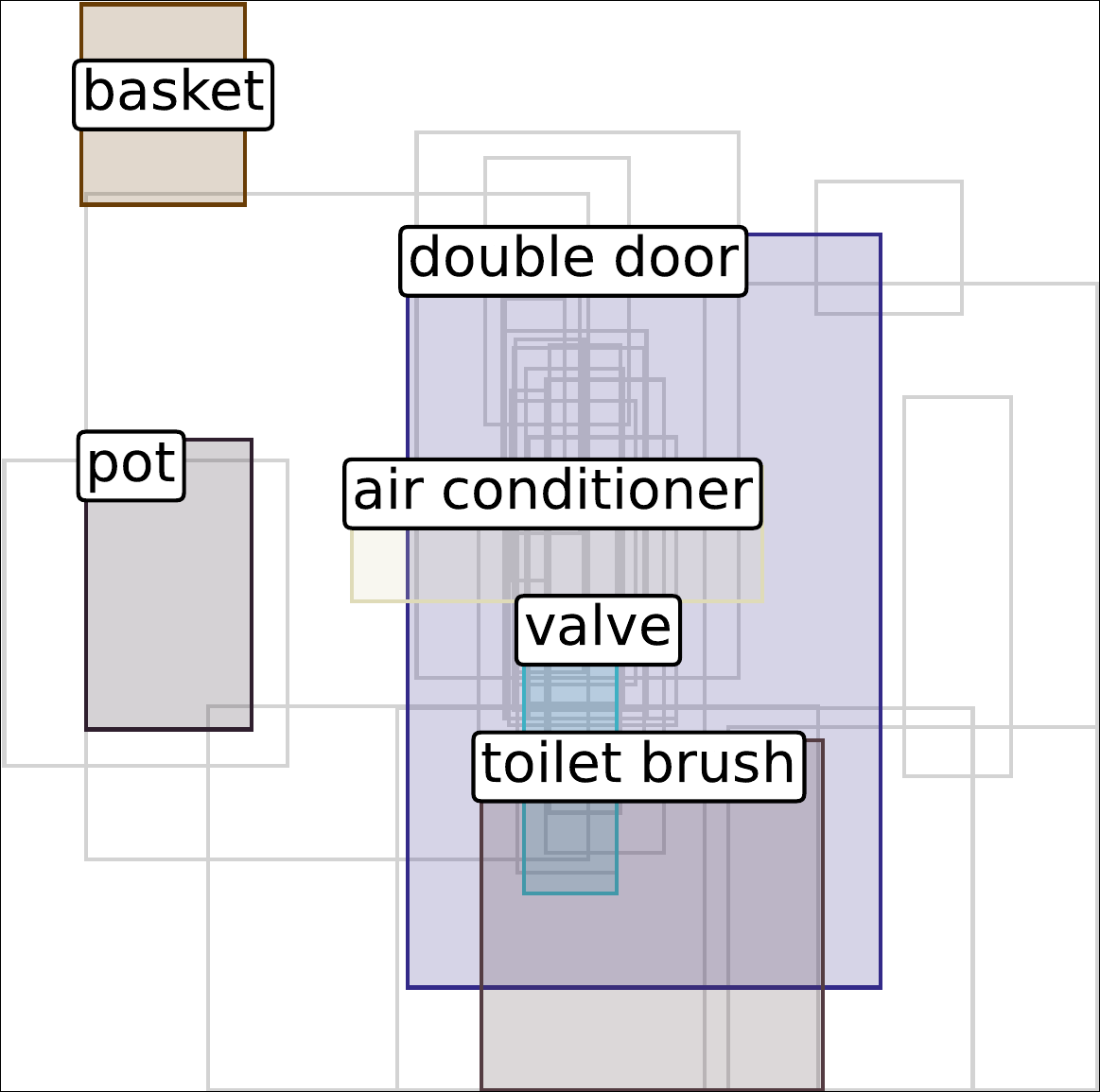} &
        \includegraphics[width={\mainResultsGraphicsWidth}]{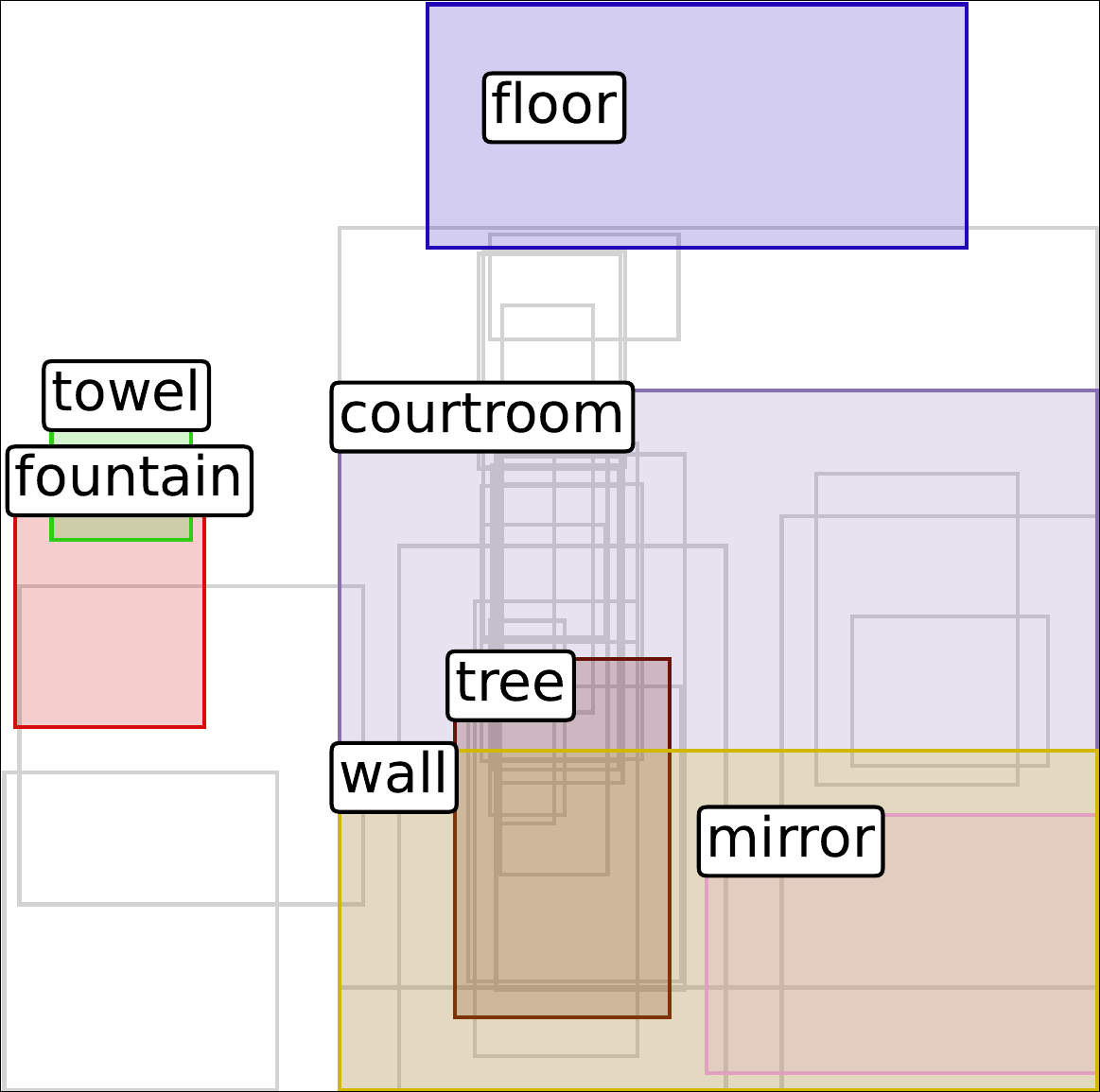} &

        \includegraphics[width={\mainResultsGraphicsWidth}]{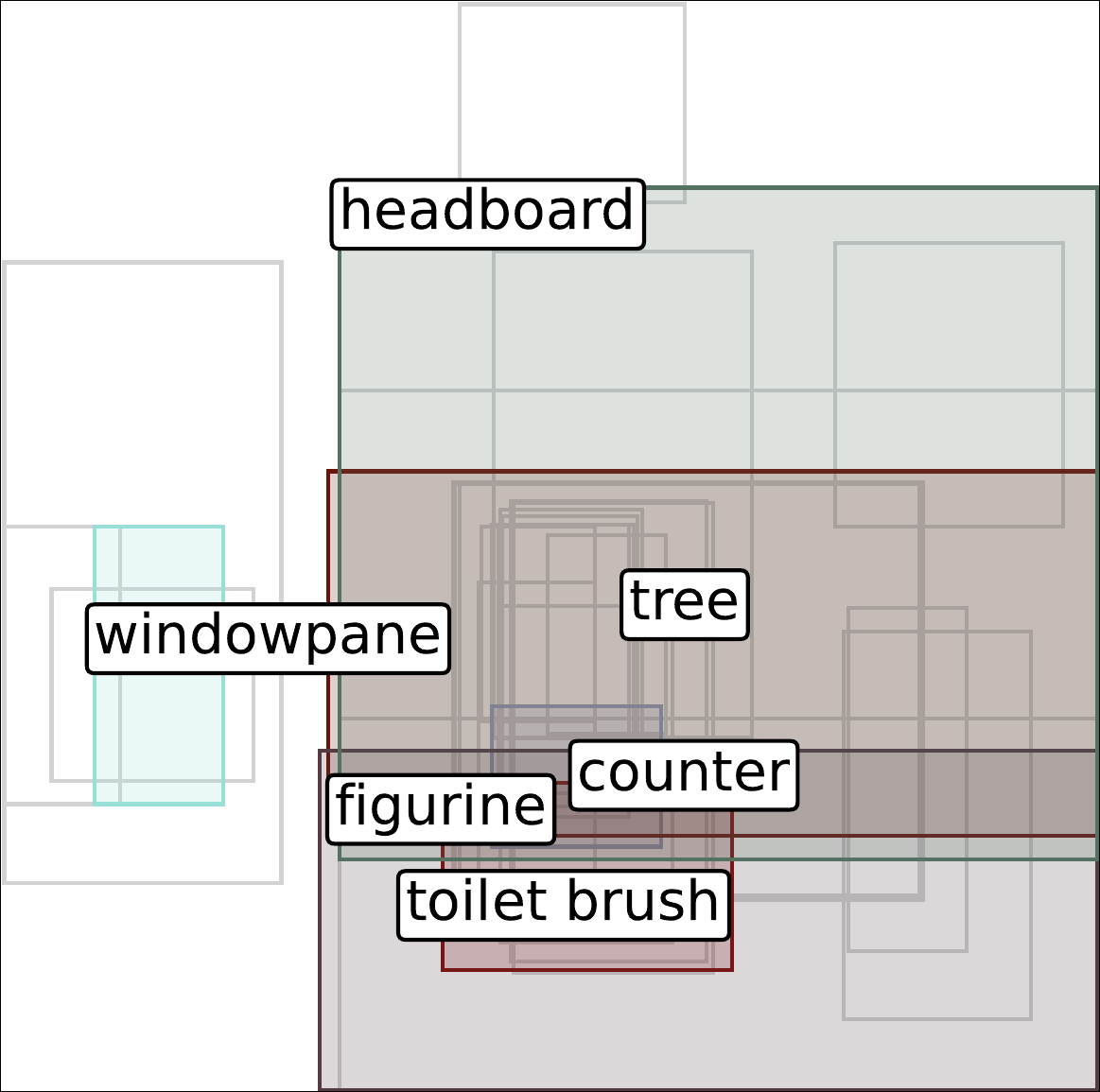} &
        \includegraphics[width={\mainResultsGraphicsWidth}]{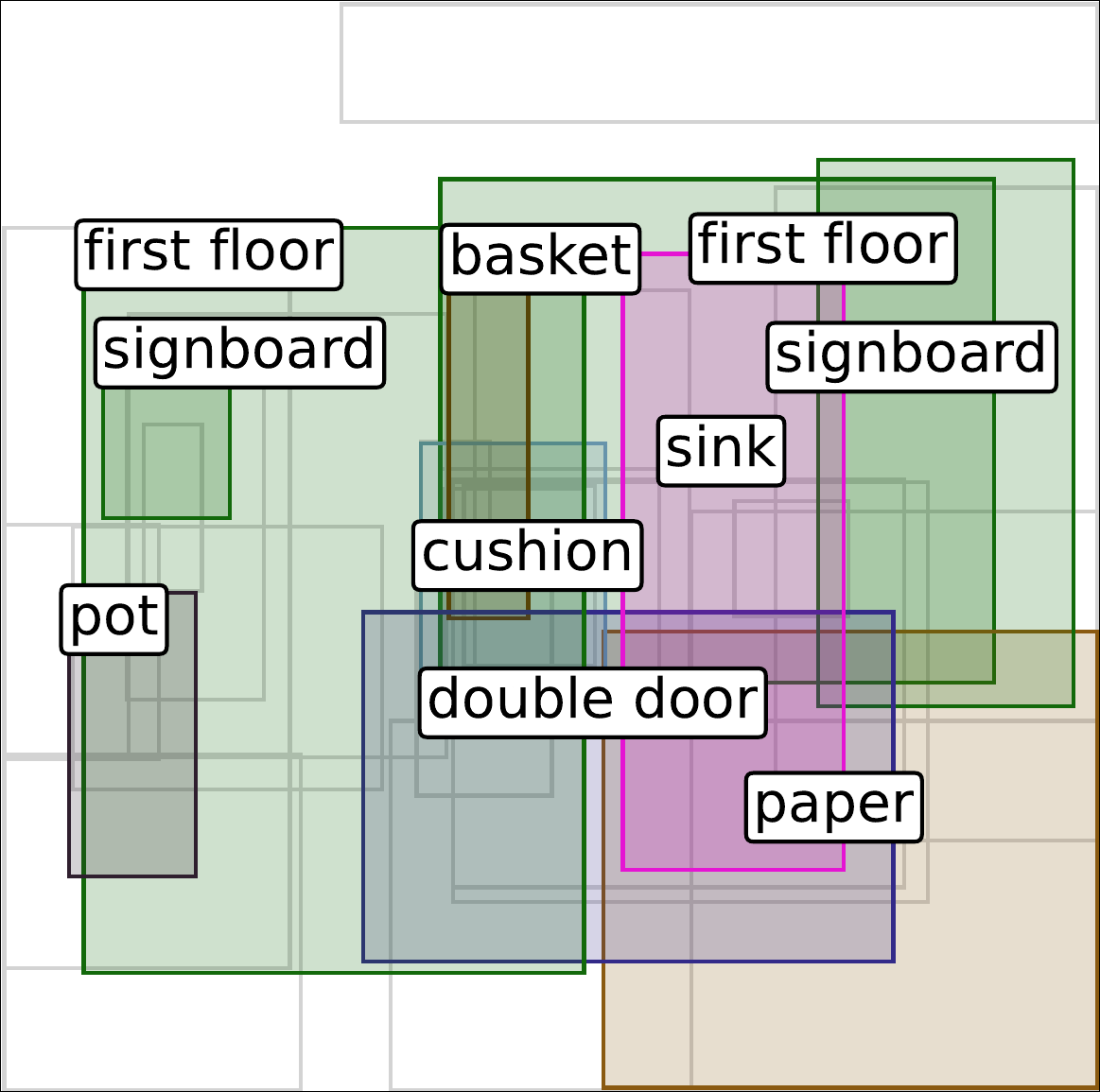} &
        \includegraphics[width={\mainResultsGraphicsWidth}]{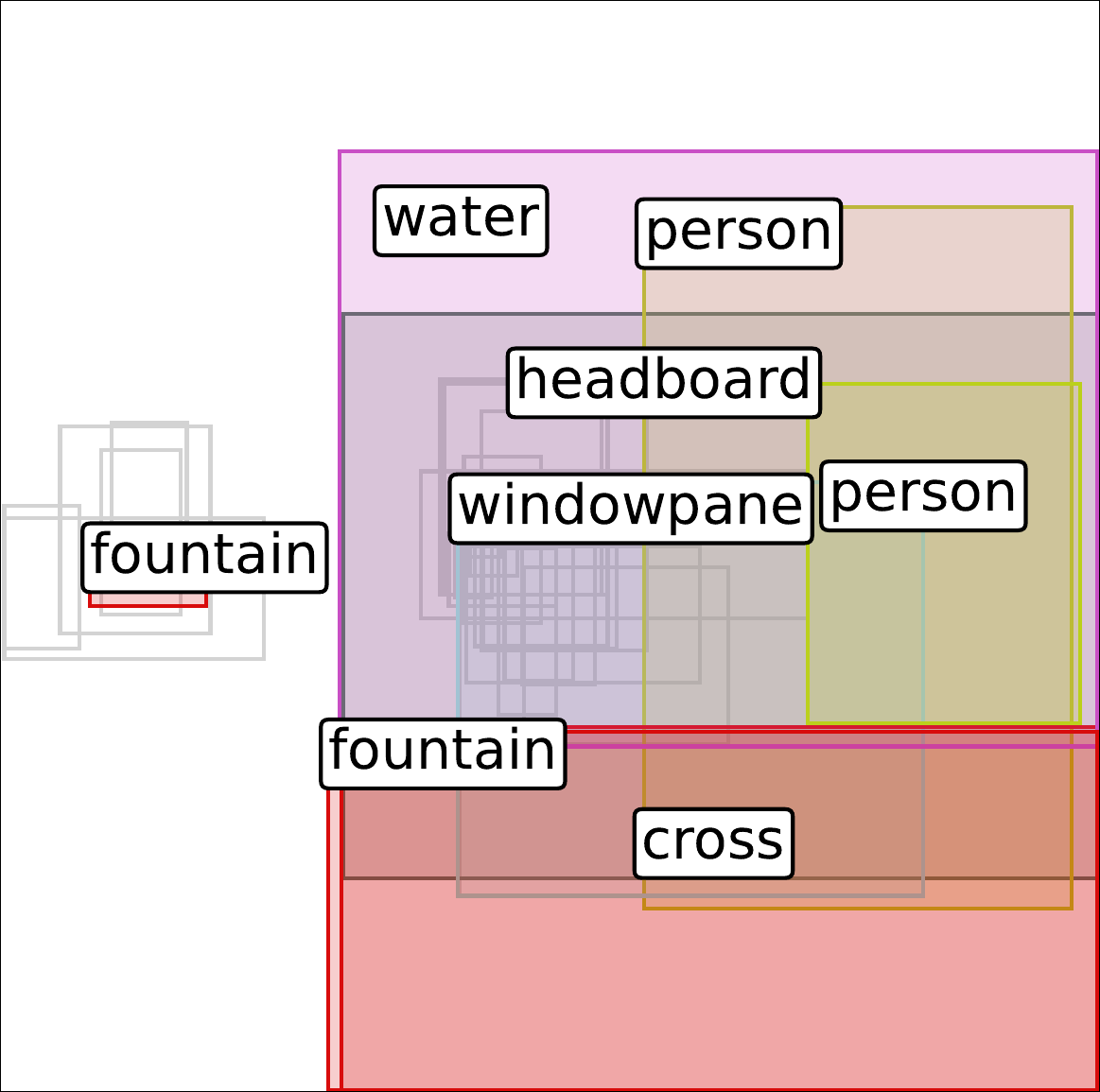} \\

        \midrule
        \multirow{-1.5}{*}{\rotatebox[origin=c]{90}{GPT4o}}&
        \includegraphics[width=\mainResultsGraphicsWidth]{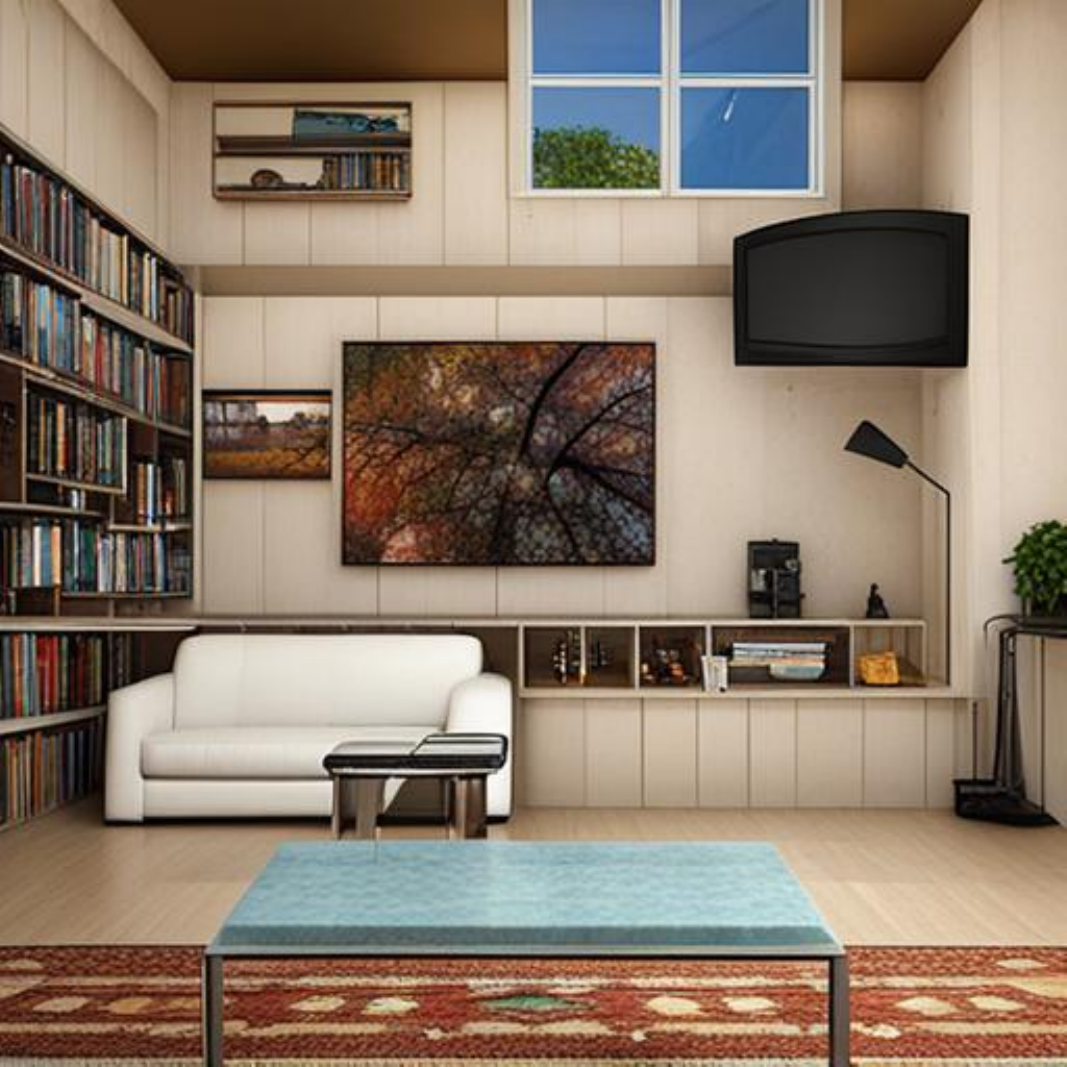} &
        \includegraphics[width=\mainResultsGraphicsWidth]{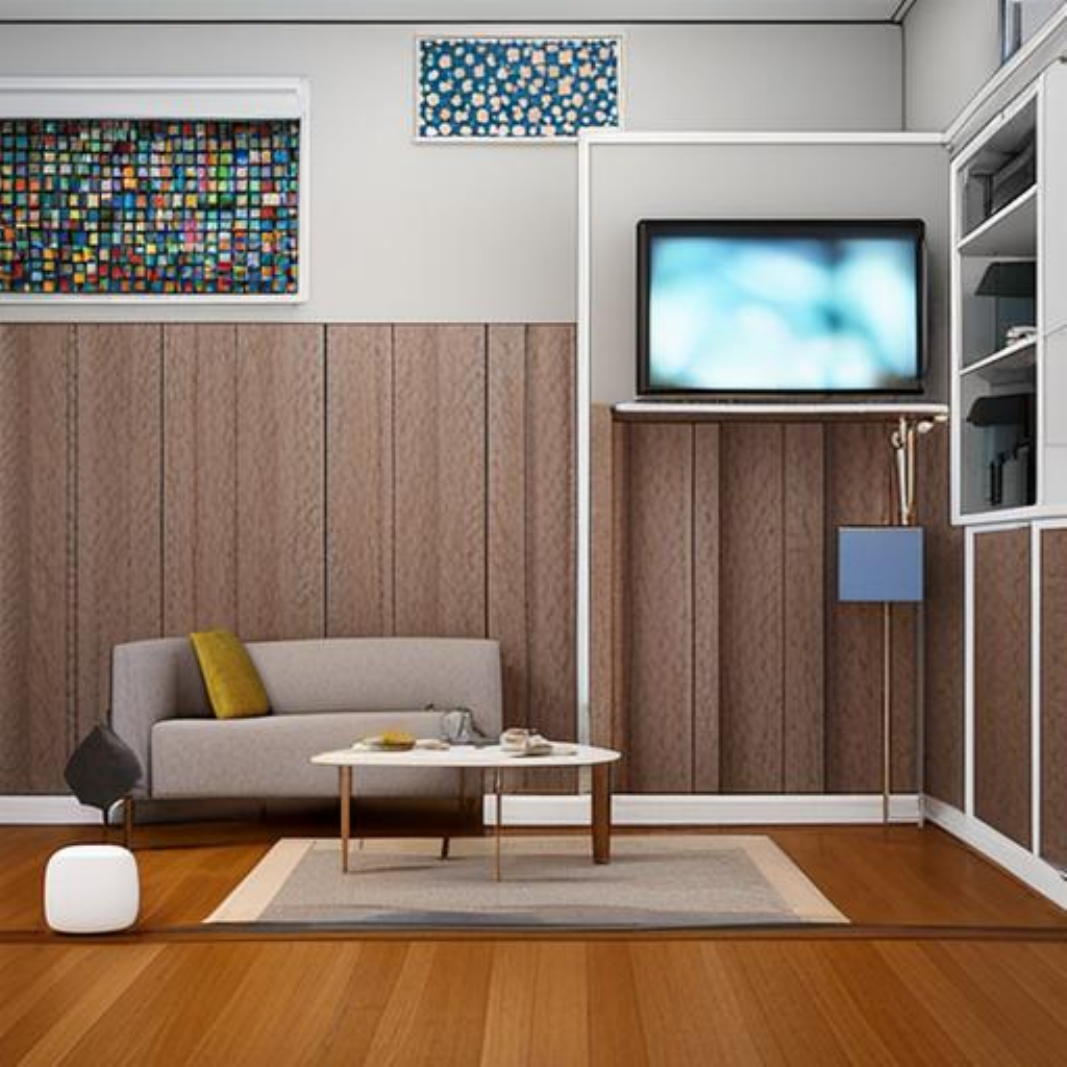} &
        \includegraphics[width=\mainResultsGraphicsWidth]{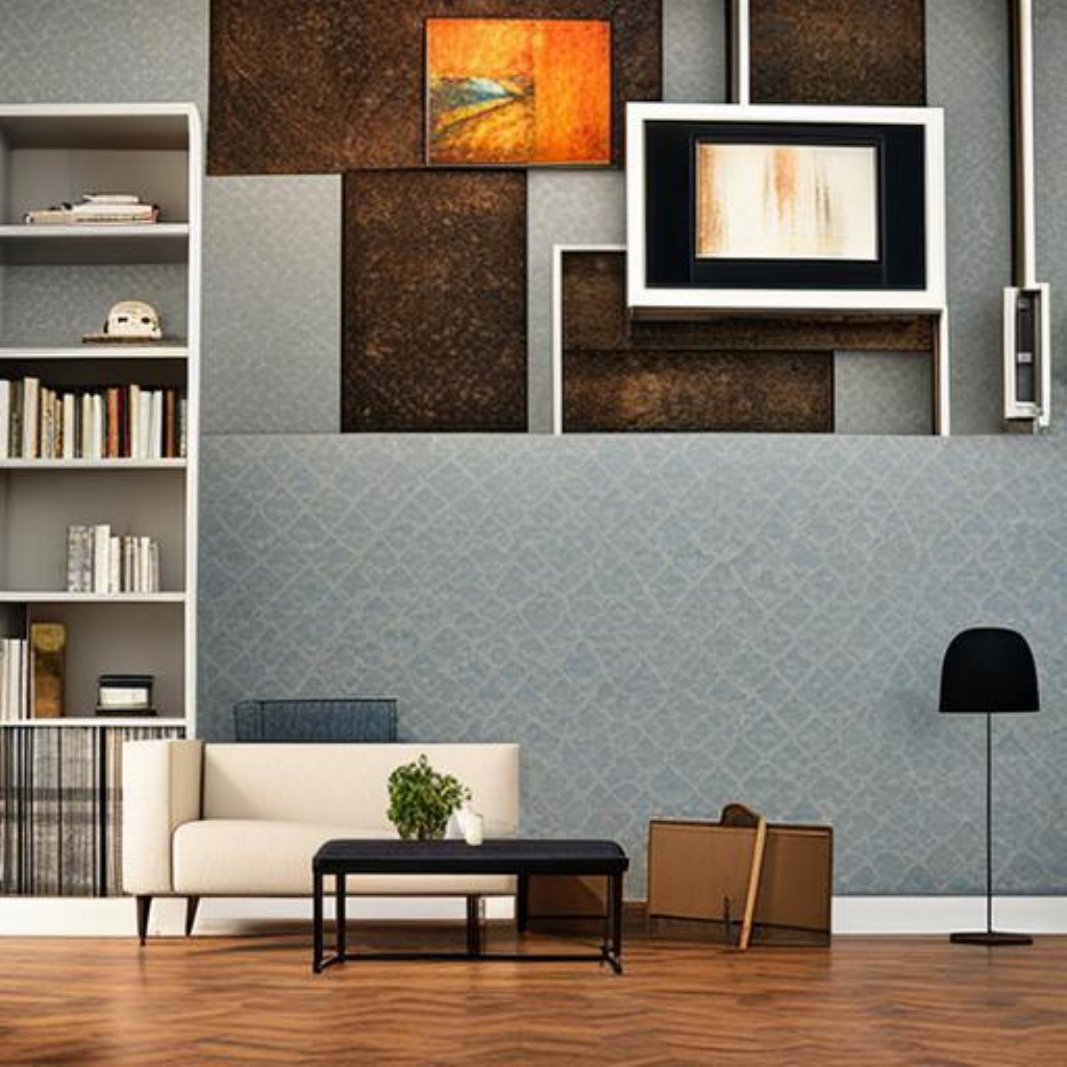} &

        \includegraphics[width=\mainResultsGraphicsWidth]{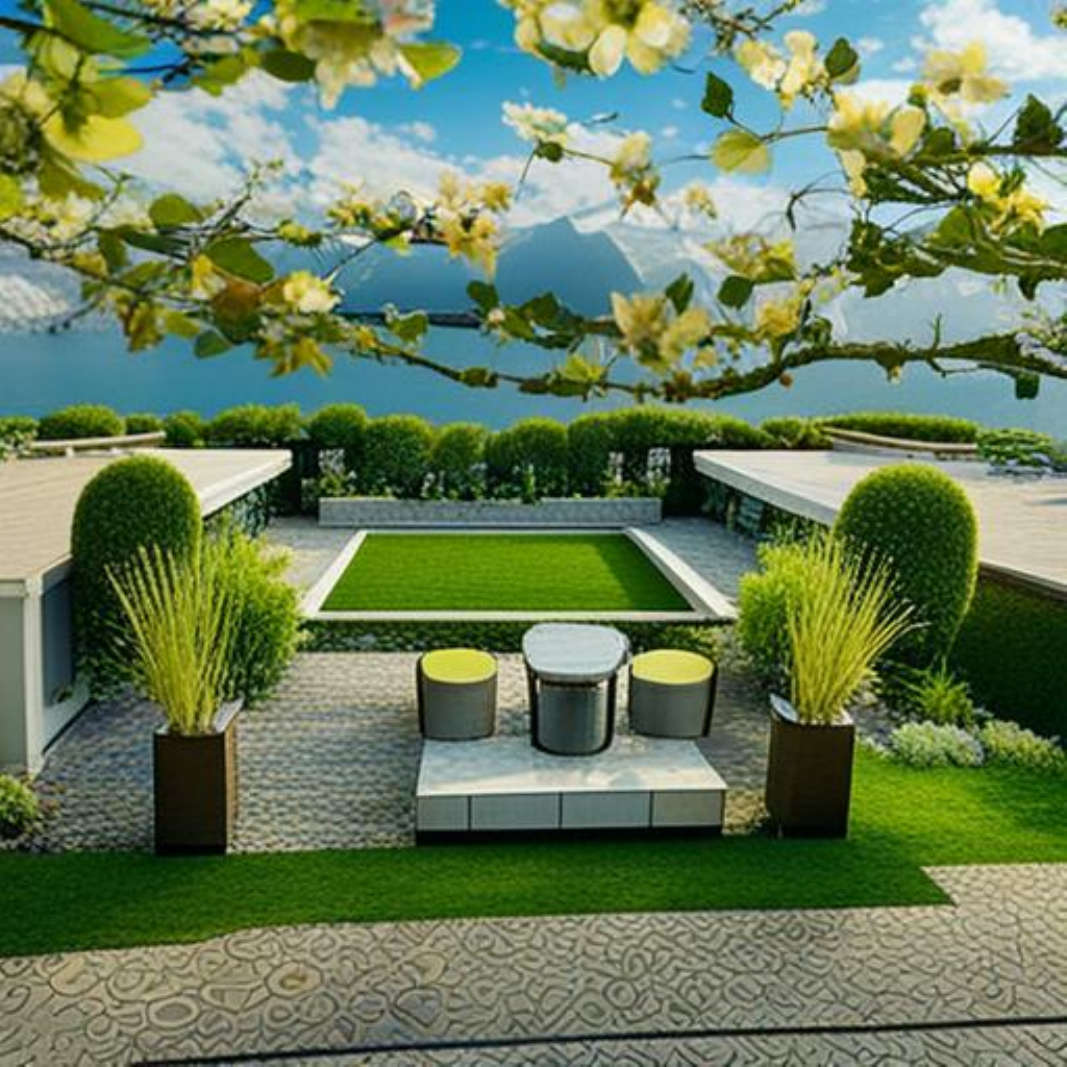} &
        \includegraphics[width=\mainResultsGraphicsWidth]{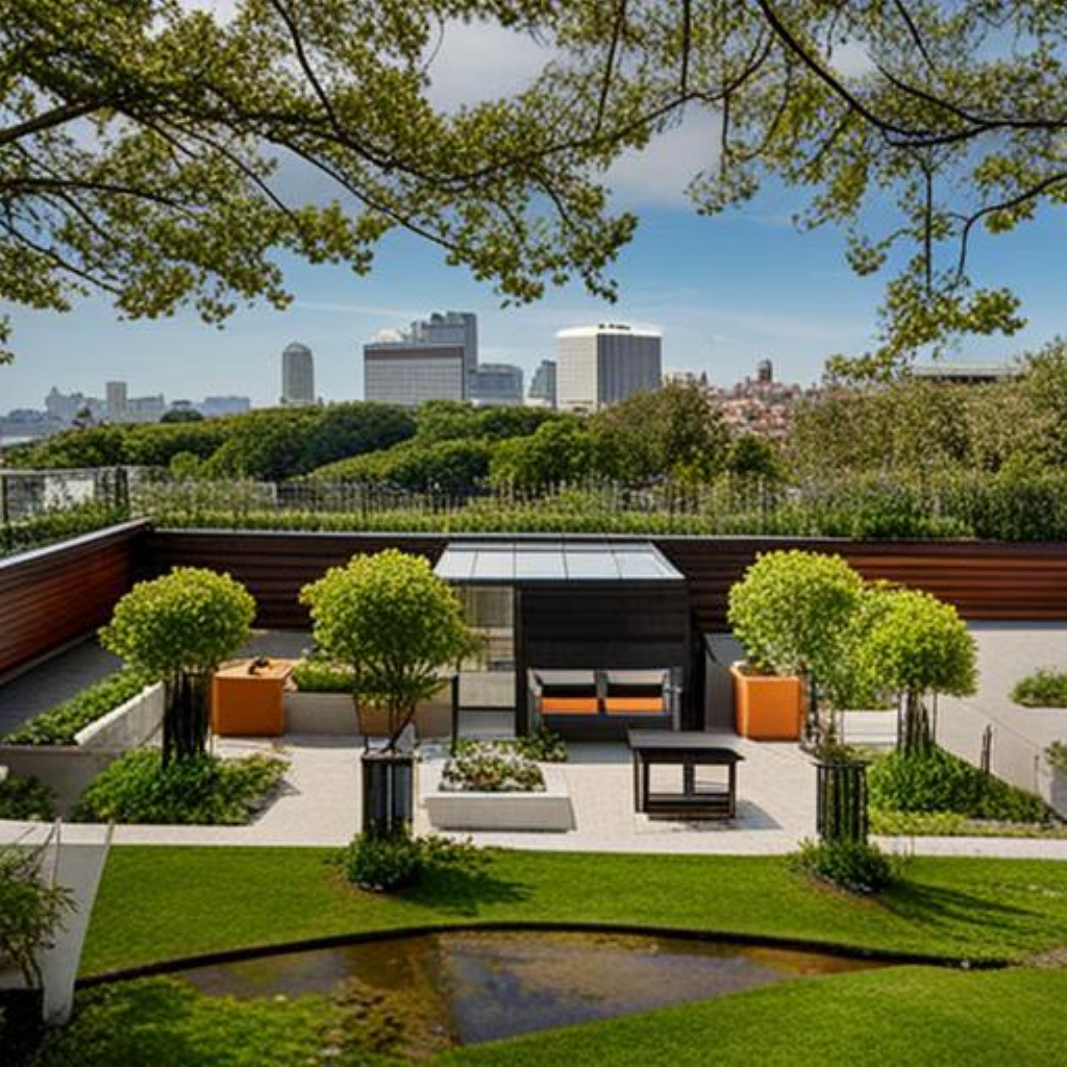} &
        \includegraphics[width=\mainResultsGraphicsWidth]{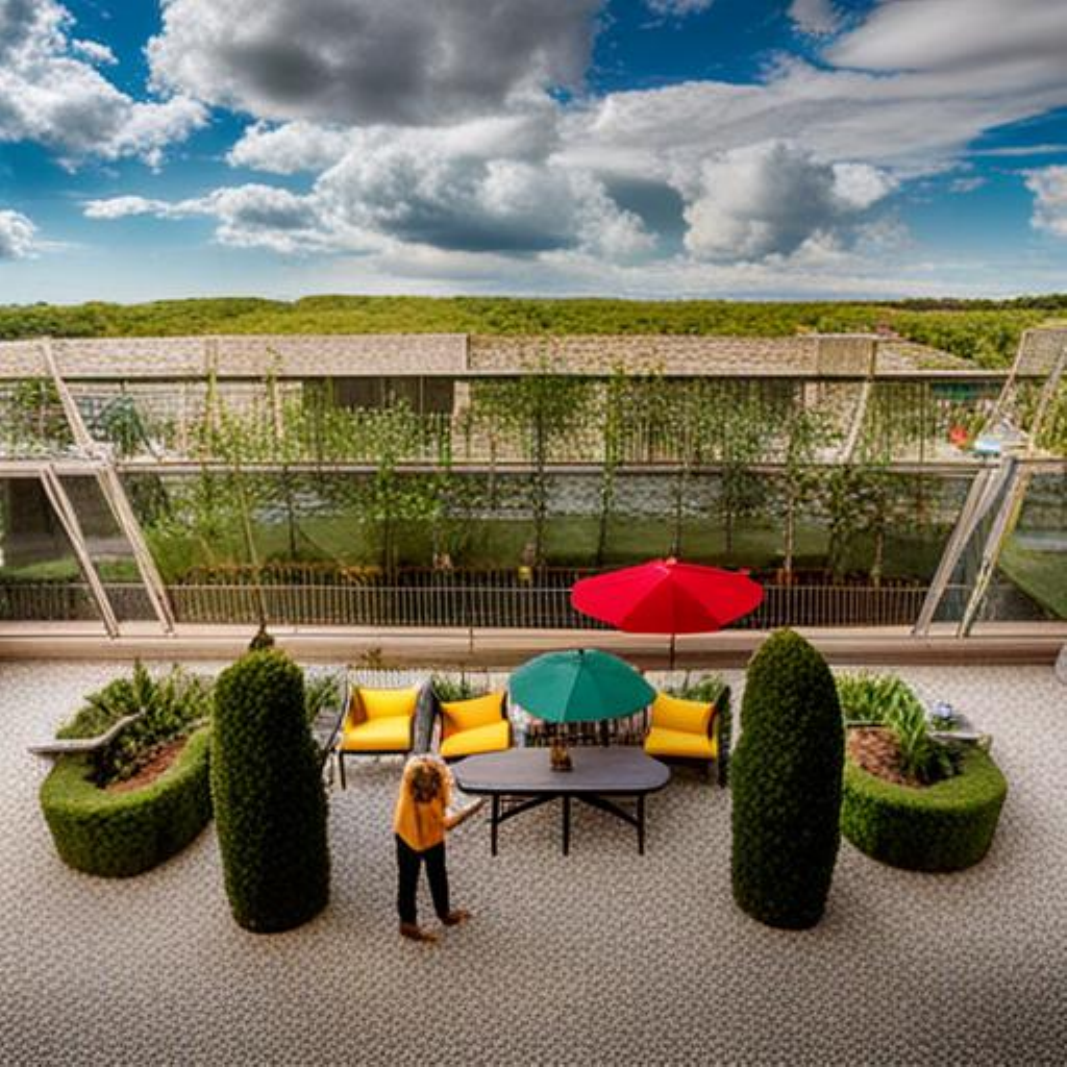} &

        \includegraphics[width=\mainResultsGraphicsWidth]{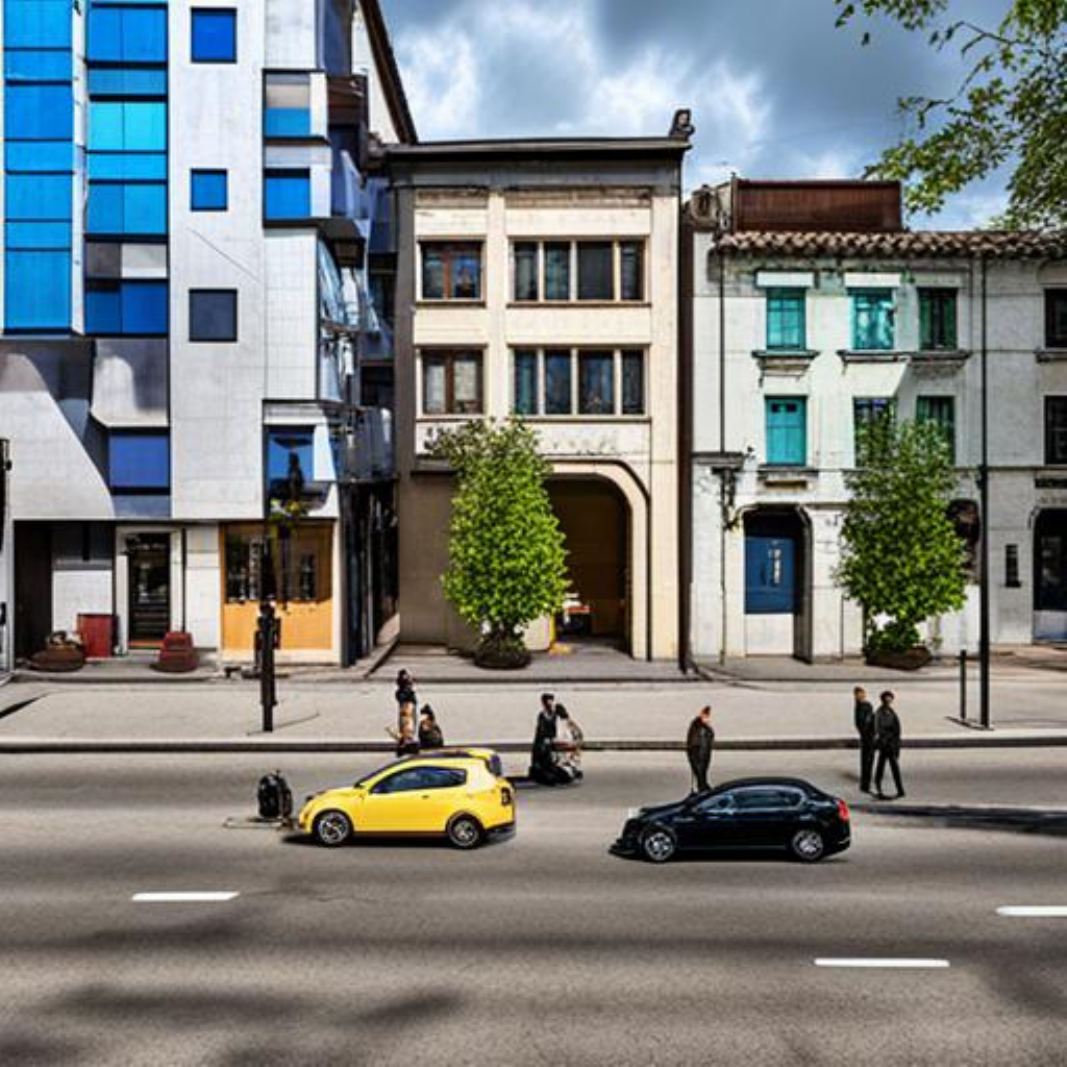} &
        \includegraphics[width=\mainResultsGraphicsWidth]{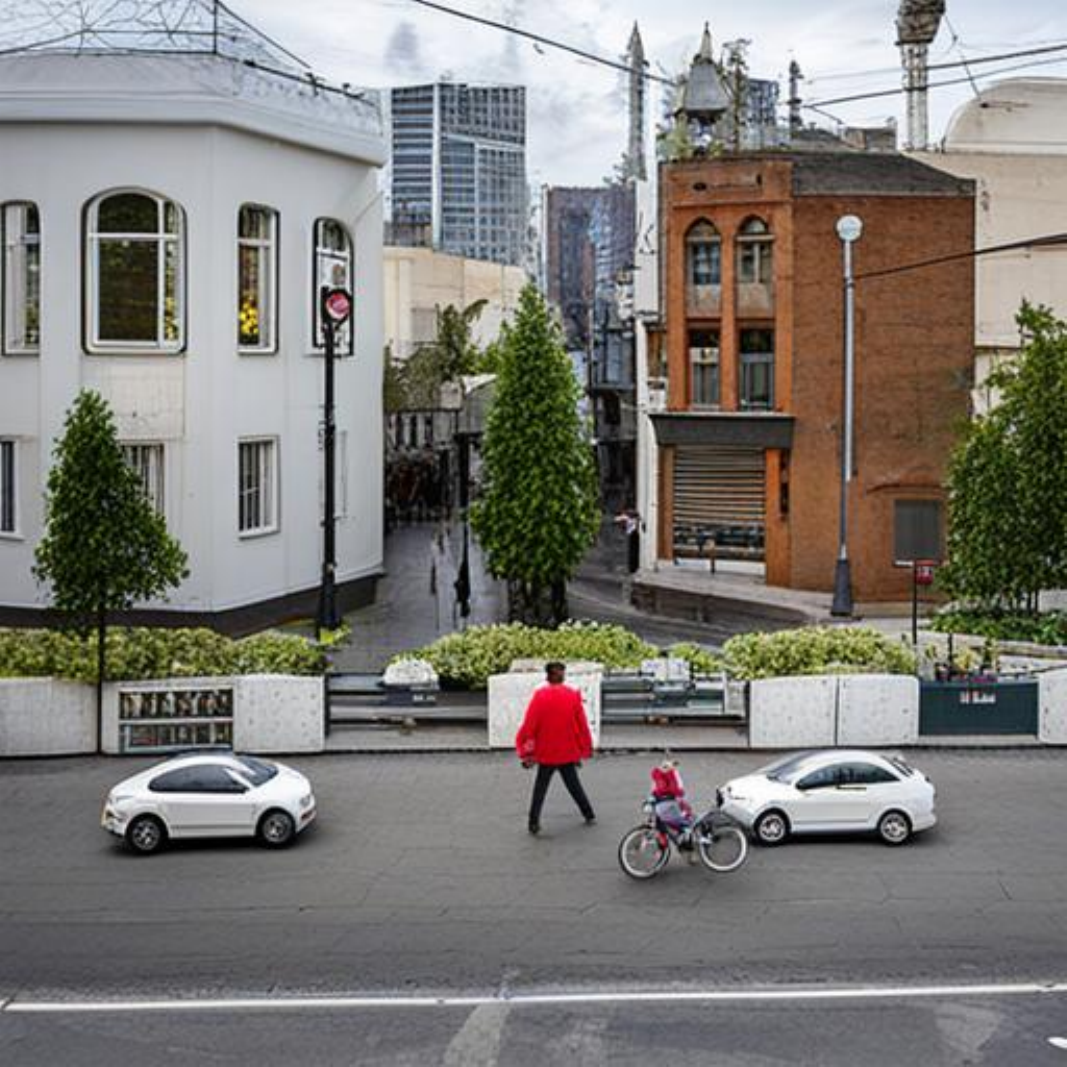} &
        \includegraphics[width=\mainResultsGraphicsWidth]{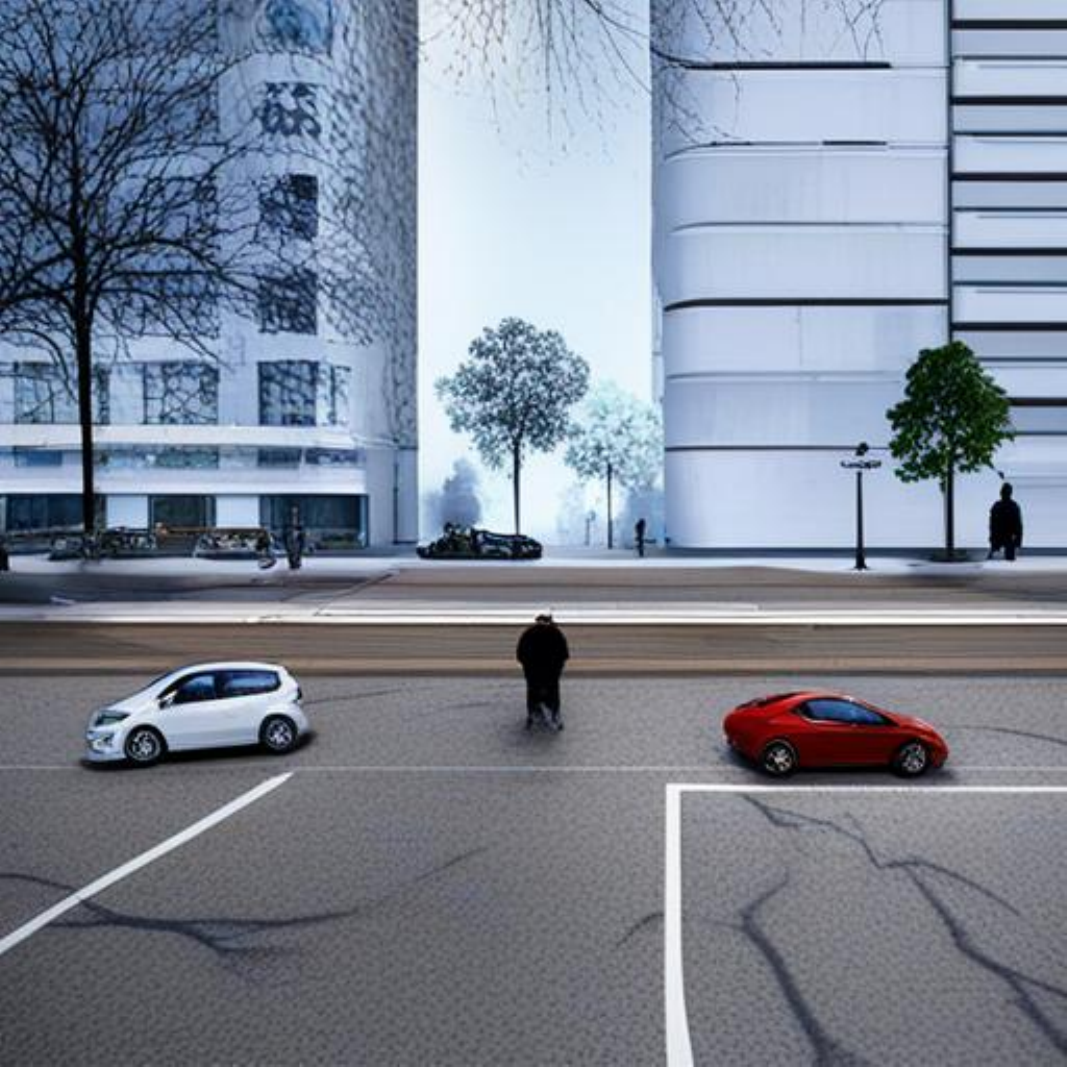} \\

        &
        \includegraphics[width={\mainResultsGraphicsWidth}]{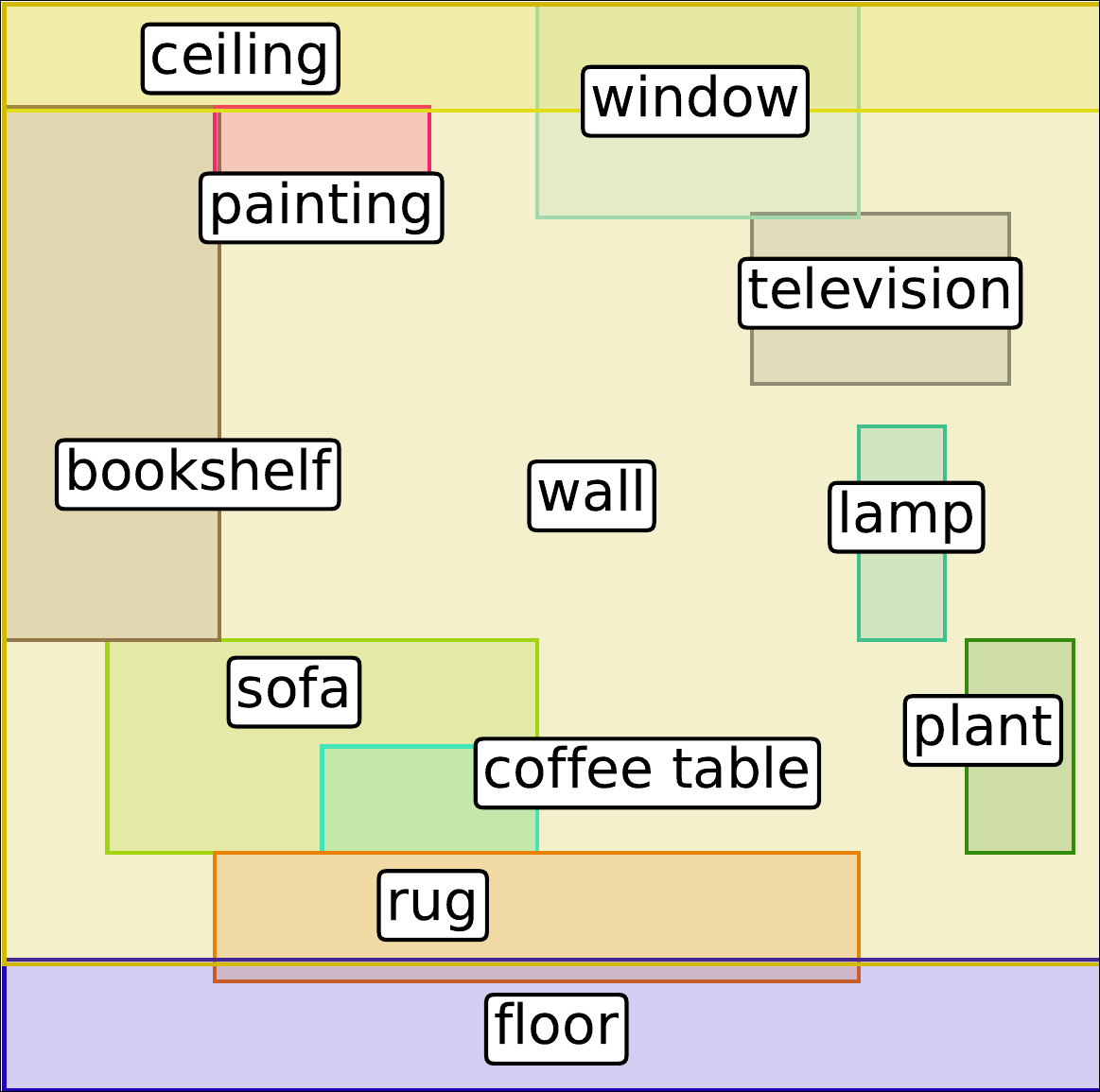} &
        \includegraphics[width={\mainResultsGraphicsWidth}]{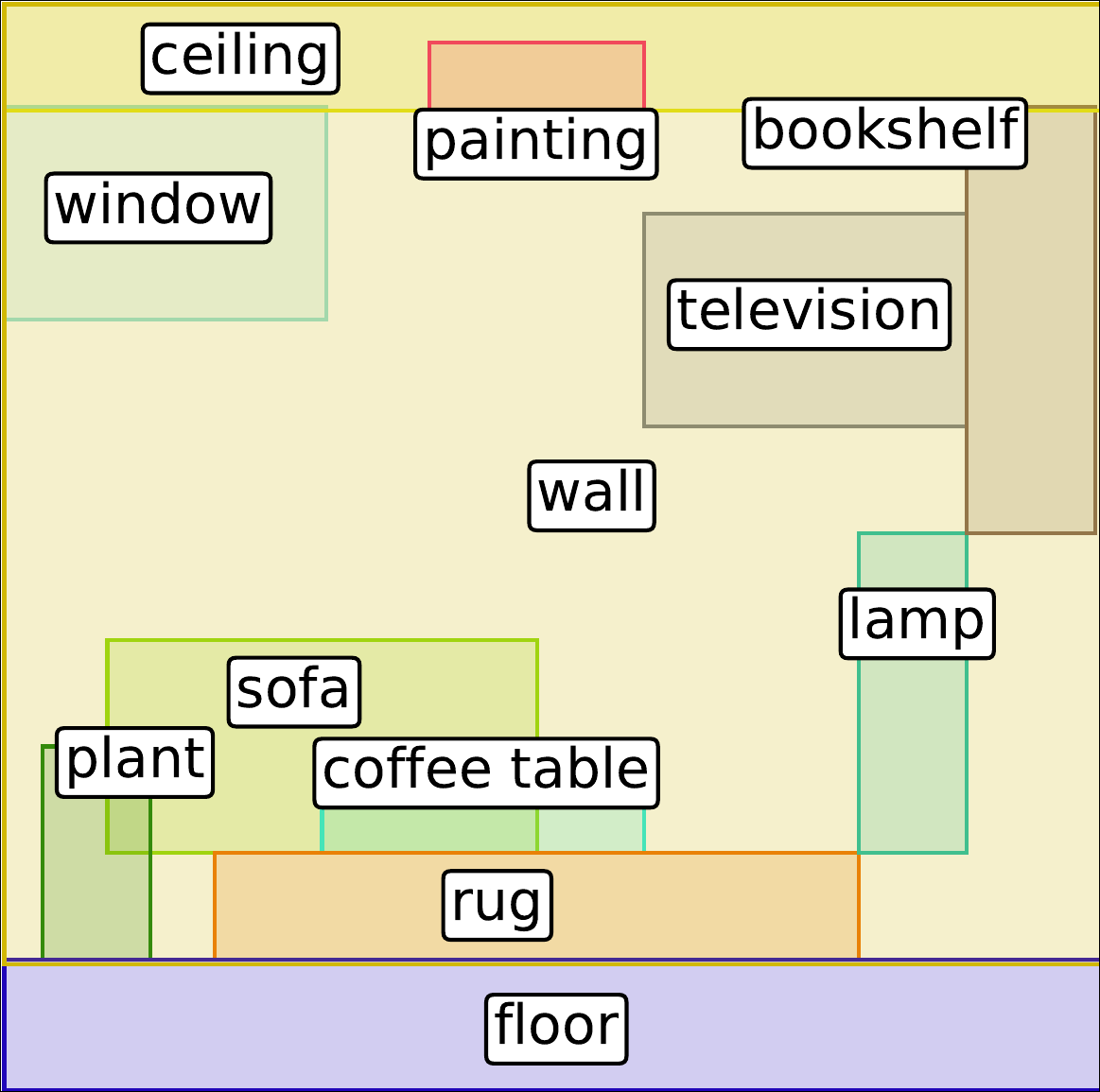} &
        \includegraphics[width={\mainResultsGraphicsWidth}]{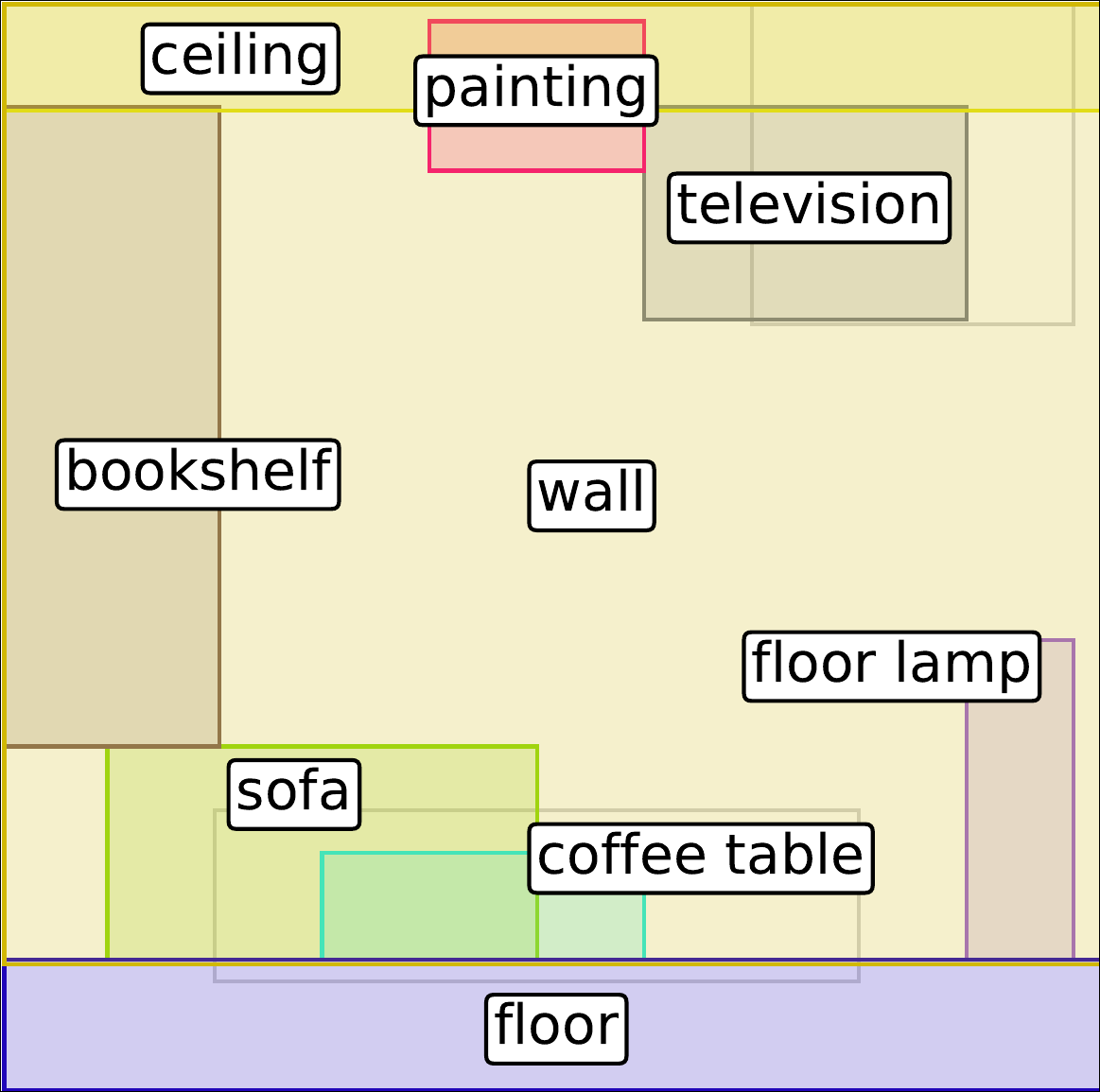} &

        \includegraphics[width={\mainResultsGraphicsWidth}]{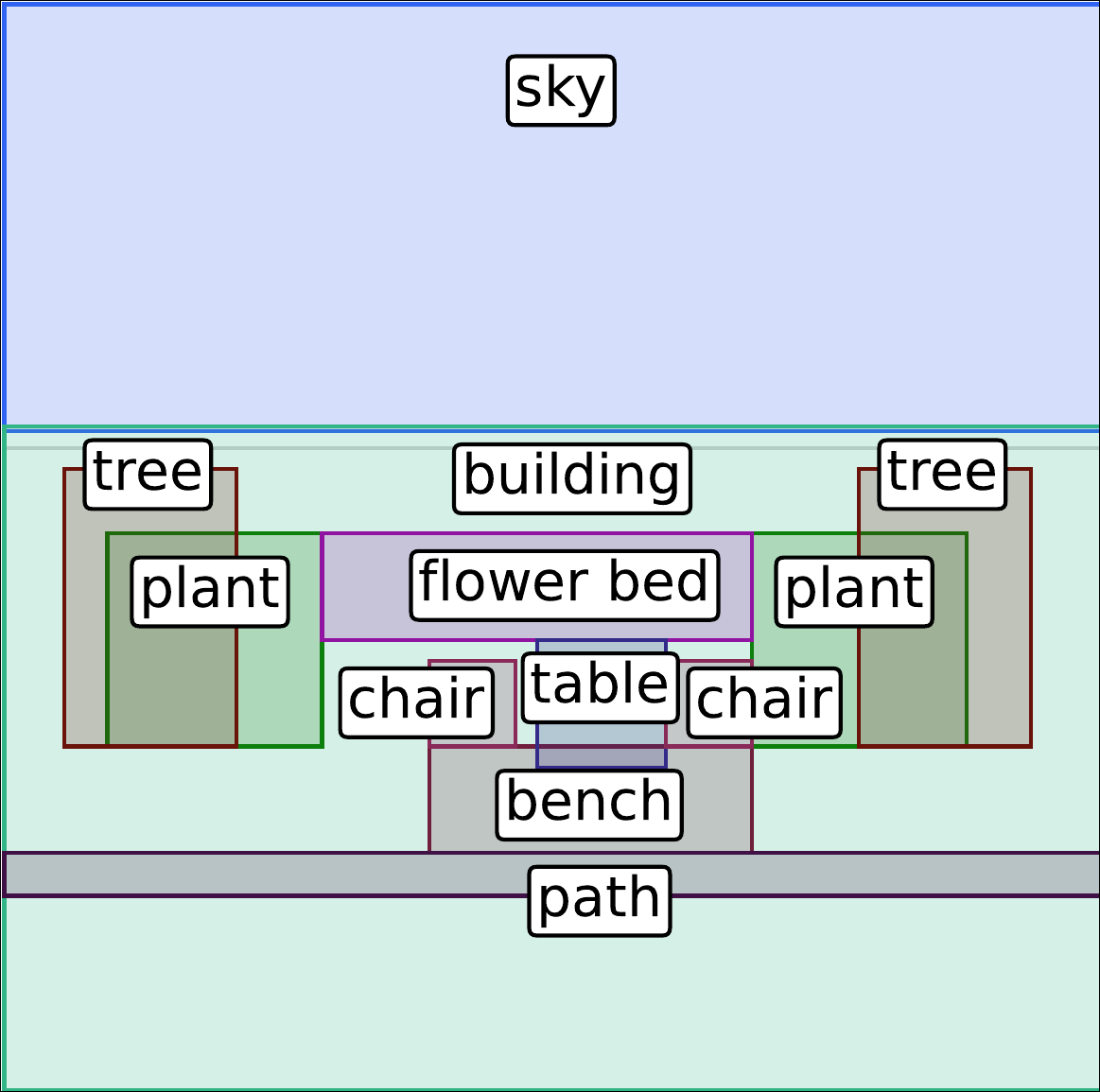} &
        \includegraphics[width={\mainResultsGraphicsWidth}]{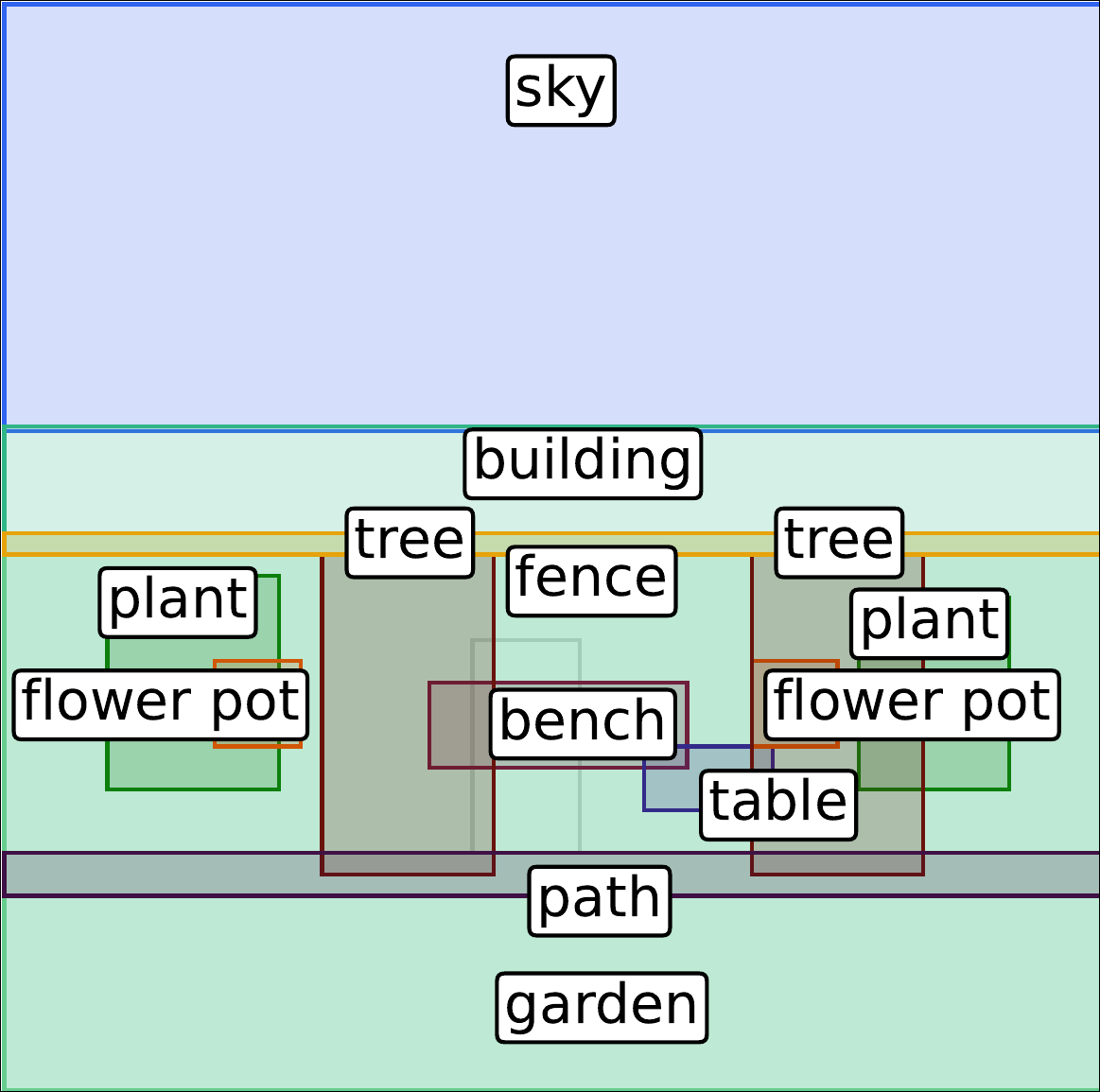} &
        \includegraphics[width={\mainResultsGraphicsWidth}]{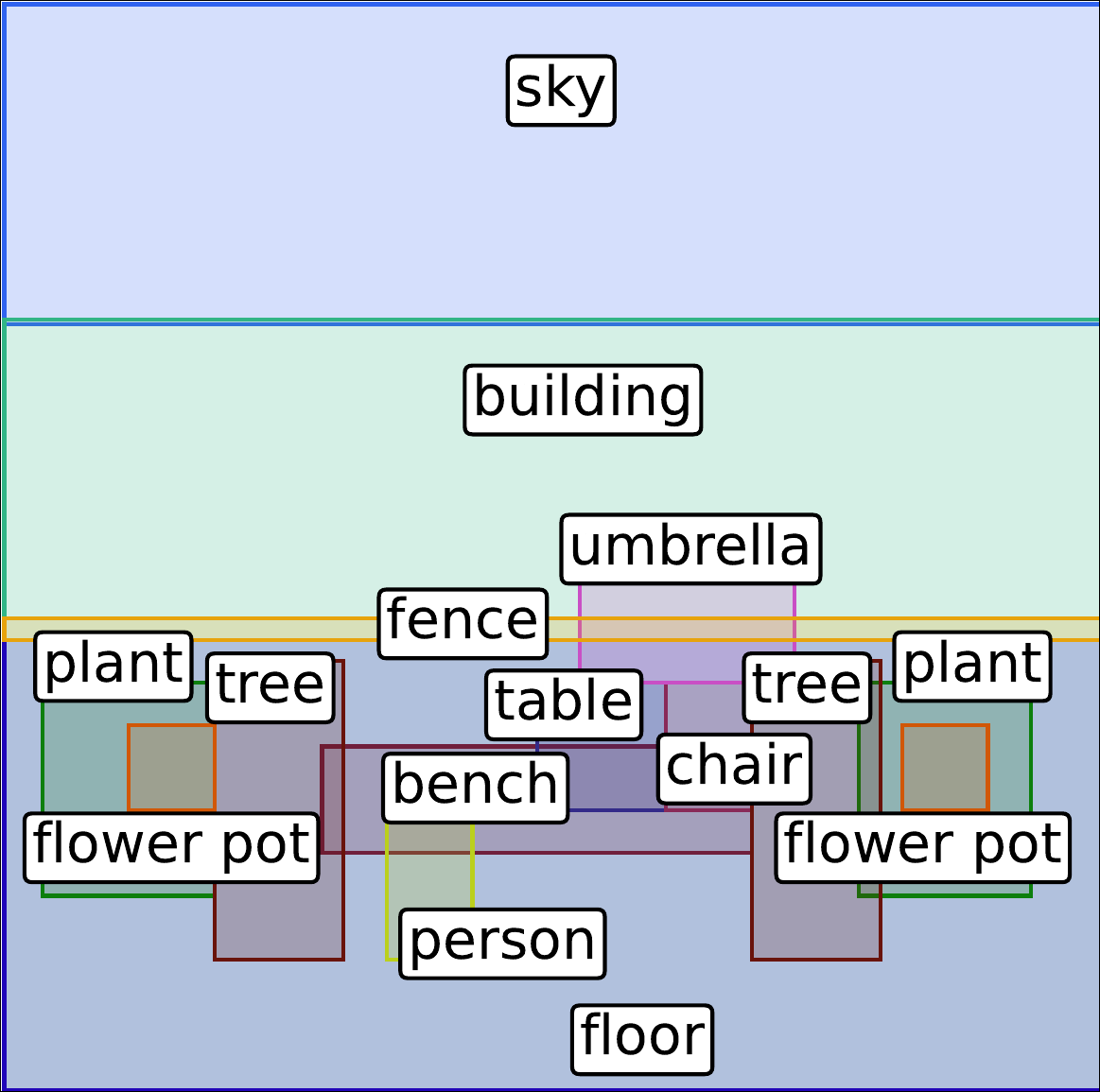} &

        \includegraphics[width={\mainResultsGraphicsWidth}]{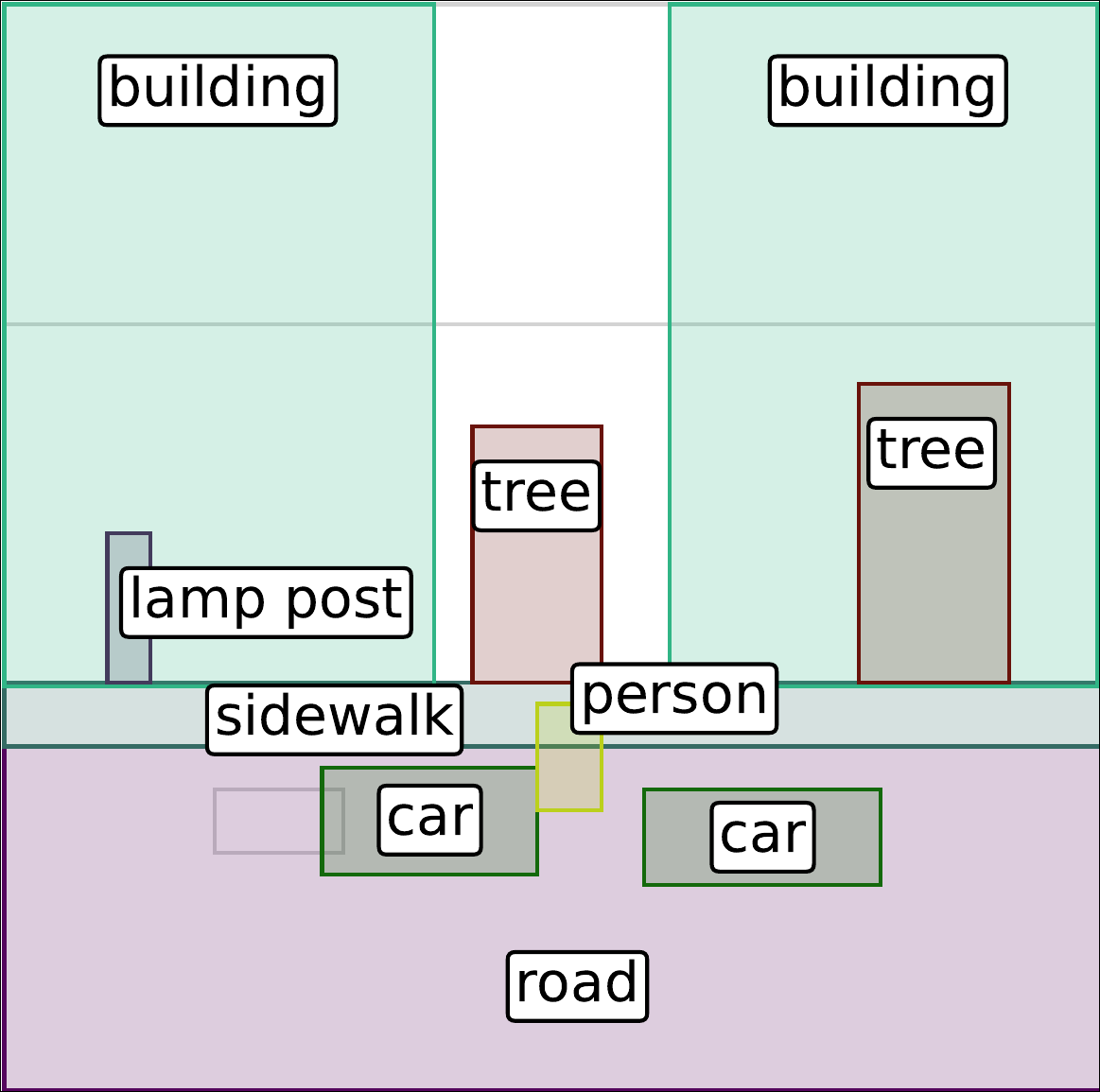} &
        \includegraphics[width={\mainResultsGraphicsWidth}]{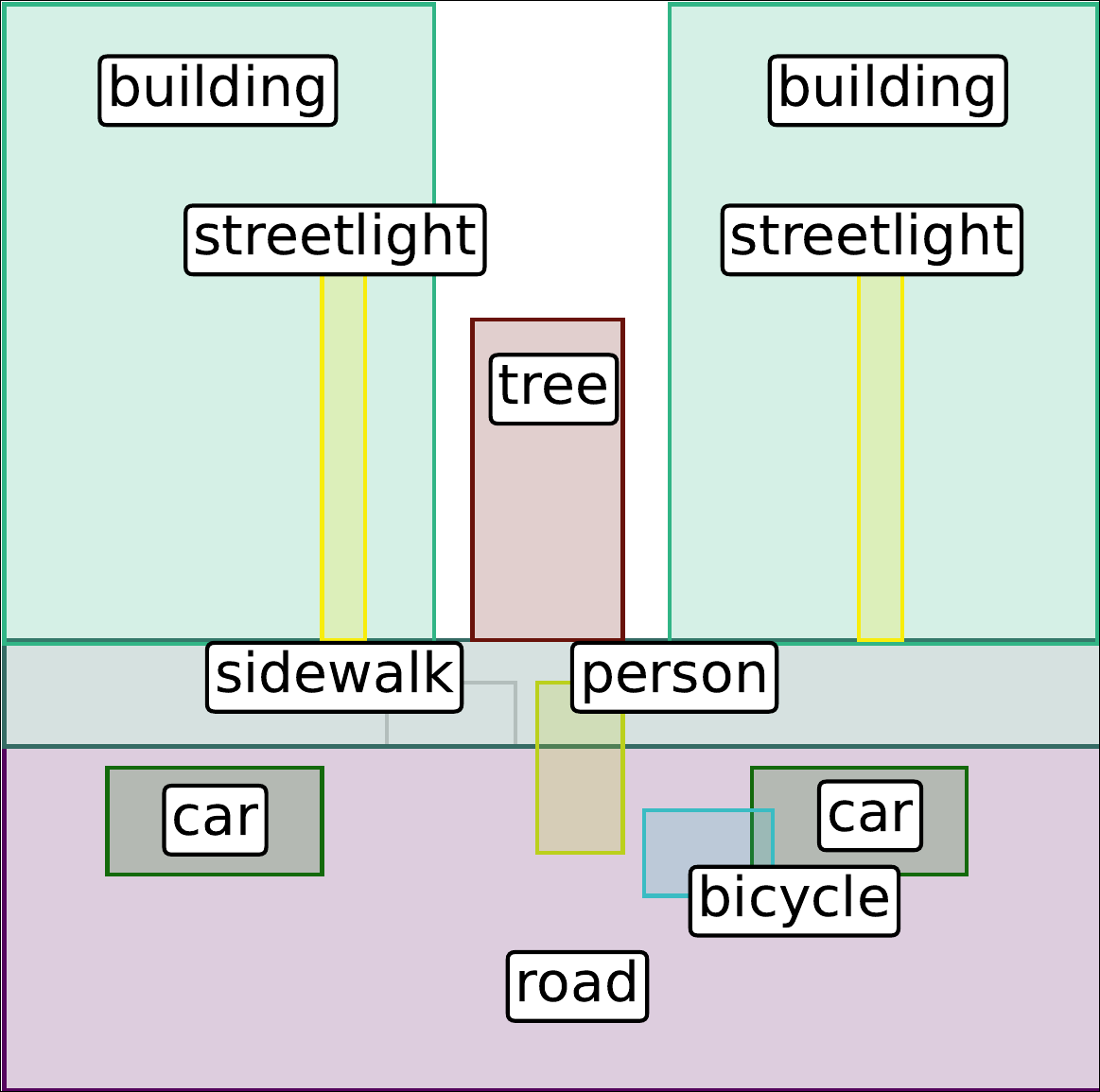} &
        \includegraphics[width={\mainResultsGraphicsWidth}]{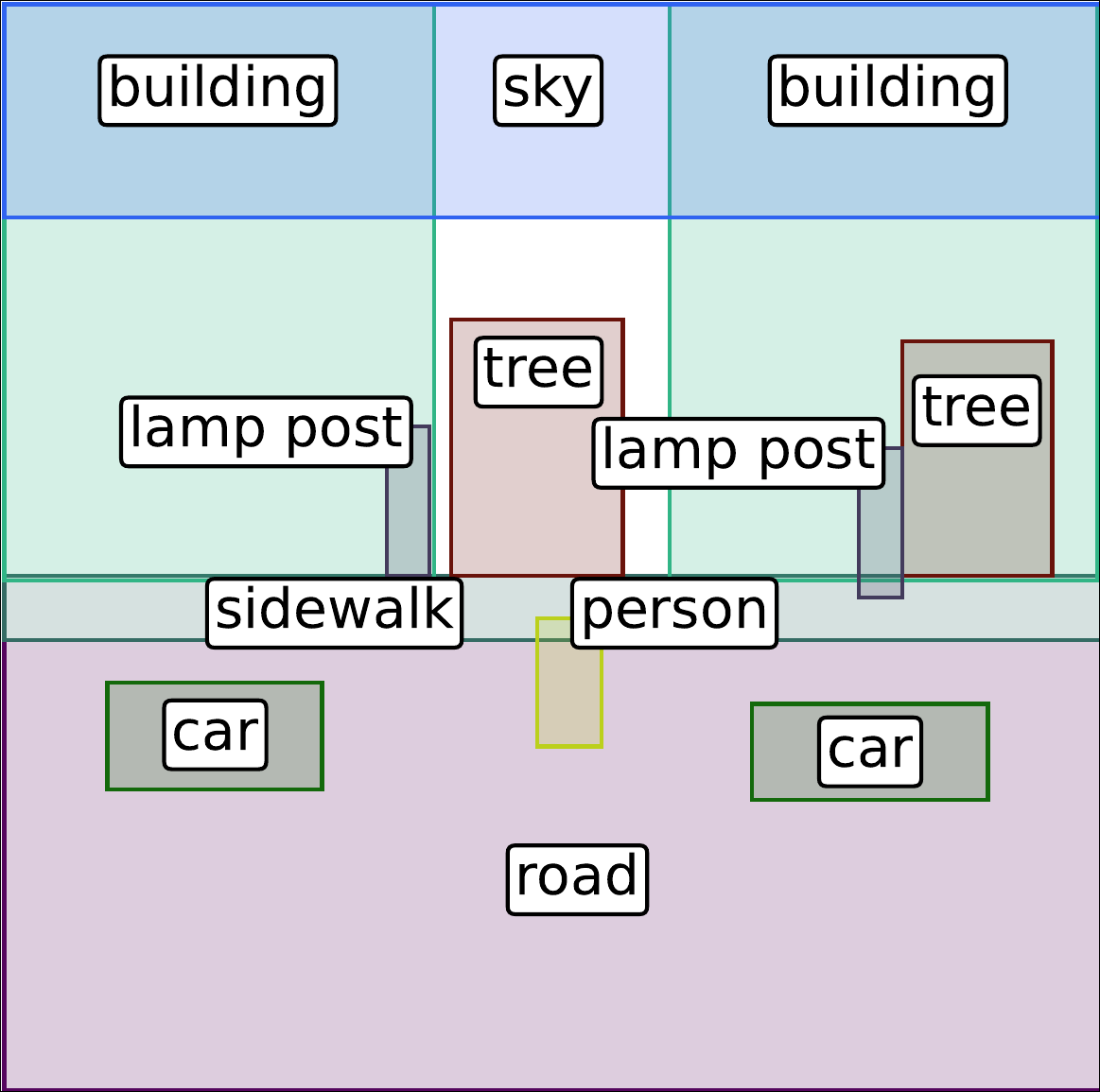} \\

        \midrule

        \multirow{-3.5}{*}{\rotatebox[origin=c]{90}{LayoutTransformer}}&
        \includegraphics[width={\mainResultsGraphicsWidth}]{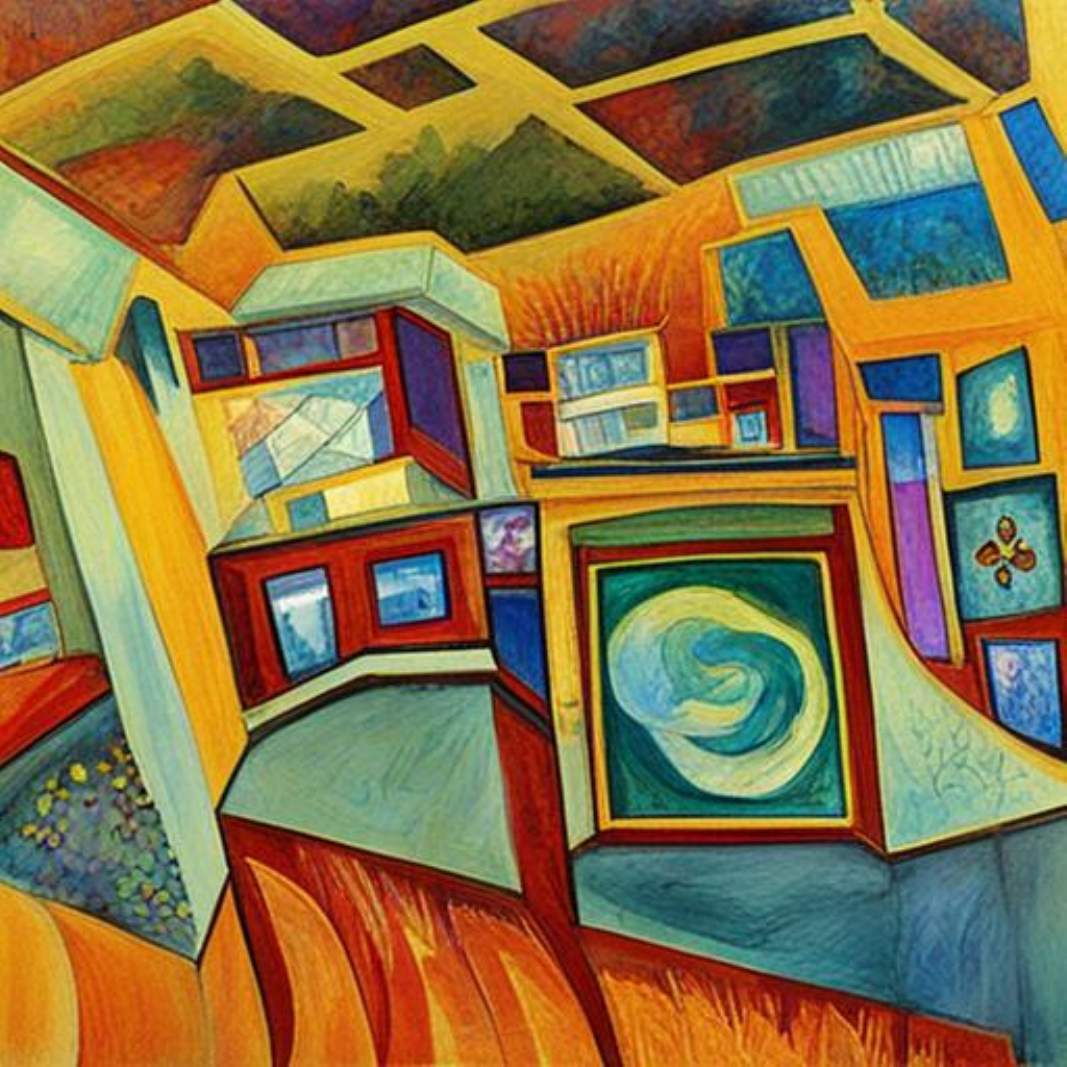} &
        \includegraphics[width={\mainResultsGraphicsWidth}]{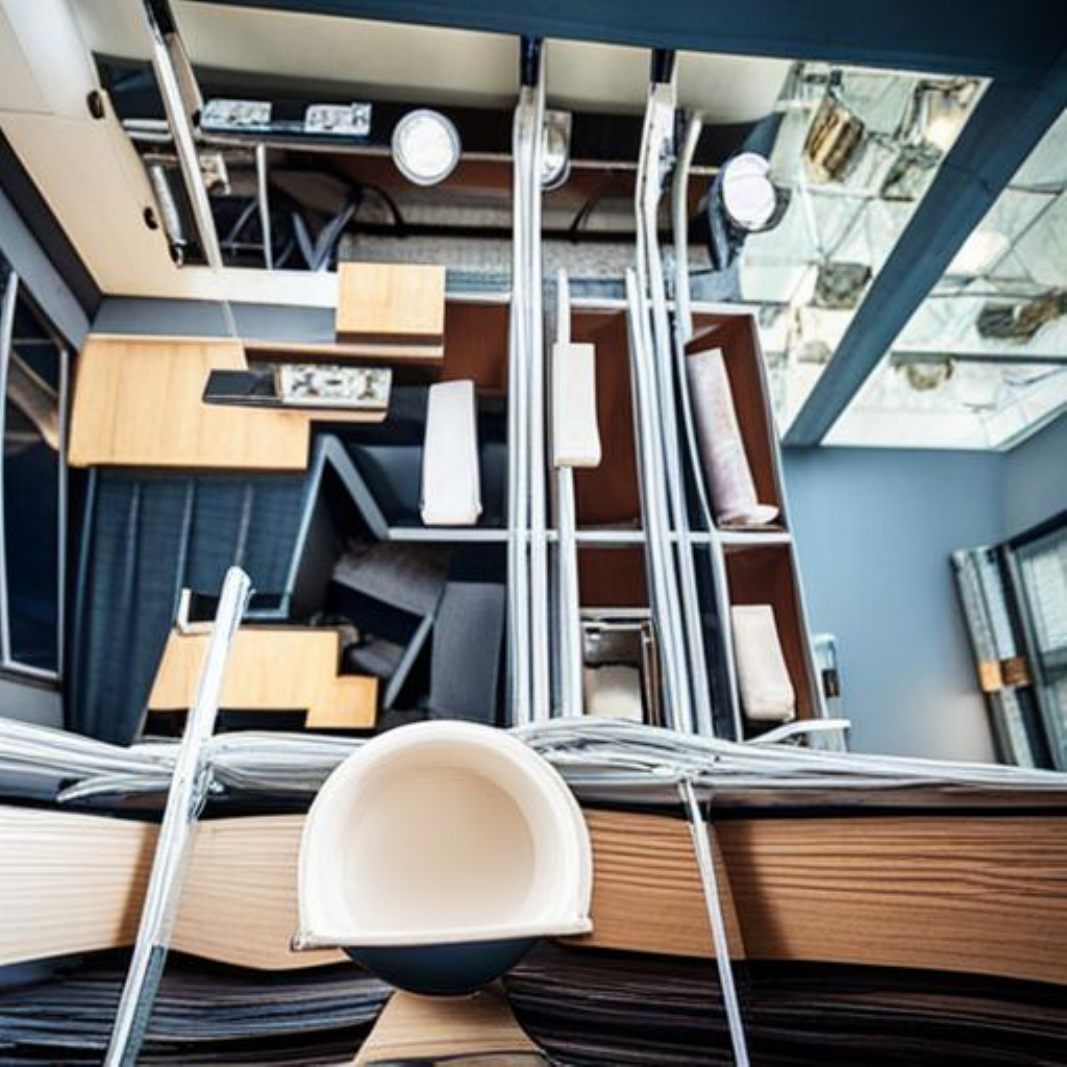} &
        \includegraphics[width={\mainResultsGraphicsWidth}]{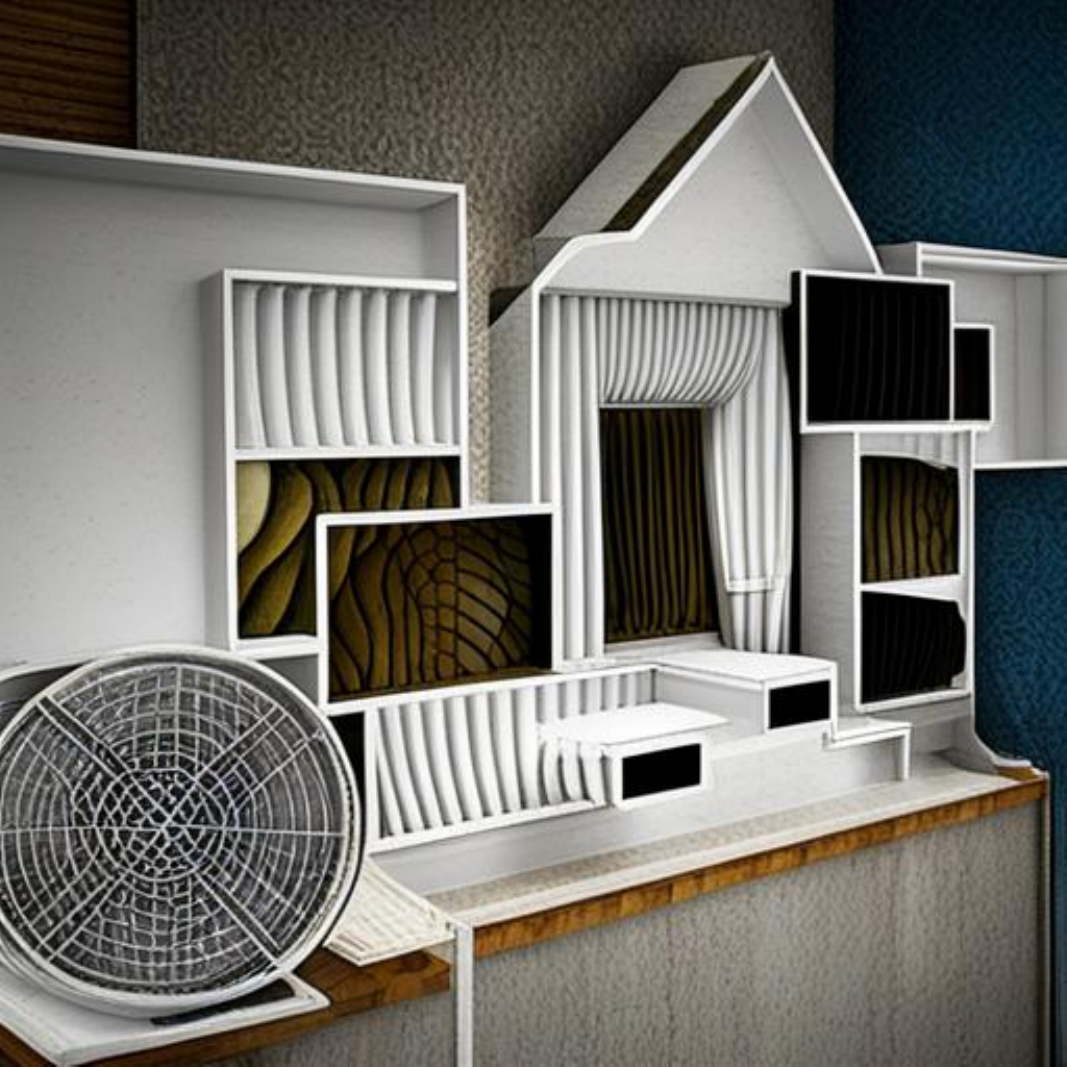} &

        \includegraphics[width={\mainResultsGraphicsWidth}]{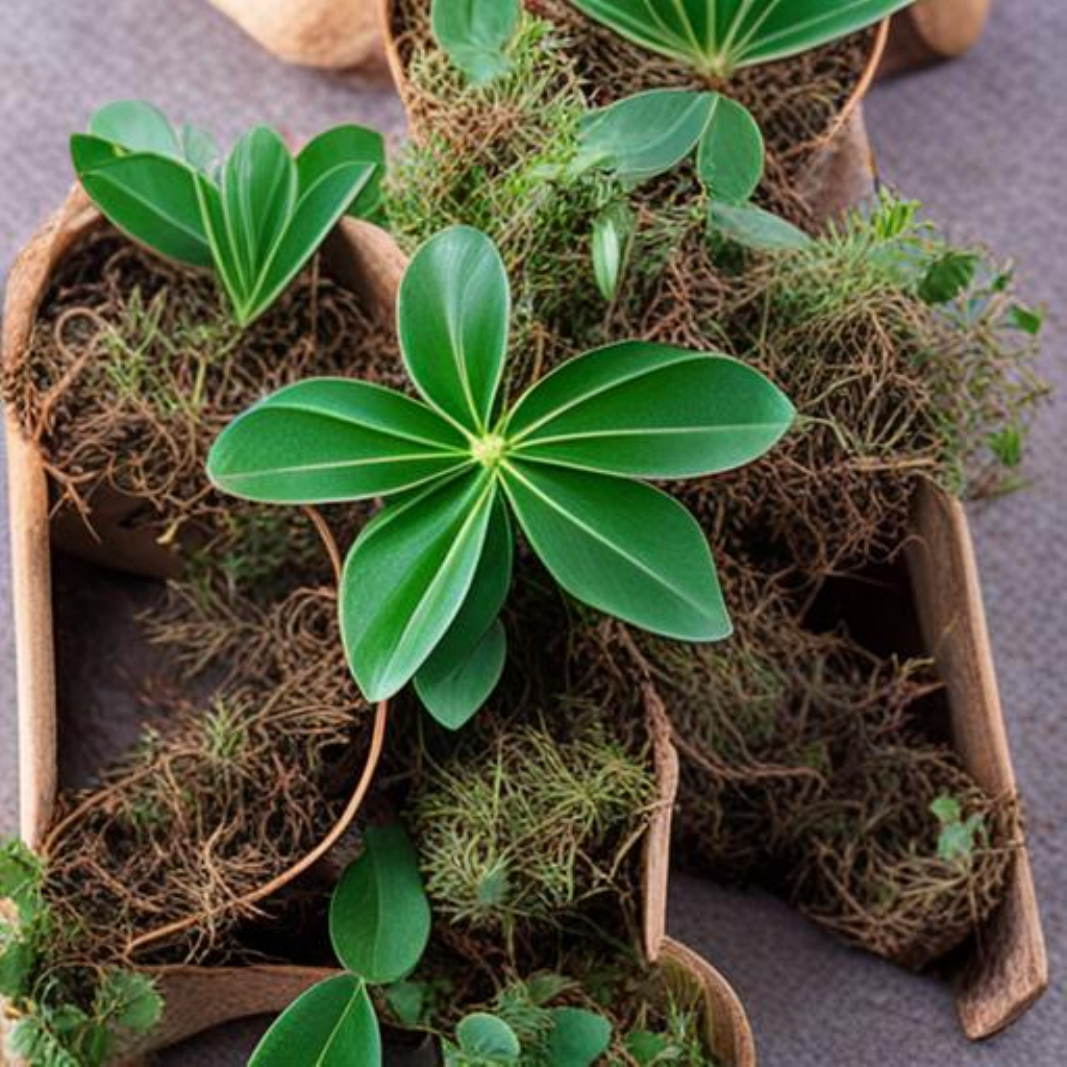} &
        \includegraphics[width={\mainResultsGraphicsWidth}]{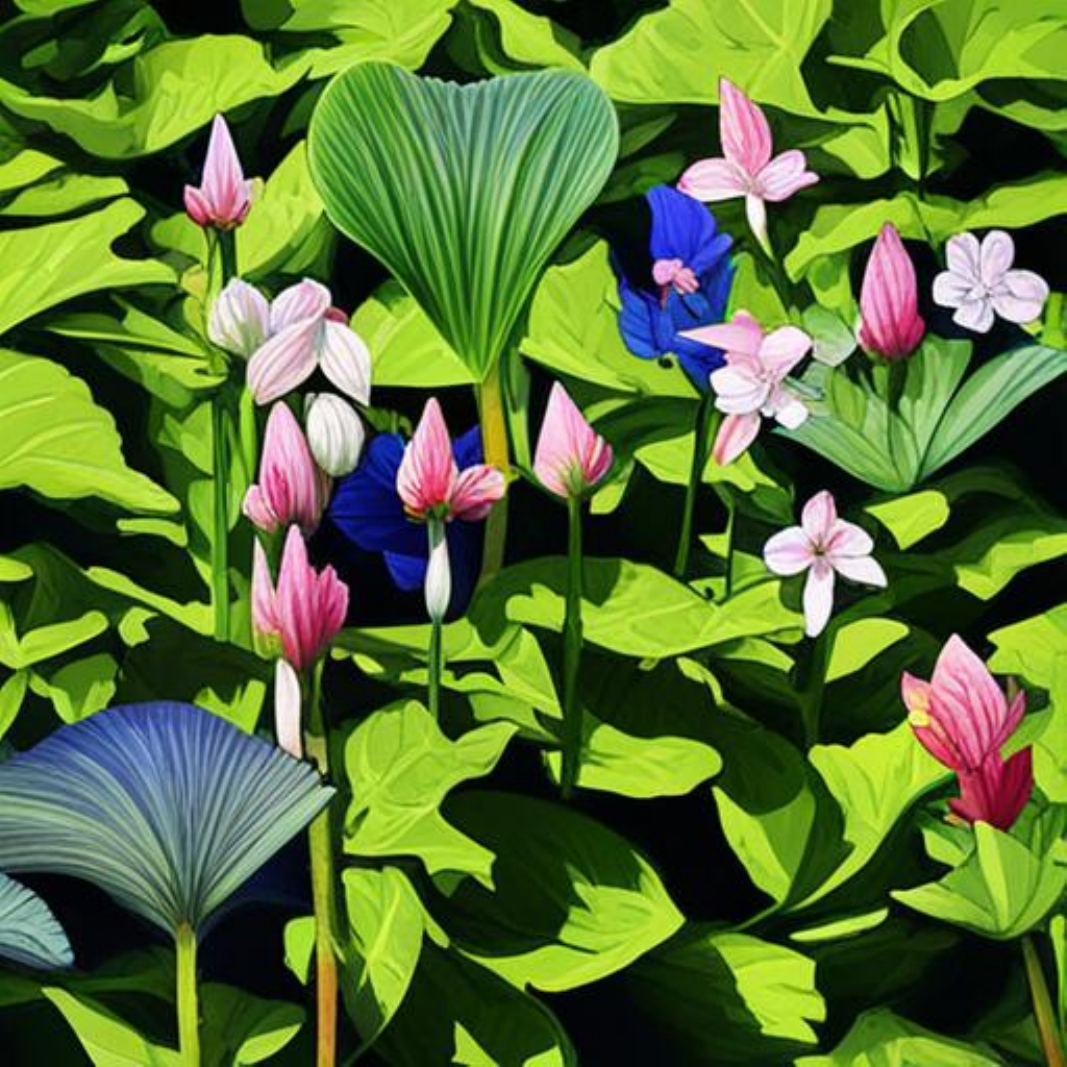} &
        \includegraphics[width={\mainResultsGraphicsWidth}]{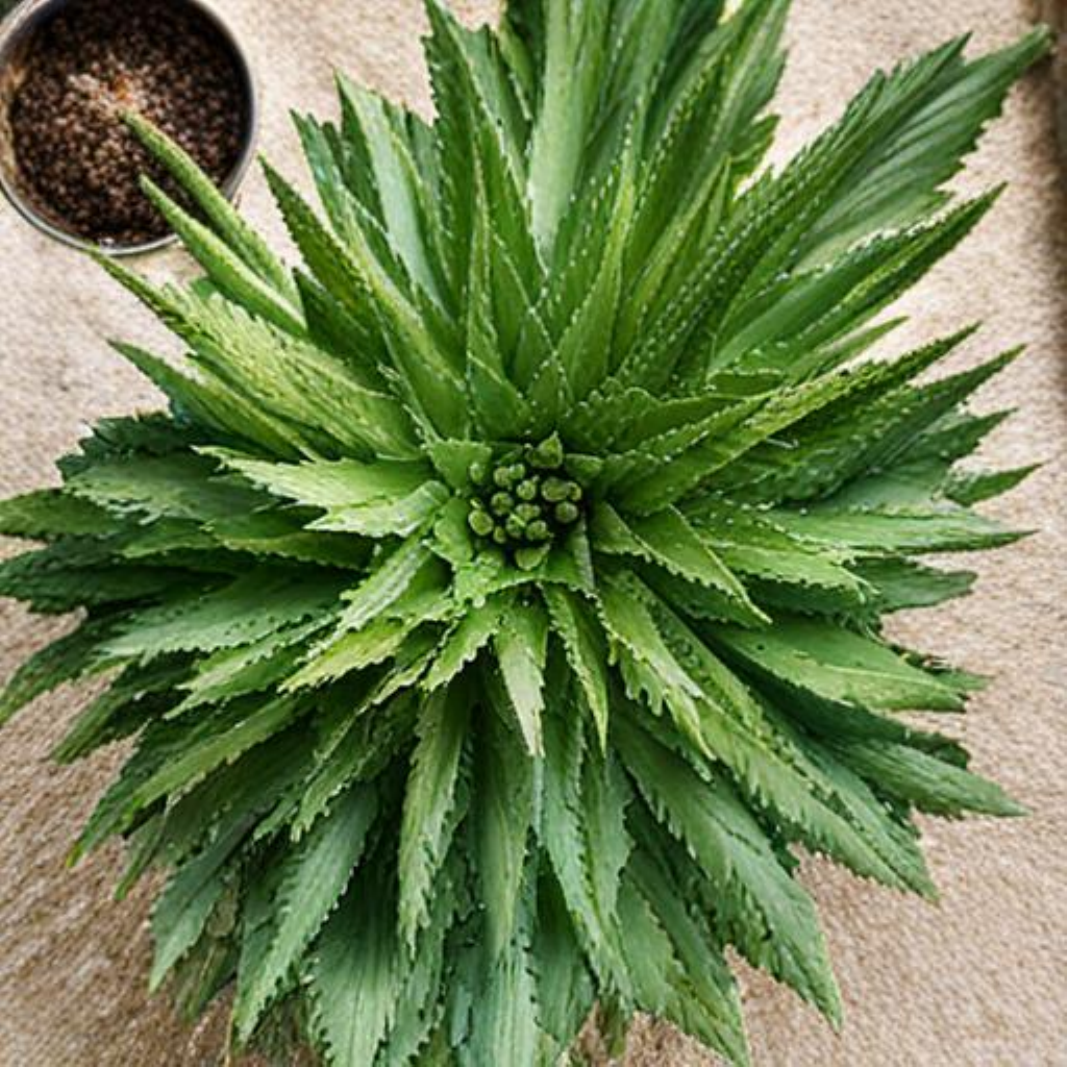} &

        \includegraphics[width={\mainResultsGraphicsWidth}]{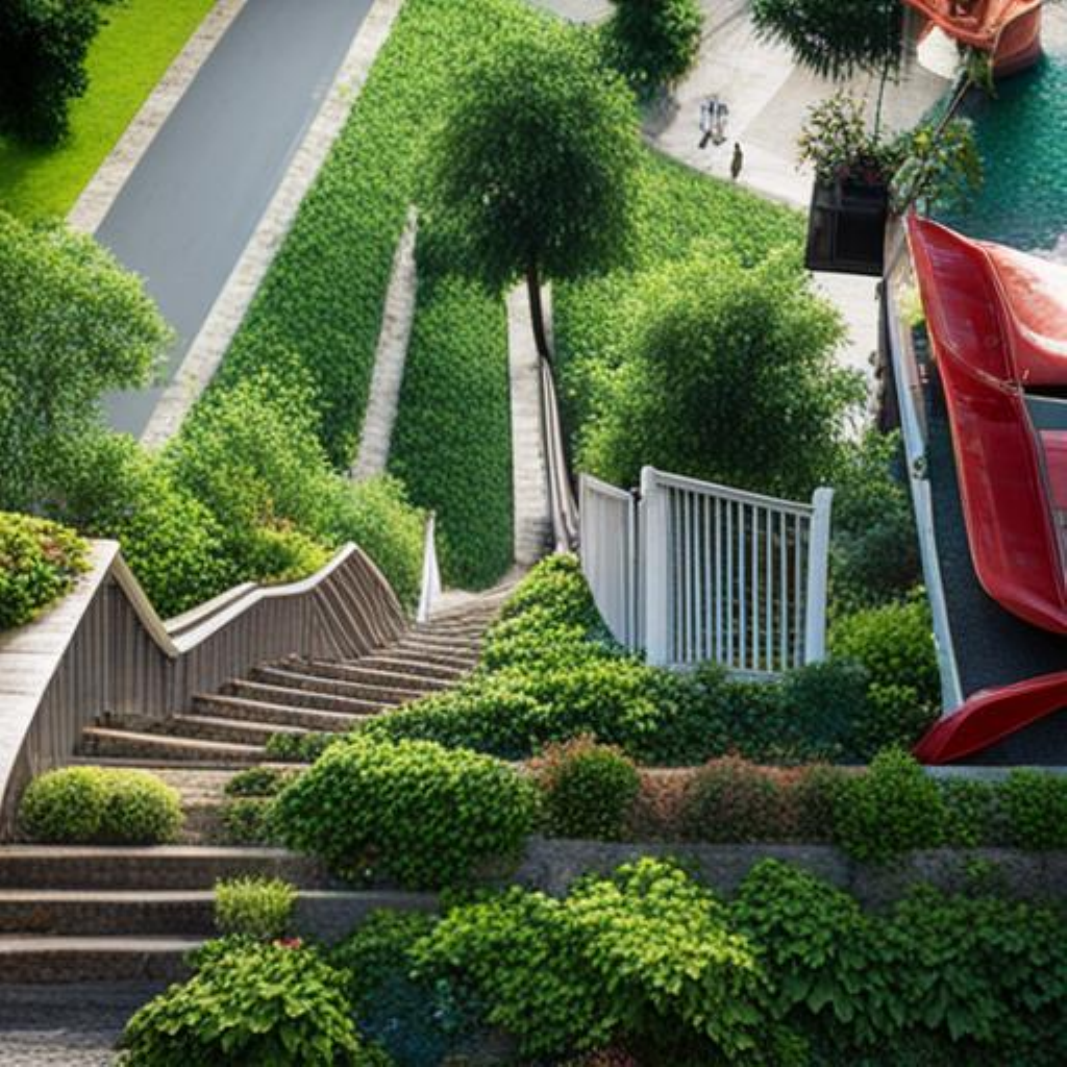} &
        \includegraphics[width={\mainResultsGraphicsWidth}]{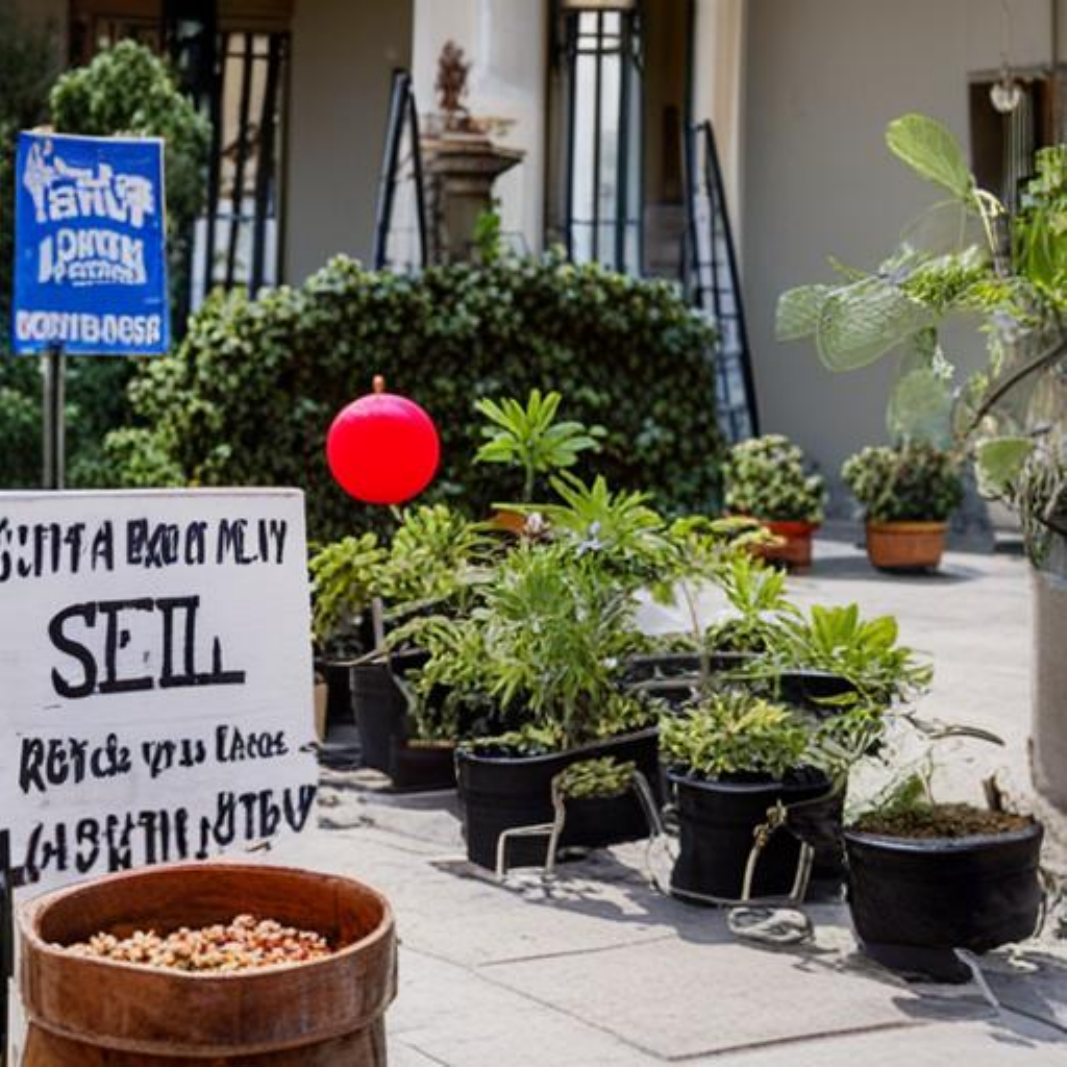} &
        \includegraphics[width={\mainResultsGraphicsWidth}]{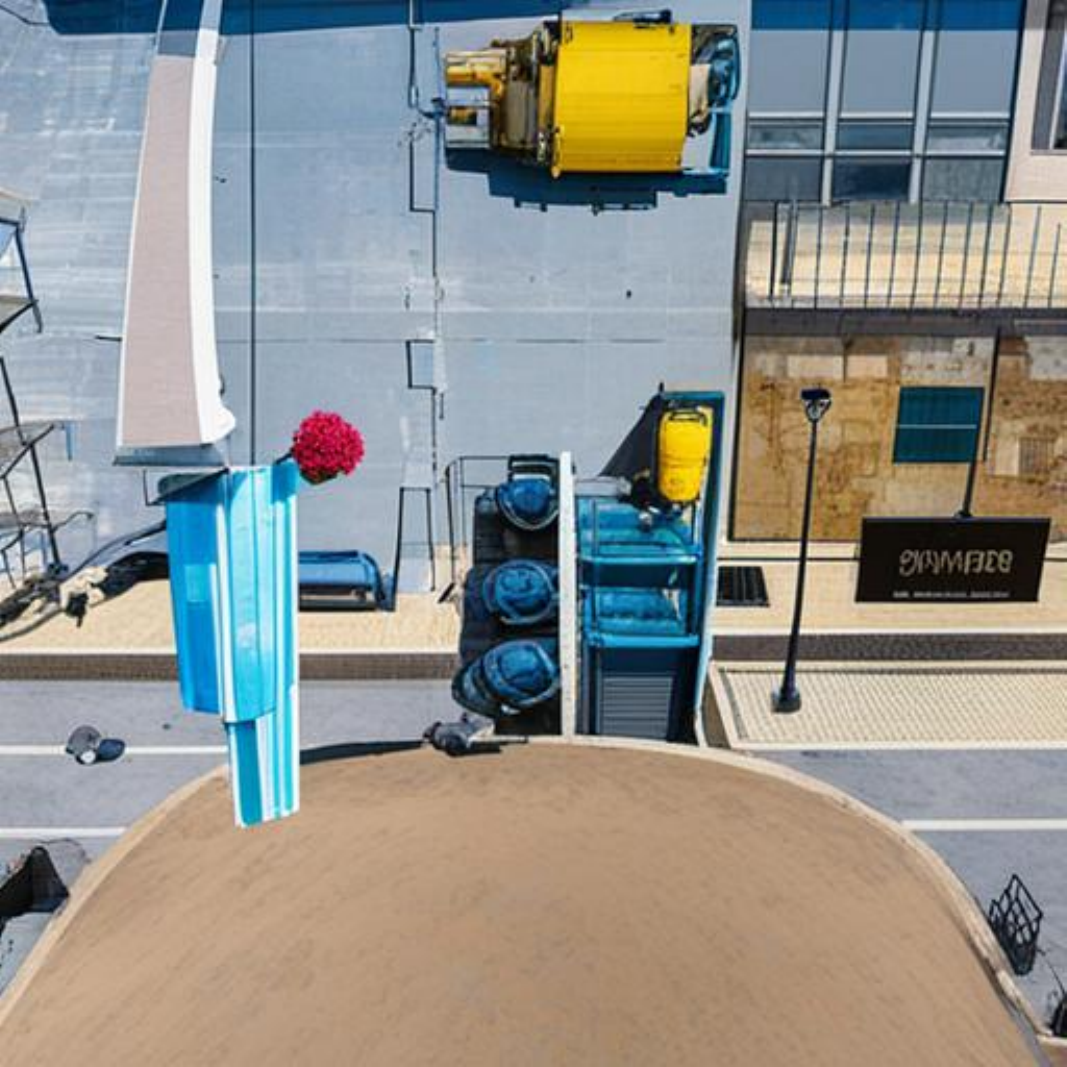} \\

        &
        \includegraphics[width={\mainResultsGraphicsWidth}]{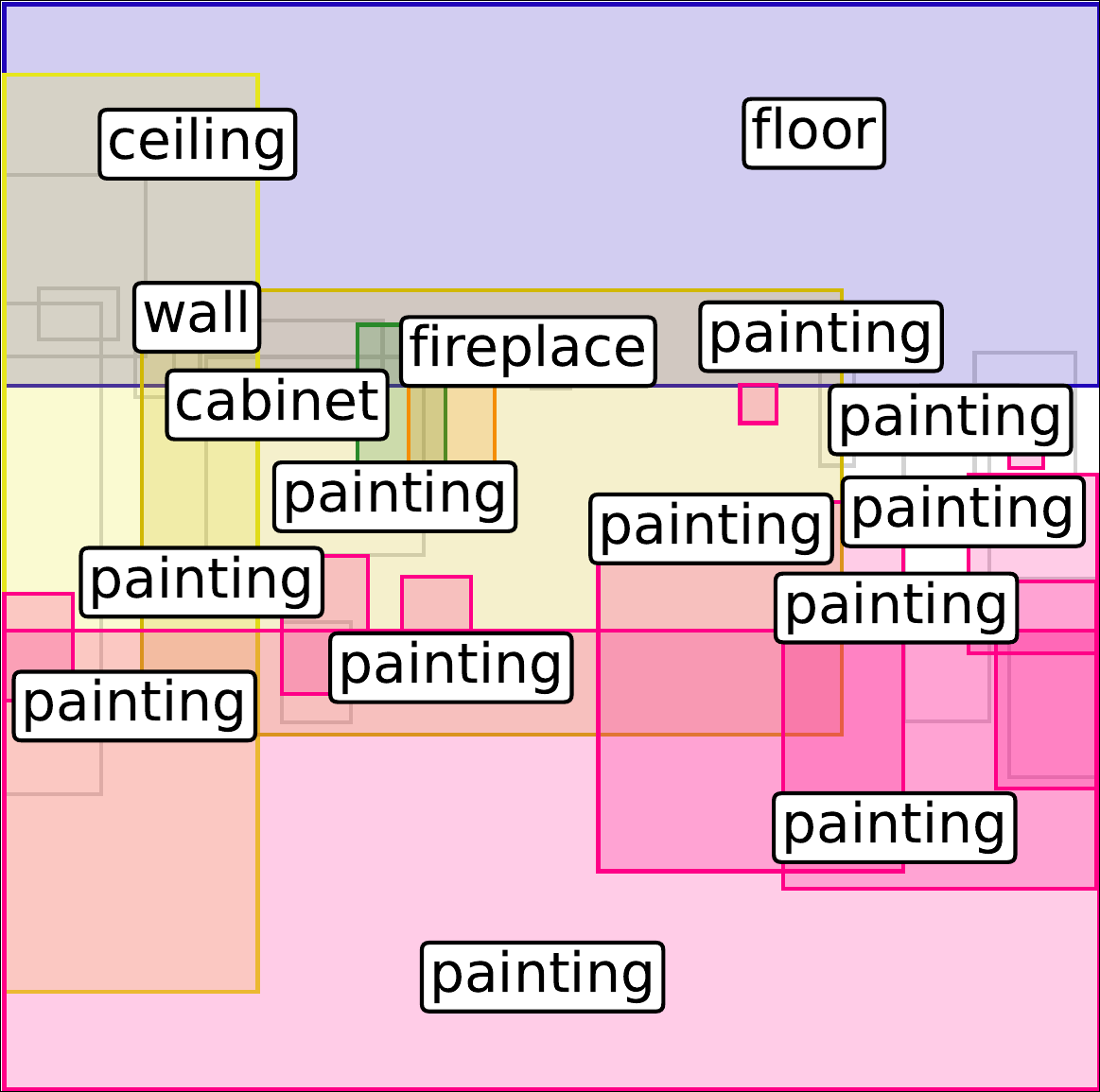} &
        \includegraphics[width={\mainResultsGraphicsWidth}]{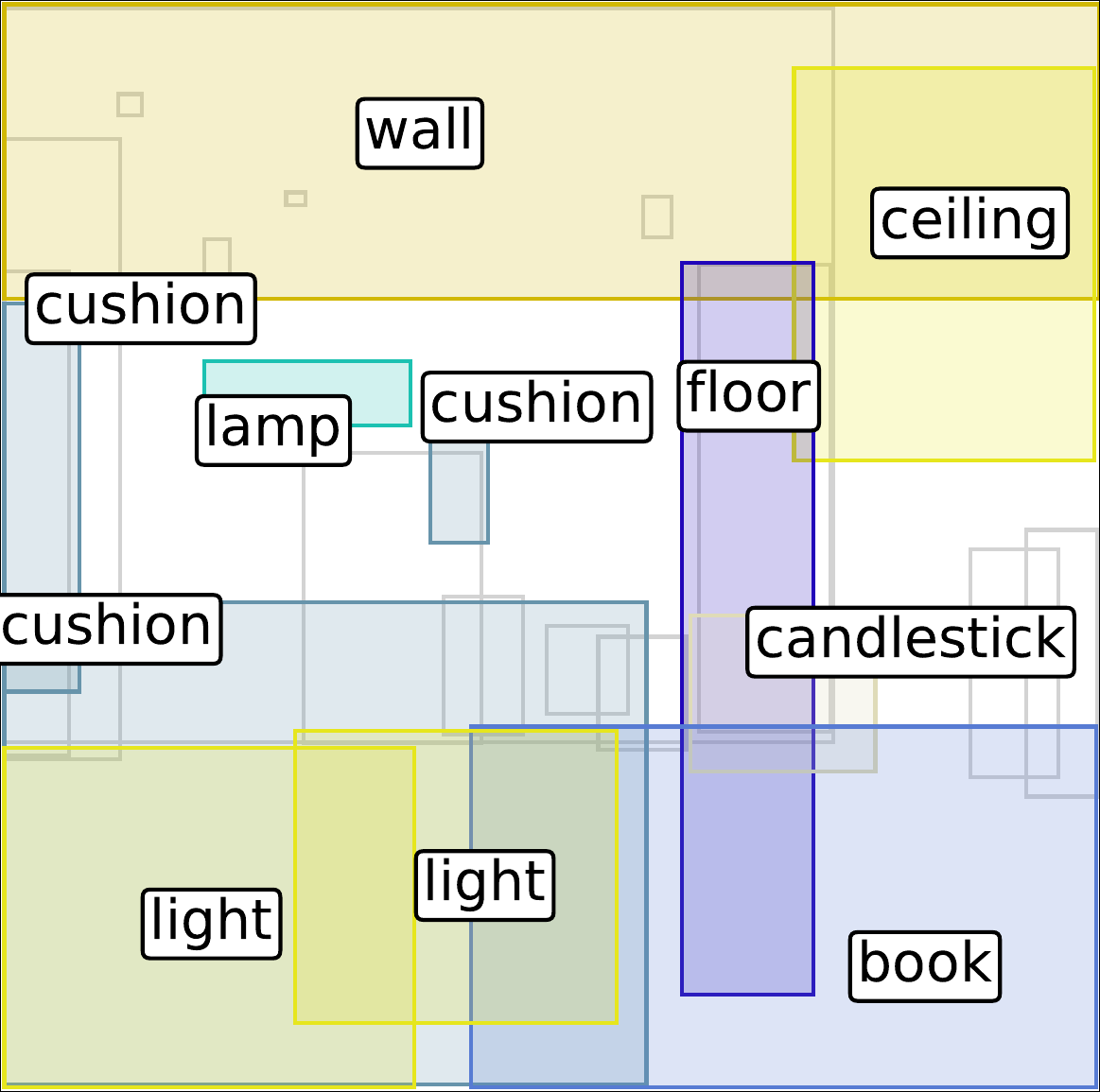} &
        \includegraphics[width={\mainResultsGraphicsWidth}]{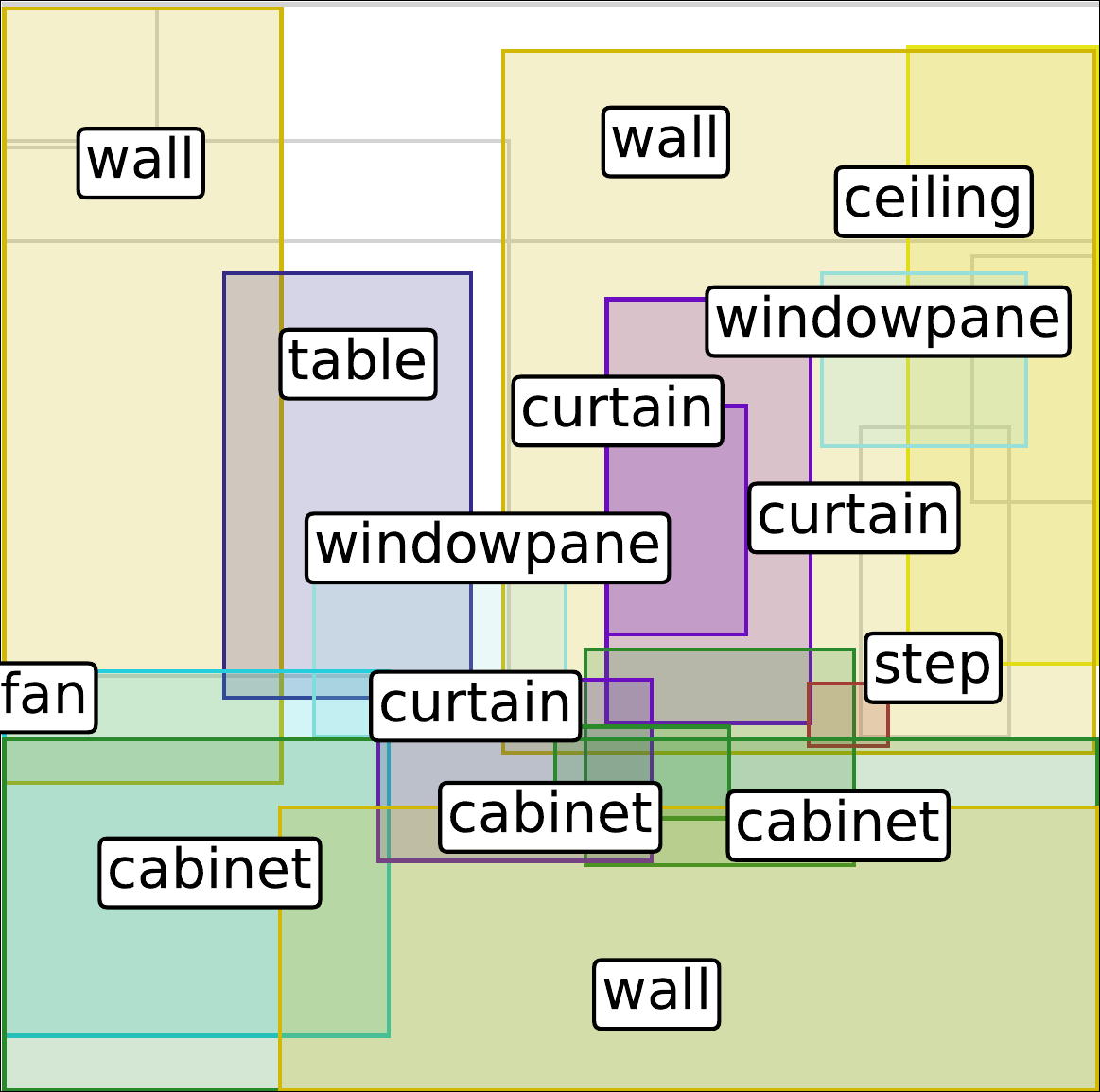} &

        \includegraphics[width={\mainResultsGraphicsWidth}]{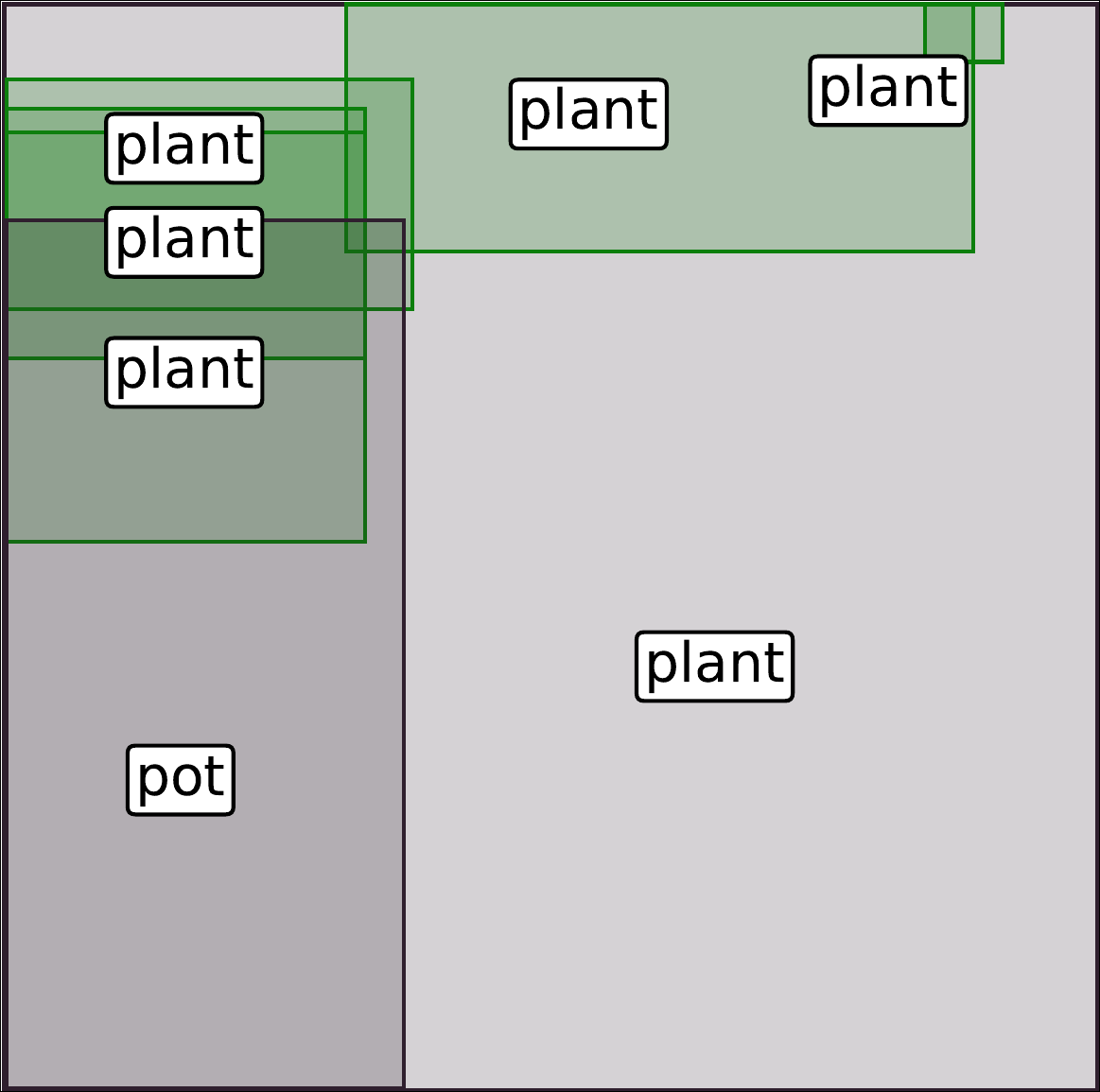} &
        \includegraphics[width={\mainResultsGraphicsWidth}]{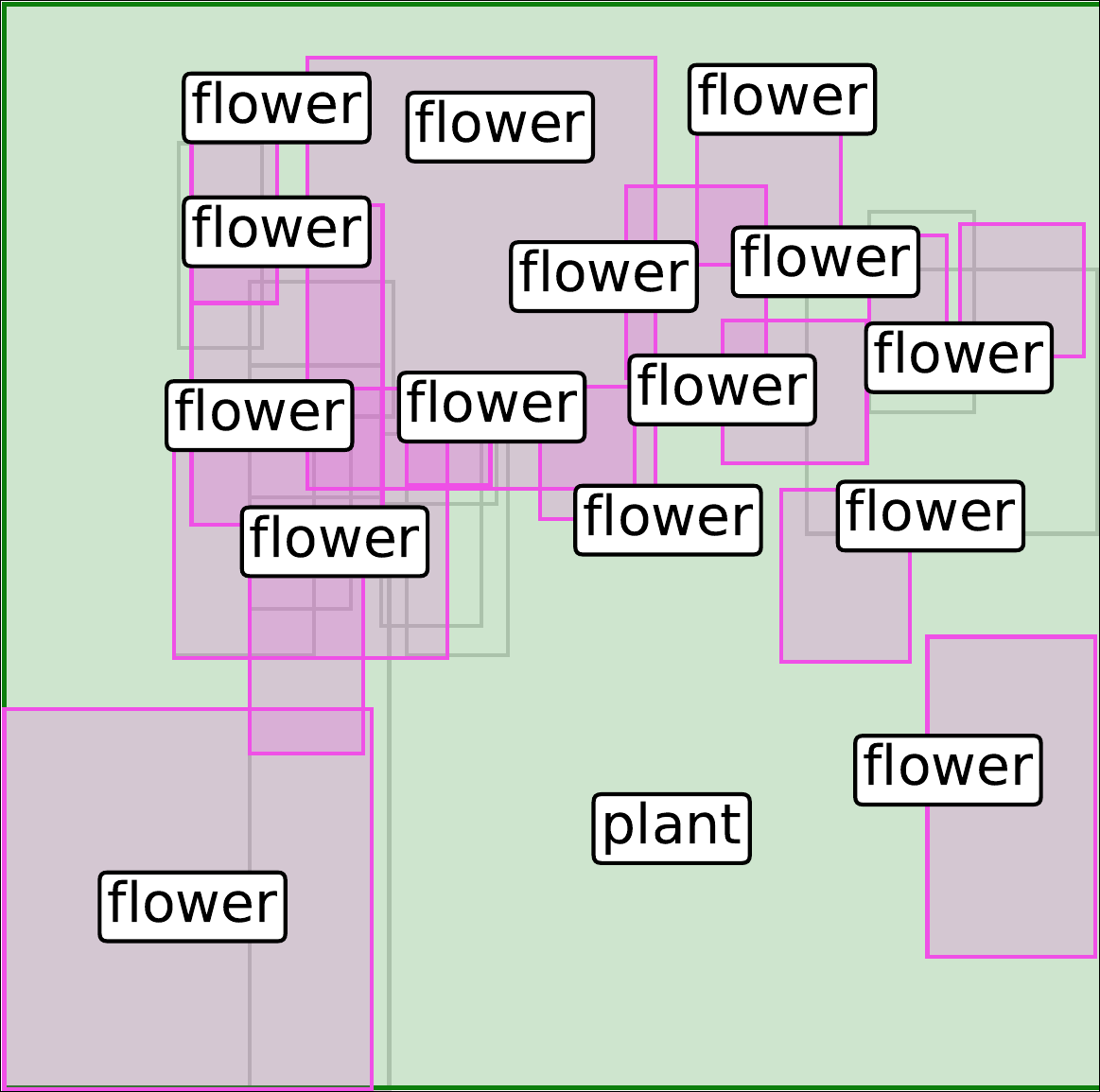} &
        \includegraphics[width={\mainResultsGraphicsWidth}]{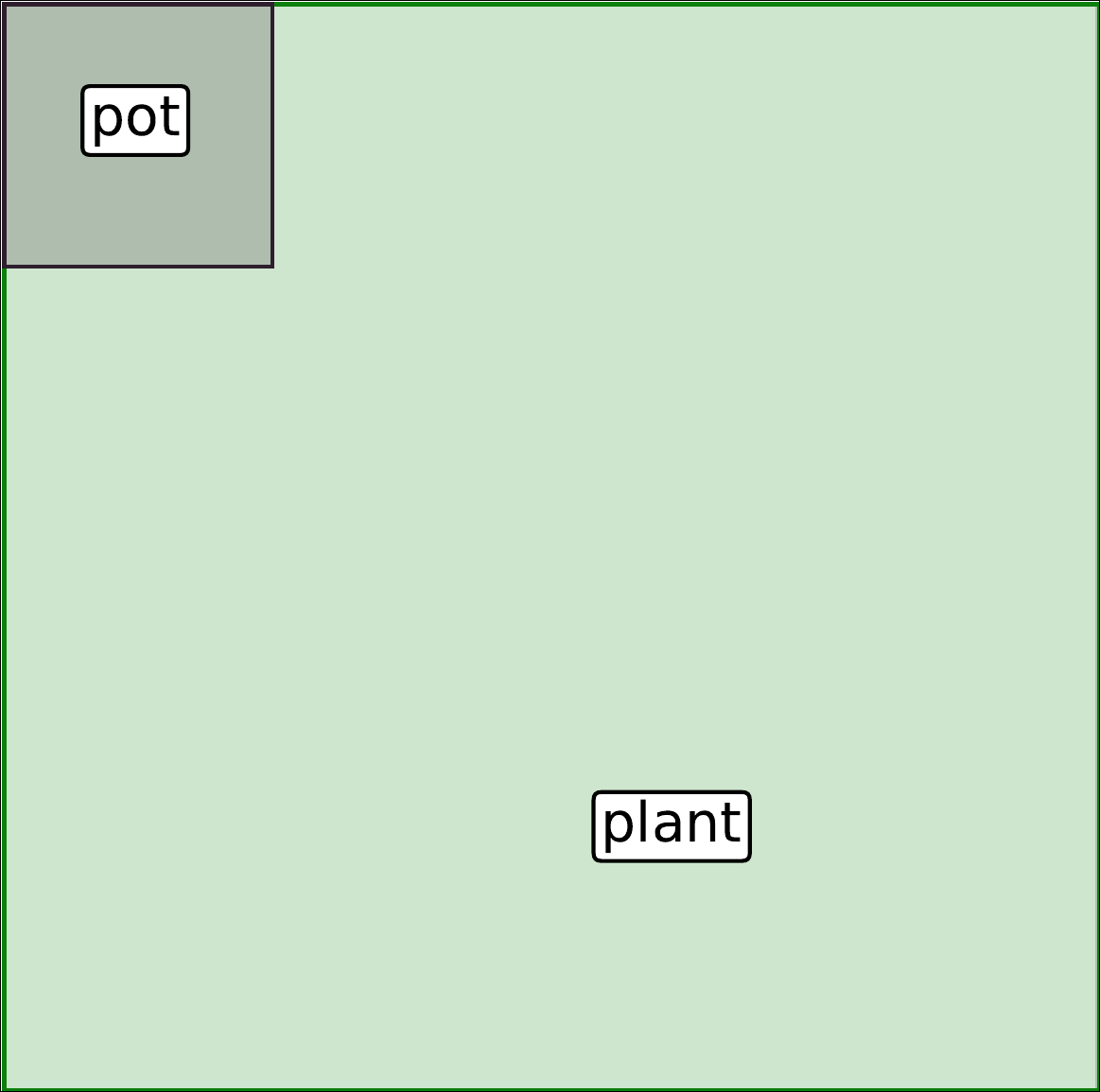} &

        \includegraphics[width={\mainResultsGraphicsWidth}]{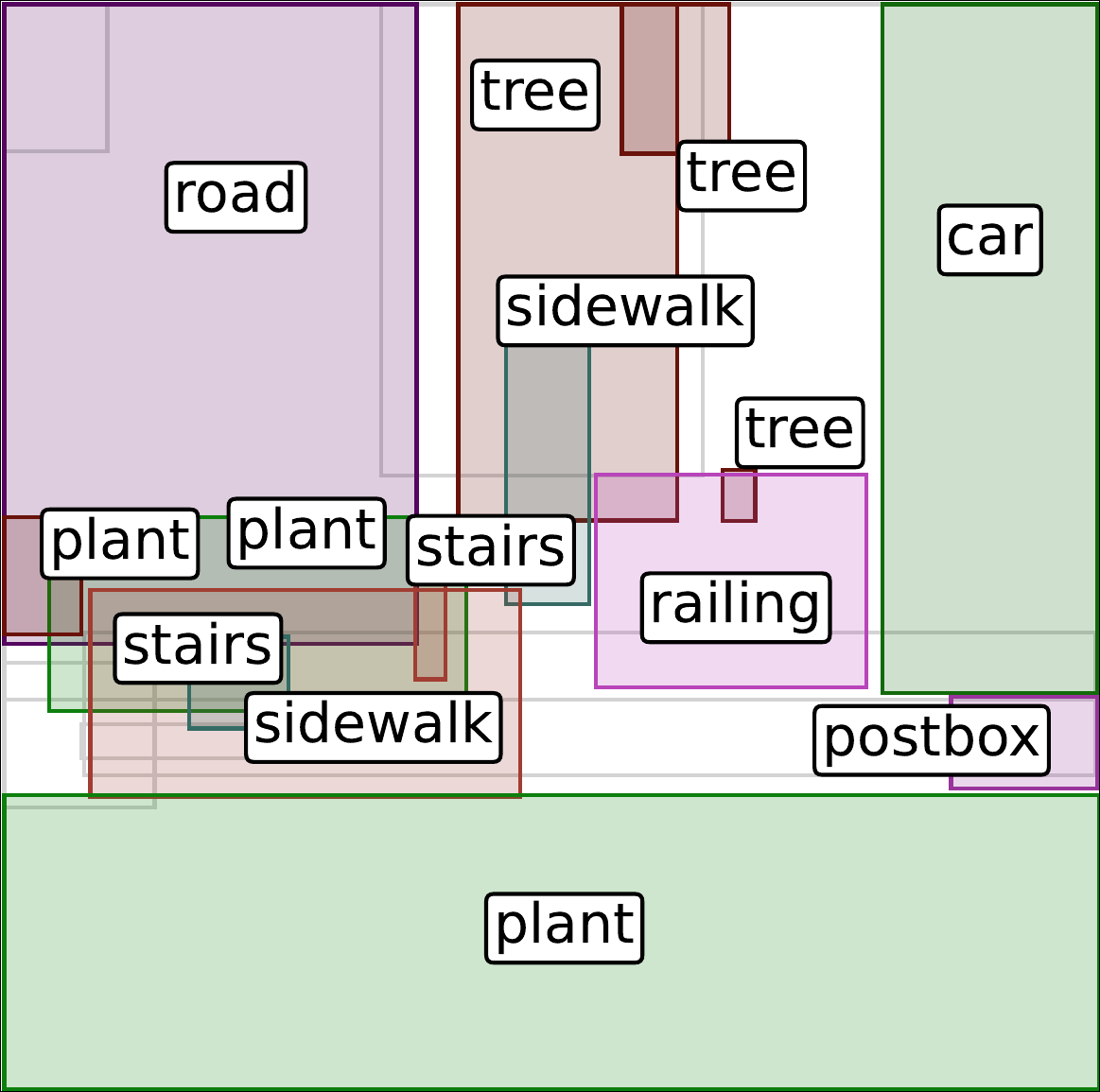} &
        \includegraphics[width={\mainResultsGraphicsWidth}]{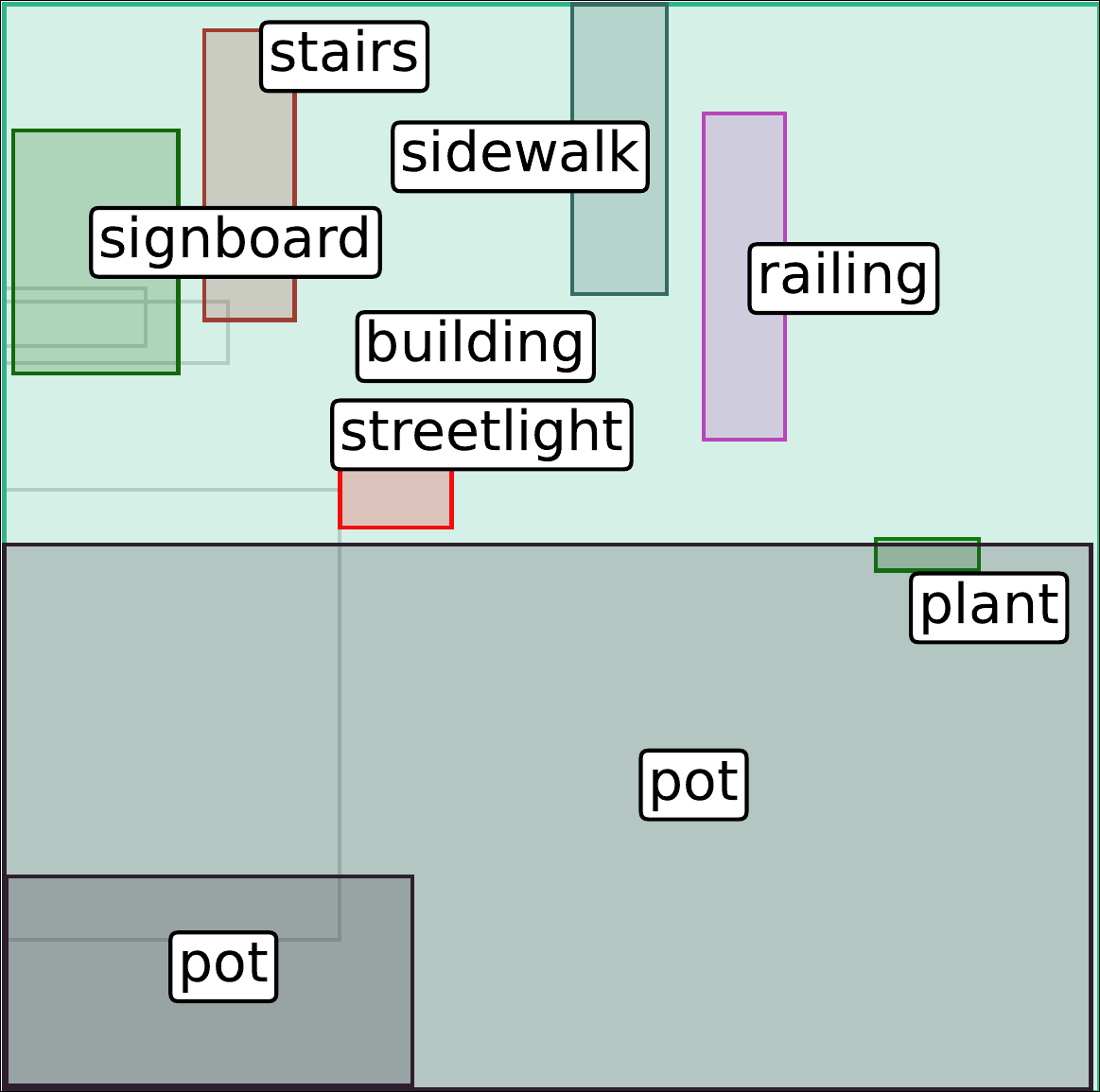} &
        \includegraphics[width={\mainResultsGraphicsWidth}]{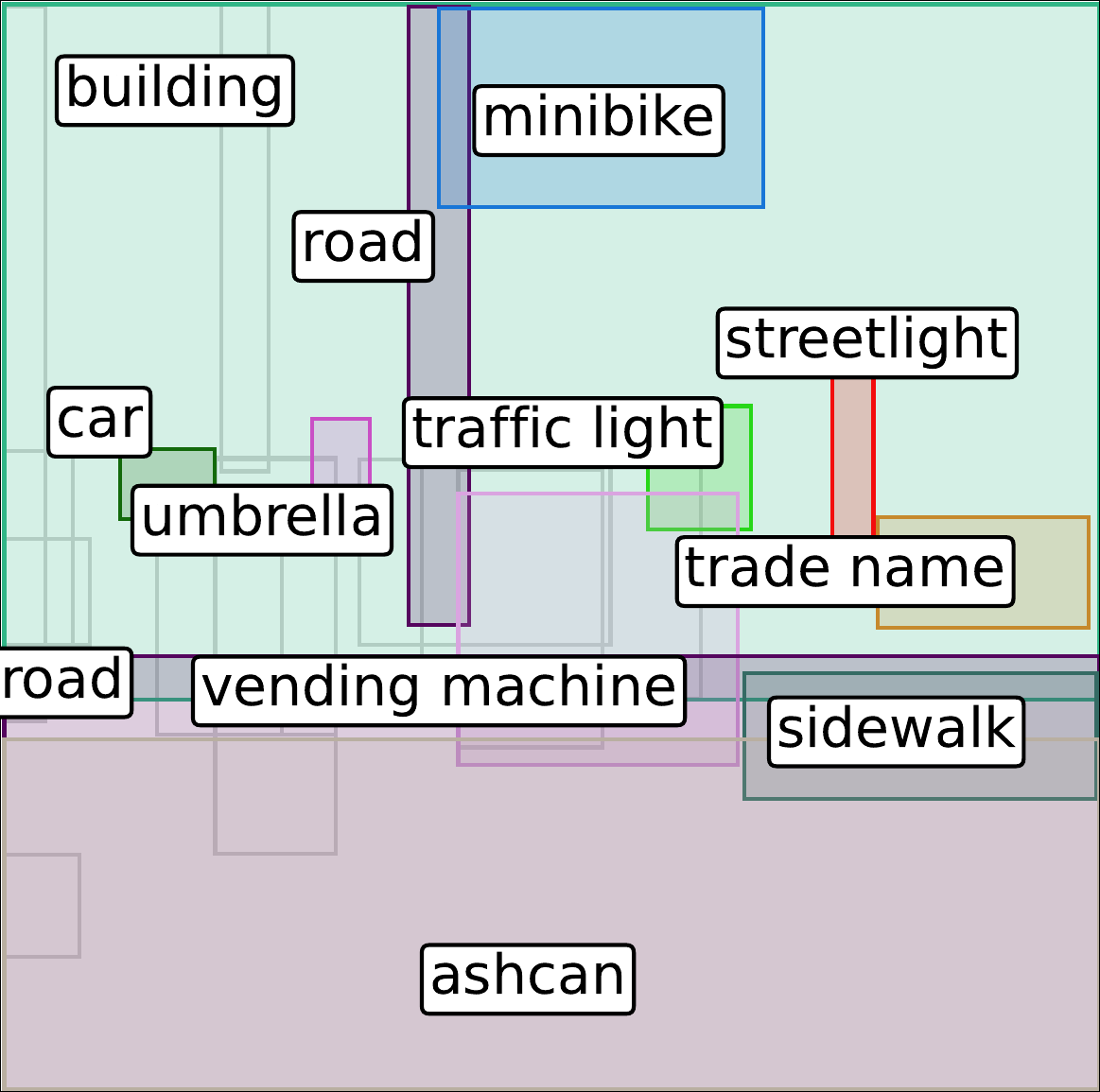} \\

\midrule

\multirow{-1}{*}{\rotatebox[origin=c]{90}{Ours}}&
        \includegraphics[width={\mainResultsGraphicsWidth}]{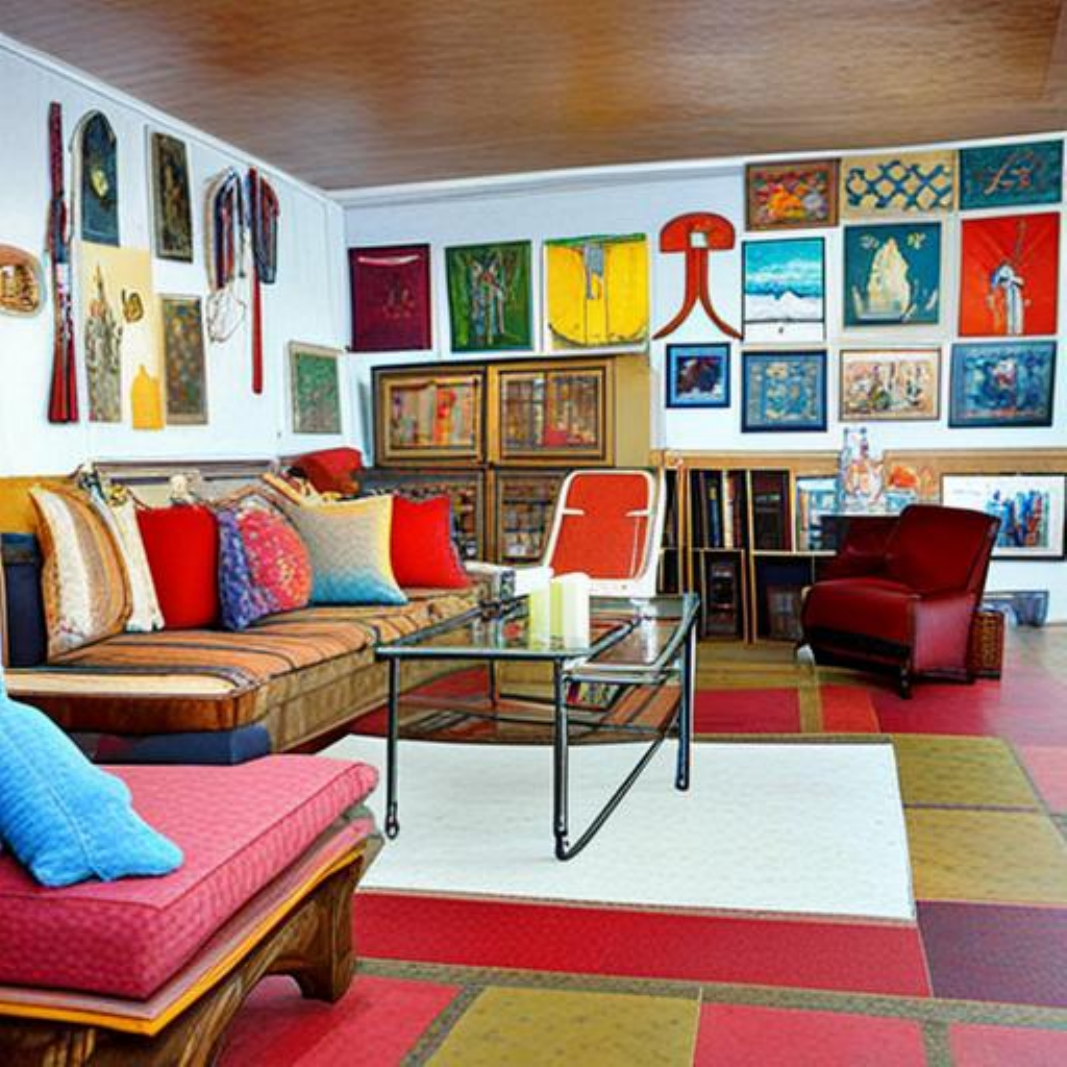} &
        \includegraphics[width={\mainResultsGraphicsWidth}]{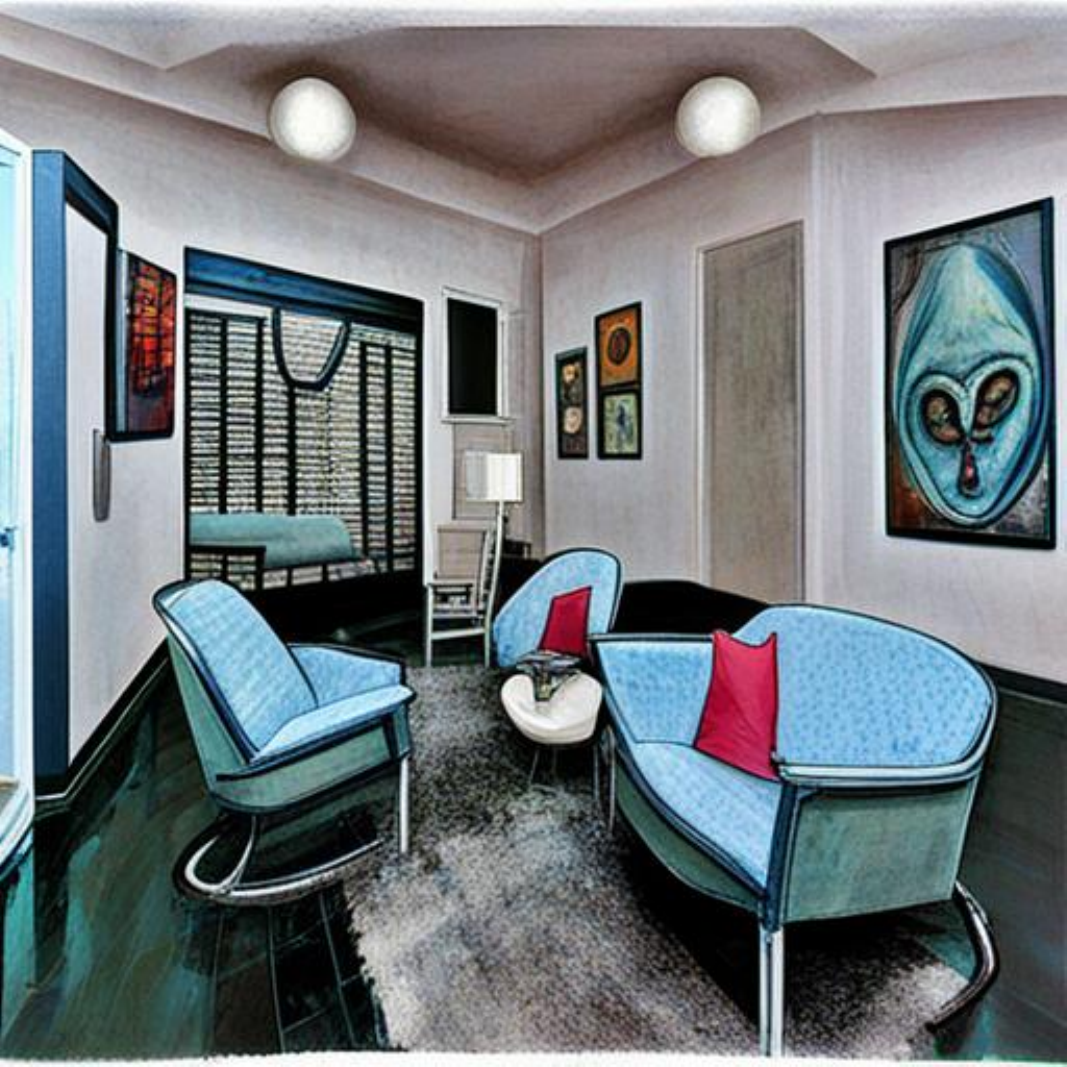} &
        \includegraphics[width={\mainResultsGraphicsWidth}]{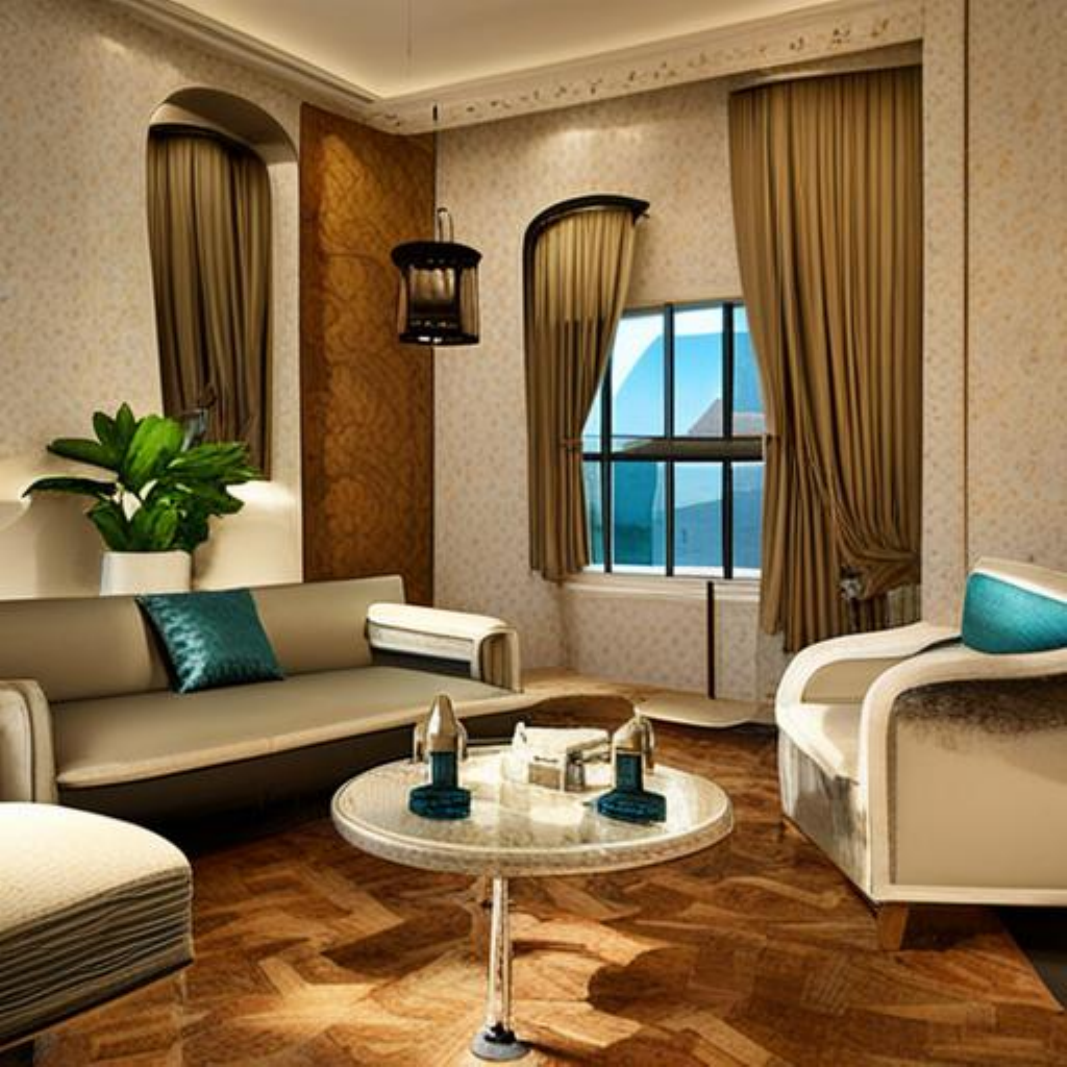} &

        \includegraphics[width={\mainResultsGraphicsWidth}]{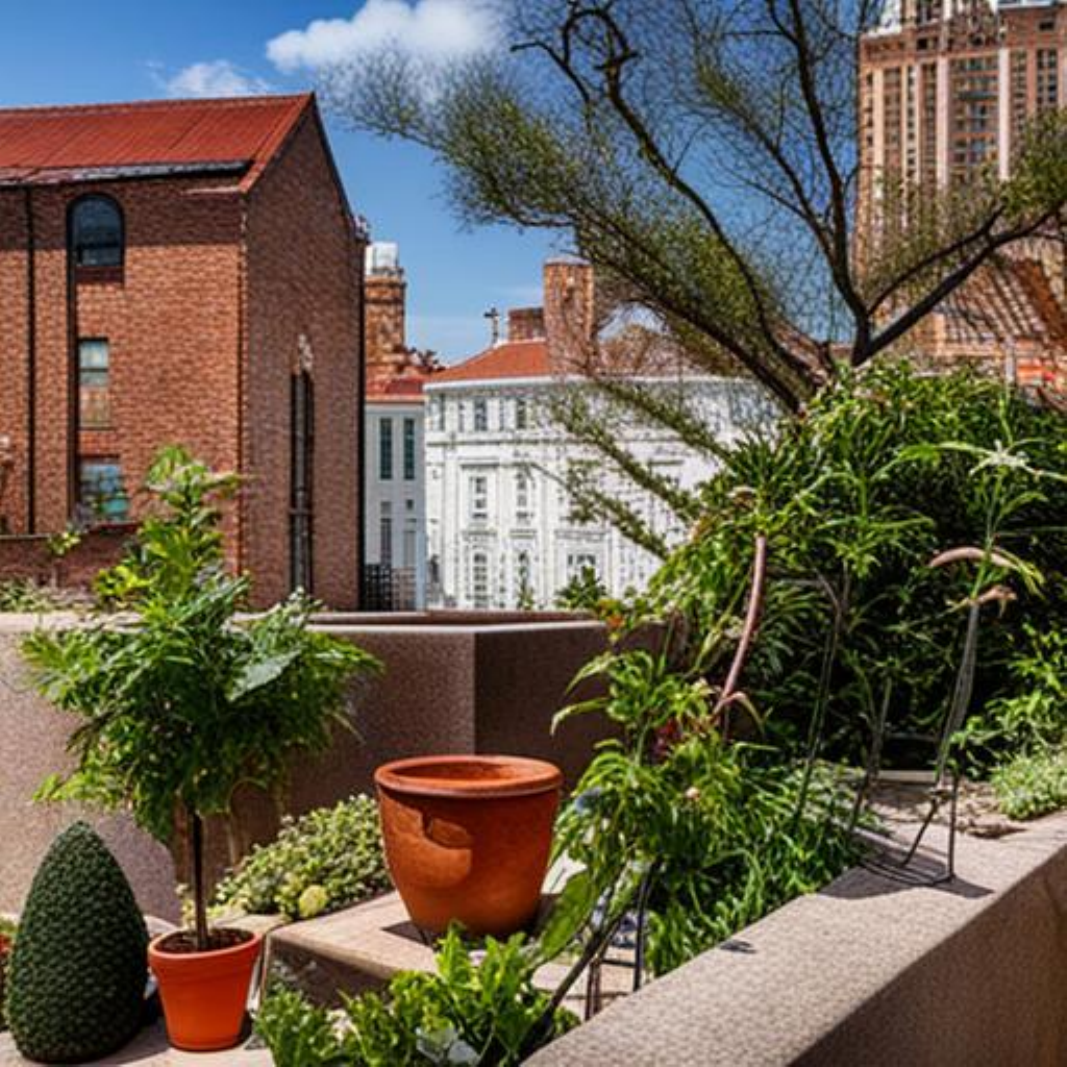} &
        \includegraphics[width={\mainResultsGraphicsWidth}]{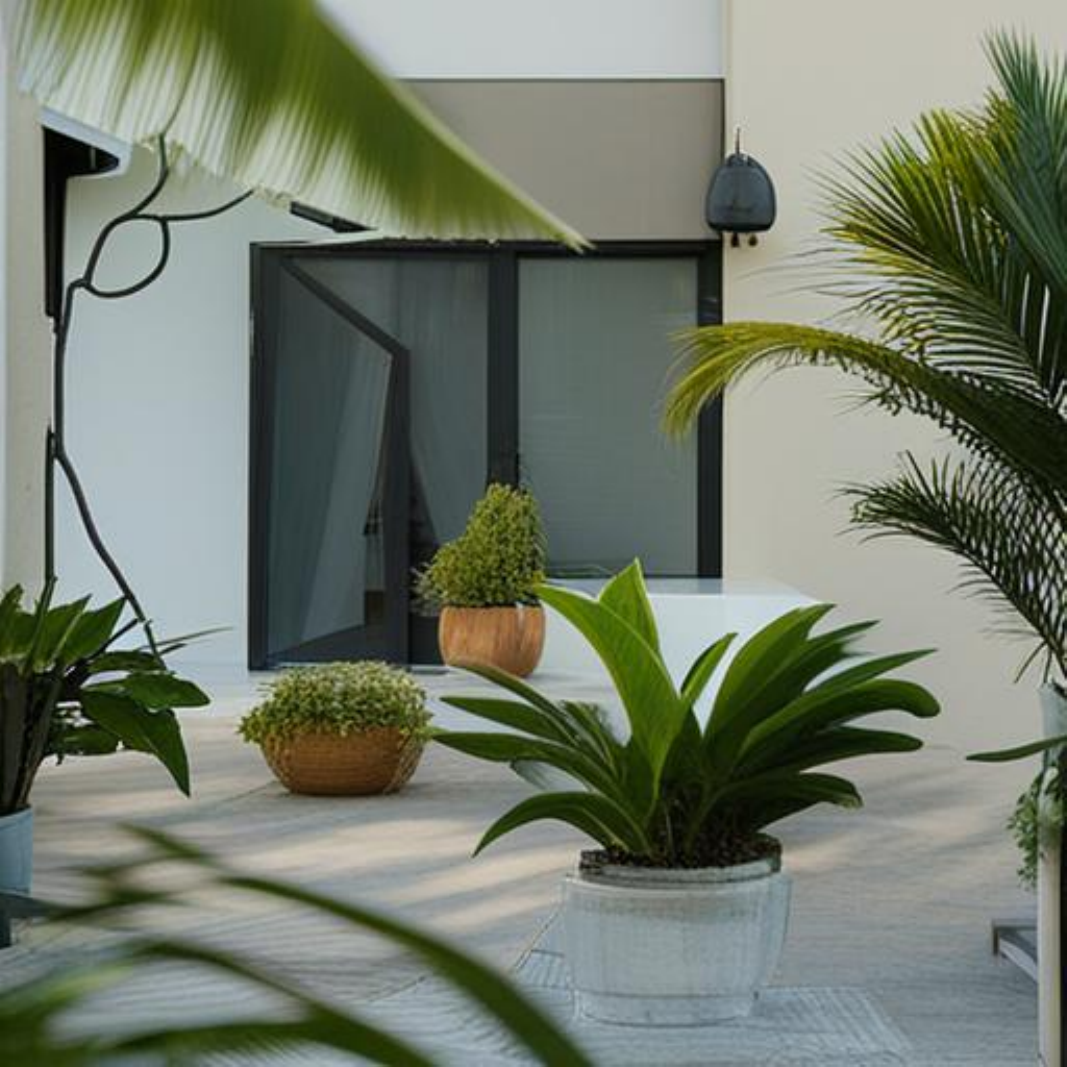} &
        \includegraphics[width={\mainResultsGraphicsWidth}]{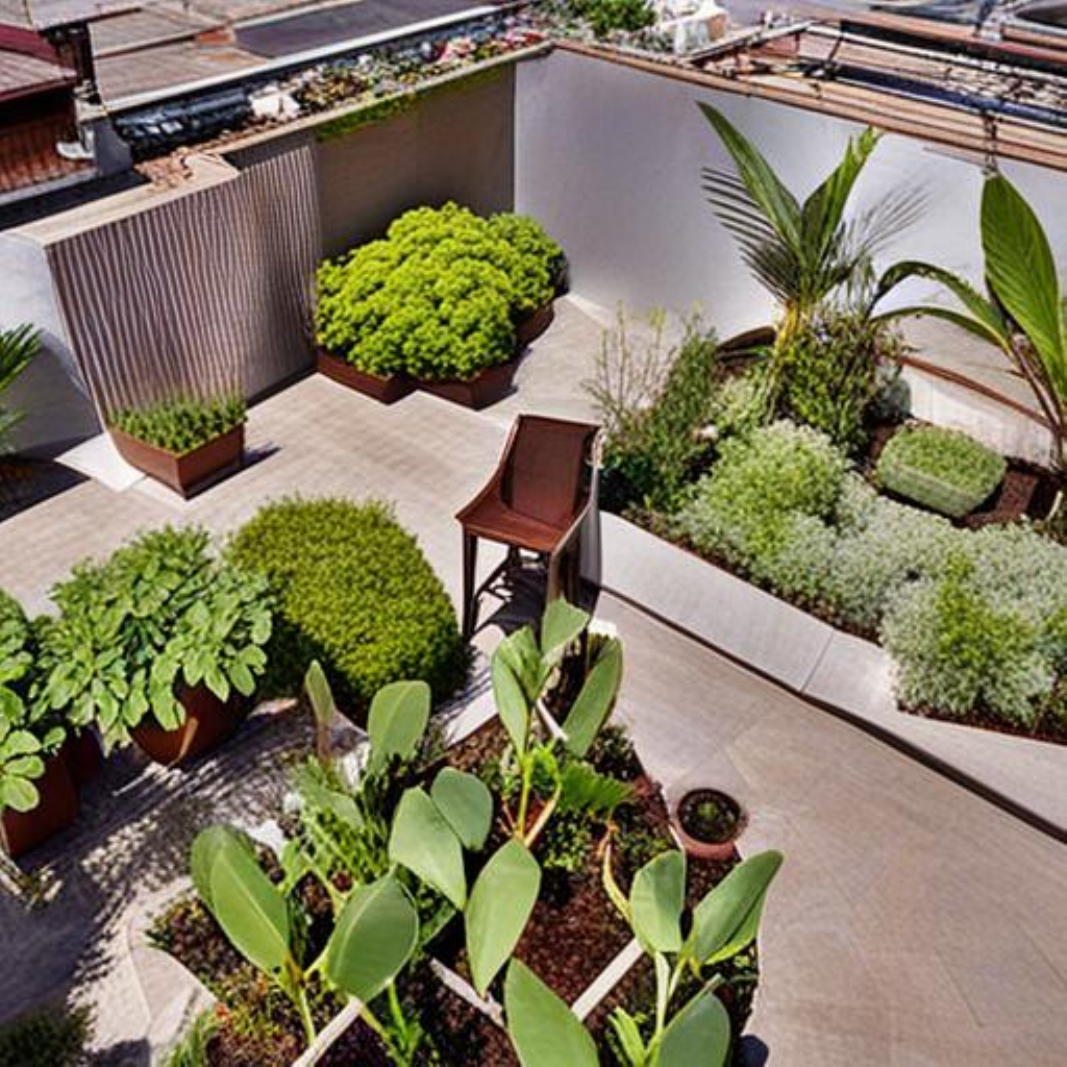} &

        \includegraphics[width={\mainResultsGraphicsWidth}]{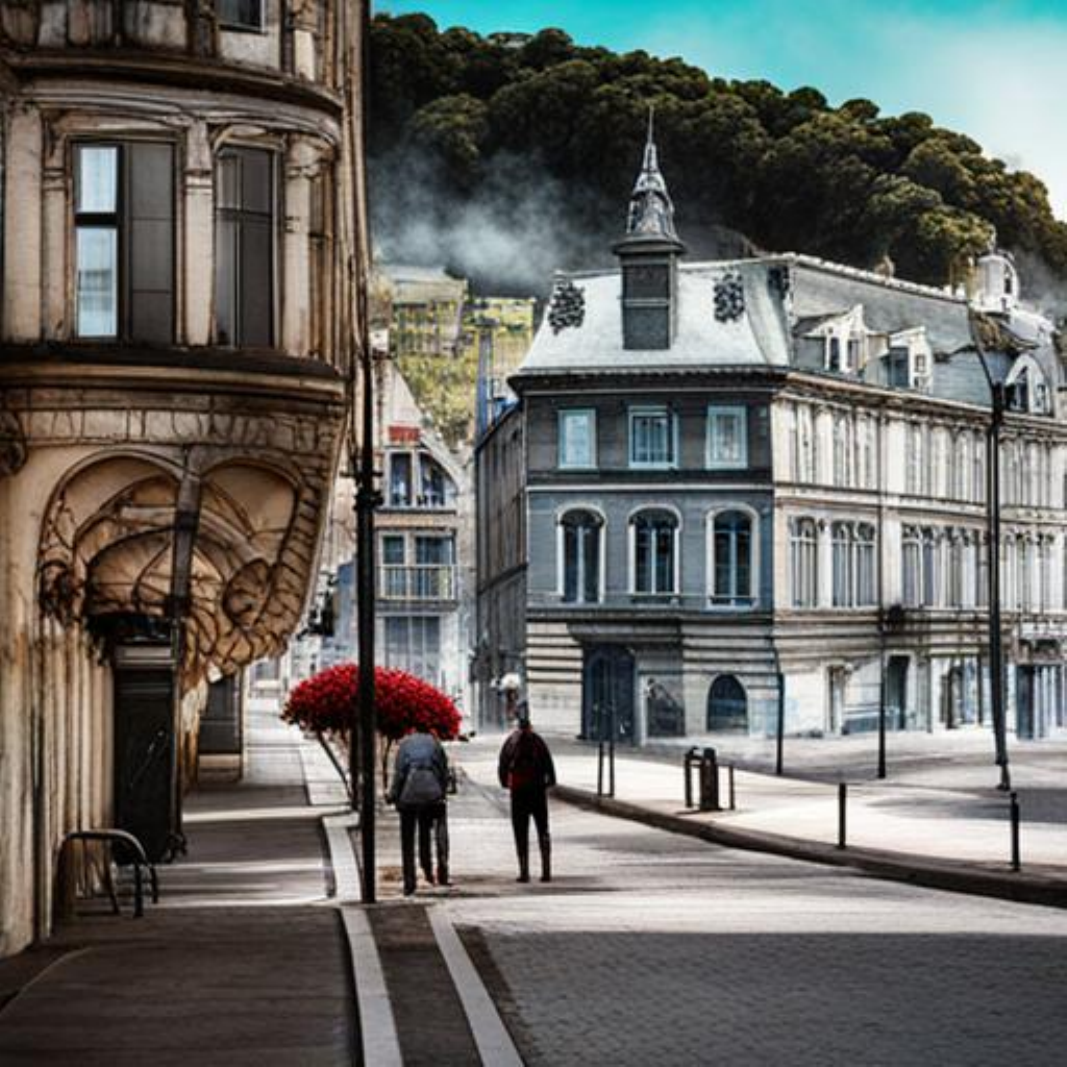} &
        \includegraphics[width={\mainResultsGraphicsWidth}]{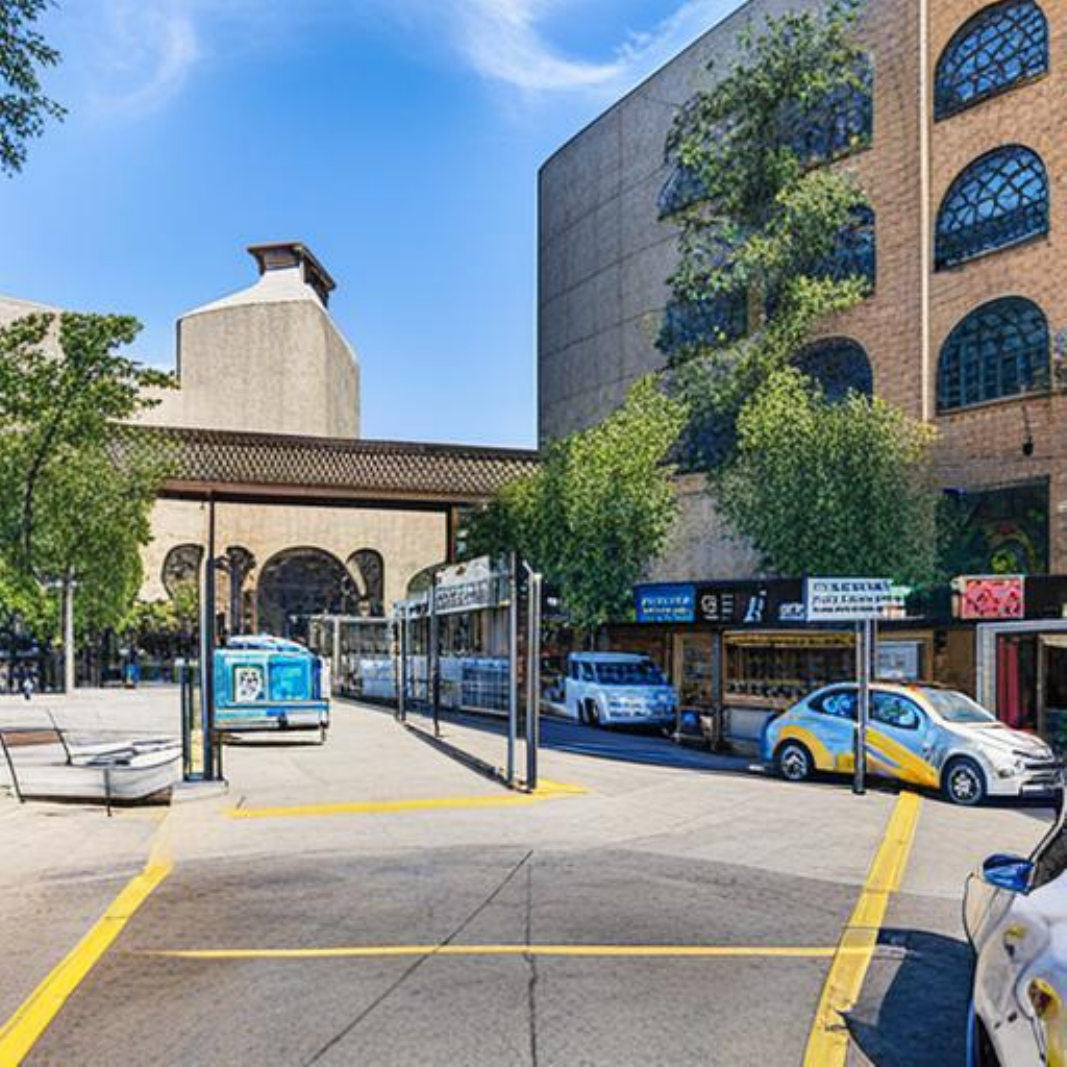} &
        \includegraphics[width={\mainResultsGraphicsWidth}]{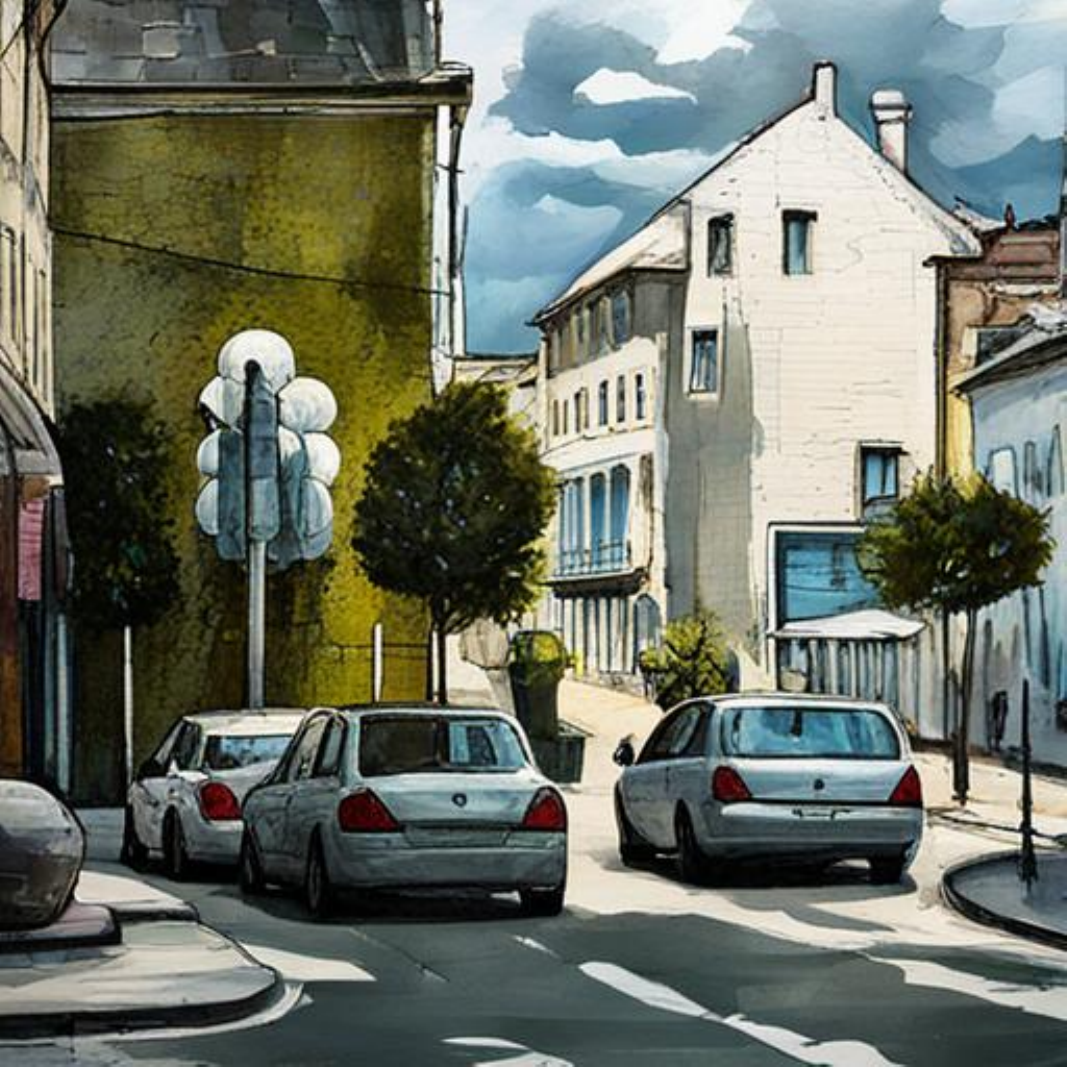} \\
        
        &
        \includegraphics[width={\mainResultsGraphicsWidth}]{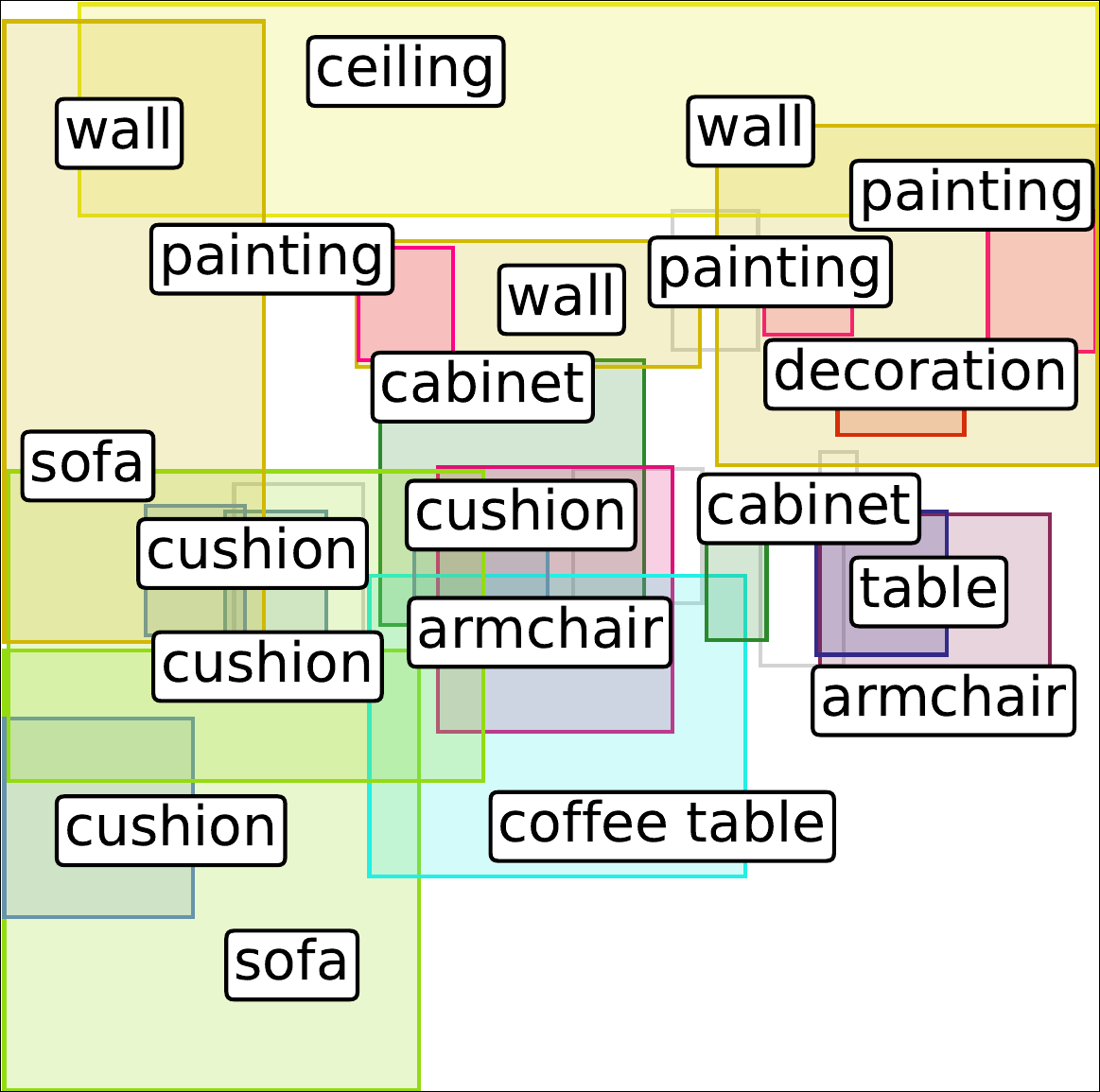} &
        \includegraphics[width={\mainResultsGraphicsWidth}]{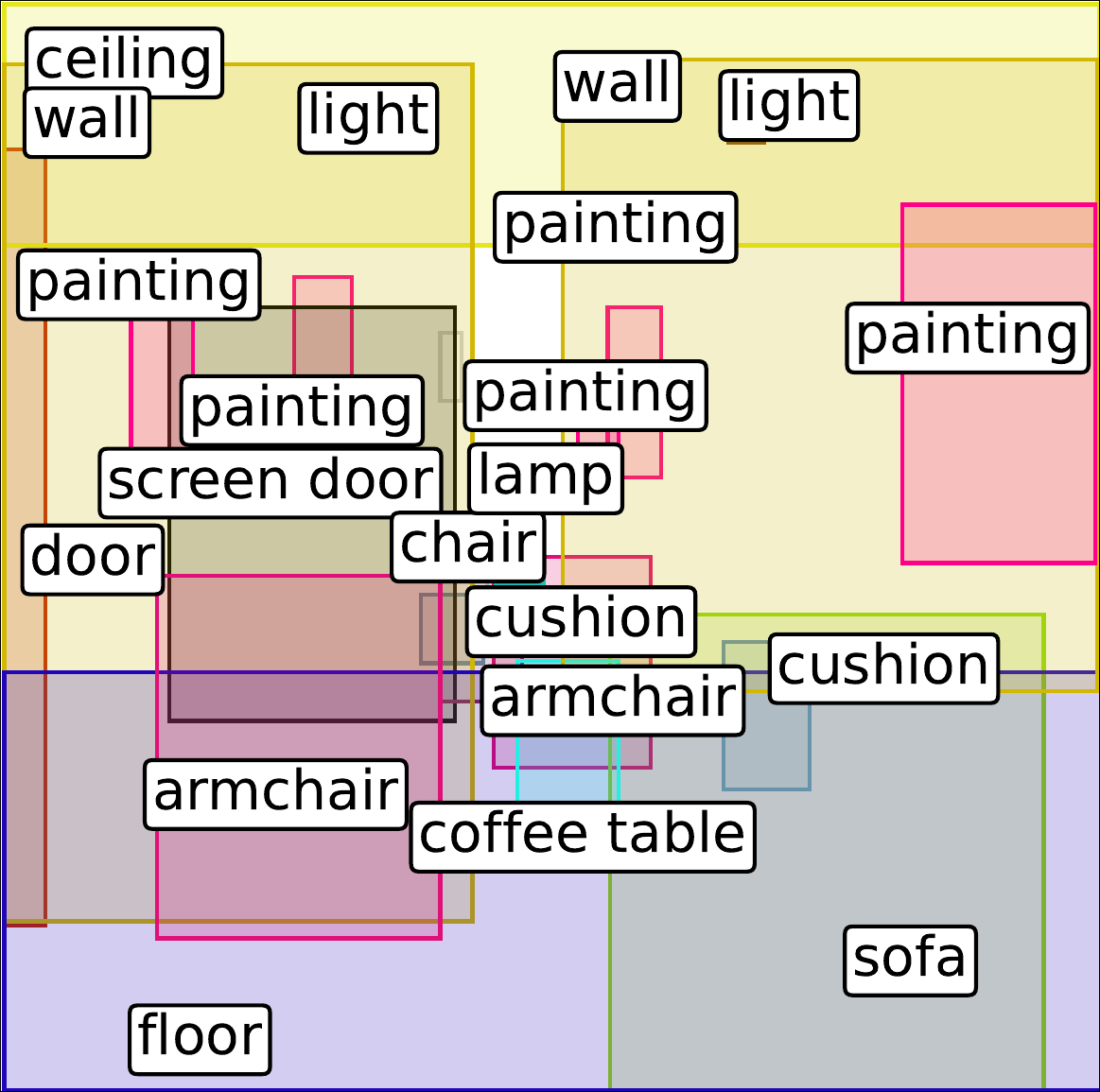} &
        \includegraphics[width={\mainResultsGraphicsWidth}]{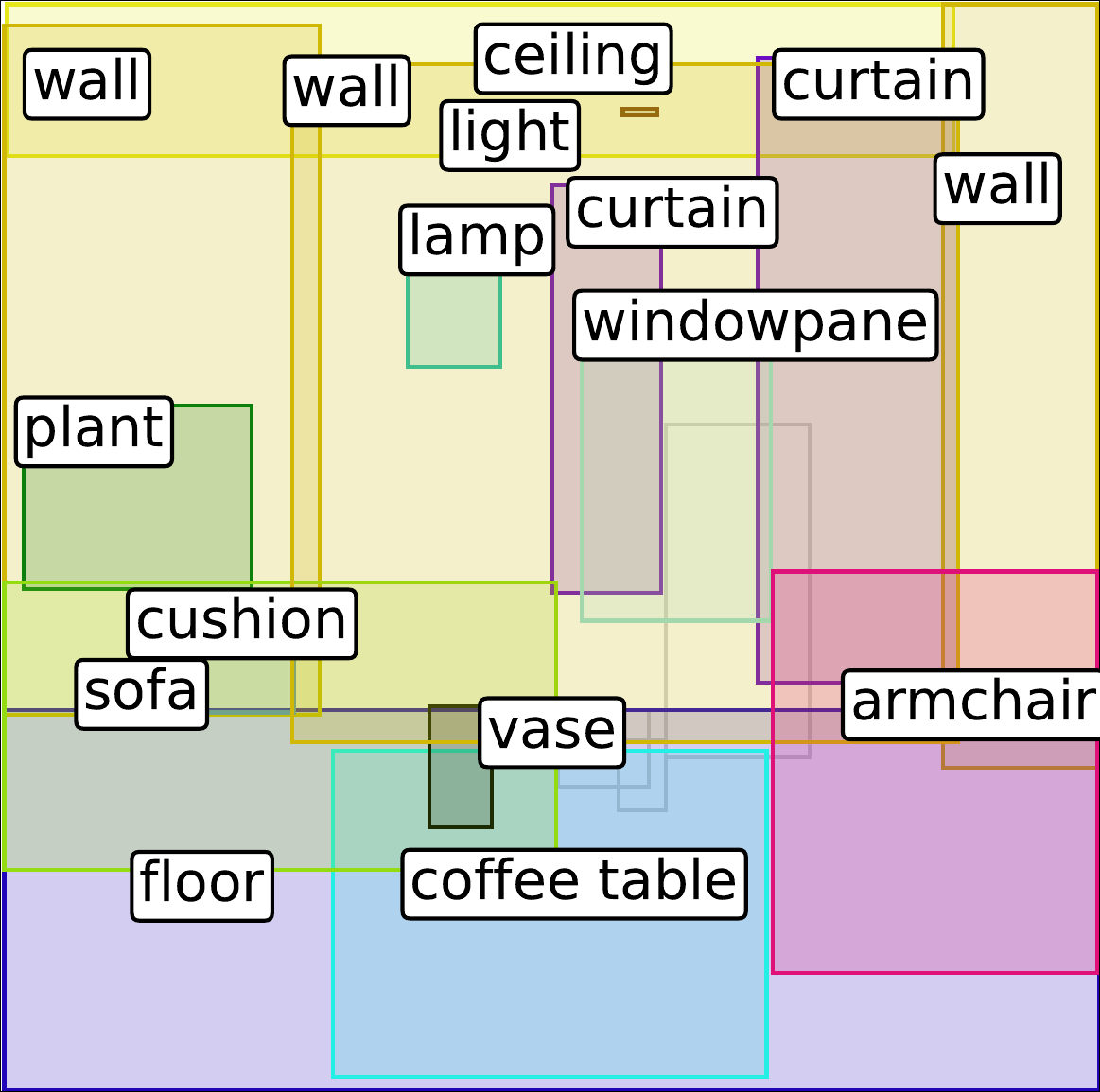} &

        \includegraphics[width={\mainResultsGraphicsWidth}]{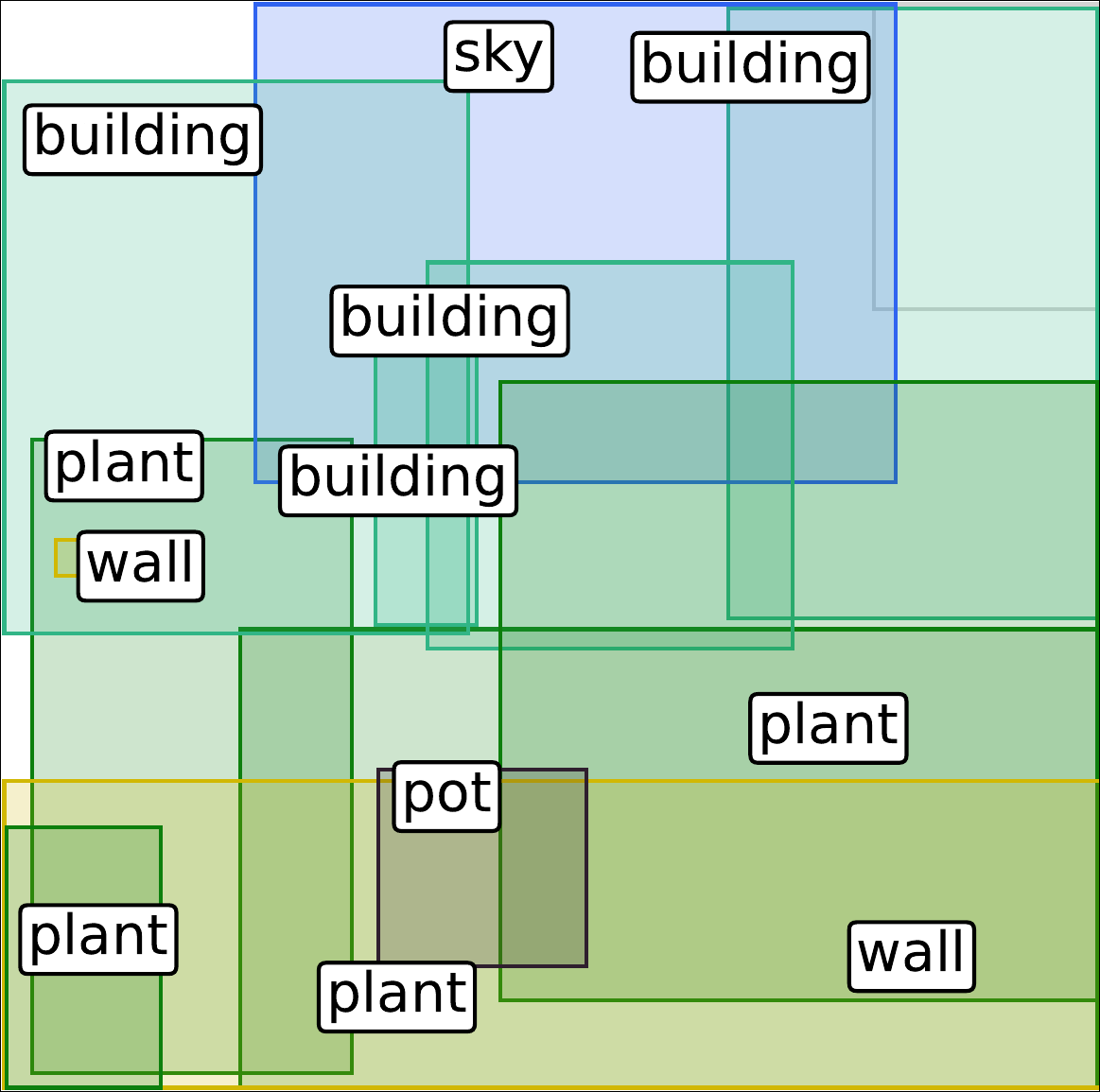} &
        \includegraphics[width={\mainResultsGraphicsWidth}]{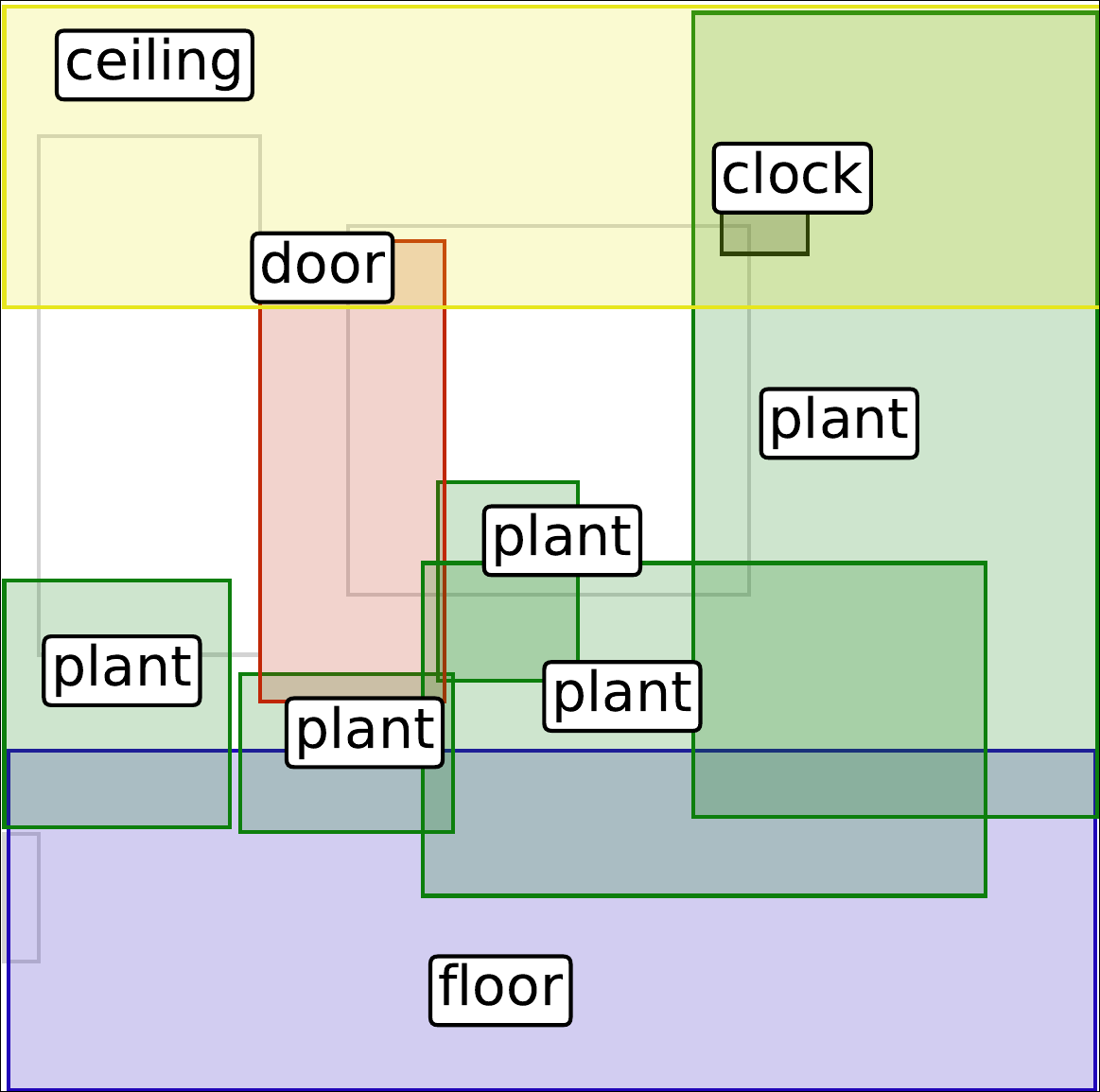} &
        \includegraphics[width={\mainResultsGraphicsWidth}]{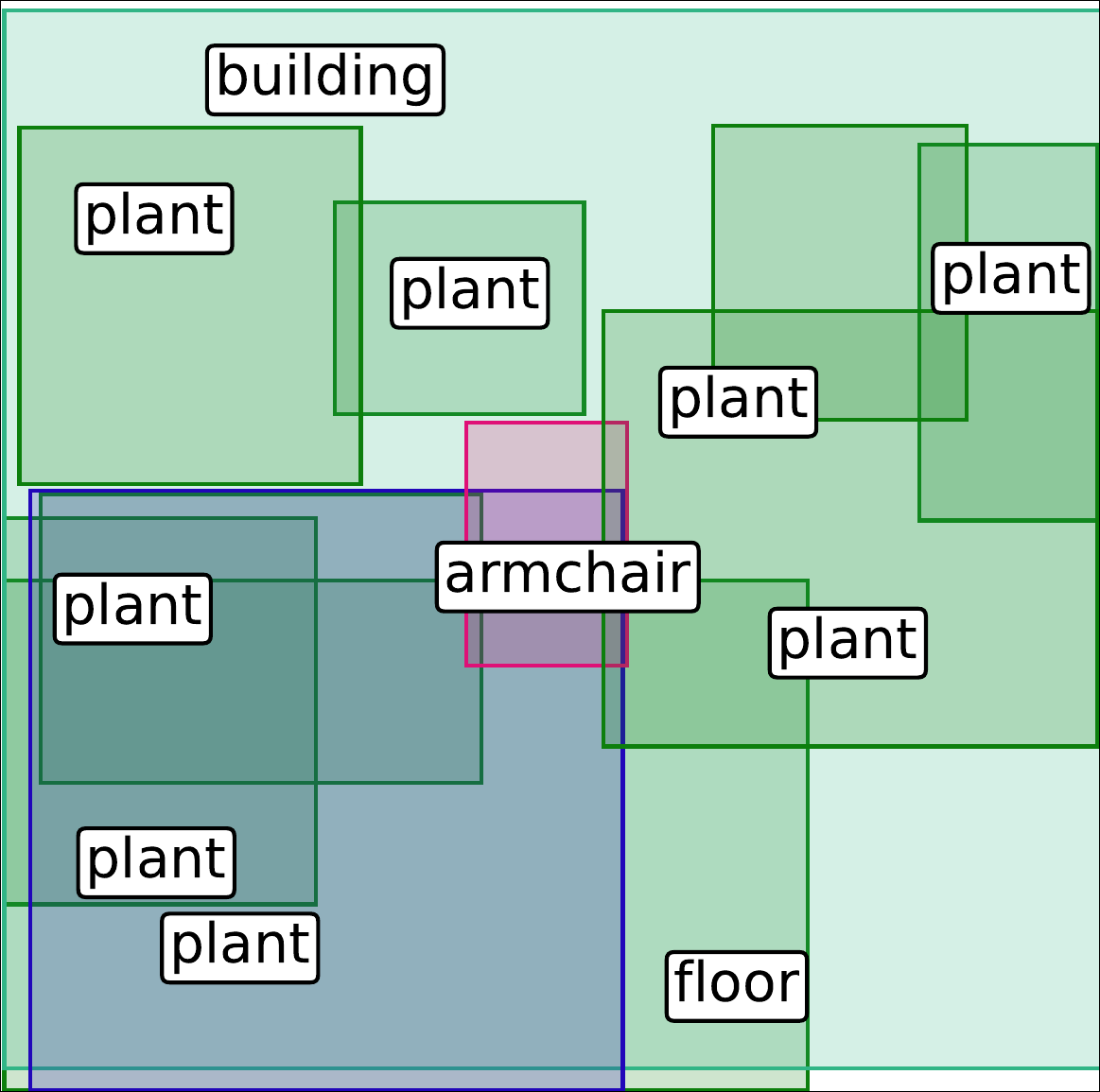} &

        \includegraphics[width={\mainResultsGraphicsWidth}]{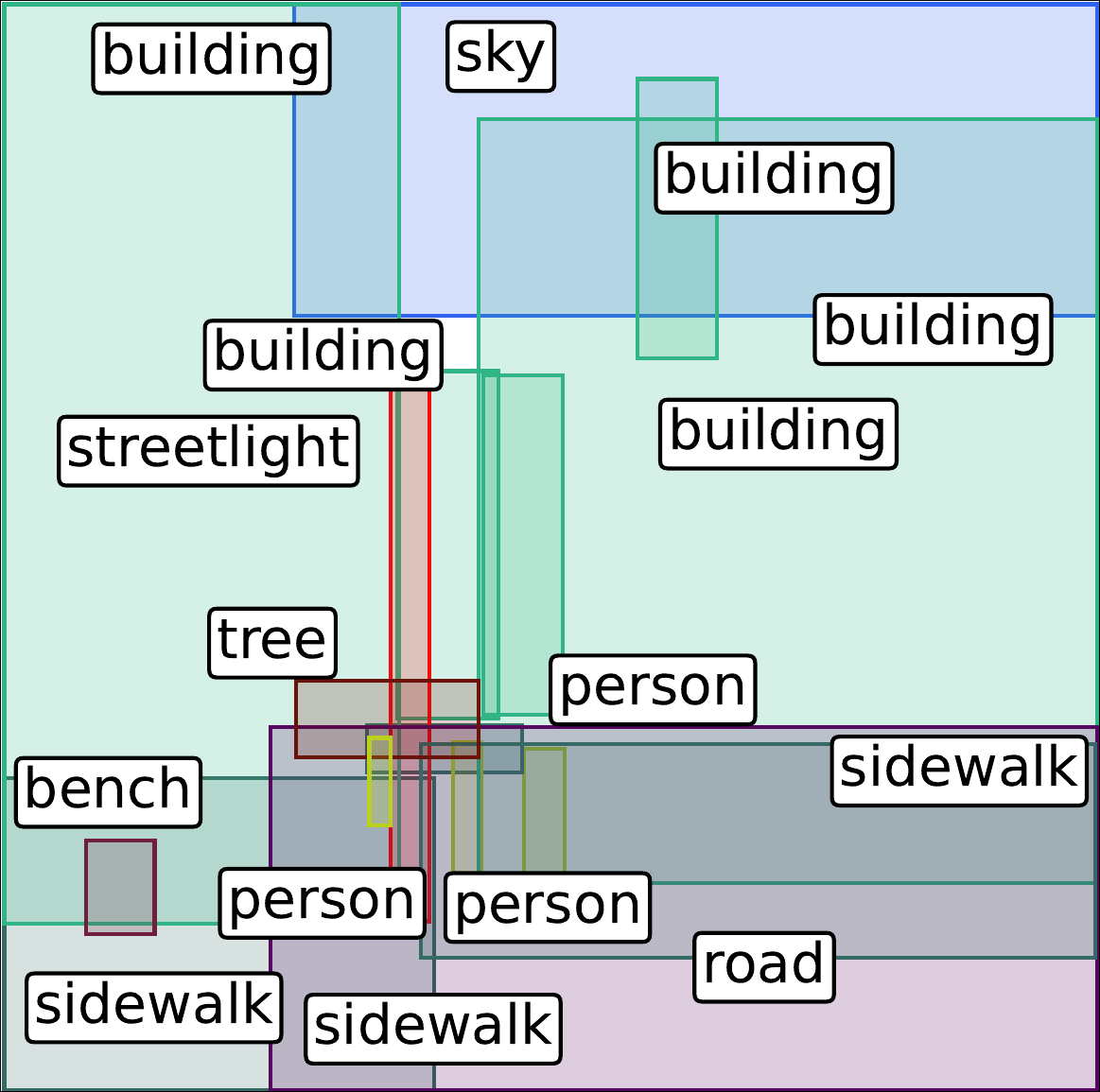} &
        \includegraphics[width={\mainResultsGraphicsWidth}]{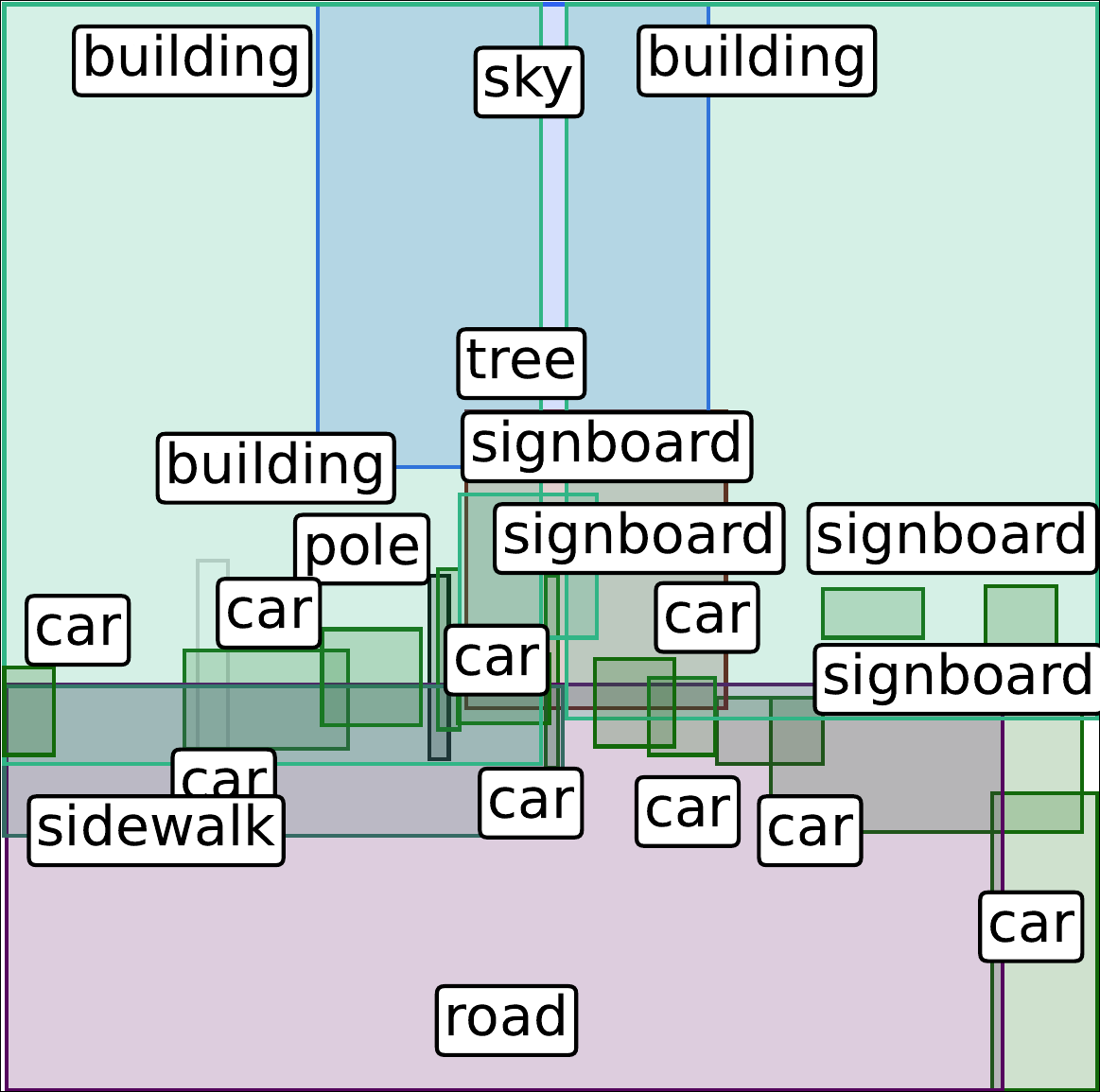} &
        \includegraphics[width={\mainResultsGraphicsWidth}]{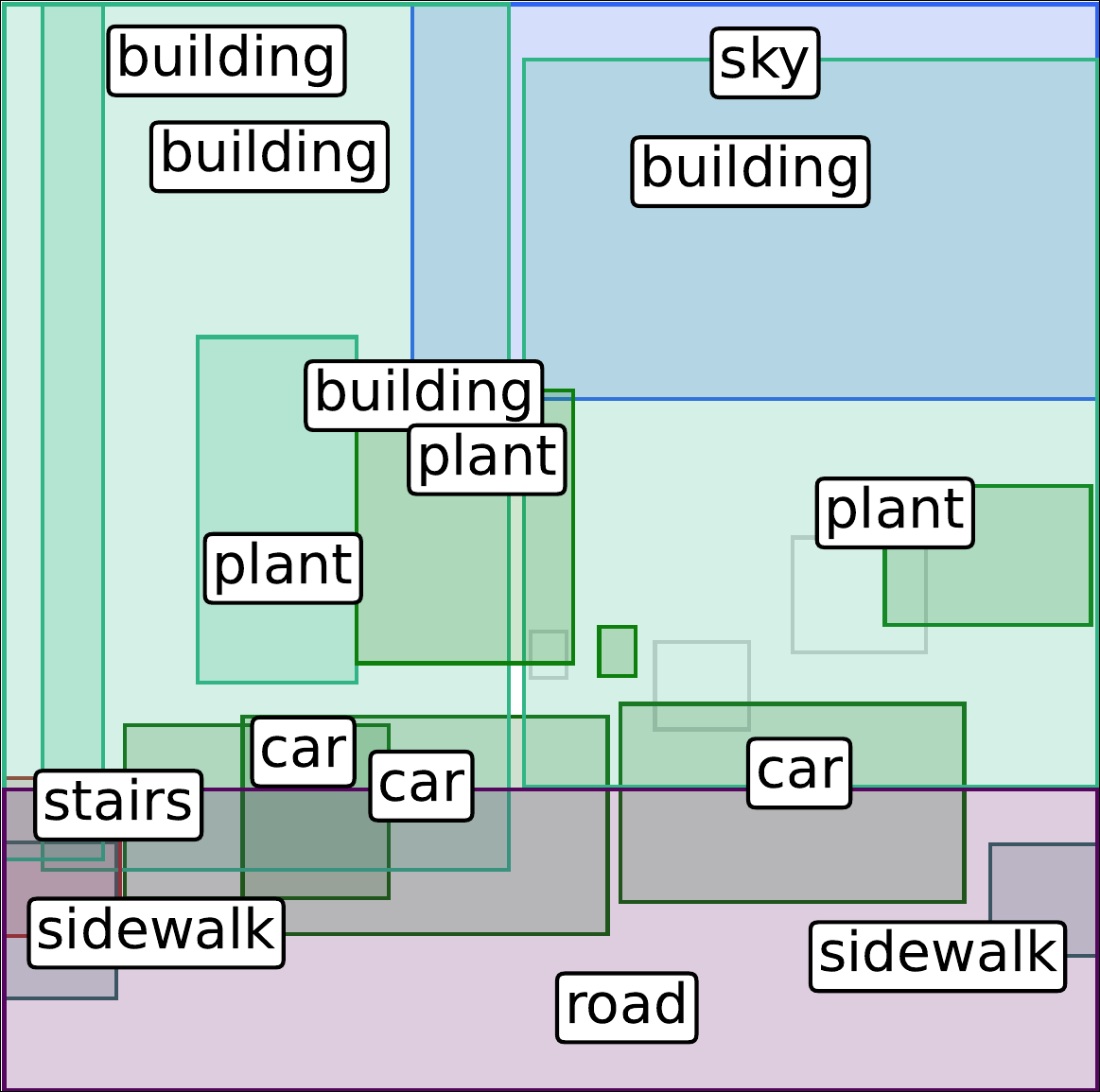} \\

        \midrule

    \end{tabular}
    \caption{\textbf{Qualitative comparison} (Best viewed up close). Layout objects that are depicted in the generated image are highlighted and labeled. From a visual inspection, having no layout produces scenes of little variation in content. LayoutFlow's layouts do not appear to capture scene structure. {\gptFourO}'s layouts lack variety. Layout Transformer produces layouts with implausible arrangements of objects, leading to images which do not depict the global prompt accurately.  Our method creates plausible and varied layouts, leading to images that are diverse and look realistic. These observations are supported by our human evaluation in ~\cref{fig:human_survey}. Zoomed-in versions of these layouts for printing are available in the supplemental.  }
    \label{fig:visual_results}
\end{figure*}

\begin{table*}[ht]
\footnotesize
\newrobustcmd\B{\DeclareFontSeriesDefault[rm]{bf}{b}\bfseries}  
\def\Uline#1{#1\llap{\uline{\phantom{#1}}}}

\sisetup{detect-weight=true,
         mode=text,
         table-format=2.2,      
         add-integer-zero=false,
         table-space-text-post={*},
         table-align-text-post=false
         }

    \centering

        \begin{tabular}{
            l
            S
            S
            S
            S
            S
            S
            S
            S
            S
            S
            S
            S
            S
            c   
            }
            \toprule
            \toprule
            {\textbf{Model}}  &  {\textbf{CMMD} ($\downarrow$)}  &  {\textbf{FID} ($\downarrow$)}  &  {\textbf{KID} ($10^{-2}$) ($\downarrow$)}   &  {\textbf{VQA} ($\uparrow$)}  &  {\textbf{HSPv2} ($\uparrow$)}  &  {\textbf{ImageReward} ($\uparrow$)} \\ 
             \midrule 
            LayoutDiffusion  & 0.61  & 1.08  & 1.92  & 0.34  & 0.19  & -2.11 \\ 
             \midrule 
            LayoutFlow  & 0.25  & 0.80  & 0.88  &  0.80  & 0.23  & -1.01 \\
             \midrule 
            LayoutTransformer  & 0.06  & 0.44  & 0.30  & 0.75  & 0.23  & -1.00 \\ 
             \midrule 
            GPT4  & 0.09  & 0.94  & 0.45 & \B 0.88  & \B 0.25  & -0.51 \\ 
             \midrule
            \rowcolor{lightgray}
            Ours  & \B 0.03  &  \B 0.17  & \B 0.27  & \B 0.88  & \B 0.25  & \B -0.32 \\ 
            \bottomrule
            \bottomrule
        
        \end{tabular}
        
    \caption{\textbf{Image Metrics Comparison} We evaluate traditional metrics and compare the images generated from layouts of different layout generators. To avoid biases of the image generator, we show the best score among the layout-to-image genators {\instanceDiffusion} ~\cite{wang2024instancediffusioninstancelevelcontrolimage}, {\gligen}~\cite{li2023gligenopensetgroundedtexttoimage}, {\boxdiff} \cite{xie2023boxdifftexttoimagesynthesistrainingfree}, and {\lmdPlus} \cite{lian2024llmgroundeddiffusionenhancingprompt} for each layout generator.    
     Our method can achieve optimal performance across all measured metrics and achieves state of the art. }
    \label{tab:traditional_metric_comparison}
\end{table*}

\subsection{Experimental Setting}\label{subsec:experimental_setting}
For a fair comparison, we train all models on the ADE20K dataset \cite{zhou2018semanticunderstandingscenesade20k}, which contains approximately 27K images and ground-truth layouts for indoor and outdoor scenes and a rich collection of object arrangements. The sample captions reflect the scene category with no additional constraints  (e.g., \emph{beach}, \emph{lecture room}).
We use the top $30$ largest bounding boxes per sample, as this is the default maximum number of bounding boxes supported by {\instanceDiffusion} ~\cite{wang2024instancediffusioninstancelevelcontrolimage}, padding samples with fewer bounding boxes. For evaluation, we use the $15$ highest represented categories and add in five randomly selected categories to include the dataset's long tail distribution. For each evaluated model, we generate $30$ layouts for all $20$ selected prompts, and an image conditioned on each layout and corresponding global prompt. The size of this collection of images is feasible to assess with our human evaluation.

We adapt {\layoutTransformer} ~\cite{gupta2021layouttransformerlayoutgenerationcompletion}, {\layoutDiffusion} ~\cite{inoue2023layoutdmdiscretediffusionmodel} and {\layoutFlow} ~\cite{guerreiro2024layoutflowflowmatchinglayout} from the UI generation setting by treating the global caption as a scene-wide bounding box and conditioning the model on this bounding box during inference.
As we demonstrate in ~\cref{fig:layout_gpt_comparison}, existing LLM-based approaches \cite{lian2024llmgroundeddiffusionenhancingprompt,feng2023layoutgptcompositionalvisualplanning} do not succeed off-the-shelf in our task. However, by adjusting the template of LLM-Grounded Diffusion \cite{lian2024llmgroundeddiffusionenhancingprompt} with relevant in-context-learning examples from ADE20K, we avoid degenerate solutions to have a fair comparison. For the underlying LLM, we select the recent large-scale LLM {\gptFourO}~\cite{openai2024gpt4technicalreport}. 

\subsection{Generated Image Evaluation}\label{subsec:generated_image_metrics_results}

\textbf{Text-to-Image Metrics.} To evaluate the generated images, we apply the established image generation metrics CMMD~\cite{jayasumana2024rethinkingfidbetterevaluation}, FID~\cite{heusel2018gans}, KID~\cite{binkowski2021demystifyingmmdgans}, VQA ~\cite{lin2024evaluatingtexttovisualgenerationimagetotext}, HPSv2 ~\cite{wu2023humanpreferencescorev2}, and ImageReward ~\cite{xu2023imagerewardlearningevaluatinghuman}. CMMD, FID and KID compare the distribution of generated images with a ground-truth distribution, while VQA, HSPv2 and ImageReward assess general image quality and alignment with a global caption. Since the conditioned image generator may itself lead to biases in image generation quality, for CMMD, FID and KID we establish the ground-truth images by running the layout-to-image generator on the ground-truth layouts. Furthermore, as different image generators are also biased towards different layout generators, for each layout generator we choose the optimal score over the possible combinations of layout and image generator.
The results are shown in~\cref{tab:traditional_metric_comparison}. As visible, our method outperforms all other methods and achieves state of the art. 

\begin{figure}[ht]
    \centering
    \includegraphics[scale=0.17]{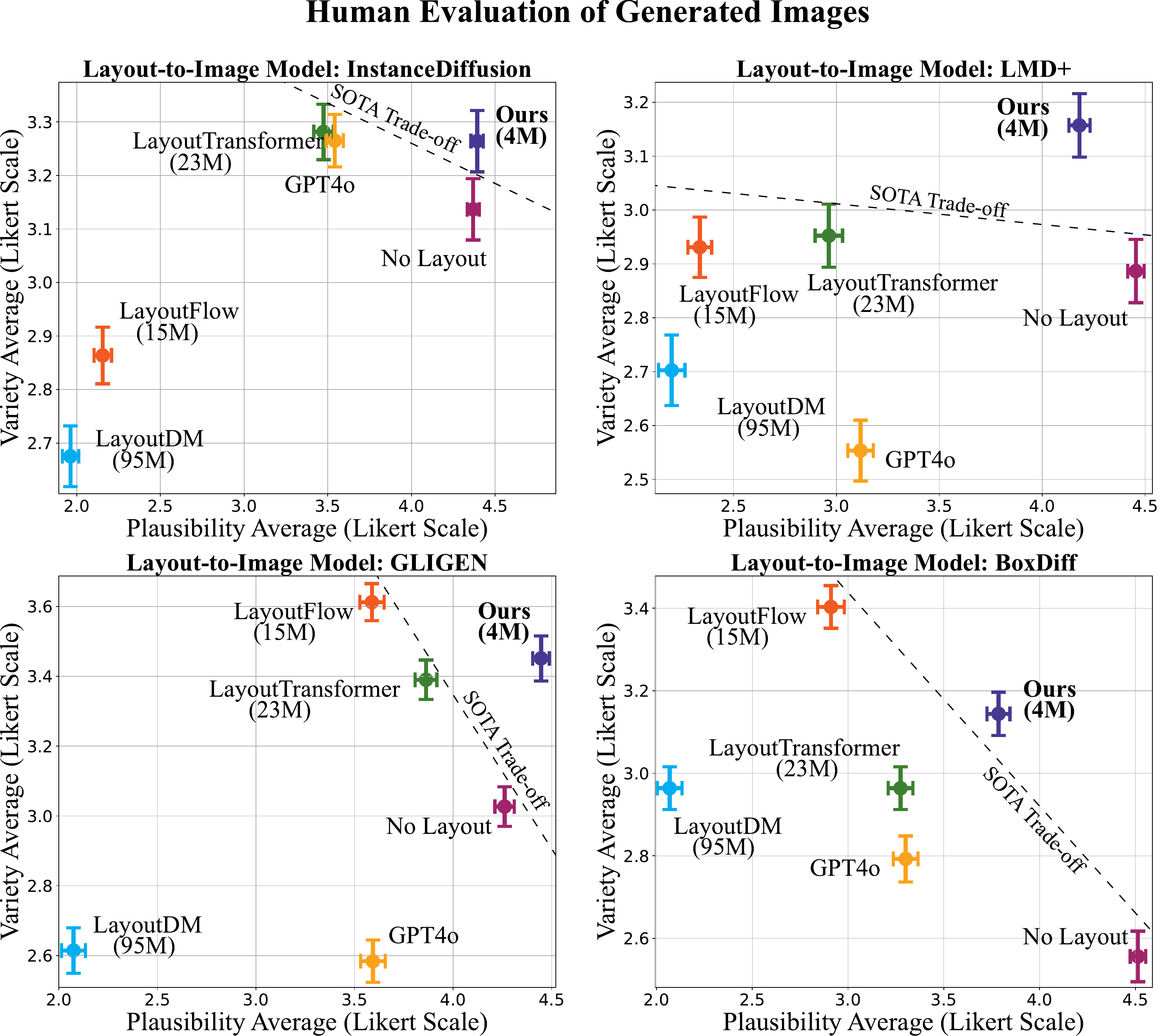}\caption{\textbf{Human Survey Results.} Our method offers a superior trade-off between plausibility and variety across all measured layout-to-image generators, while being a much smaller model. The error bars indicate the standard errors of the means and show that the results are statistically significant. 
    }
    \label{fig:human_survey}
\end{figure}

\noindent \textbf{Human Evaluation.} As shown in ~\cref{fig:human_survey}, our model achieves a state-of-the-art balance in image plausibility and variety as well. We display the approximate number of model parameters added to the full text-to-layout-to-image pipeline by the layout generators that can be locally run. Our model is the smallest by over a factor $3$.

\noindent \textbf{Visual Results.}\label{subsec:visual_results}
We provide a qualitative overview of the generated layouts and the final images in~\cref{fig:visual_results}.
Due to space limitations, we display the baselines that performed best in ~\cref{tab:traditional_metric_comparison}. We label bounding boxes by querying with all text labels present within ADE20K. From a visual inspection, {\layoutTransformer} struggles with arranging objects spatially, leading to images with unclear structure that do not depict the global prompt. {\gptFourO} layouts appear somewhat flat, while struggling to make a variety of  layouts. Our method appears to produce both plausible and diverse images across a range of global prompts of indoor and outdoor settings.

\begin{figure*}[ht]
\setlength{\tabcolsep}{2pt}

        \begin{tabular}{ c c c | c c c}
        {\small \textbf{Partial Conditioning}} &
        {\small \textbf{Generated Layout}} &
        {\small \textbf{Generated Image}}&
        {\small \textbf{Partial Conditioning}} &
        {\small \textbf{Generated Layout}} &
        {\small \textbf{Generated Image}}\\
        \includegraphics[width=0.16\textwidth]{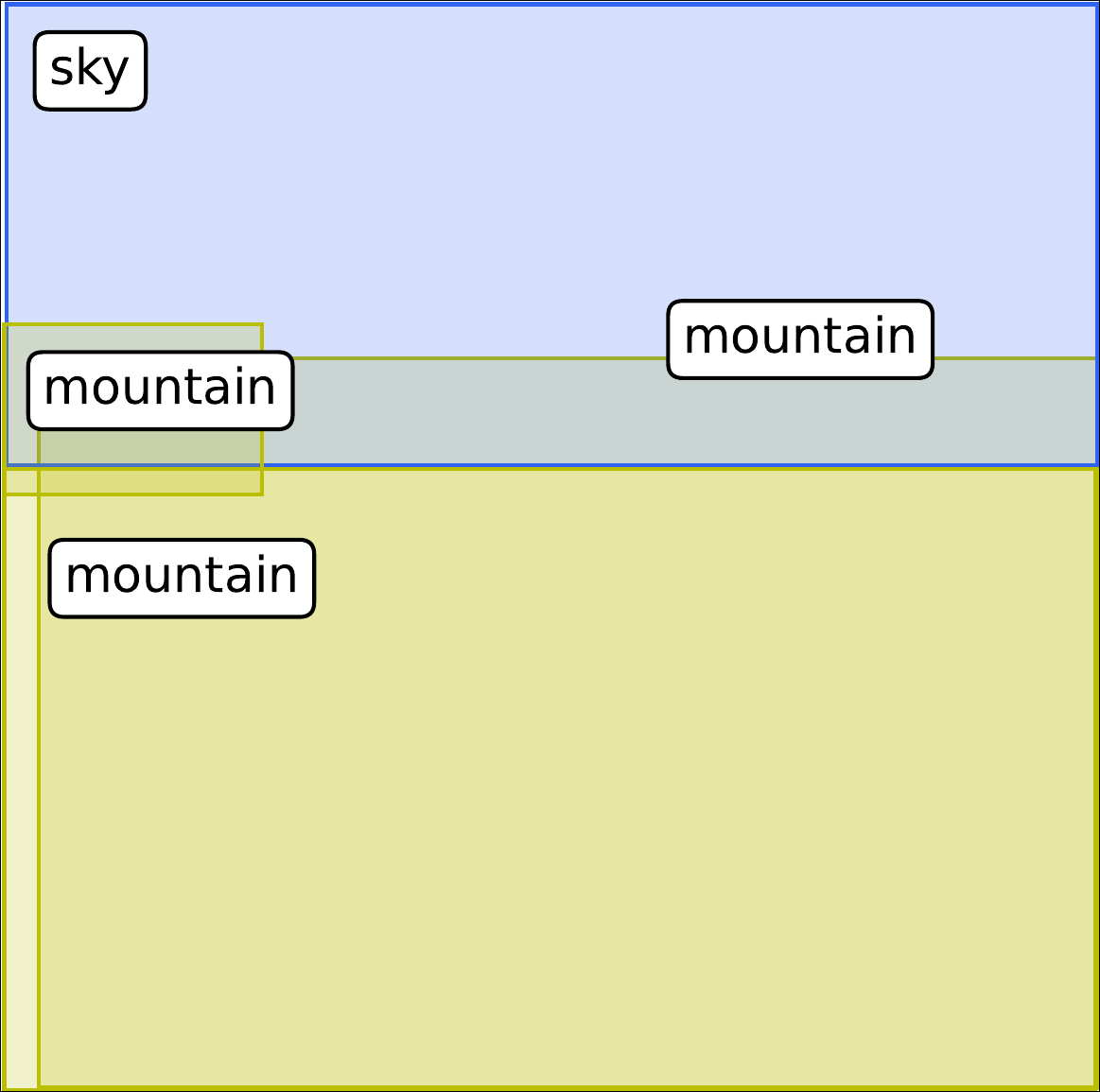} & %
        \includegraphics[width=0.16\textwidth]{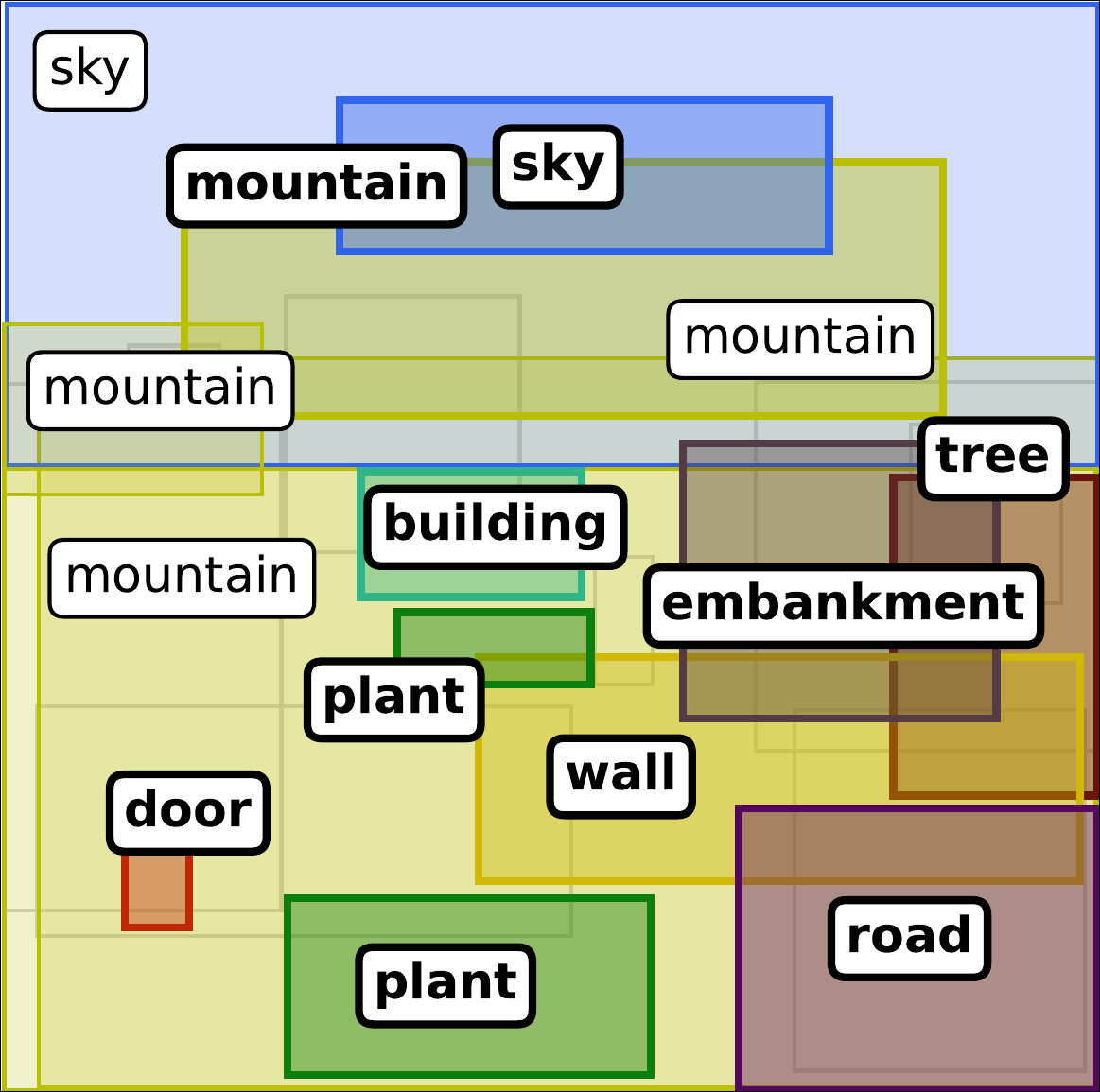} & %
        \includegraphics[width=0.16\textwidth]{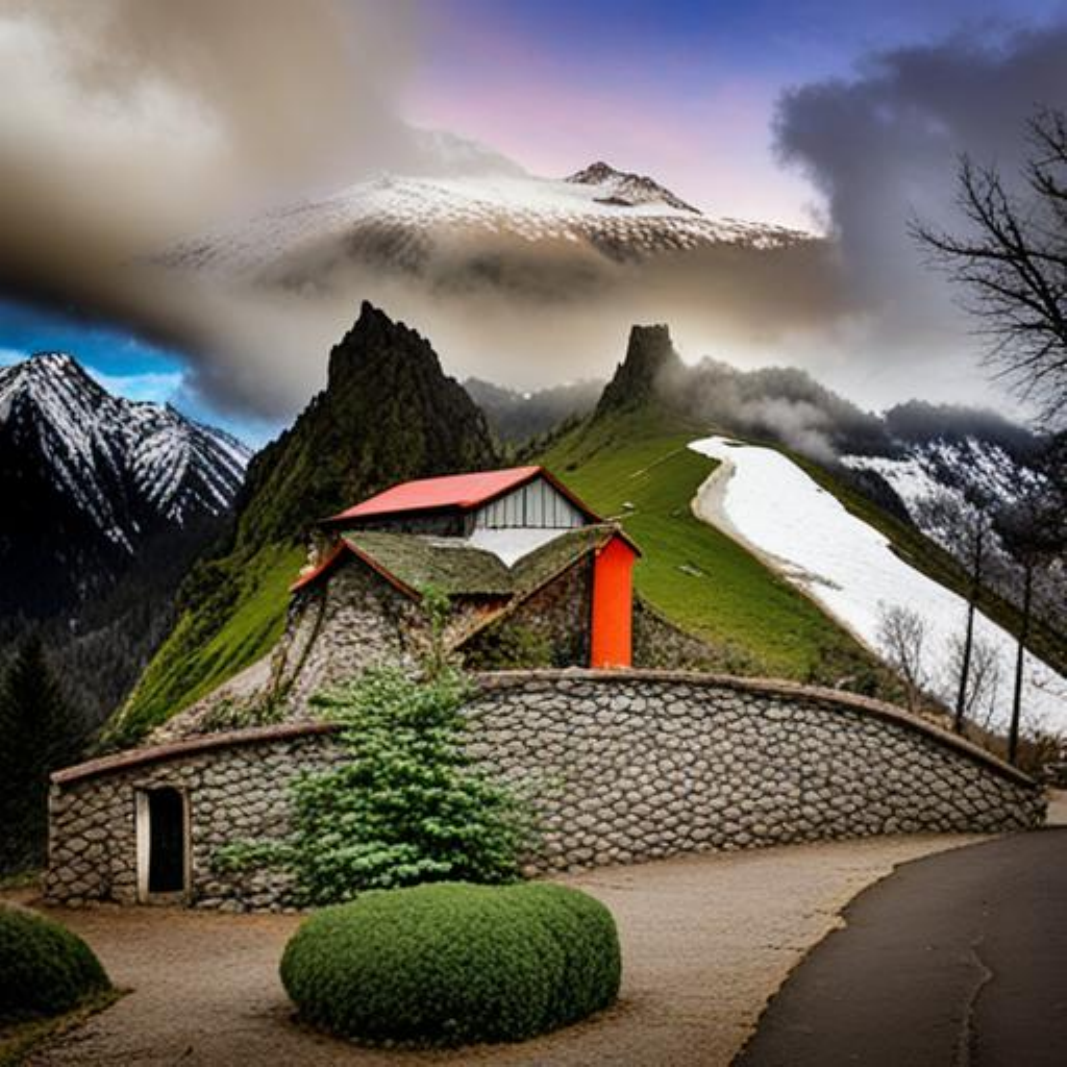} & %
        \begin{minipage}[b]{0.16\textwidth}{\footnotesize
        The scene should contain these objects:\vfill
        \begin{tabular}[b]{l l}
            \tabitem \textbf{ceiling}&
            \tabitem \textbf{wall} (2x) \\
            \tabitem \textbf{curtain}&
            \tabitem \textbf{sink}\\
            \tabitem \textbf{floor}&
            \tabitem \textbf{bathtub}\\
            \multicolumn{2}{l}{\tabitem \textbf{windowpane}}\\

        \end{tabular}}
        \vspace{0.2cm}
        \end{minipage} & %
        \includegraphics[width=0.16\textwidth]{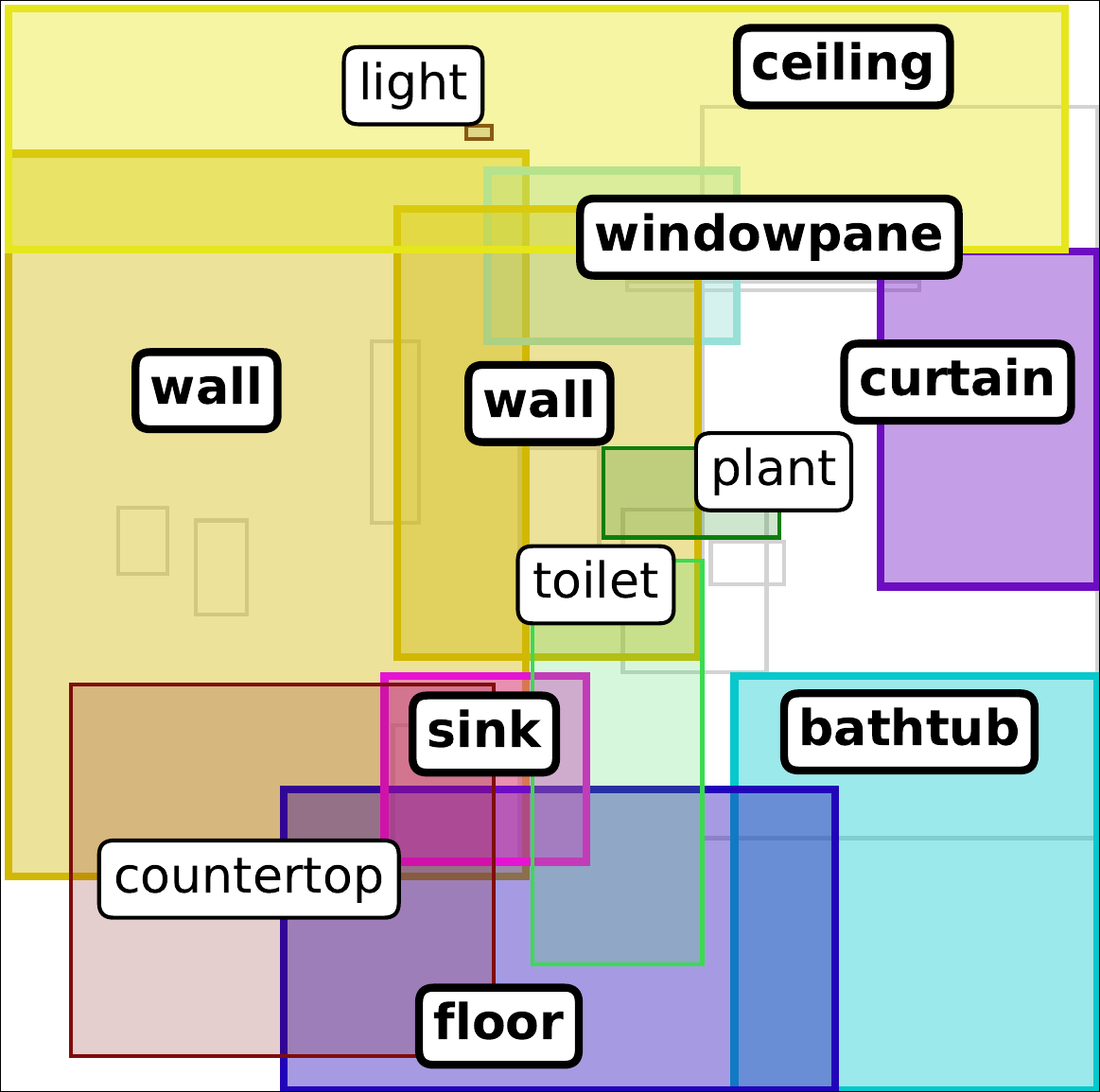} &
        \includegraphics[width=0.16\textwidth]{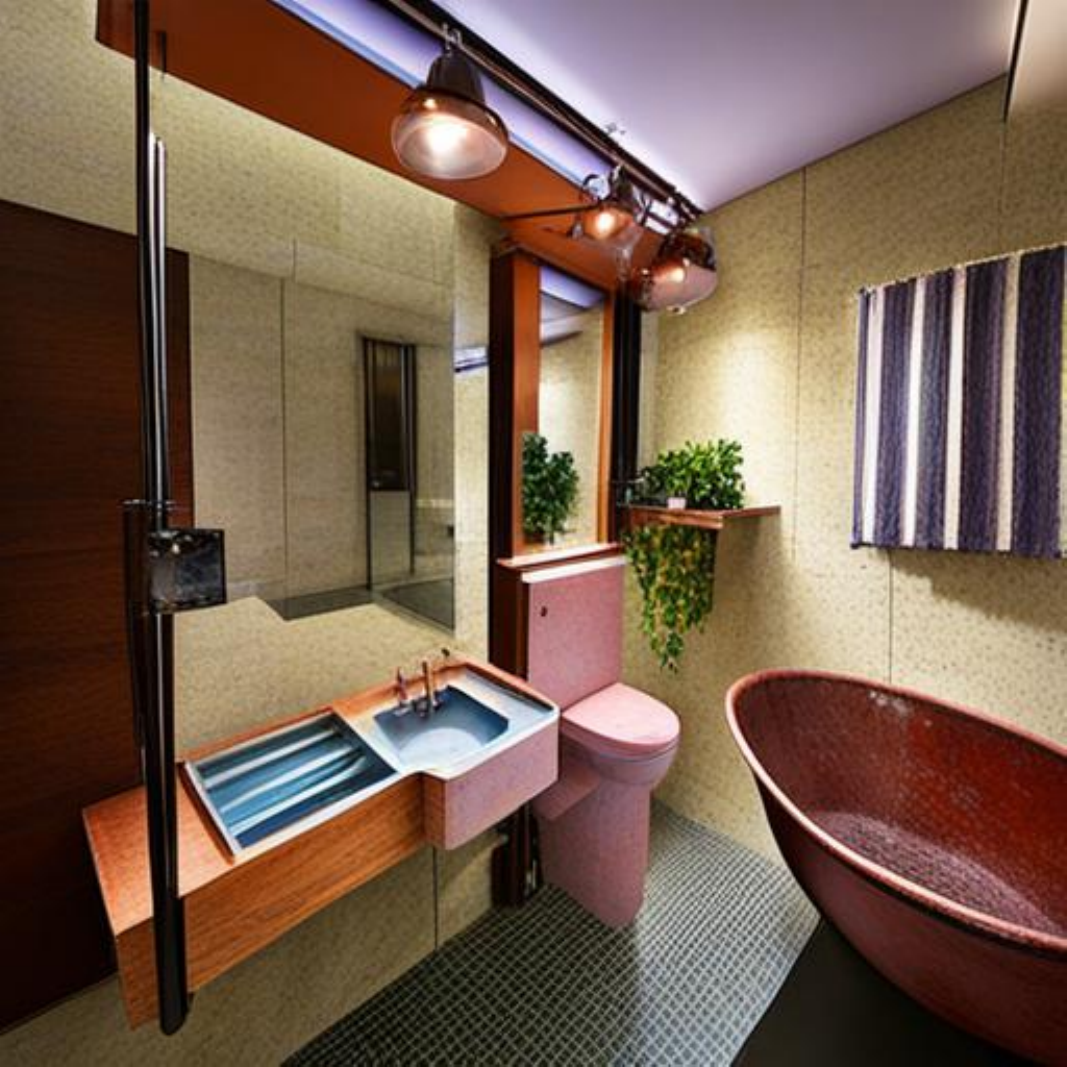} \\ 
    \end{tabular}
    \caption{\textbf{Disentangled Generation.} We show examples of two modes of disentangled generation for scenes with the prompt \emph{Snowy Mountain}, and \emph{Bathroom}, respectively. 
    As visible, our model is able to complete the initial layouts well. 
    }
    \label{fig:disentanglement}
\end{figure*}

\subsection{Layout Evaluation and Inference Speed}\label{subsubsec:numerical_metrics_results}

We evaluate our method on the layout metrics proposed in section~\cref{subsubsec:numerical_metrics} in~\cref{tab:layout_only_metric_comparison}.  
While {\layoutTransformer} outperforms our model on the variety measure $\positionalVarianceScore$, we observe in \cref{fig:visual_results} that the layouts lack plausibility. For example, the floor in the leftmost example appears at the top and the ashcan on the rightmost example is significantly too large. 

We also measure the average time it takes for each model to generate a batch of $30$ layout samples on an Nvidia A6000 GPU in~\cref{tab:layout_only_metric_comparison}. As we do not have direct access to {\gptFourO}, we query it sequentially through an API. 
Our method ranks second in speed. Although {\layoutFlow}'s default settings enable faster inference, we observe no definitive improvement in its layout statistics when the number of inference steps are raised to match our model's speed.

\begin{table}
    \vspace*{-5pt}
    \centering
    \footnotesize
    \setlength{\tabcolsep}{3pt} 
    \renewcommand{\arraystretch}{0.9} 
    \begin{tabular}{l | r r r r r| r}
            {\textbf{Model}} 
             &  {$\objectNumeracyScore$}
             &  {$\positionalVarianceScore$}
             &  {$\firstOrderPositionalLikelihood$}
             &  {$\secondOrderPositionalLikelihood$}
             &  {$\mIoU$}
             &  {Time (s)}\\
             \midrule 
            GPT4    & 3.71  & 93  & 4.17  & 1.42 & 0.10   & 111.0 \\ 
             \midrule 
            LayoutDiffusion   & 2.12  & 65  & 1.47  & 0.71 & 0.00  & 138.0 \\ 
             \midrule 
            LayoutFlow    & 3.01  & 142  & 1.48  & 0.72  & 0.01   & \textbf{0.5} \\ 
             \midrule 
            LayoutFlow (More steps)  & 2.96  & 143  & 1.44  & 0.65 & 0.01    & \underline{15.5} \\ 
             \midrule 
            LayoutTransformer    & \textbf{0.90}  & \textbf{231}  & 3.09  & 1.21 & 0.15    & 25.0 \\ 
             \midrule 
            Ours    & \underline{1.14}  & \underline{187}  & \textbf{4.76}  & \textbf{1.93} & \textbf{0.17}   & \underline{15.5}
    \vspace*{-7pt}
    \end{tabular}
    \caption{\textbf{Layout Metrics, and Inference Speed.} 
    A comparison of our metrics introduced in \cref{subsubsec:numerical_metrics}. $\objectNumeracyScore$ is scaled by $10^{2}$.
    Our method achieves the best on $\mIoU$, $\firstOrderPositionalLikelihood$, and $\secondOrderPositionalLikelihood$, and is behind {\layoutTransformer} in $\objectNumeracyScore$ and $\positionalVarianceScore$.
    The closer inspection in~\cref{subsec:generated_image_metrics_results}
    reveals that {\layoutTransformer} falls short in terms of plausibility and image quality, indicating that it generates a large variety but implausible layouts. 
    }
    \label{tab:layout_only_metric_comparison}
\end{table}

\subsection{Scaling to MSCOCO}\label{subsec:scaling_to_mscoco}

\begin{figure}
        \setlength{\tabcolsep}{1pt}
        \begin{tabular}{ c c | c c}
    
            \includegraphics[width=\mscocoGraphicsWidth]{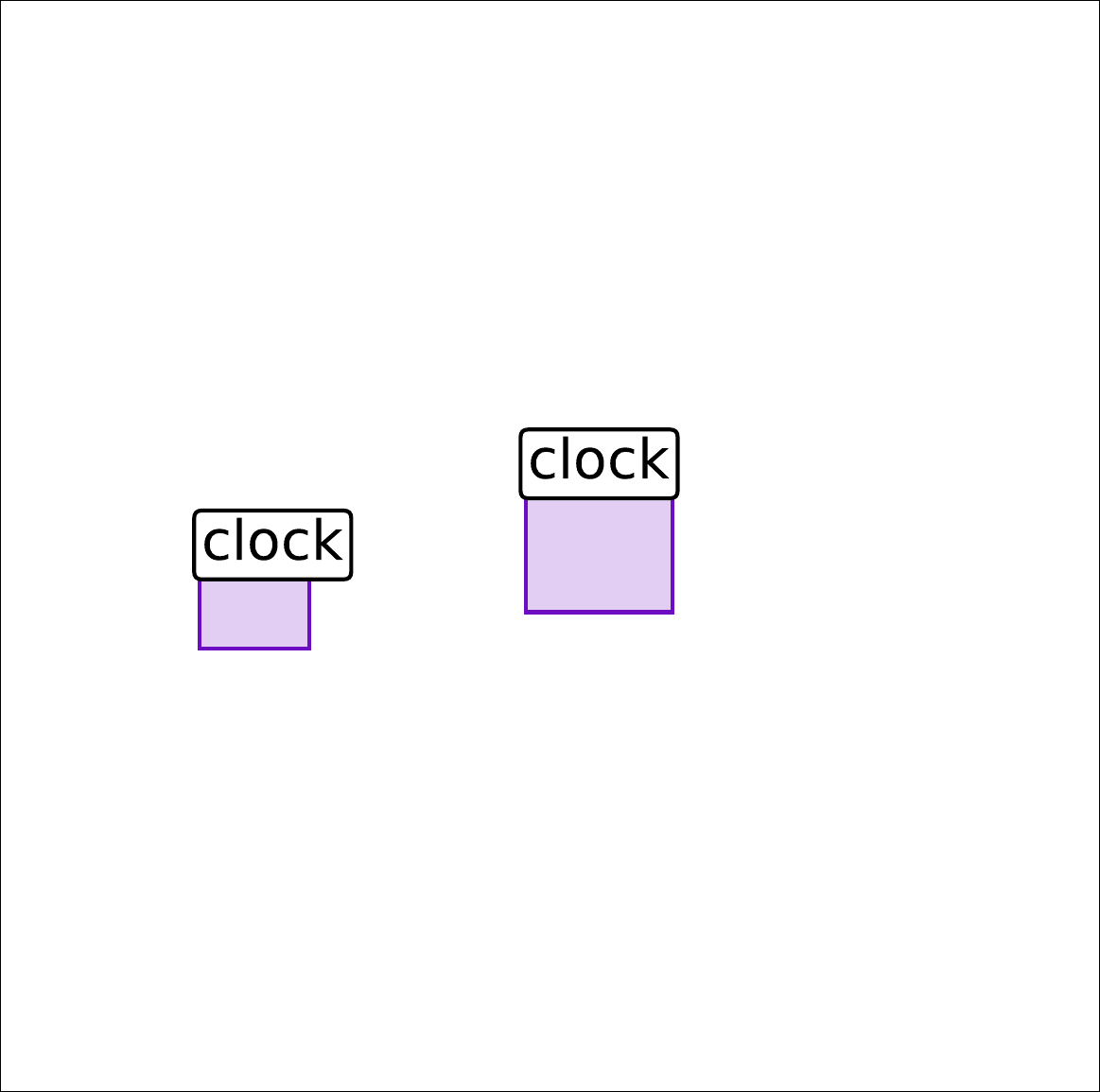} &
            \includegraphics[width=\mscocoGraphicsWidth]{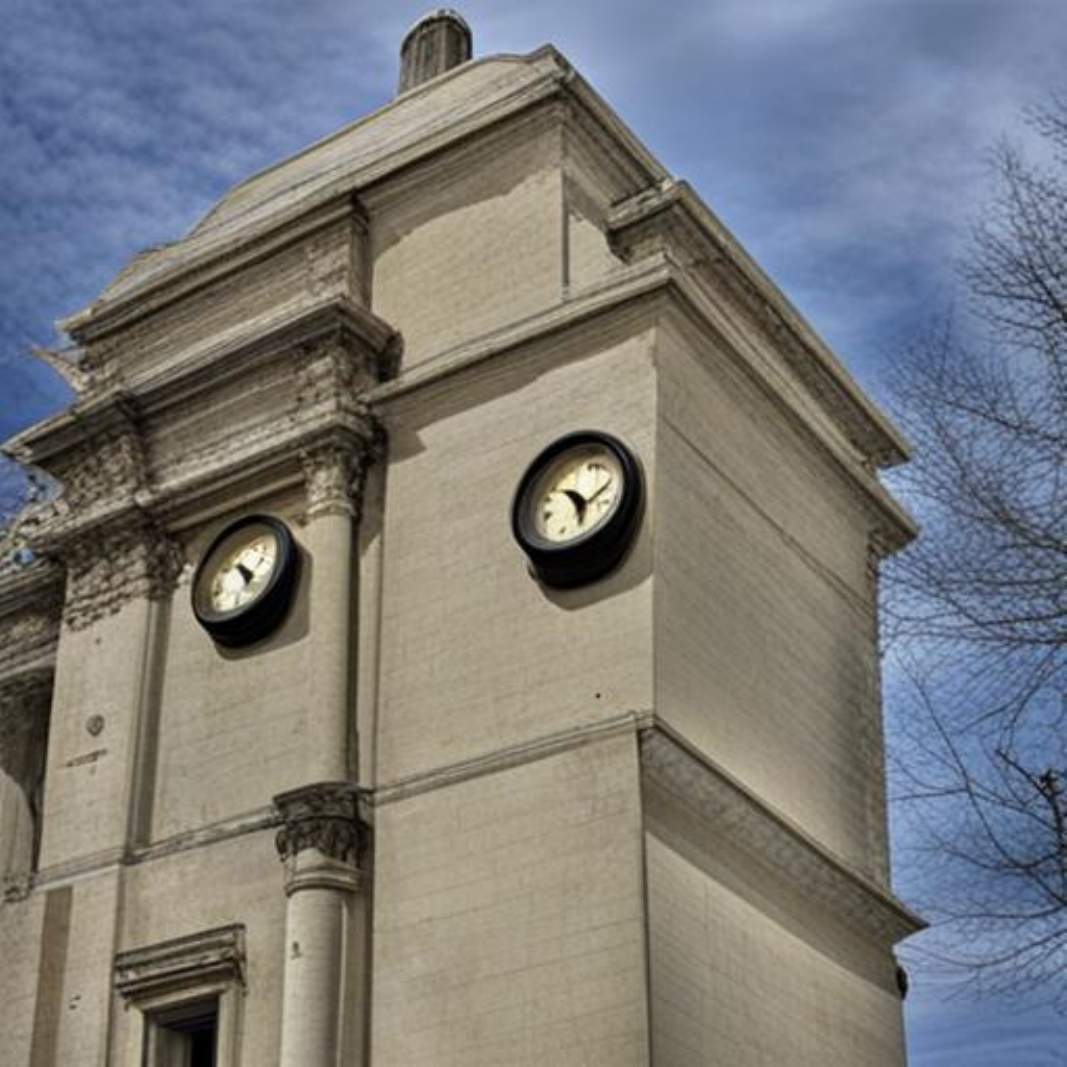} &
            \includegraphics[width=\mscocoGraphicsWidth]{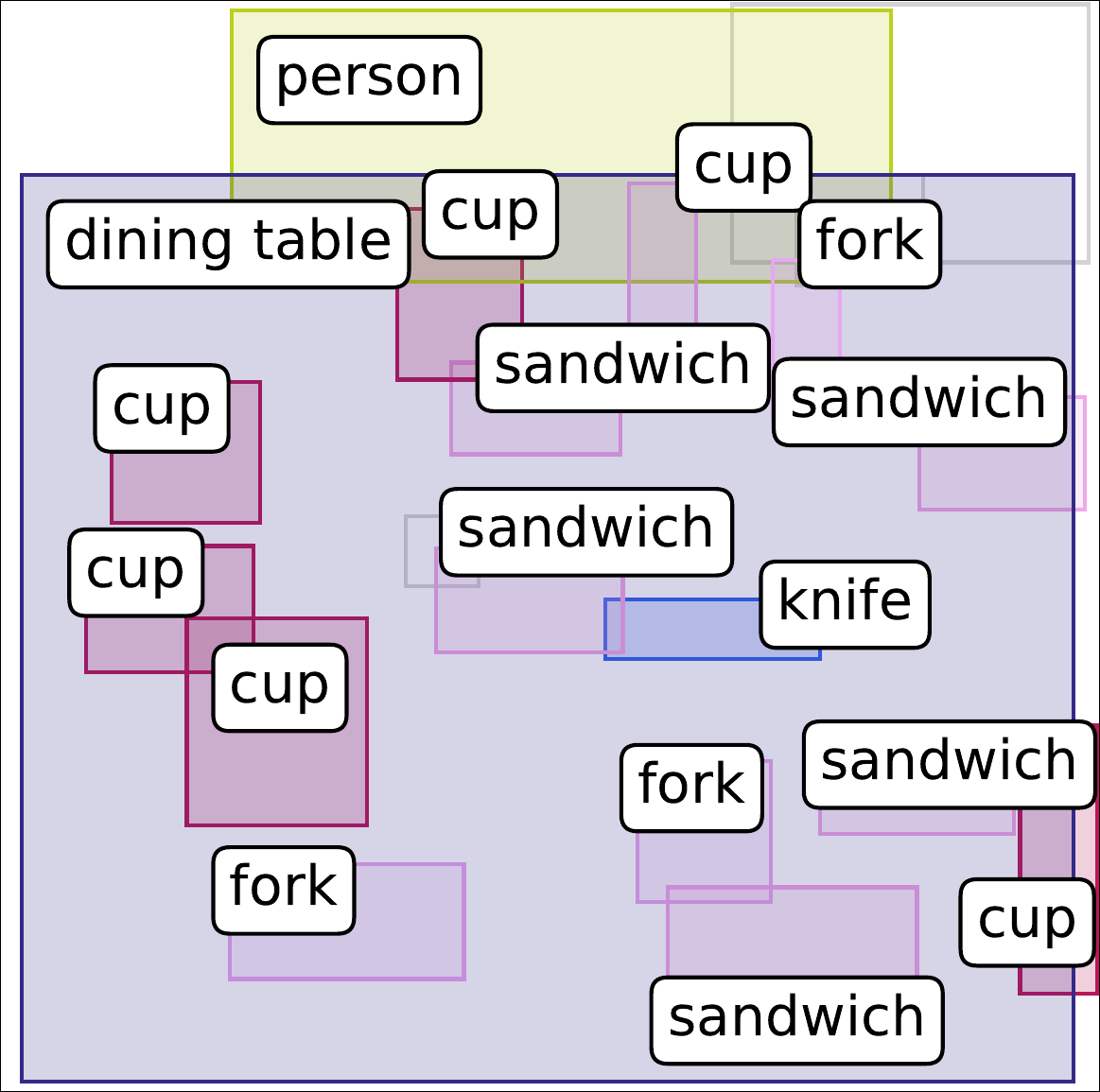} &
            \includegraphics[width=\mscocoGraphicsWidth]{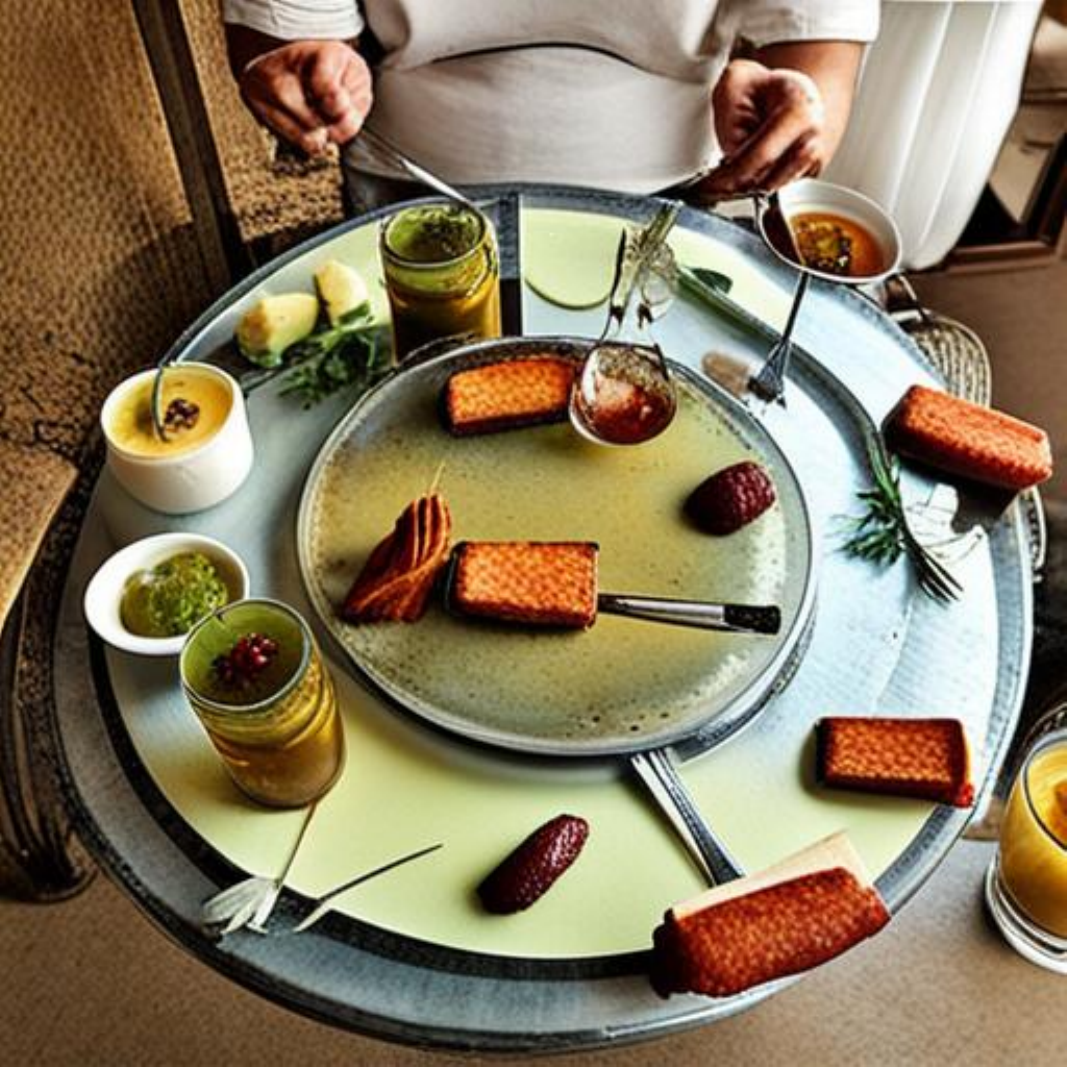} 
            \\
            \hline
            \includegraphics[width=\mscocoGraphicsWidth]{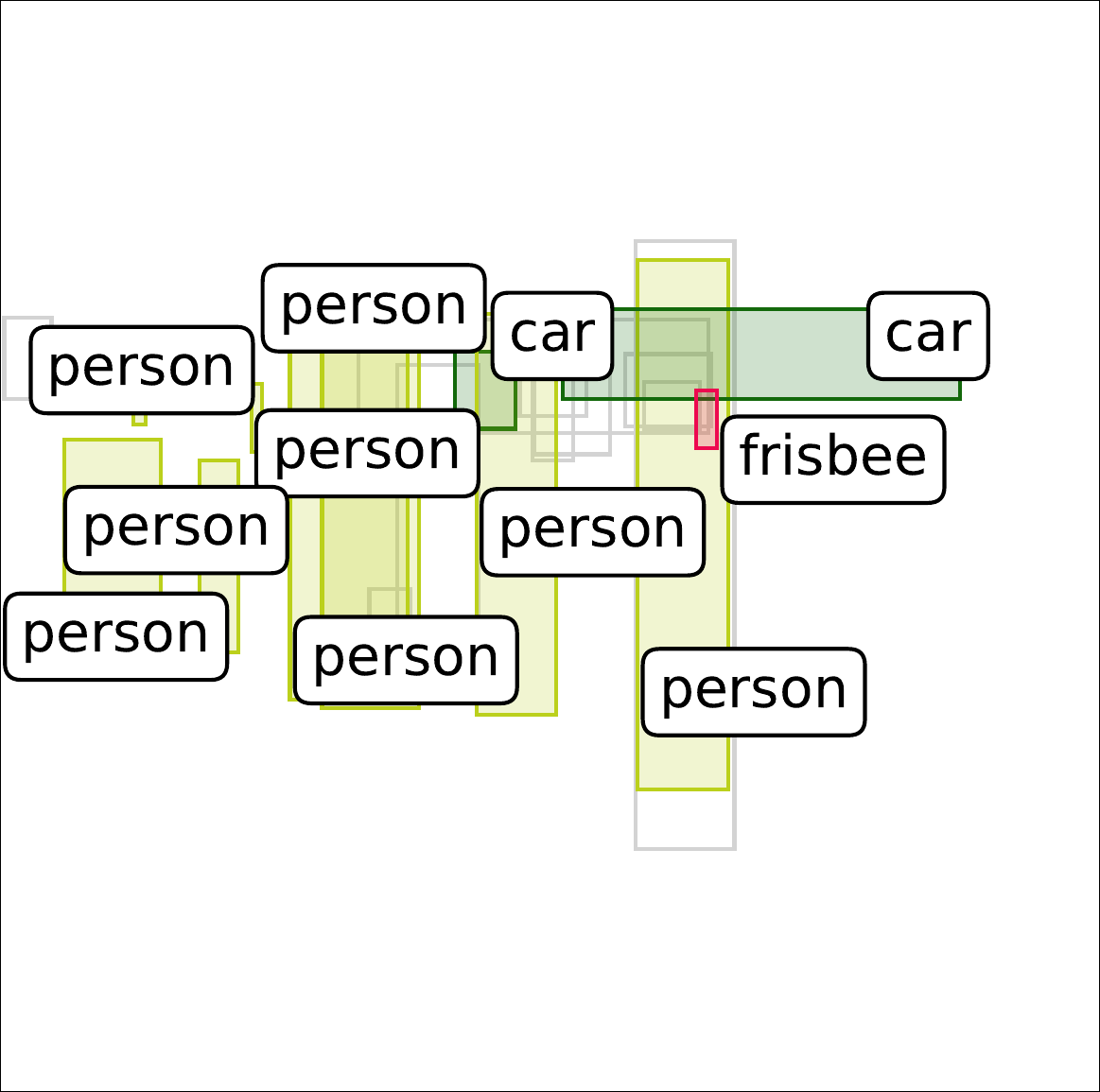} &
            \includegraphics[width=\mscocoGraphicsWidth]{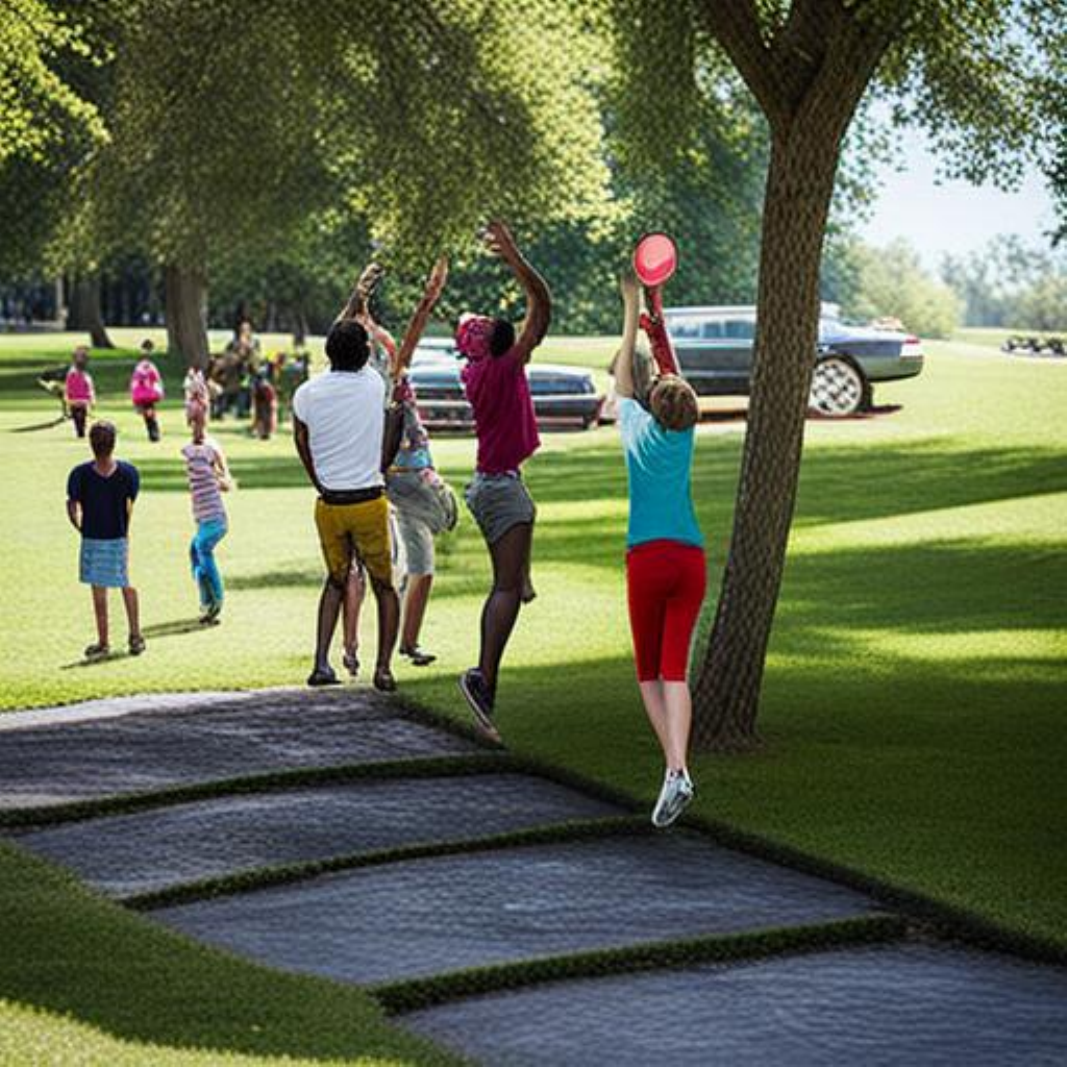} &
            \includegraphics[width=\mscocoGraphicsWidth]{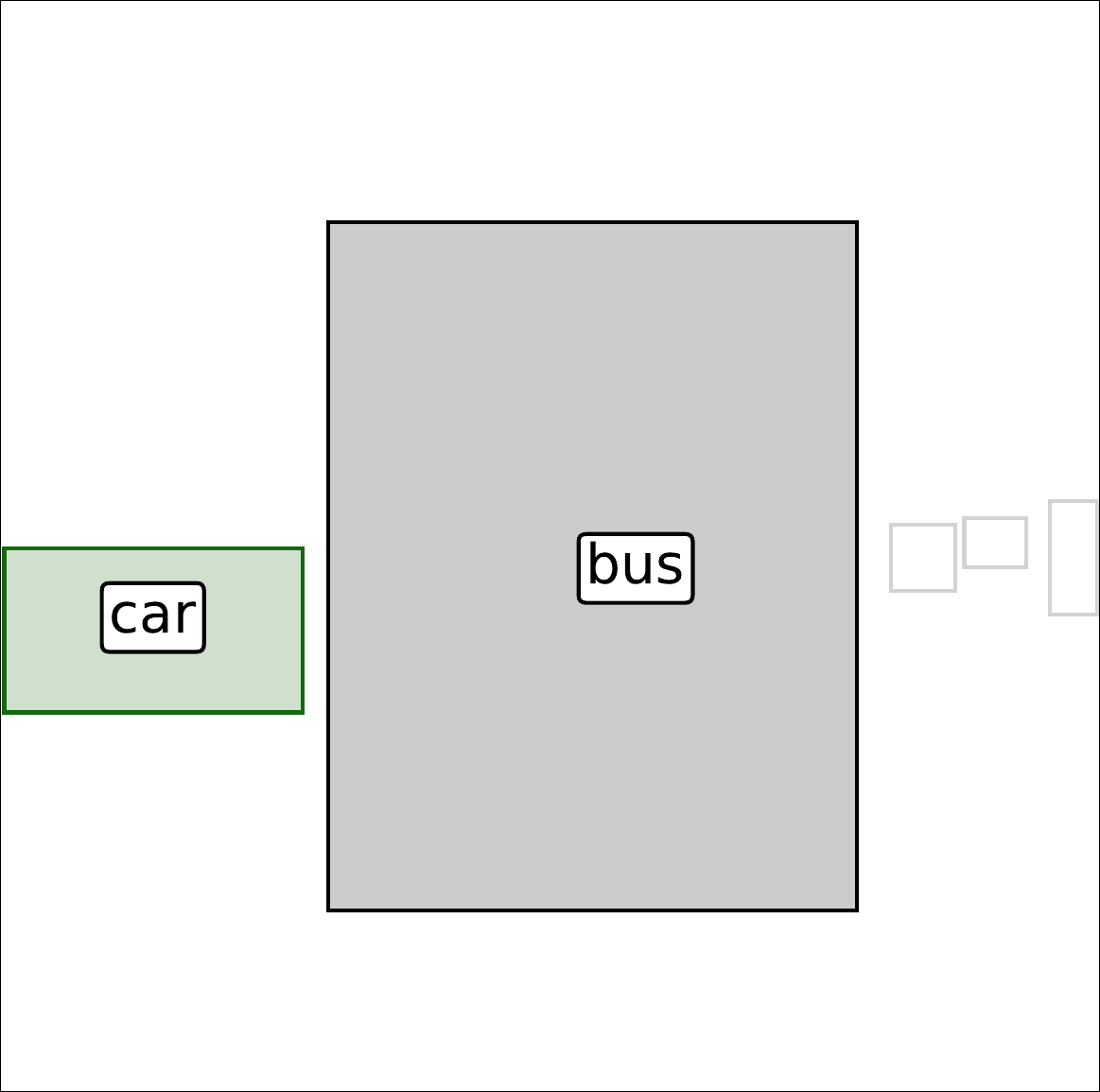} &
            \includegraphics[width=\mscocoGraphicsWidth]{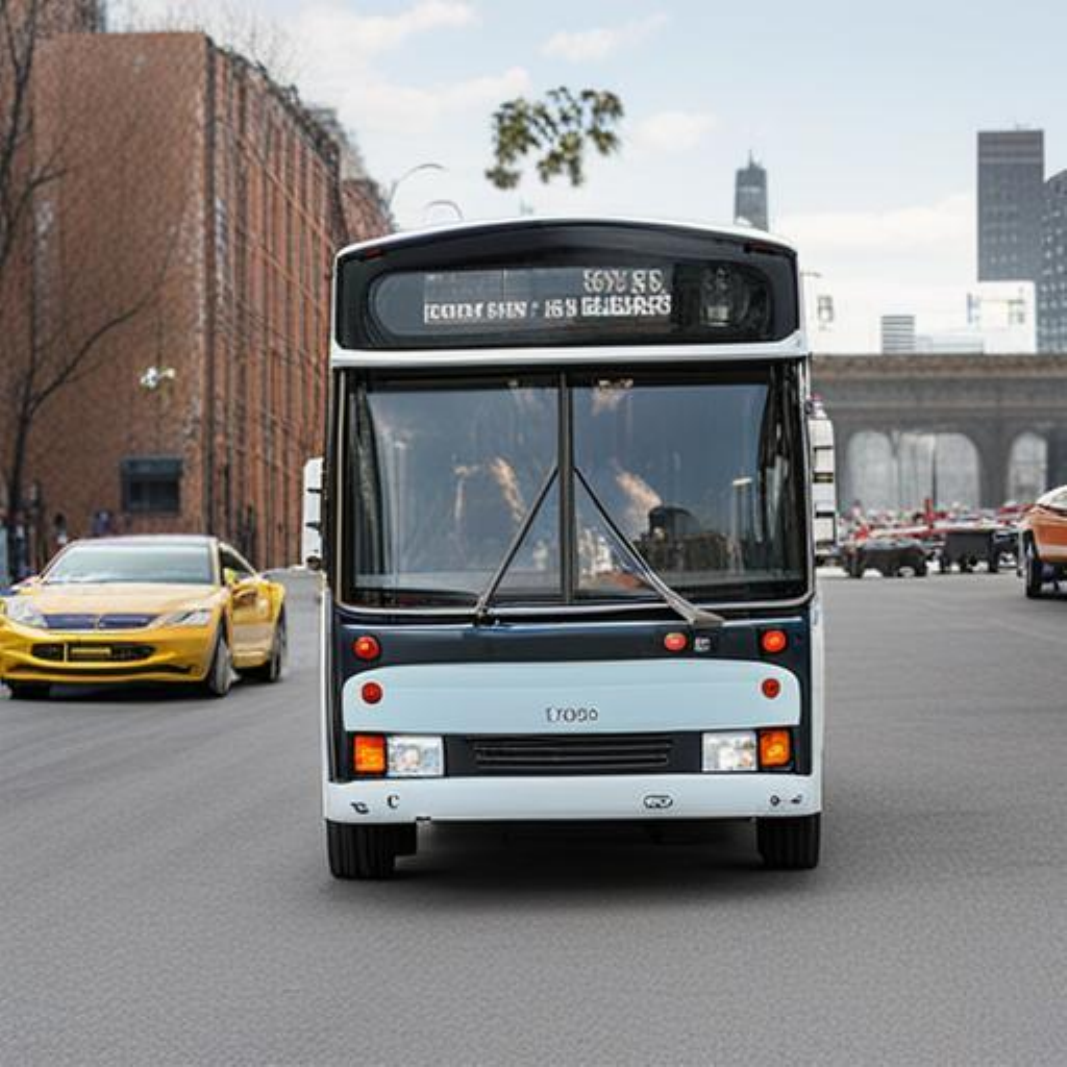}
            \\
    
        \end{tabular}
        \caption{\textbf{MSCOCO Trained Model:} Layouts and generated images from the prompts:
        \emph{An image with two clocks},
        \emph{The food is ready to be eaten on the table},
        \emph{A group of teens playing frisbee in a park}, and \emph{A car to the left of a bus}}
        \label{fig:mscoco_results}
    \end{figure}

MSCOCO contains approximately 125K samples that consist of an image, the annotated scene layout, and five alternative global captions that are full sentence descriptions (e.g. "A black cat laying on top of a bed next to pillows"). We explore how our model scales to MSCOCO \cite{lin2015microsoftcococommonobjects} by training on this dataset using the same architecture and hyperparameters as our main model. We provide some qualitative examples in~\cref{fig:mscoco_results} and show that we are able to generate appealing images for the complex prompts with our method.  

\subsection{Additional Model Features}\label{subsection:additional_model_features}

At last, we present the additional features that our model offers that make it appealing for use in an end-to-end image generation pipeline. In ~\cref{fig:disentanglement}, we show examples of our model's performance in different partial layout generation settings. This feature gives users even more fine-grained control over the image generation process. 

\begin{figure}[ht]
    \begin{tabular}{c c c}
        \textbf{User Action} &
        \textbf{Image Layout} &
        \textbf{Generated Image} \\
        \toprule
        \begin{minipage}[b]{\editingResultsGraphicsWidth}{\footnotesize A layout for a ``Conference Room'' with a ``plant'' is generated. The generated image has a plant in the correct position.\vfill\vspace{0.25cm}}\end{minipage} & \includegraphics[width=\editingResultsGraphicsWidth]{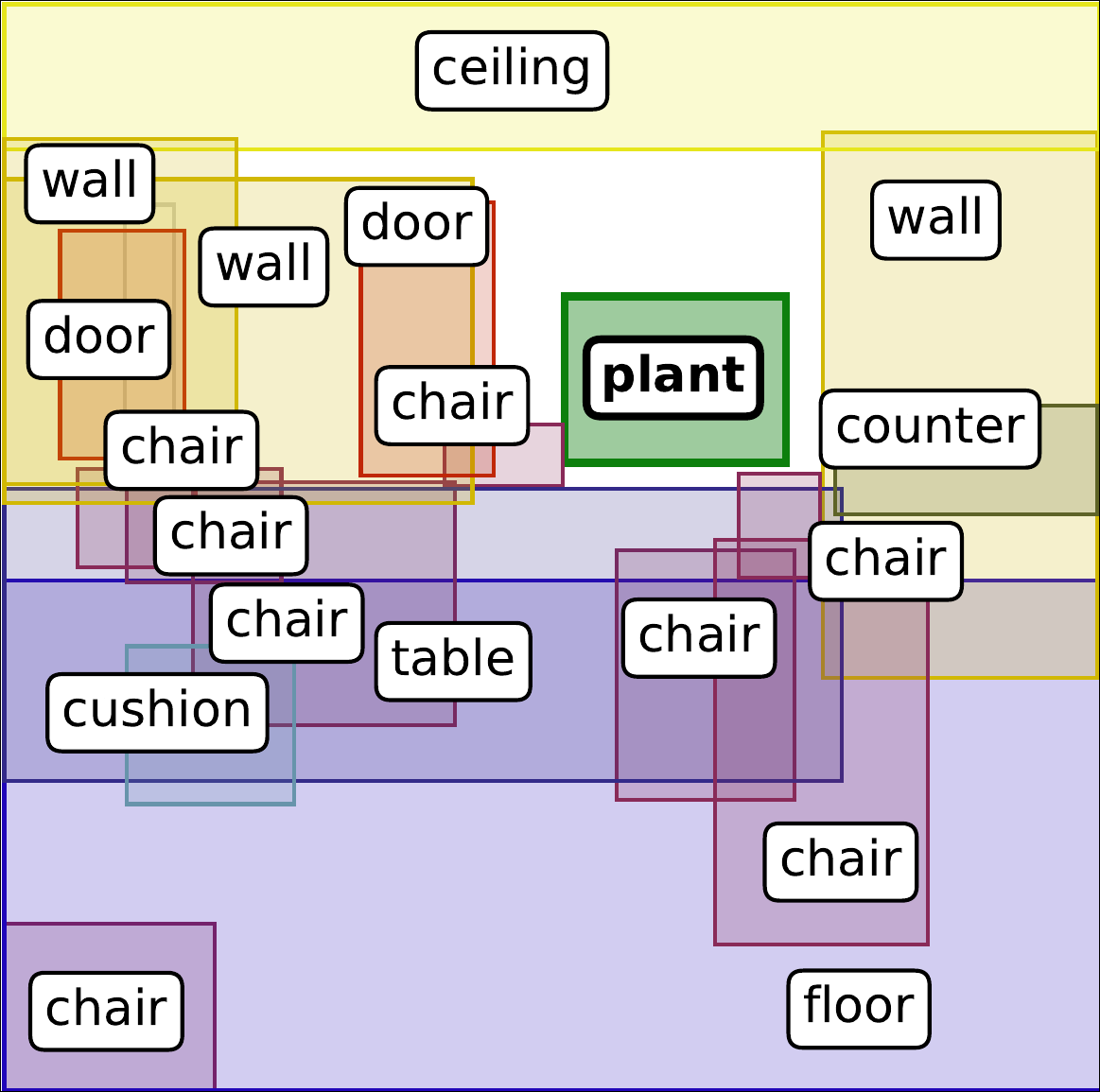} &
        \includestandalone[width=\editingResultsGraphicsWidth]{sections/results/figures/editing/tikz_visuals/original_plant} \\
        \midrule 
        \begin{minipage}[b]{\editingResultsGraphicsWidth}{\footnotesize The ``plant'' is moved. The generated image is faithful to the new layout.\vfill \vspace{0.55cm}}\end{minipage} &
        \includegraphics[width=\editingResultsGraphicsWidth]{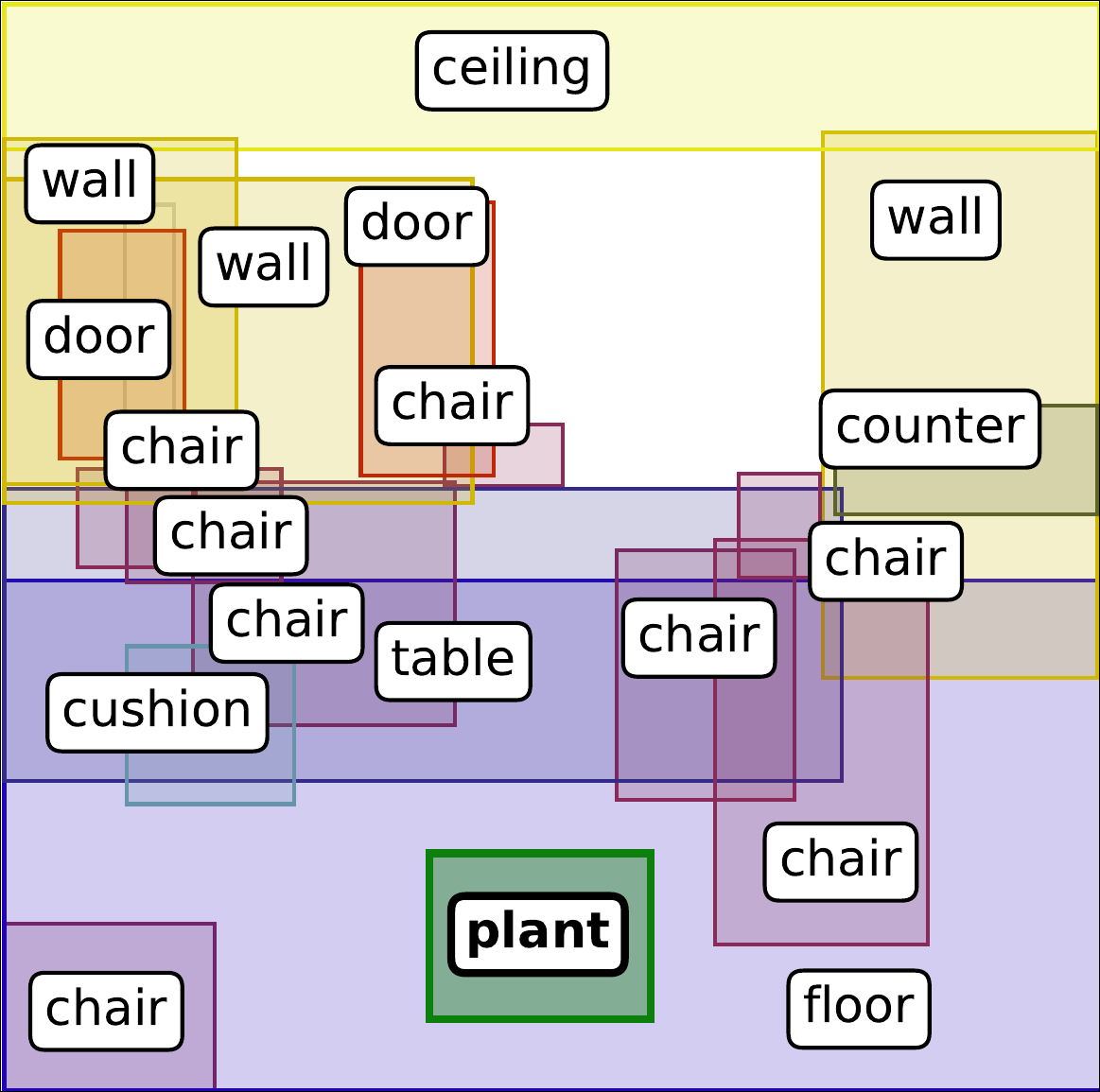} &
        \includegraphics[width=\editingResultsGraphicsWidth]{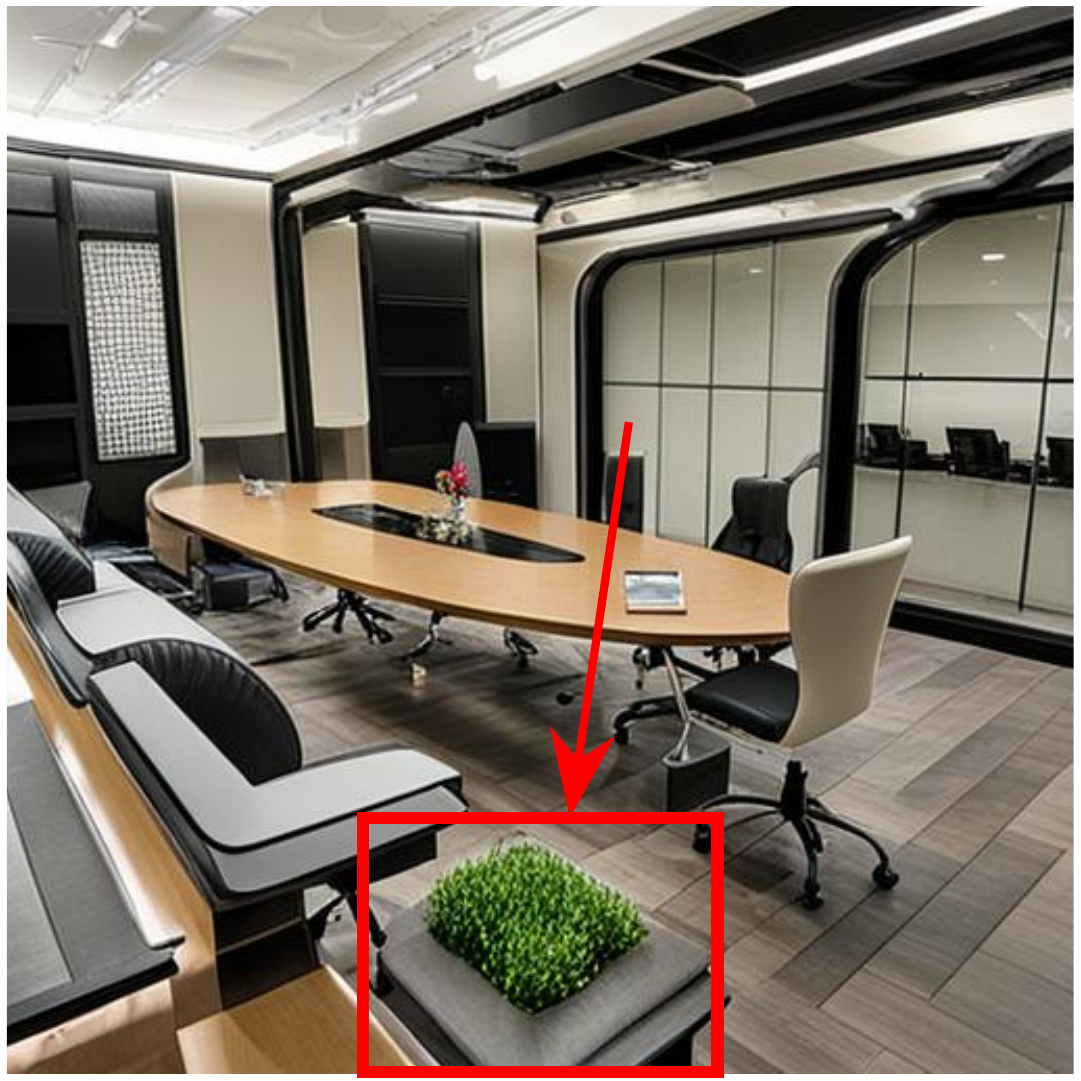} \\
        \midrule 
        \begin{minipage}[b]{\editingResultsGraphicsWidth}{\footnotesize The ``plant'' bounding box is removed entirely. The plant  is also removed in the generated image.\vfill\vspace{0.5cm}}\end{minipage} &
        \includegraphics[width=\editingResultsGraphicsWidth]{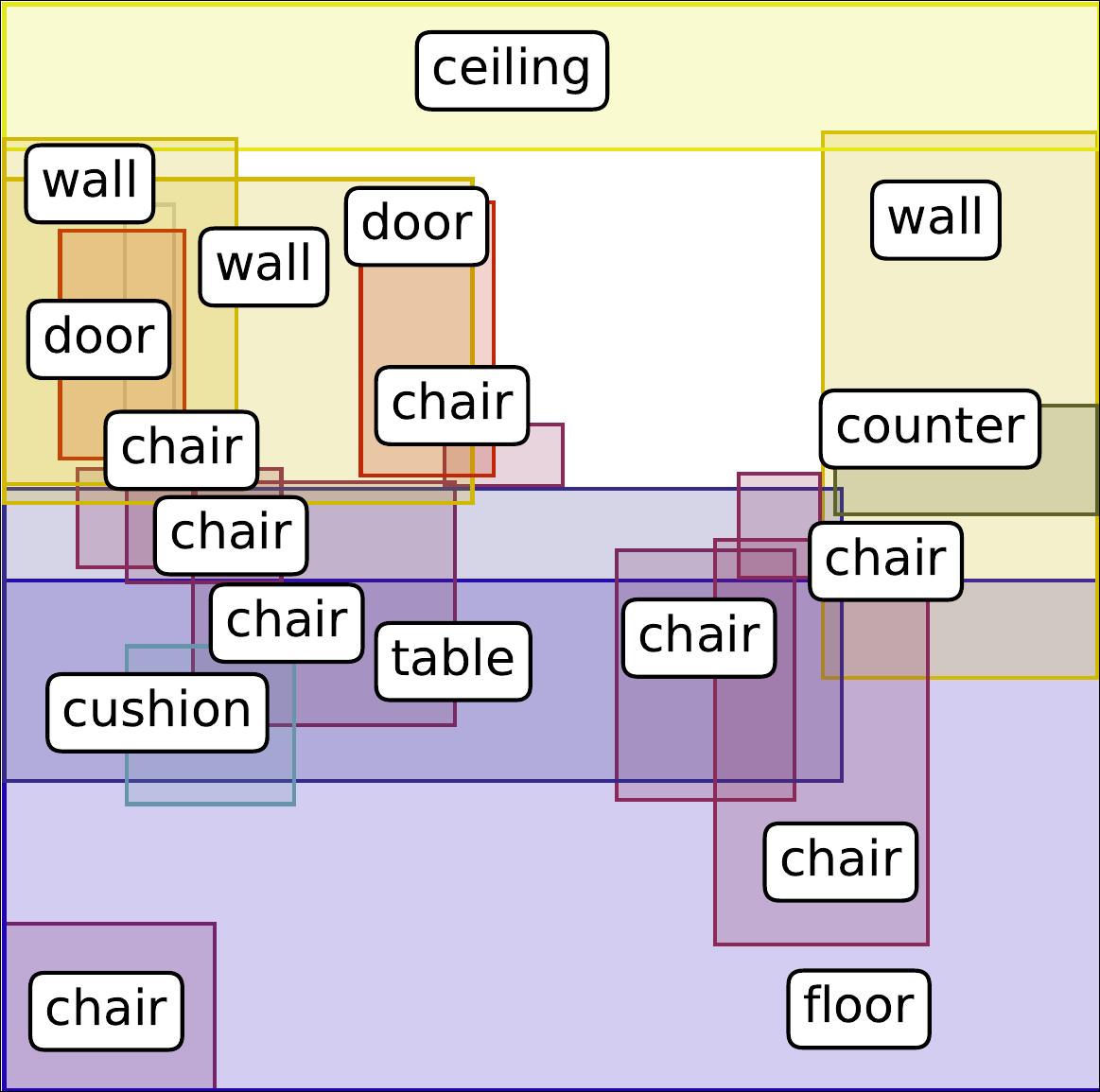} &
        \includegraphics[width=\editingResultsGraphicsWidth]{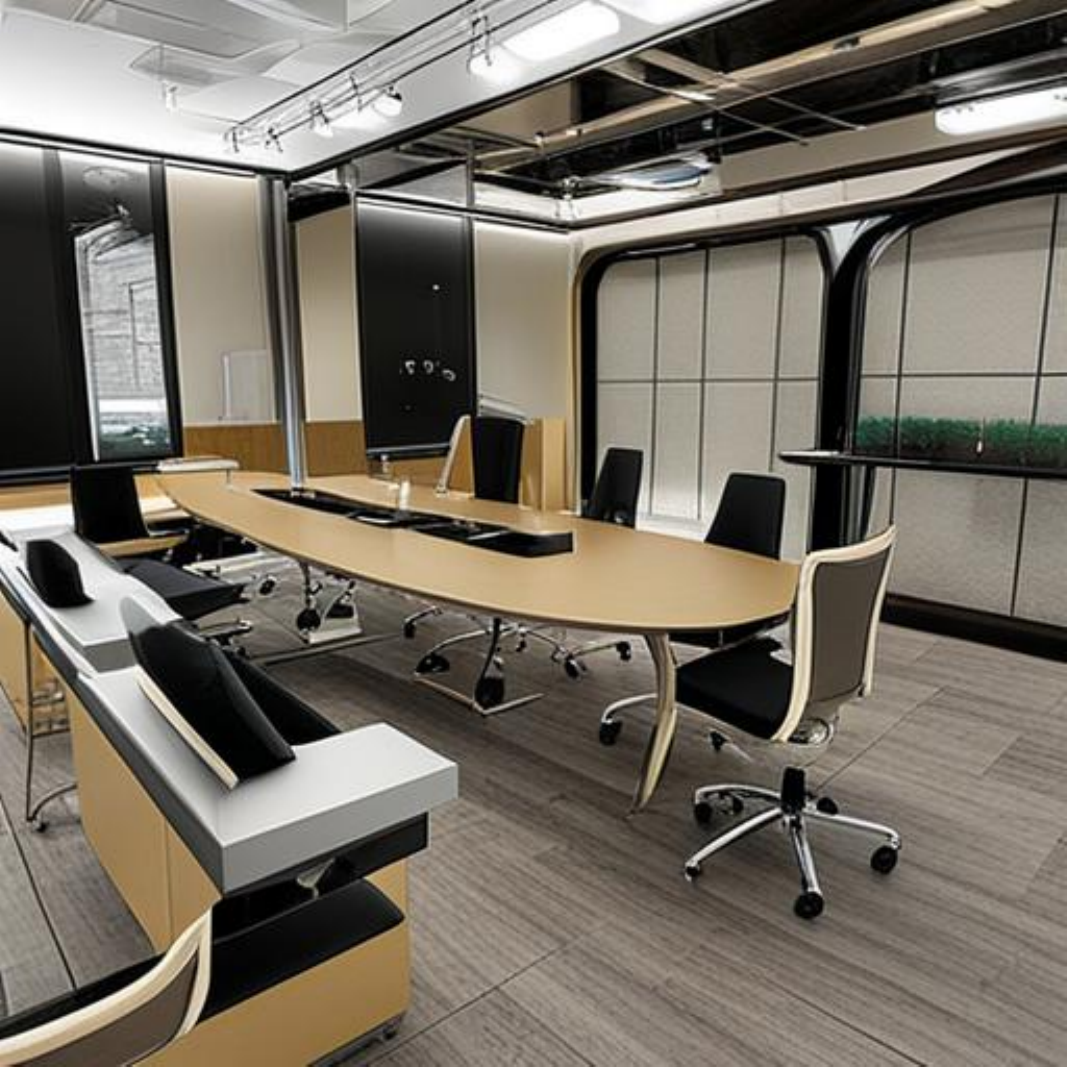} \\
        \midrule
        \begin{minipage}[b]{\editingResultsGraphicsWidth}{\footnotesize We place a ``painting'' in place of the ``plant''. The generated image now contains a painting where the plant has been.\vfill \vspace{0.2cm}}\end{minipage} &
        \includegraphics[width=\editingResultsGraphicsWidth]{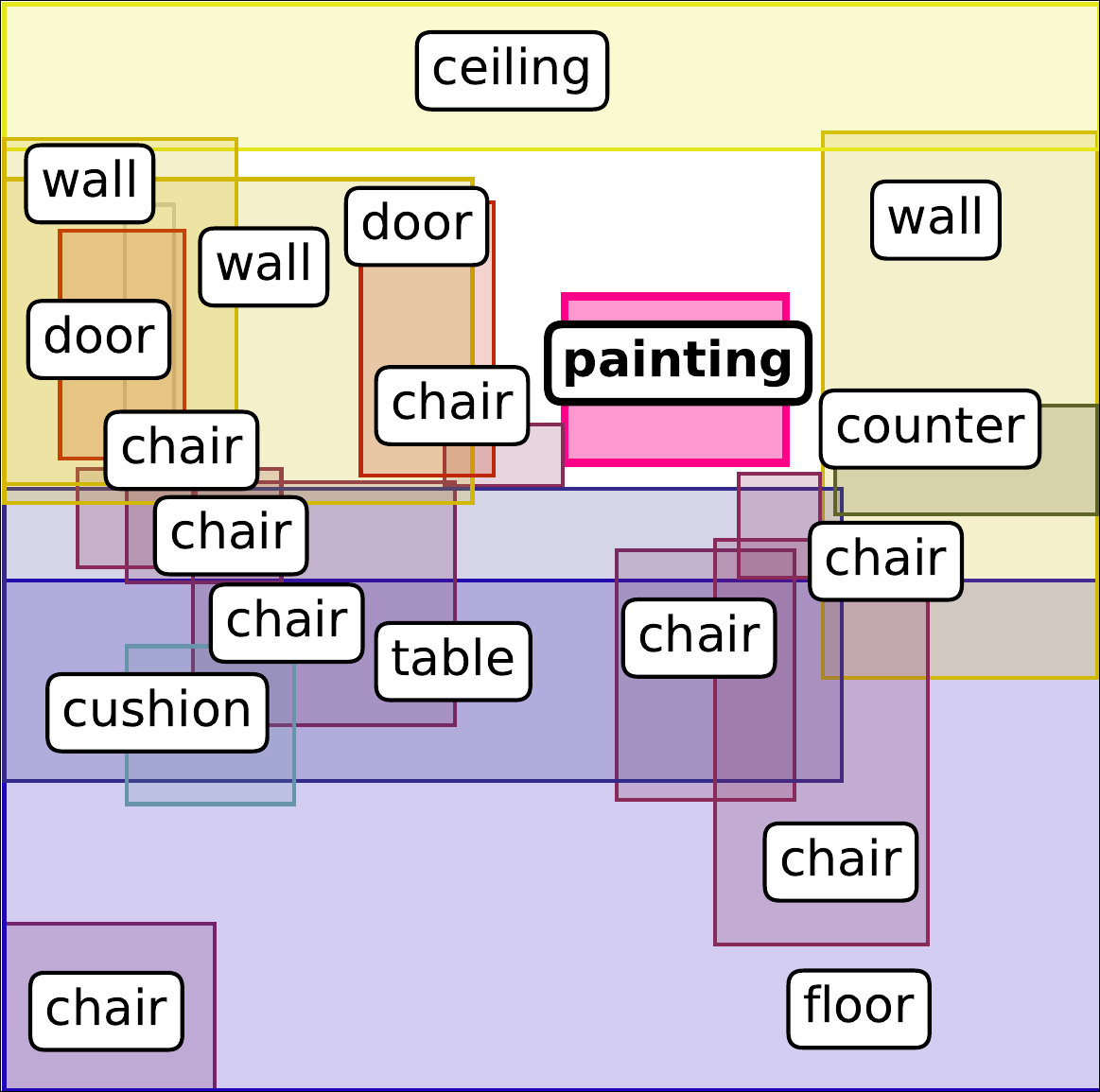} &
        \includestandalone[width=\editingResultsGraphicsWidth]{sections/results/figures/editing/tikz_visuals/painting} \\  
    \end{tabular}
    \caption{\textbf{Editing.} We show how our pipeline enables user editing of images by altering the intermediate scene layout representation. Individual objects can be easily moved, removed, and replaced.}
    \label{fig:editing}
\end{figure}

Furthermore, we demonstrate how our pipeline allows for editing of generated images in ~\cref{fig:editing}. This is accomplished through modifying the intermediate scene layout, and rerunning InstanceDiffusion with the original seed and global prompt. We observe that the edits are not totally isolated to the altered objects and leave it for future work to develop improved conditional diffusion models.

\section{Conclusion}

We have introduced a text-to-image generation pipeline with an intermediate human-interpretable and controllable layout representation.
With a substantially smaller model, we achieve state-of-the-art performance in generated image quality metrics among competing baselines, as well as state of the art when considering both of the axes plausibility and variety. In addition, we have introduced a suite of metrics for scene layouts and generated images for reliable evaluation. Both of these contributions together enable further exploration of image generation pipelines with intermediate explicit representations in the future. 

\section{Acknowledgements}
We thank Tom Fischer, Raza Yunus, and Julian Chibane for their feedback on the draft. Funded in part by the Deutsche Forschungsgemeinschaft (DFG, German Research Foundation) -- GRK 2853/1 “Neuroexplicit Models of Language, Vision, and Action” - project number 471607914.

{
    \small
    \bibliographystyle{ieeenat_fullname}
    \bibliography{main}
}

\twocolumn[
\begin{center}
    \begin{minipage}{\linewidth}
    \begin{center}
        { \Large \textbf{Supplementary Material} }
    \end{center}
    \end{minipage}
\end{center}
]

The supplementary material is structured as follows. First, we present the full details of the human evaluation study performed to judge the generation quality in Sec. ~\ref{sec:study_details}. Next, we introduce details about our partial conditioning procedure in Sec.~\ref{sec:partial_conditioning_details}. We further provide detailed information about training data and hyperparameters in Sec.~\ref{sec:training_data_and_hyperparameters}, discussions about {\layoutTransformer} and GPT4o temperatures in Sec.~\ref{sec:lt_temp} and \ref{sec:gpt_temp}, respectively, and the GPT4o query template in Sec.~\ref{sec:query_template}. Last, we provide a comparison between rectified flow and DDIM in Sec.~\ref{sec:ddim_comp} and high-resolution results in Sec.~\ref{sec:bonus_images} and Sec.~\ref{sec:high_res}.

\section{Human Evaluation Details} 
\label{sec:study_details}
\noindent\textbf{Study Goal.}
Although our method achieves optimal performance in ~\cref{tab:traditional_metric_comparison}, we aim to confirm that these metrics, which were designed for measuring the quality of text-to-image models, are applicable to text-to-layout-to-image models. We also want to control for the effect which the layout-to-image model could have on the final quality, and assess how effective the underlying layouts are in the image generation process. To this end, we provide a human-evaluation study that can be repeated by others. 

In general, we want a text-to-layout model to generate layouts that appear plausible and are also of a large variety. However, assessing human opinions for these criteria directly on layouts is challenging: the evaluators require time to understand the layout diagrams and explain them, and furthermore, assessments are hard to make without actually seeing the image. To make the study effective, we measure the effect of our model on the downstream generated images.
Image qualities that are assessed in other studies (for example, the overall quality and aesthetic appeal of the image in Liang \etal ~\cite{liang2024richhumanfeedbacktexttoimage}) are highly dependent on the conditioned image generator. Therefore, we consider these misleading for our case and introduce a suitable study in the following. 

\noindent\textbf{Design Principles.}
We follow the design principles presented by Otani \etal  ~\cite{otani2023verifiablereproduciblehumanevaluation} in their work on human evaluation of text-to-image generation: 1) \emph{the (evaluation) task should be simple}, and 2) \emph{results should be interpretable}. Following these principles, we show participants only images, and omit the underlying image layouts entirely, which may take some effort to understand. To make the results interpretable, participants rate these images for their plausibility and variety on a Likert scale (as specifically recommended in Otani \etal ~\cite{otani2023verifiablereproduciblehumanevaluation}) from $1$ to $5$. This way, average ratings for different layout generation models can be meaningfully compared to each other, which would be more difficult in other systems (e.g. using non-numbered ranking). To ensure cost efficiency, our survey must be small enough that the data can be collected quickly and repeatedly throughout the model development, and thus we show participants collections of images rather than individual pairs. We kept all of these constraints in mind when designing our study, which is explained in further detail below. 

\noindent\textbf{Study Description.}
Our study was developed using Qualtrics, a standard survey platform. Each participant answers ten plausibility questions and ten variety questions, meaning they rate $80$ image collections in total. We survey $60$ participants. The prompts, image collection index, and the order in which the collections are displayed to participants is randomized to control for any potential effects of a fixed ordering.

\begin{figure*}[ht]
    \centering

    \includegraphics[height=0.9\textwidth]{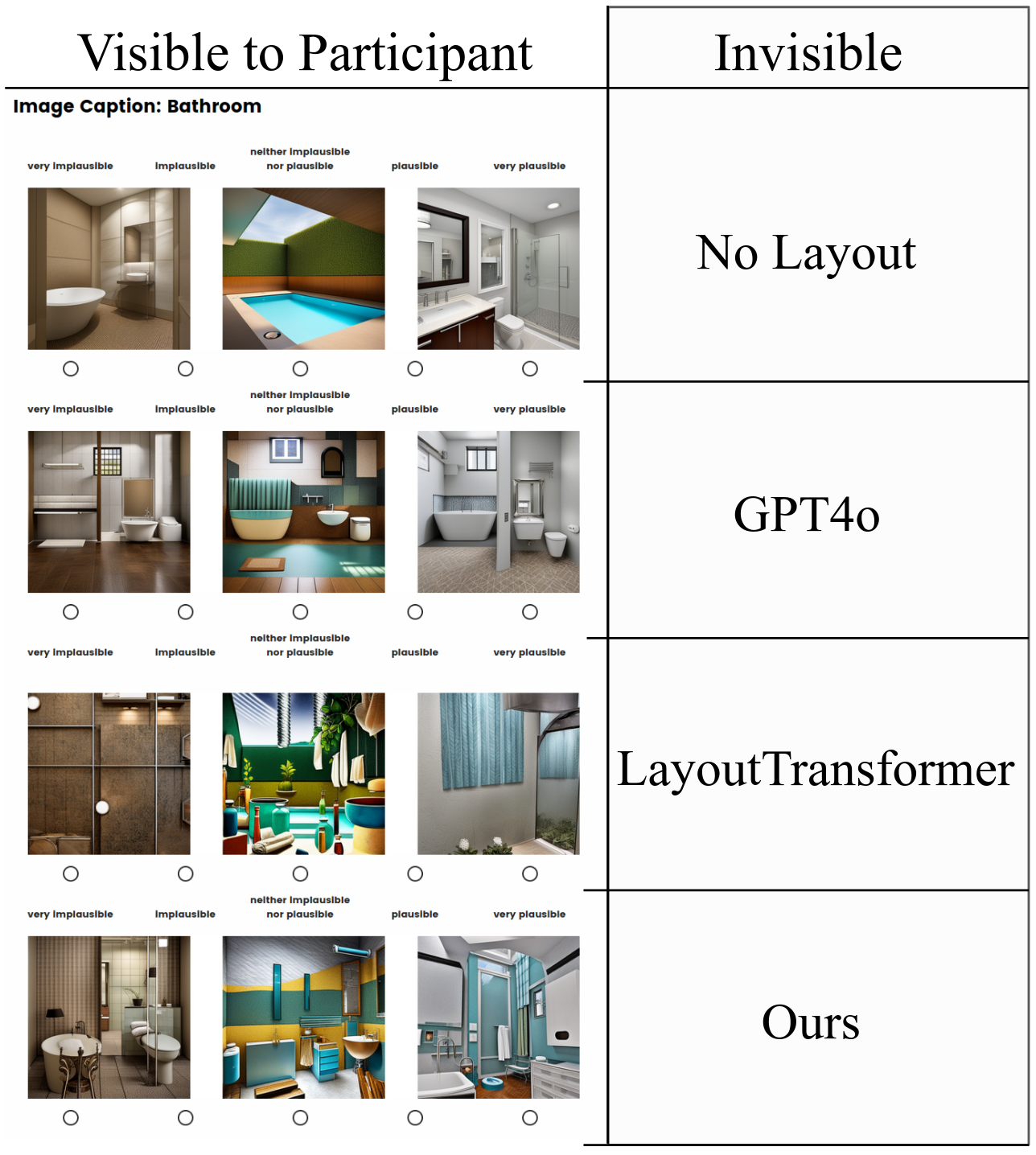} \\
    
    \caption{An example question page from our survey. Users must rate collections of 3 images from very implausible to very plausible. The underlying layout generators for the collections shown are (from top to bottom): No Layout, GPT4o, LayoutTransformer, and our method. Collection order was randomized for each question presented to the participant. Users click the button }
    \label{fig:survey_exerpt}
\end{figure*}

Survey data is selected as described in~\cref{subsec:experimental_setting}. 
As shown in~\cref{fig:survey_exerpt}, each survey question shows collections of three images from each of the four methods listed above, where every image on the screen has the same global prompt. Given the instructions from~\cref{fig:survey_full_instructions}, the participant must rate each collection for either their plausibility or variety. Ratings are on a Likert scale ($1$ to $5$, where $1$ corresponds to very implausible/very low variance, and $5$ corresponds to very plausible/very high variance). For plausibility, we instructed participants to consider the overall realism of the collection, as well as how effectively it depicts the global text prompt. For variety, we instructed users to consider the spatial arrangement of objects in an image and implied camera angle in addition to overall image appearance. 

\noindent\textbf{Participant Selection and Ethics.}
Participants were recruited through Connect CloudResearch, a crowdsourcing service built on Amazon Mechanical Turk that implements rigorous quality control procedures to enhance the reliability of the participant pool in line with the study recommendations given by Otani \etal ~\cite{otani2023verifiablereproduciblehumanevaluation}. The study was approved by the Ethics Review Board of our institution, ensuring compliance with ethical standards. Prior to engaging in the tasks, all participants provided informed consent. The study was designed to be completed within $15$ minutes by each participant, who were compensated at an hourly rate of $13.02$ USD. 
This results in a total cost of 245 USD per run. Participants were anonymized, and we did not collect any personally-identifiable information.

\begin{figure*}[ht]
    \centering
    \begin{minipage}{0.9\textwidth}
    \textbf{Section 1: Plausibility}
    \newline
    \fbox{
        \parbox{0.9\textwidth}{
            ``For the following section of the survey, you will be asked to rate collections of images based on how \textbf{plausible} they appear to be, from \textbf{very implausible} to \textbf{very plausible}. An image is considered plausible if objects within the image are \textbf{realistically and organically} placed, and it is a \textbf{reasonable match to the presented caption}. The images do not have to be photorealistic to be considered plausible. You will perform ratings on 10 categories of images, and each page will contain 4 collections that you must rate separately.''
        }
    }
    \newline
    \newline
    \newline
    \textbf{Section 2: Variety}
    \newline
    \fbox{
        \parbox{0.9\textwidth}{
            ``For the following section of the survey, you will be asked to rate collections of images based on their perceived \textbf{variance}, from \textbf{very low variance} to \textbf{very high variance}. When judging the variance, consider criteria such as the differences in the \textbf{spatial arrangement} of objects, the differences in \textbf{camera perspective}, and the differences in the \textbf{overall image appearance} across the collection. You will perform ratings on 10 categories of images, and each page will contain 4 collections that you must rate separately.''
        }
    }
    \end{minipage}

    \caption{Full instructions to participants for both sections of the survey. Our instructions clearly define the task and give users detailed information on what to assess }
    \label{fig:survey_full_instructions}
\end{figure*}

\clearpage
\clearpage
\section{Implementation of Partial Conditioning} 
\label{sec:partial_conditioning_details}
We explain our adaptation of the RePaint \cite{lugmayr2022repaintinpaintingusingdenoising} technique mentioned in ~\cref{subsec:architecture}, which was used for the partial layout conditioning examples in ~\cref{fig:disentanglement}, An overview is presented in ~\cref{alg:our_alg}. At every timestep, the intermediate sample $\xit$ is first updated with the rate of change provided by our model ($v$). Then the sample is slightly adjusted to conform to a path which will yield the values of the partial conditioning layout $\yinone$ at non-null entries after inference.

Some additional algorithm notation: The partial layout representation $\yinone = \{\ylij \}_{j \in J}$ is defined like the layout representation in ~\cref{subsec:layout_representation} extended by null values $\varnothing$, a placeholder value for entries of $\yinone$ tokens where no conditioning is provided. 
To give an example, consider a partial conditioning layout where the token $\ylij$ enforces that a bounding box with the label \emph{chair} must be present in the final layout, but can have any coordinates. We set $\xlijbbox = (\varnothing, \varnothing, \varnothing, \varnothing)$, $\xlijclip$ to be the PCA-reduced CLIP embedding of the word \emph{chair} and $\xlijopacity = 1$, and write:
\begin{equation}
    \ylij = (\xlijbbox \mathbin\Vert \xlijclip \mathbin\Vert \xlijopacity) \, .
\end{equation}
The mask variable $M$ on line $5$ of our algorithm tracks which values of $\yinone$ are null-values, and masks these values out during the update on line $12$. To perform this masking, we define the arithmetic on $\varnothing$ as follows:
\begin{equation}
    \begin{array}{l}
    \varnothing + a = \varnothing  \text{ for } a \in \mathbb{R} ,\\
    \varnothing * a = \varnothing  \text{ for } a \in \mathbb{R} - \{0\} , \\
    \varnothing * 0 = 0 .
    \end{array}
\end{equation}

We construct the drift vector $\driftvector$ which encodes the directional constraints. We begin by initializing $\driftvector$ to $0$ in all entries. Then, we add constraints. For example, if there is a constraint that bounding box $j$ must be left of bounding box $j'$, then 

\begin{equation}
    \vec{d}_i^j \leftarrow \vec{d}_i^j + (\lambda, 0,0,0\mathbin\Vert 0 \mathbin\Vert 0) \, ,
\end{equation}
\begin{equation}
    \vec{d}_i^{j'} \leftarrow \vec{d}_i^{j'} + (-\lambda, 0,0,0\mathbin\Vert 0 \mathbin\Vert 0) \, ,
\end{equation}
where $\lambda$ is a small constant.

In the special case when no conditioning is provided or directional constraints are provided ($\yinone \equiv \varnothing, \driftvector = 0$), this algorithm is identical to the rectified flow inference presented in ~\cref{subsec:architecture} of our main paper.

\renewcommand{\algorithmiccomment}[1]{\hfill #1}

\begin{algorithm*}
\begin{algorithmic}[1]
    \STATE \texttt{conditionedInference(} $P_i$ \texttt{,} $\yinone$ \texttt{,} $\directionset$ \texttt{):}
    \STATE $T \leftarrow 1200$
    \STATE $\Delta t \leftarrow 1/T$
    \STATE $t \leftarrow 0$

    \STATE $M \leftarrow 0 \text{ where } \yinone = \varnothing \text{ otherwise } 1$ \COMMENT \textit{//Create a binary mask for the conditioning layout}

    \STATE $\xiz \sim \mathcal{N}(0, I)$ \COMMENT{\textit{//Sample the starting noise}}
    \WHILE{$t<1$}
        
        \STATE $\frac{d \xit}{dt} \leftarrow v(\xit, t, P_i)$ \COMMENT{\textit{//Calculate the rate of change of $\xit$ at timestep $t$}}
        \STATE $t \leftarrow t + \Delta t$ \COMMENT{\textit{//Update timestep $t$}}
        \STATE $\xit \leftarrow \xitminusdelta + \frac{d \xitminusdelta}{dt} \cdot \Delta t$ \COMMENT{\textit{//Calculate $\xit$ for the next timestep}}
        \STATE $ \yit \leftarrow \yinone \cdot t + \xiz \cdot (1-t)$ \COMMENT{\textit{//Calculate conditioning update $\yit$}}
        \STATE $\xit \leftarrow \yit \odot M + \xit \odot (1 - M)$ \COMMENT{\textit{//Update $\xit$ with conditioning in masked area}}
        \STATE $\xit \leftarrow \xit + \driftvector$  \COMMENT{\textit{//Apply drift for all given directional constraints}}
    \ENDWHILE
    \STATE Return $\xione$
\end{algorithmic}
\caption{Partially Conditioned Layout Generation
}
\label{alg:our_alg}
\end{algorithm*}

\clearpage
\section{Training Data and Hyperparameters} \label{sec:training_data_and_hyperparameters}

Our model consists of $20$ AdaLN transformer blocks with $12$-headed attention. For a token $\xlj$, we sinusoidally encode $\xljbbox$ into $\mathbb{R}^{72}$, and $\xljopacity$ into $\mathbb{R}^{18}$. $\xljclip$ consists of the $30$ top principal components of the object-label's CLIP embedding, which accounts for $77.35\%$ of the explainable variance of our embeddings found in our training data. The timestep $t$ is sinusoidally encoded into $\mathbb{R}^9$, while the CLIP embedding of a global prompt $\ell$ is down-projected by a trainable linear layer into $\mathbb{R}^{17}$ before interfacing with the AdaLN block.

When reporting model parameters, we include all transformer block weights and attached linear layers, including the PCA projection matrices. Given that CLIP dominates the number of parameters, it is a necessary subcomponent for InstanceDiffusion, and needed to form any complete text-to-image pipeline, we factor it out. 

We train our model for $2000$ epochs using stochastic gradient descent with learning rate  $\lambda=0.0005$ and a batch size of $32$. 

\clearpage
\section{LayoutTransformer Temperature}
\label{sec:lt_temp}

Throughout our main paper, we maintained {\layoutTransformer} defaulttemperature parameter equal to one. However, the question arises whether the generated layouts would be higher quality at lower temperatures, where the model's output is more stable. As shown in ~\cref{tab:temp_equals_zero} even when we select the lowest temperature of zero for optimal stability, we are still not measuring a decisive improvement across numerical metrics, therefore we kept the temperature at its original setting of one to remain as faithful as possible to the prior work. 

\begin{table*}[ht]
\footnotesize
\newrobustcmd\B{\DeclareFontSeriesDefault[rm]{bf}{b}\bfseries}  
\def\Uline#1{#1\llap{\uline{\phantom{#1}}}}

\sisetup{detect-weight=true,
         mode=text,
         table-format=2.2,      
         add-integer-zero=false,
         table-space-text-post={*},
         table-align-text-post=false
         }

    \centering
        \begin {tabular}{
        l
        S
        S
        S
        S
        S
        S
        S
        S
        S
        S
        S
        c   
        }
        \toprule
        \toprule
        {\textbf{Model}} & {\textbf{FID} ($\downarrow$)} & {\textbf{KID} ($10^{-2}$)($\downarrow$)} & {\textbf{CMMD} ($\downarrow$)}
        & {$\objectNumeracyScore(\downarrow)$} & 
        {$\firstOrderPositionalLikelihood (10^{-11})(\uparrow)$} &
        {$\secondOrderPositionalLikelihood (10^{-11}) (\uparrow)$} &    
        {$\positionalVarianceScore(\uparrow)$}
        \\
        
        \midrule

        {\layoutTransformer} temp= 1 & \Uline{0.44} & 0.94 & \Uline{1.34} & \B 0.90 & 3.09 & 1.21 & \B 231  \\
        \midrule
        {\layoutTransformer} temp= 0 & 0.48 & \Uline{0.92} & 1.77 & 4.11 & \Uline{3.73} & \Uline{1.53} & 0 \\
        \midrule 
        \textbf{Ours} & \B 0.17 & \B 0.27 & \B 0.03  & \Uline{1.14} & \B 4.76 & \B 2.03 & 187  \\
        \bottomrule
        \bottomrule
        
        \end{tabular}
    \caption{Comparison of metrics {\layoutTransformer} with a temperature of one (model default) and a temperature of zero. Even when the temperature is zero, we see that our method still performs better across our metrics. }
    \label{tab:temp_equals_zero}
\end{table*}

\begin{figure*}[ht]
    \centering
    \begin{tabular}{c c c}
        ``Street'' &
        ``Kitchen'' &
        ``Bathroom'' \\
        \includegraphics[width=0.3\textwidth]{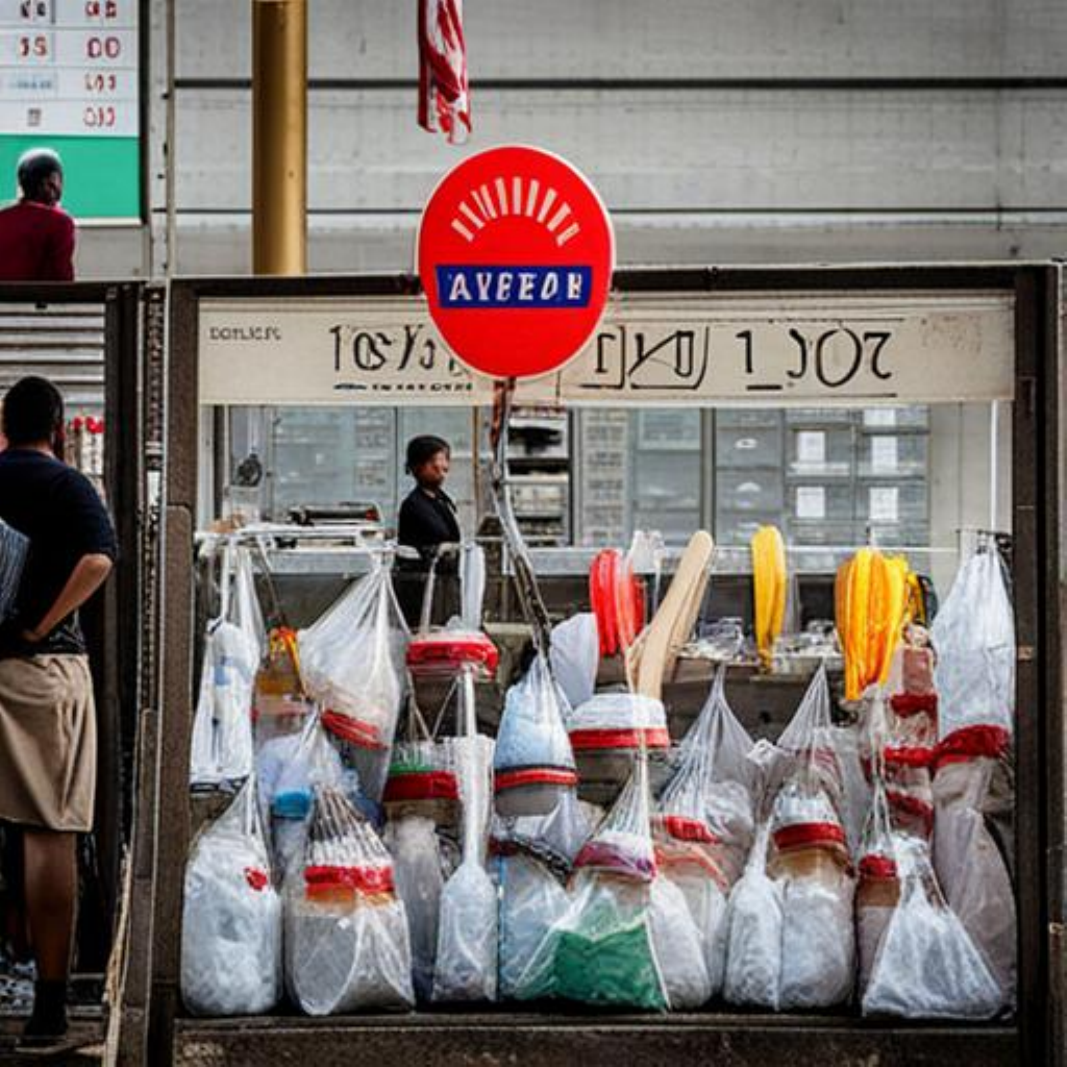} &
        \includegraphics[width=0.3\textwidth]{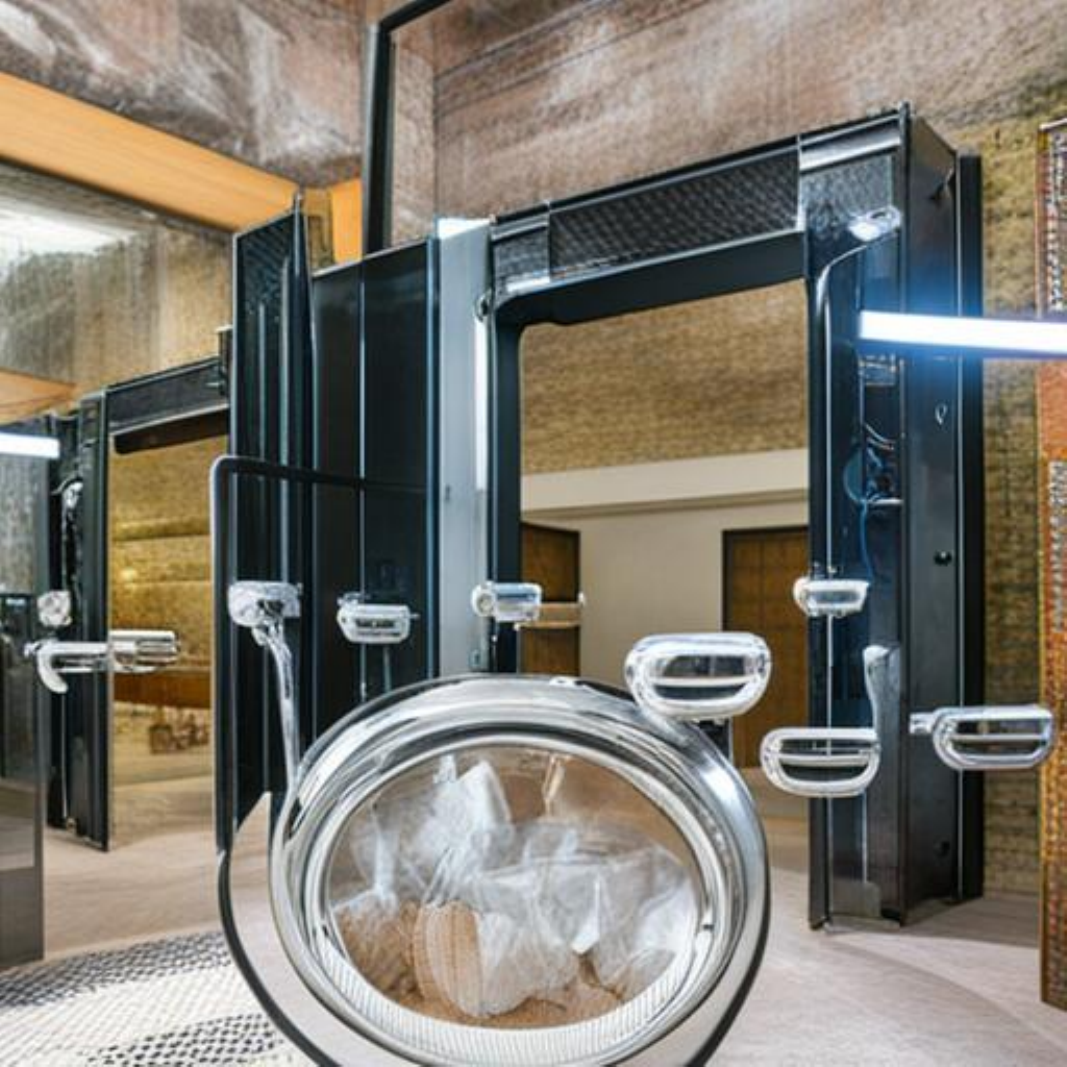} &
        \includegraphics[width=0.3\textwidth]{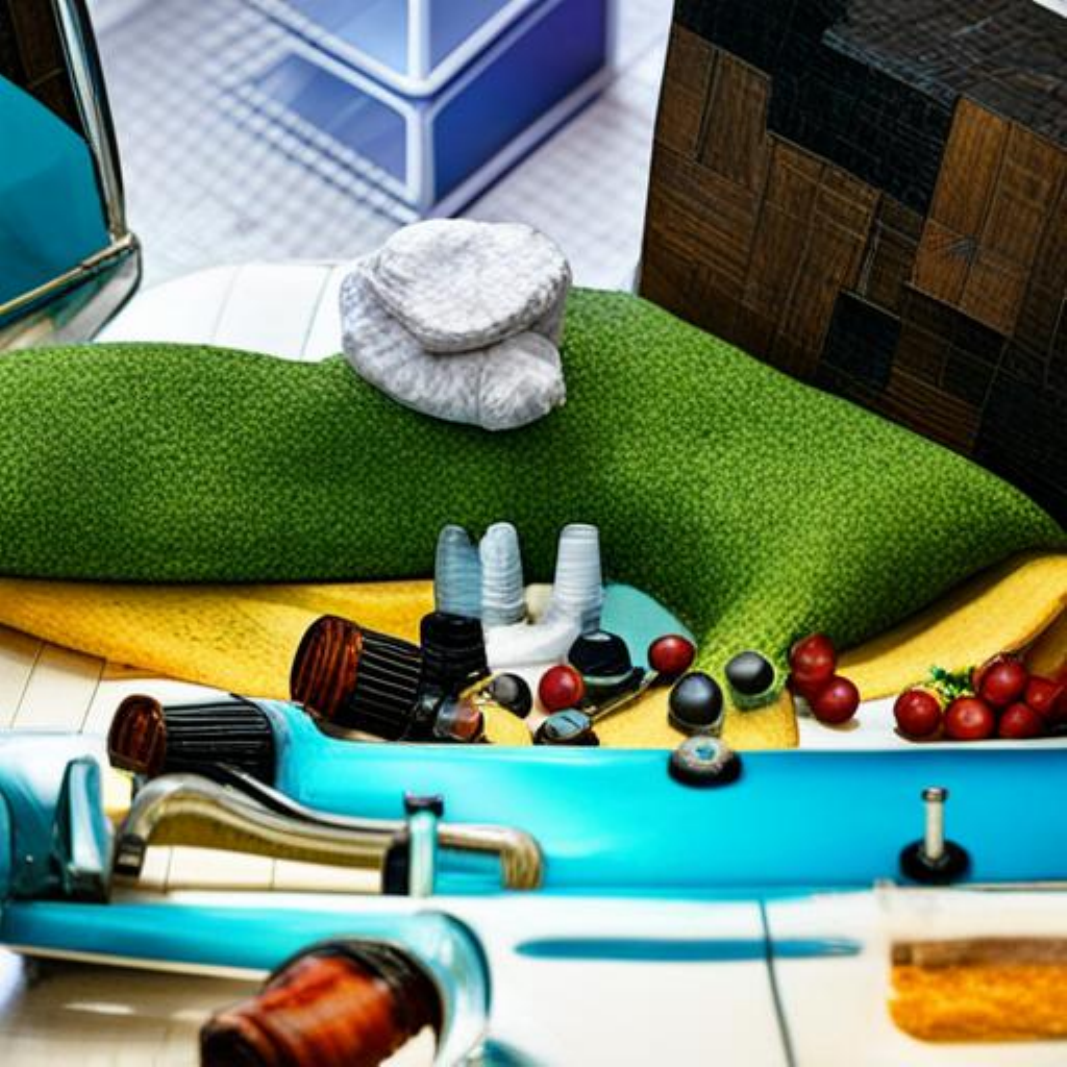} \\
        \includegraphics[width=0.3\textwidth]{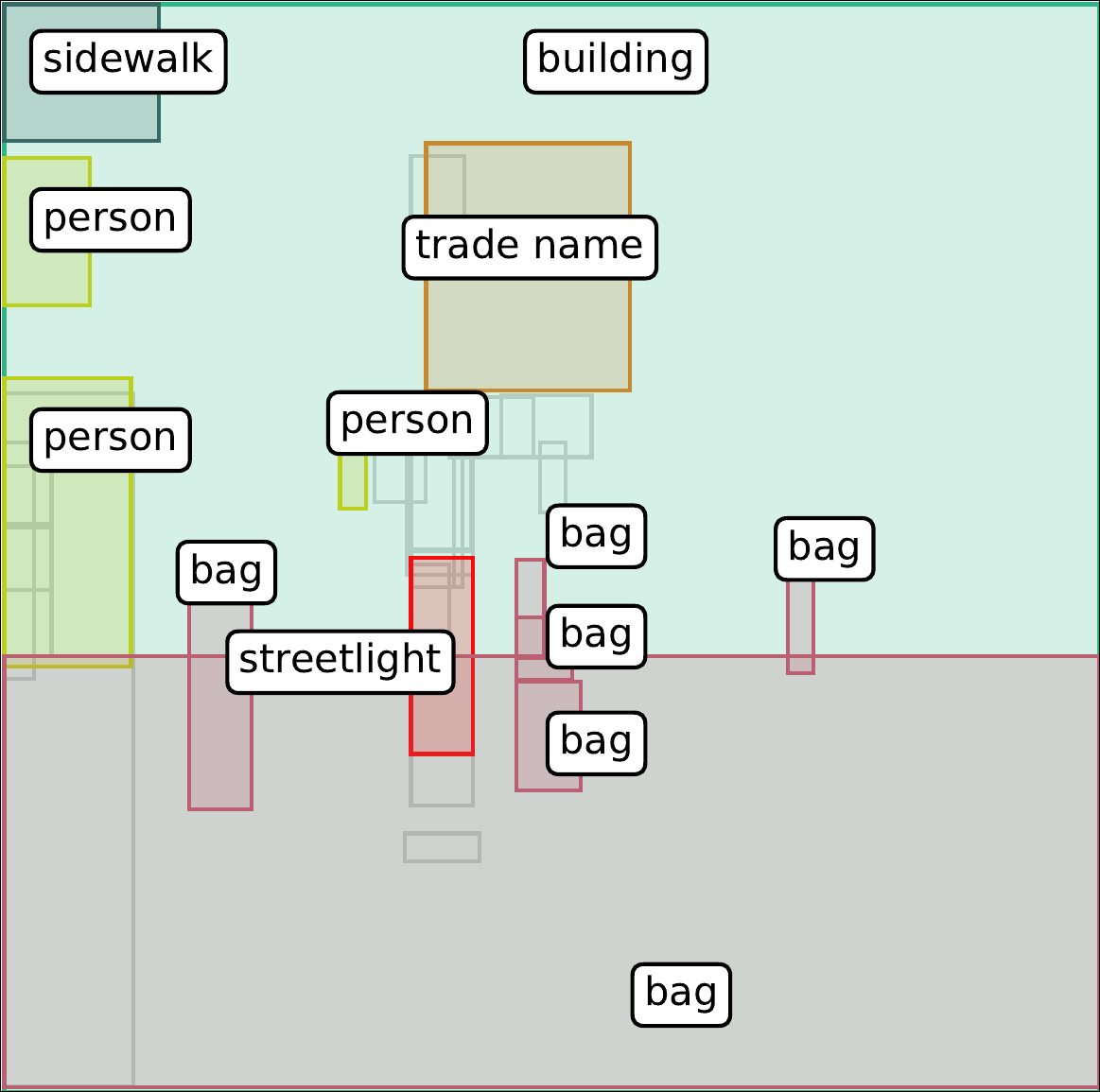} &
        \includegraphics[width=0.3\textwidth]{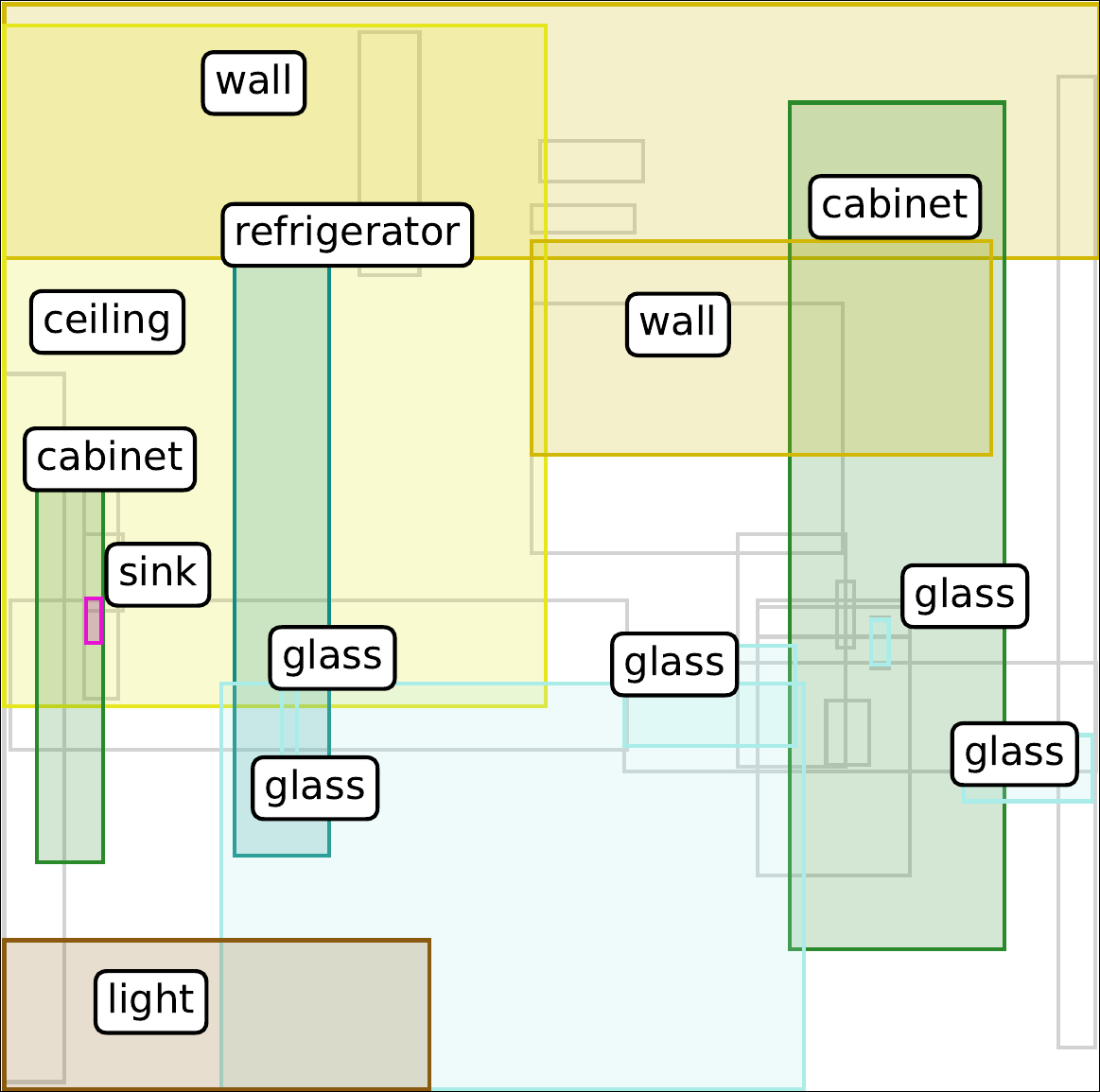} &
        \includegraphics[width=0.3\textwidth]{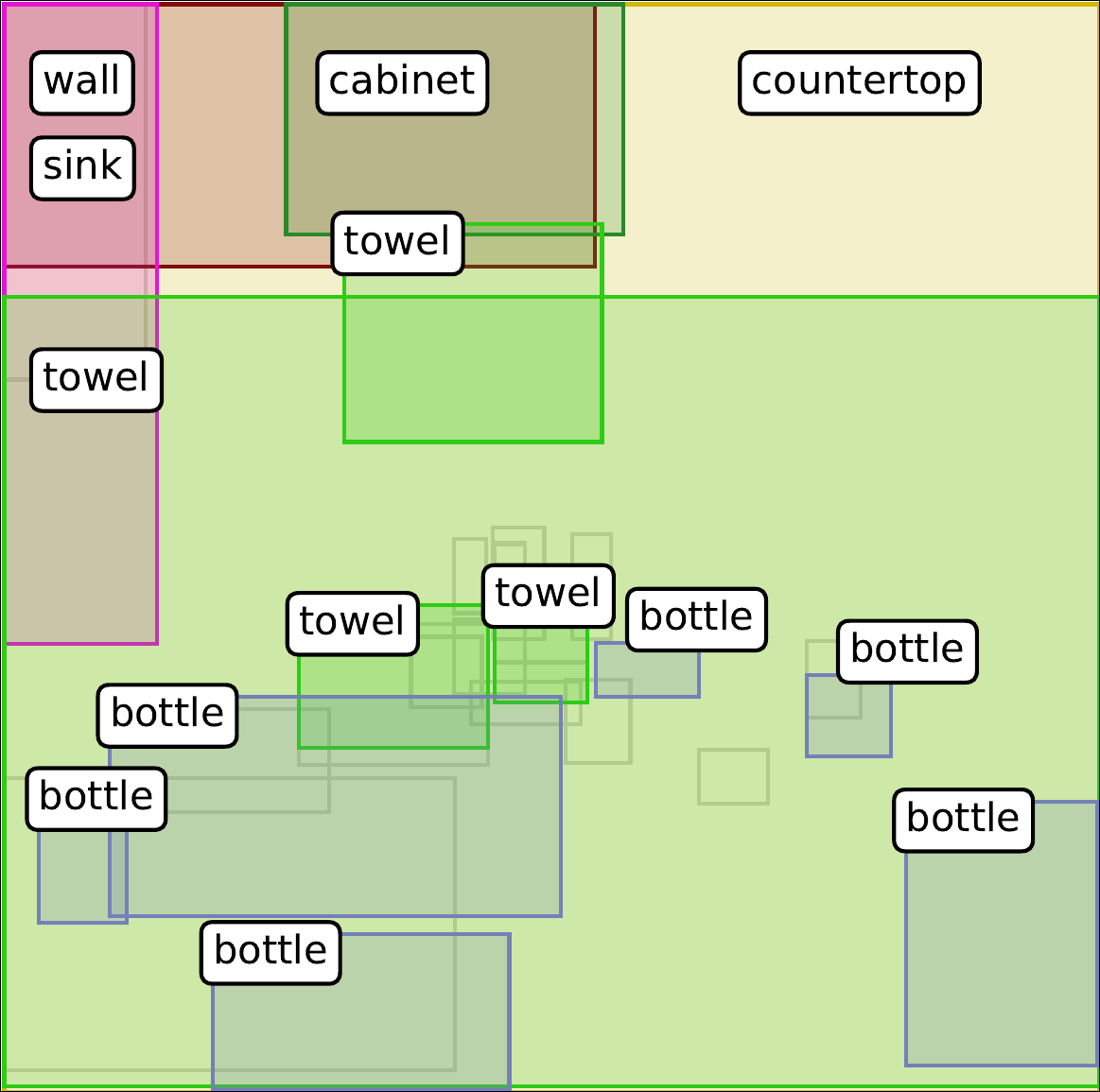} \\
    \end{tabular}
    \caption{Example layouts and images for {\layoutTransformer} when temp=$0$. Even at the most stable setting, the images appear implausible. Objects that are typically small details, such as \emph{bag}, \emph{glass}, or \emph{bottle} repeated many times across the layout.}
    \label{fig:lt_zero_temp}
\end{figure*}

\clearpage
\section{GPT4o Temperature}
\label{sec:gpt_temp}
Because we observed low image variance for {\gptFourO} layouts, we also considered what would happen if we raised the temperature of {\gptFourO} from the default $0.25$ as set in {\LLMGroundedDiffusion} to achieve more variety.

We experimented with increasing the temperature from $0.2$ in increments of $0.1$. We found that at a temperature of $1$, GPT4o failed to produce a parsable layout $14\%$ of the time. However, these mistakes were easy to catch and query the model again. Temperatures higher than $1$ caused more frequent parsing failures, and began to produce long, tangential sentences rather than proper object labels. Without a method to heuristically filter these responses, we settled on a temperature of one as a reasonable upper limit for operation temperature of {\gptFourO} on this task.

We compare the performance of {\gptFourO} with a temperature of one with our method, and {\gptFourO} with the default temperature in ~\cref{tab:gpt_temp_equals_one}. Our model still outperforms {\gptFourO} when the temperature is one in FID and KID. While raising the temperature improves the object numeracy score $\objectNumeracyScore$ and the positional variance score $\positionalVarianceScore$ improve in {\gptFourO} when the temperature is raised, they are still worse than our method, and come at the cost of decreased performance in the positional likelihood scores $\firstOrderPositionalLikelihood$ and $\secondOrderPositionalLikelihood$. Therefore, raising the temperature does not offer a clear advantage on our numerical metrics.

\begin{figure*}[ht]
    \centering
    \begin{tabular}{c c c} 
        \includegraphics[width=0.3\textwidth]{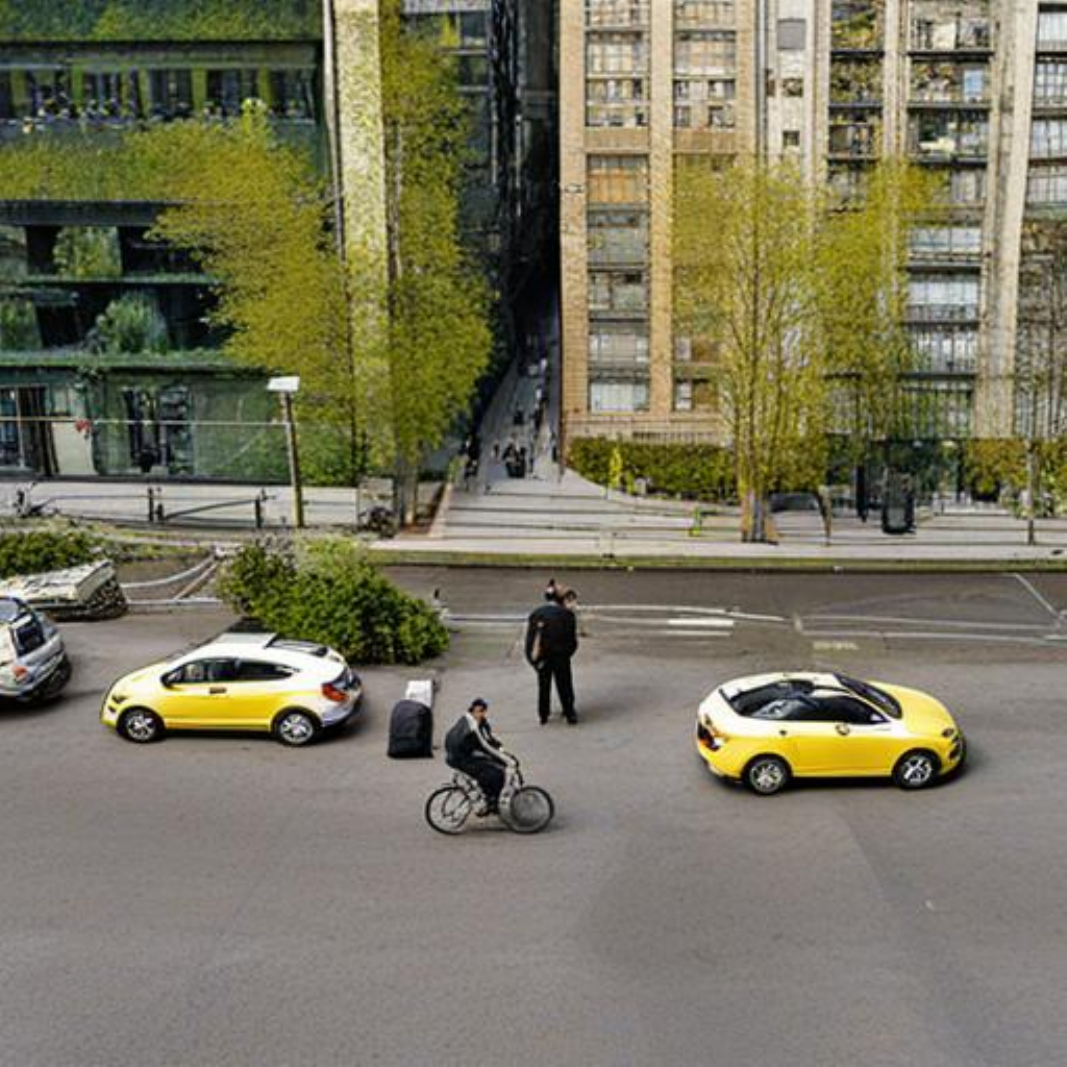} &
        \includegraphics[width=0.3\textwidth]{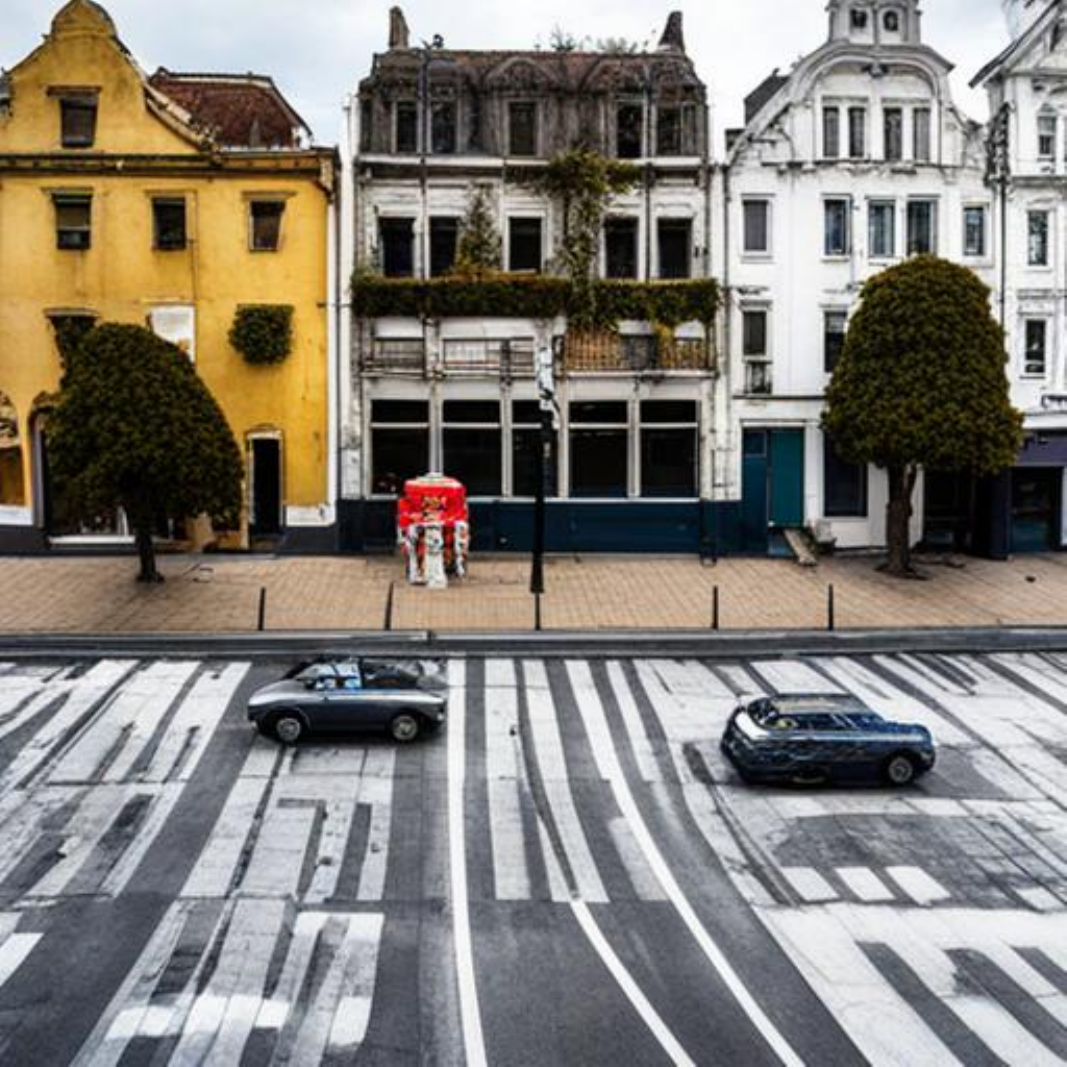} &
        \includegraphics[width=0.3\textwidth]{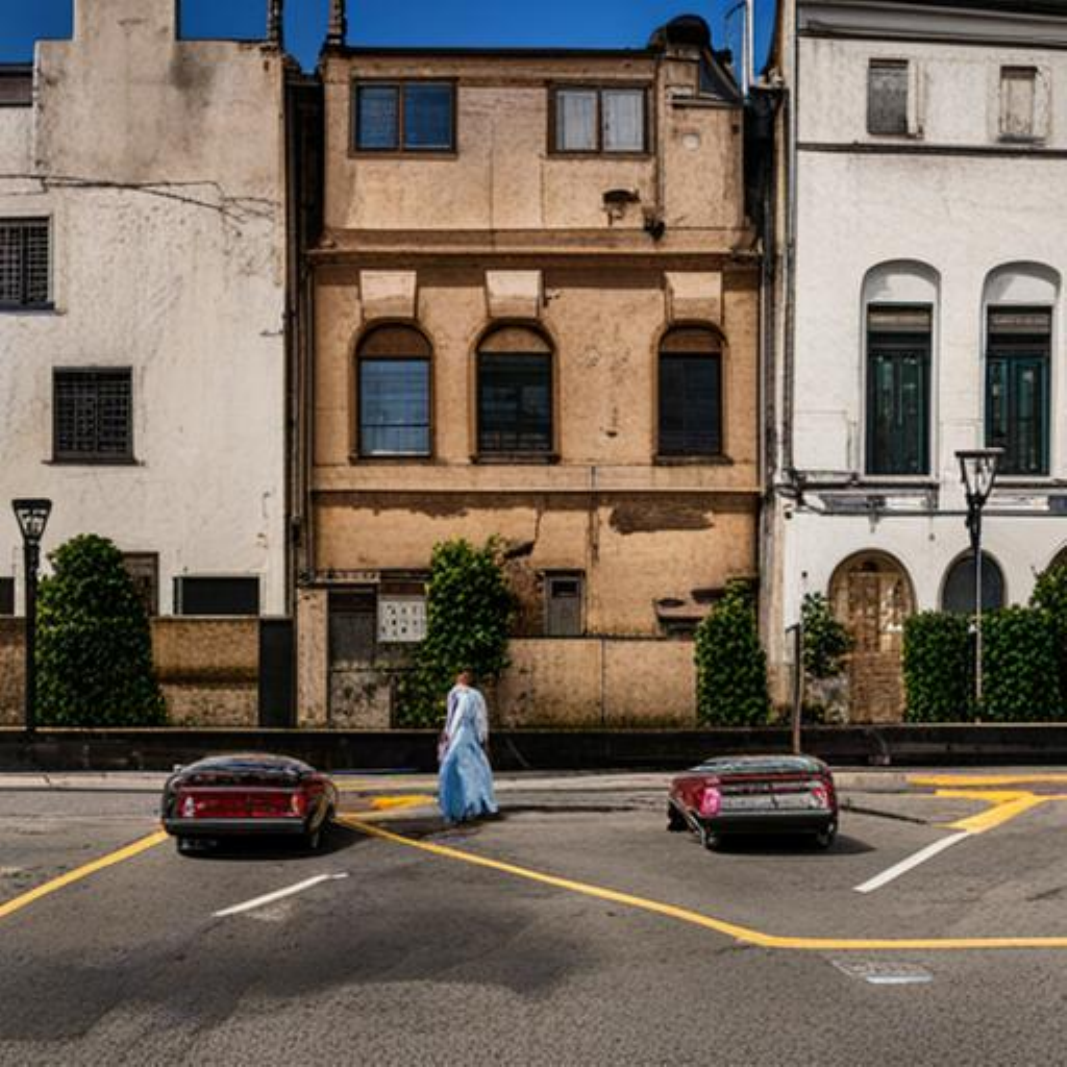} \\
        \includegraphics[width=0.3\textwidth]{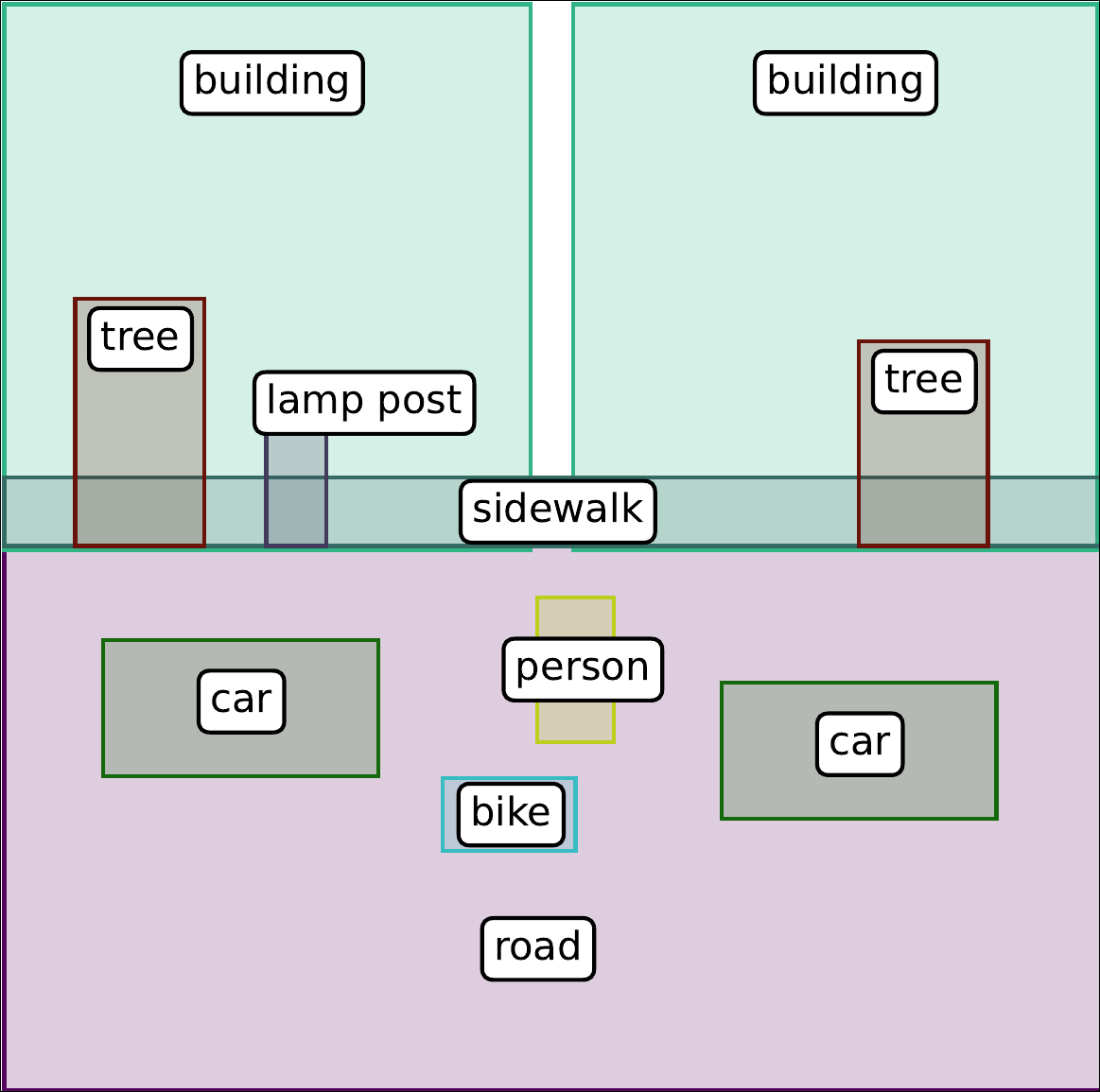} &
        \includegraphics[width=0.3\textwidth]{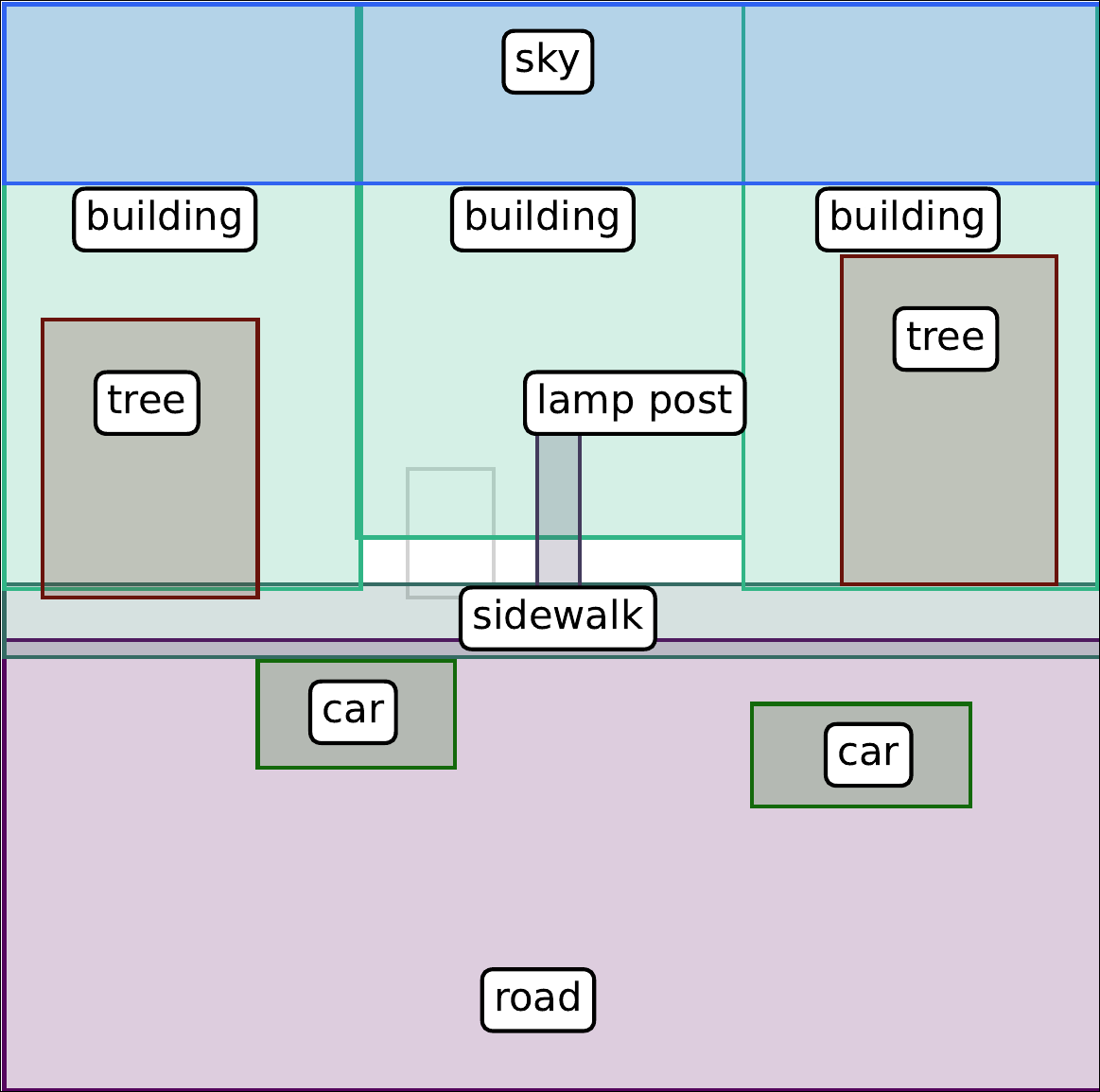} &
        \includegraphics[width=0.3\textwidth]{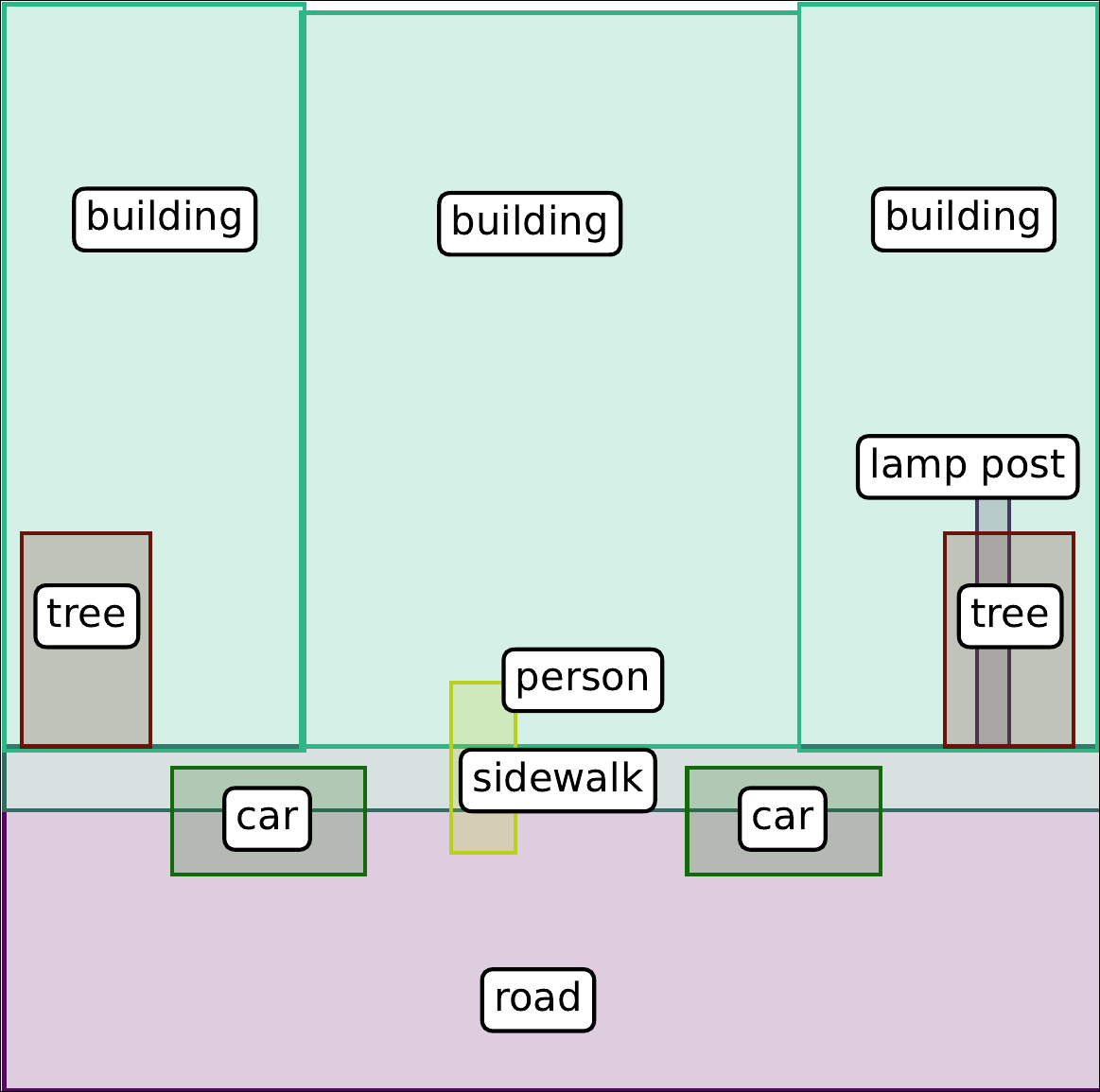} \\
    \end{tabular}
    \caption{Example images and underlying layouts of the prompt \emph{street} for {\gptFourO} when the temperature is one, the highest stable temperature. Visually, there is slightly more variation than at a  temperature of $0.25$, (see ~\cref{fig:gpt4o_street}), but this is not a pronounced effect: positions and quantities of objects, and implied camera angle, are still very repetitive.}
    \label{fig:gpt_one_temp}
\end{figure*}

We also visualized outputs of {\gptFourO} with the raised temperature in  ~\cref{fig:gpt_one_temp}. Although there is some increase in the variation of scenes, the effect does not appear to be noticeably pronounced. Therefore, we choose to stick with a temperature of $0.25$ for our human evaluation, as this is the most faithful adaptation of our {\LLMGroundedDiffusion} baseline, without neglecting a clear optimization.

\begin{table*}[ht]
\footnotesize
\newrobustcmd\B{\DeclareFontSeriesDefault[rm]{bf}{b}\bfseries}  
\def\Uline#1{#1\llap{\uline{\phantom{#1}}}}

\sisetup{detect-weight=true,
         mode=text,
         table-format=2.2,      
         add-integer-zero=false,
         table-space-text-post={*},
         table-align-text-post=false
         }

    \centering
        \begin {tabular}{
        l
        S
        S
        S
        S
        S
        S
        S
        S
        S
        S
        S
        c   
        }
        \toprule
        \toprule
        {\textbf{Model}} & {\textbf{FID} ($\downarrow$)} & {\textbf{KID} ($10^{-2}$)($\downarrow$)} & {\textbf{CMMD} ($\downarrow$)}
        & {$\objectNumeracyScore(\downarrow)$} & 
        {$\firstOrderPositionalLikelihood (10^{-11})(\uparrow)$} &
        {$\secondOrderPositionalLikelihood (10^{-11}) (\uparrow)$} &    
        {$\positionalVarianceScore(\uparrow)$}
        \\
        
        \midrule

        {\gptFourO} temp=$0.25$ & \Uline{0.94} & \Uline{0.99} & \Uline{1.34} & 3.71 & \Uline{4.37} & \Uline{1.49} & 93  \\
        \midrule
        {\gptFourO} temp=1 & 1.47 & 1.62 & 1.35 & \Uline{2.86} & 4.02 & 1.35  & \Uline{142} \\
        \midrule 
        \textbf{Ours} & \B 0.17 & \B 0.27 & \B 0.03  & \B 1.14 & \B 4.76 & \B 2.03 & \B 187  \\
        \bottomrule
        \bottomrule
        
        \end{tabular}
    \caption{Comparison of metrics \gptFourO{} with a temperature of $0.25$ (adapted model default) and one (highest stable temperature). At increased temperatures, {\gptFourO} performs worse on the FID and KID metrics. Although increasing the temperature of {\gptFourO} improves $\objectNumeracyScore$ (the object frequencies are closer to the ground truth) and $\positionalVarianceScore$ (the layouts are more varied overall), performance on $\firstOrderPositionalLikelihood$ and $\secondOrderPositionalLikelihood$ drops (the positions of the objects are less plausible). Our method still performs better in all displayed metrics.   }
    \label{tab:gpt_temp_equals_one}
\end{table*}

\clearpage
\section{GPT4o Query Template} 
\label{sec:query_template}
We adapted the prompt template from {\LLMGroundedDiffusion} by replacing the old scene caption and layout examples with ground truth data from ADE20k and encourage chain of thought reasoning ~\cite{wei2023chainofthoughtpromptingelicitsreasoning}. Our LLM prompt is listed in ~\cref{tab:llm_prompt}. Our in-context examples are listed in ~\cref{tab:incontext}.

\begin{figure*}[ht]
\centering 

\begin{tabular}{p{0.5cm} p{15cm}}
\hline

\scriptsize 1 & \texttt{Task Description and Rules } \\

\scriptsize 2 & \leftskip=2cm \parindent=-2cm \texttt{You are a smart program for automatic image layout generation. I provide you with a global prompt which describes the entire image. The image layout has a height of 512 and a width of 512. The coordinate system assumes the origin (0,0) is in the top left corner. Bounding box coordinates are specified in the format (x,y,w,h), where x and y are the top left corner coordinate, and w and h are the full width and height of the box. Your task is to imagine which objects reasonably belong in an image with a global prompt, and arrange these objects in into a layout which could plausibly be for a real image.} \\

\scriptsize 3 & \texttt{} \\

\scriptsize 4 & \texttt{Meta Command} \\

\scriptsize 5 & \leftskip=2cm \parindent=-2cm \texttt{Reason about the objects added to the layout For each object reason about its position in the layout relative to the other objects, and why it is likely. In general maintain a plausible configuration of the objects within the image layout such that the coordinates obey our coordinate convention. Do not number the objects, instead put them in a list in the exact format shown below. Remember to include the caption, background prompt and negative prompt in the layout. } \\

\scriptsize 6 & \texttt{} \\

\scriptsize 7 & \texttt{[ In - context Examples ]} \\

\scriptsize 8 & \texttt{} \\

\scriptsize 9 & \texttt{Question} \\
\scriptsize 10 & \texttt{Provide the layout for a "\{prompt\}"} \\

\hline

\end{tabular}
\captionsetup{width=25cm, justification=centering}
\caption{ Our full prompt to the LLM for layout generation. LLM starts completions from ``Objects.''}\label{tab:llm_prompt}
\end{figure*}
\clearpage  

\onecolumn
\begin{longtable}{p{0.1\textwidth} p{0.9\textwidth}}
\hline

\scriptsize 1 & \texttt{Provide a layout for an "airport terminal”} \\

\scriptsize 2 & \texttt{Answer: } \\

\scriptsize 4 & \texttt{Caption: An airport terminal} \\

\scriptsize 5 & \leftskip=2cm \parindent=-2.2cm \texttt{ Objects: [('ceiling',[1, 0, 510, 292]), ('floor',[0, 360, 468, 151]), ('wall',[0, 337, 152, 152]), ('wall',[3, 193, 182, 100]),('first floor',[353, 262, 158, 96]),('wall',[353, 207, 158, 86]),('first floor',[0, 264, 136, 93]),('plant',[177, 398, 158, 78]),('fountain',[197, 376, 122, 79]),('first floor',[134, 310, 265, 30]),('wall',[398, 343, 68, 117]),('wall',[467, 330, 44, 172]),('wall',[190, 270, 157, 42]),('column',[65, 159, 30, 157]),('column',[0, 76, 19, 242]),('column',[442, 153, 30, 141]),('wall',[275, 336, 123, 33]),('wall',[134, 340, 129, 31]),('fence',[331, 394, 44, 83]),('first floor',[110, 282, 86, 39]),('column',[404, 171, 23, 139]),('wall',[239, 499, 272, 12]),('column',[113, 175, 22, 144]),('seat',[187, 478, 134, 17]),('column',[380, 193, 19, 117]),('fence',[134, 403, 40, 53]),('column',[145, 196, 18, 116]),('tree',[345, 324, 34, 45]),('tree',[43, 404, 30, 49]),('tree',[150, 329, 32, 45])]} \\

\scriptsize 6 & \texttt{Background prompt: an airport terminal} \\

\scriptsize 7 & \texttt{Negative prompt: empty} \\

\scriptsize 8 & \leftskip=2cm \parindent=-2cm \texttt{Reasoning: Airport terminals contain many walls and columns, and have a floor and ceiling. They also contain seats for passengers to wait in as well as decorative trees} \\

\scriptsize 9 & \texttt{} \\

\scriptsize 10 & \texttt{Provide a layout for an "schoolhouse”} \\

\scriptsize 11 & \texttt{Answer: } \\

\scriptsize 12 & \texttt{Caption: schoolhouse} \\

\scriptsize 13 & \leftskip=2cm \parindent=-2cm \texttt{Objects: [('sky',[0, 0, 510, 431]),('building',[22, 23, 460, 465]),('tree',[1, 0, 173, 200]),('grass',[0, 449, 510, 61]),('tree',[422, 129, 83, 281]),('tree',[1, 202, 44, 254]),('path',[0, 478, 308, 27]),('grass',[0, 469, 201, 27]),('plant',[42, 412, 71, 76]),('plant',[399, 417, 56, 69]),('person',[229, 412, 33, 64]),('car',[3, 444, 40, 29]),('tree',[0, 431, 41, 23]),('tree',[472, 426, 37, 23])]} \\

\scriptsize 14 & \texttt{Background prompt: schoolhouse} \\

\scriptsize 15 & \texttt{Negative prompt: empty} \\

\scriptsize 16 & \leftskip=2cm \parindent=-2cm \texttt{Reasoning: A schoolhouse is typically a building. The layout could include a path, students, trees, plants, and a car in the schoolyard.} \\

\scriptsize 17 & \texttt{} \\

\scriptsize 18 & \texttt{Provide a layout for an "ball pit”} \\

\scriptsize 19 & \texttt{Answer: } \\

\scriptsize 20 & \texttt{Caption: ball pit} \\

\scriptsize 21 & \leftskip=2cm \parindent=-2cm \texttt{Objects: [('inflatable park',[1, 0, 510, 510]),('person',[85, 42, 313, 398]),('ball',[451, 292, 48, 69]),('ball',[77, 253, 46, 61]),('ball',[416, 278, 40, 58]),('ball',[475, 265, 34, 68]),('ball',[371, 240, 39, 55]),('ball',[430, 246, 40, 47])]} \\

\scriptsize 22 & \texttt{Background prompt: ball pit} \\

\scriptsize 23 & \texttt{Negative prompt: empty} \\

\scriptsize 24 & \leftskip=2cm \parindent=-2cm \texttt{Reasoning: A ball pit is an inflatable park with balls and people. The layout could include a person playing in the ball pit and colorful balls scattered around the inflatable park.} \\

\scriptsize 25 & \texttt{} \\

\scriptsize 26 & \texttt{Provide a layout for an "jail cell”} \\

\scriptsize 27 & \texttt{Answer: } \\

\scriptsize 28 & \texttt{Caption: jail cell} \\

\scriptsize 29 & \leftskip=2cm \parindent=-2cm \texttt{Objects: [('bar',[0, 0, 510, 512]),('floor',[24, 304, 390, 206]),('wall',[296, 16, 156, 482]),('wall',[72, 4, 232, 302]),('bed',[174, 256, 234, 196]),('cell',[462, 26, 48, 484]),('wall',[20, 4, 50, 458]),('shelf',[66, 48, 242, 20]),('sink',[152, 194, 40, 54])]} \\

\scriptsize 30 & \texttt{Background prompt: jail cell} \\

\scriptsize 31 & \texttt{Negative prompt: empty} \\

\scriptsize 32 & \leftskip=2cm \parindent=-2cm \texttt{Reasoning: A jail cell typically has bars, walls, a floor, and a bed. The layout could include a cell door, a shelf, and a sink.} \\

\scriptsize 33 & \texttt{} \\

\scriptsize 34 & \texttt{Provide a layout for an "badlands”} \\

\scriptsize 35 & \texttt{Answer: } \\

\scriptsize 36 & \texttt{Caption: badlands} \\

\scriptsize 37 & \leftskip=2cm \parindent=-2cm \texttt{Objects: [('earth',[0, 199, 334, 267]),('earth',[58, 201, 453, 144]),('hill',[0, 106, 512, 118]),('sky',[0, 0, 512, 116]),('earth',[194, 334, 316, 177]),('water',[34, 218, 301, 128]),('tree',[0, 369, 236, 142]),('rock',[0, 381, 97, 83]),('person',[463, 273, 18, 71]),('tripod',[450, 289, 5, 38]),('photo machine',[449, 283, 8, 8])]} \\

\scriptsize 38 & \texttt{Background prompt: badlands} \\

\scriptsize 39 & \texttt{Negative prompt: empty} \\

\scriptsize 40 & \leftskip=2cm \parindent=-2cm \texttt{Reasoning: Badlands are characterized by eroded rock formations, so the layout could include earth, hills, rocks, and trees. The badlands may also have water, a person, a tripod, and a photo machine.} \\

\scriptsize 41 & \texttt{} \\

\scriptsize 42 & \texttt{Provide a layout for an "art gallery”} \\

\scriptsize 43 & \texttt{Answer: } \\

\scriptsize 44 & \texttt{Caption: art gallery} \\

\scriptsize 45 & \leftskip=2cm \parindent=-2cm \texttt{Objects: [('wall',[224, 36, 287, 360]),('floor',[0, 323, 512, 188]),('wall',[0, 84, 226, 261]),('ceiling',[0, 0, 511, 112]),('board',[306, 153, 205, 140]),('board',[0, 170, 250, 102]),('double door',[251, 176, 55, 168]),('grill',[378, 260, 21, 91]),('grill',[338, 257, 20, 85]),('vent',[248, 22, 47, 19]),('drawing',[490, 196, 21, 40]),('spotlight',[453, 32, 17, 49]),('drawing',[8, 194, 18, 35]),('spotlight',[381, 54, 14, 43]),('drawing',[456, 241, 22, 25]),('spotlight',[279, 83, 15, 35]),('spotlight',[320, 71, 14, 38]),('drawing',[391, 187, 17, 30]),('drawing',[314, 201, 20, 26]),('vent',[6, 45, 36, 13]),('spotlight',[259, 88, 12, 34]),('drawing',[420, 196, 14, 28]),('drawing',[445, 204, 18, 22]),('drawing',[409, 239, 13, 28]),('spotlight',[234, 97, 11, 32]),('drawing',[43, 195, 18, 18]),('drawing',[351, 181, 12, 27]),('drawing',[135, 237, 17, 19]),('drawing',[40, 227, 14, 23]),('drawing',[205, 200, 11, 27])]} \\

\scriptsize 46 & \texttt{Background prompt: art gallery} \\

\scriptsize 47 & \texttt{Negative prompt: empty} \\

\scriptsize 48 & \leftskip=2cm \parindent=-2cm \texttt{Reasoning: An art gallery is indoors, so it has walls, a floor, and a ceiling. It can also have boards for displaying art, doors, grills, vents, and spotlights. The art gallery may have drawings on the walls and spotlights to illuminate the art.} \\

\scriptsize 49 & \texttt{Provide a layout for an "art gallery”} \\

\scriptsize 50 & \texttt{Answer: } \\

\scriptsize 51 & \texttt{Caption: window seat} \\

\scriptsize 52 & \leftskip=2cm \parindent=-2cm \texttt{Objects: [('seat',[2, 172, 507, 337]),('floor',[28, 322, 482, 187]),('wall',[102, 0, 266, 228]),('wall',[0, 0, 109, 510]),('person',[222, 20, 133, 390]),('wall',[363, 0, 146, 324]),('windowpane',[140, 0, 204, 69]),('windowpane',[0, 0, 102, 122]),('windowpane',[388, 0, 122, 75]),('hat',[375, 157, 80, 69])]} \\

\scriptsize 53 & \texttt{Background prompt: window seat} \\

\scriptsize 54 & \texttt{Negative prompt: empty} \\

\scriptsize 59 & \leftskip=2cm \parindent=-2cm \texttt{Reasoning: A window seat typically has a seat, walls, and a floor. The layout could include a person sitting on the seat, looking out the window, and wearing a hat.} \\

\hline
\captionsetup{width=15cm, justification=centering}
\caption{ Our in-context examples. We use fixed in-context examples for layout generation.}\label{tab:incontext}
 
\end{longtable}
\twocolumn

\clearpage
\clearpage
\section{Comparison to DDIM}
\label{sec:ddim_comp}
We initially considered a DDIM ~\cite{song2022denoisingdiffusionimplicitmodels} based approach rather than rectified flow. However, early experiments showed less promise in this direction. DDIM models struggled with generating the correct CLIP embeddings, leading to meaningless images that did not match the prompt, whereas rectified flow-based approaches were more successful without needing to search the hyperparameter space. 

We provide an example here, from a model with an identical architecture to our presented model (including all hyperparameters specified in ~\cref{sec:training_data_and_hyperparameters}, except it is trained with a DDIM training objective and performs DDIM inference (with a log-linear noise schedule from $\sigma = 0.02$ to $\sigma=1$). This is not an exhaustive search by any means, but is intended as a point-of-reference for other researchers.  

\begin{figure*}[ht]
    \centering
    \begin{tabular}{c c c} 
        \includegraphics[width=0.3\textwidth]{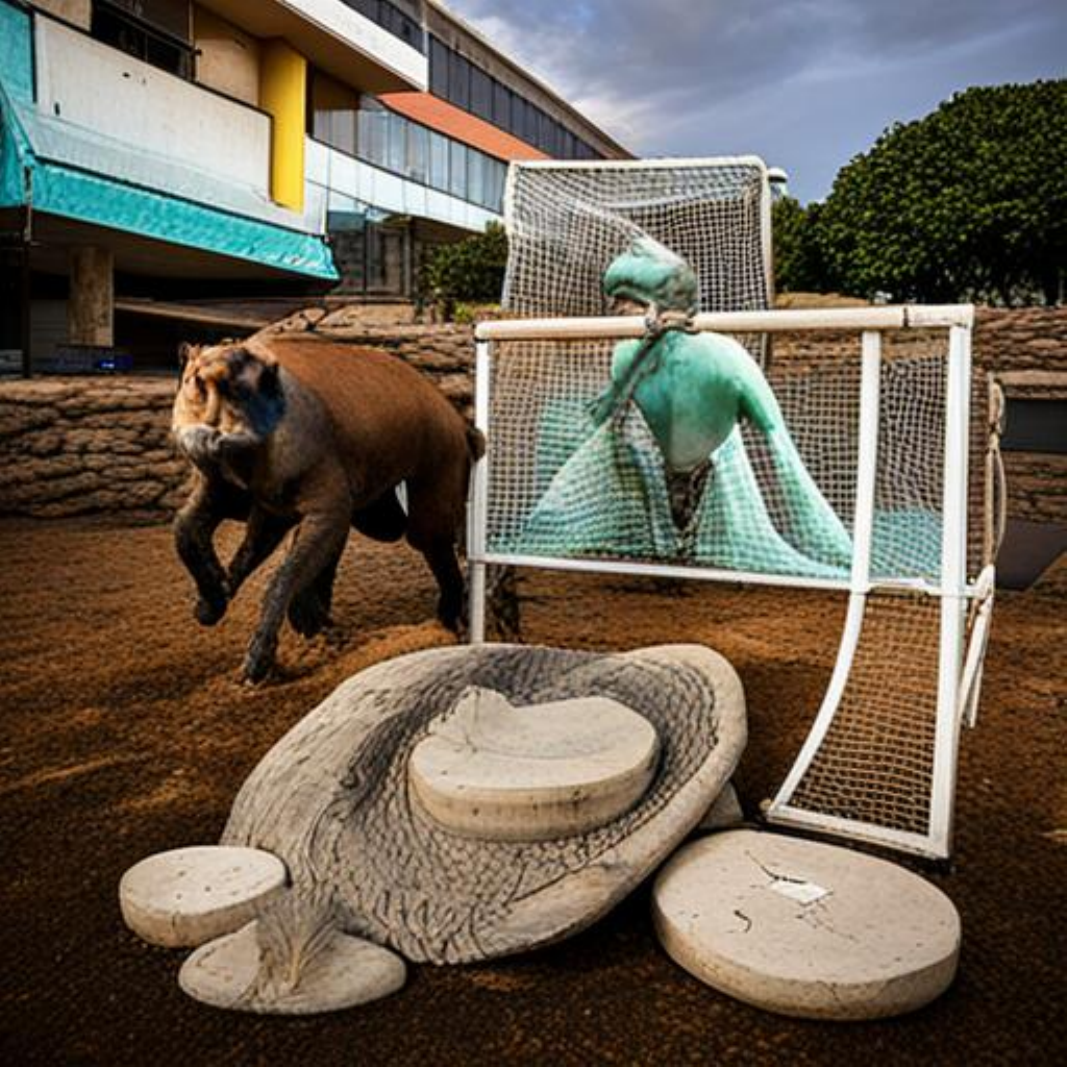} &
        \includegraphics[width=0.3\textwidth]{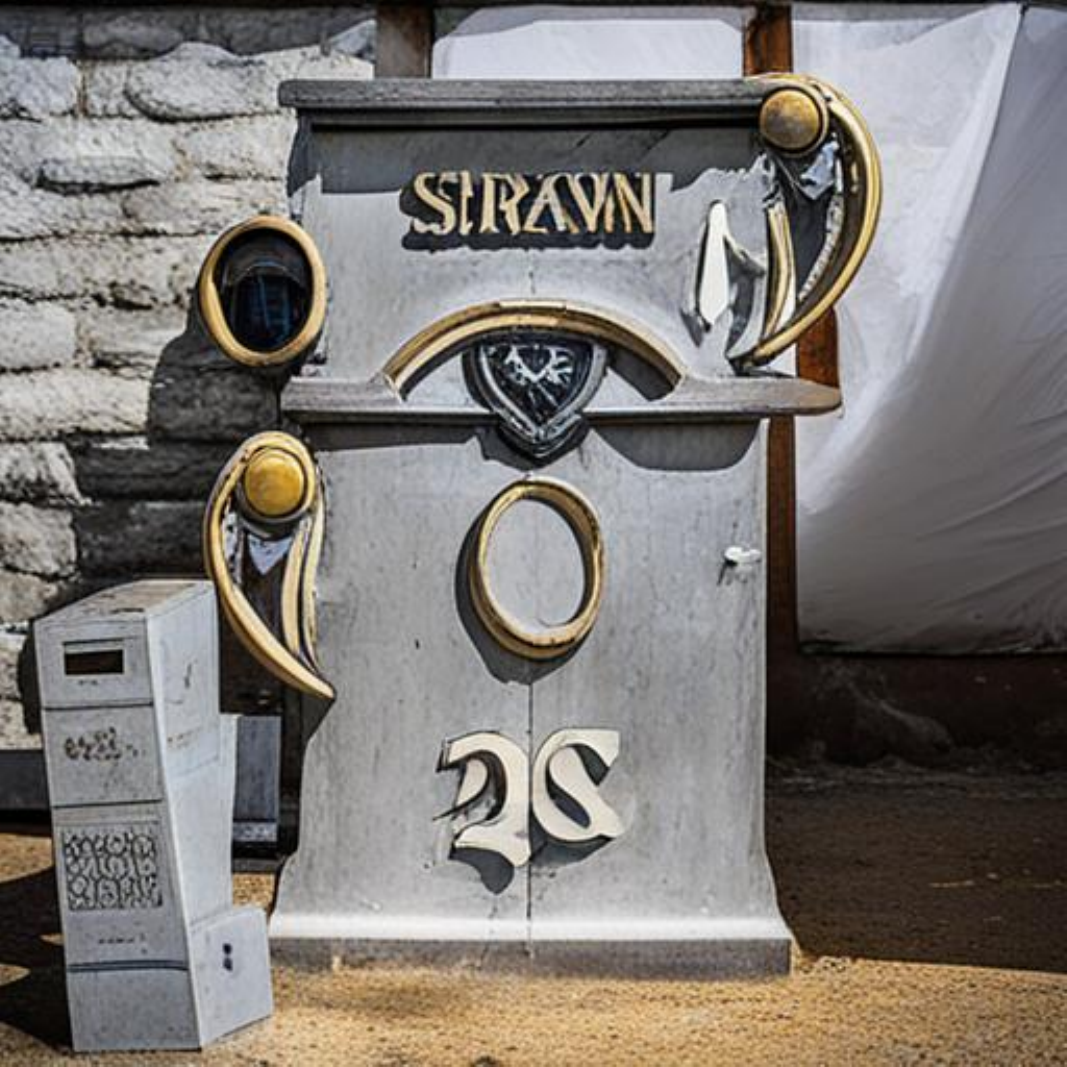} &
        \includegraphics[width=0.3\textwidth]{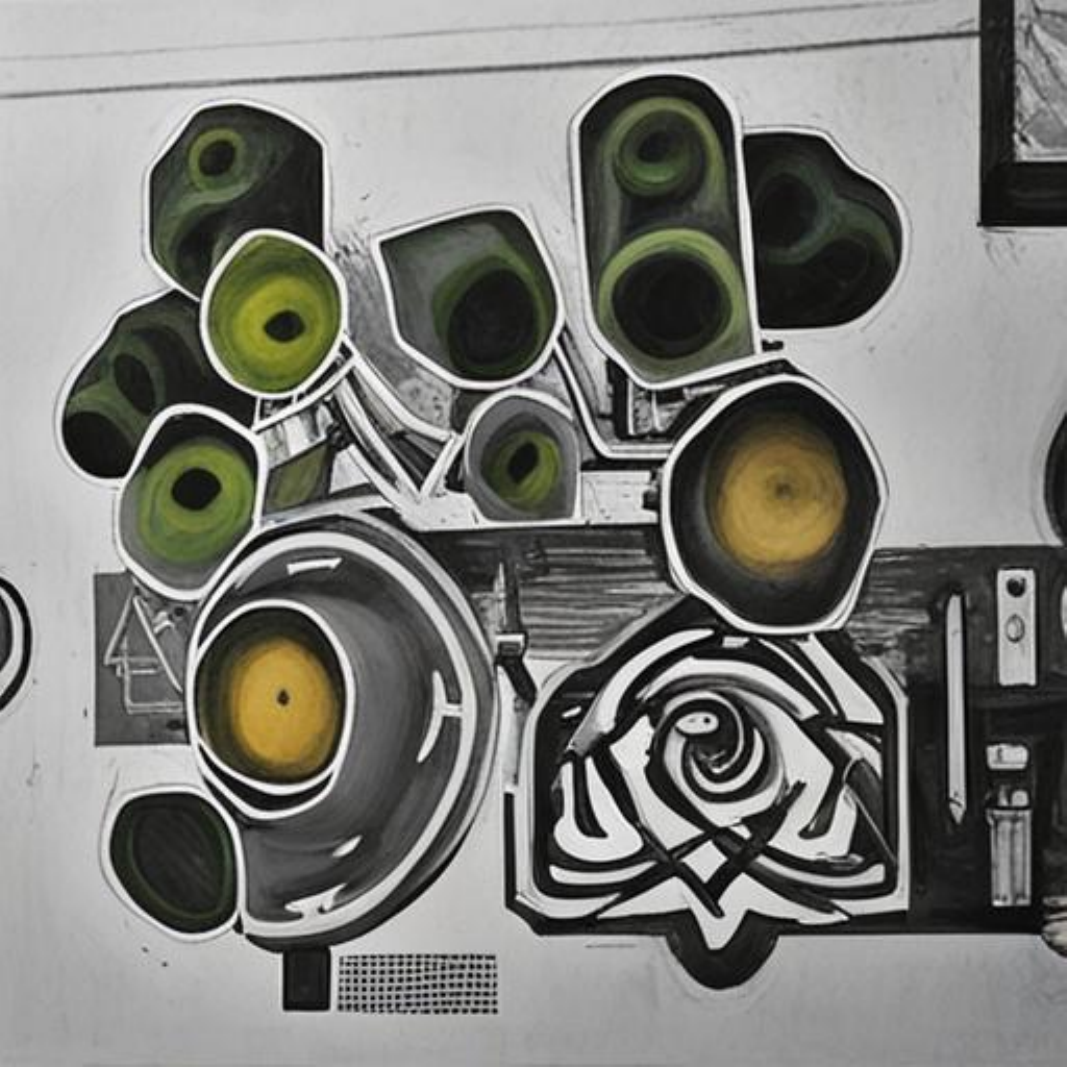} \\
        
        \includegraphics[width=0.3\textwidth]{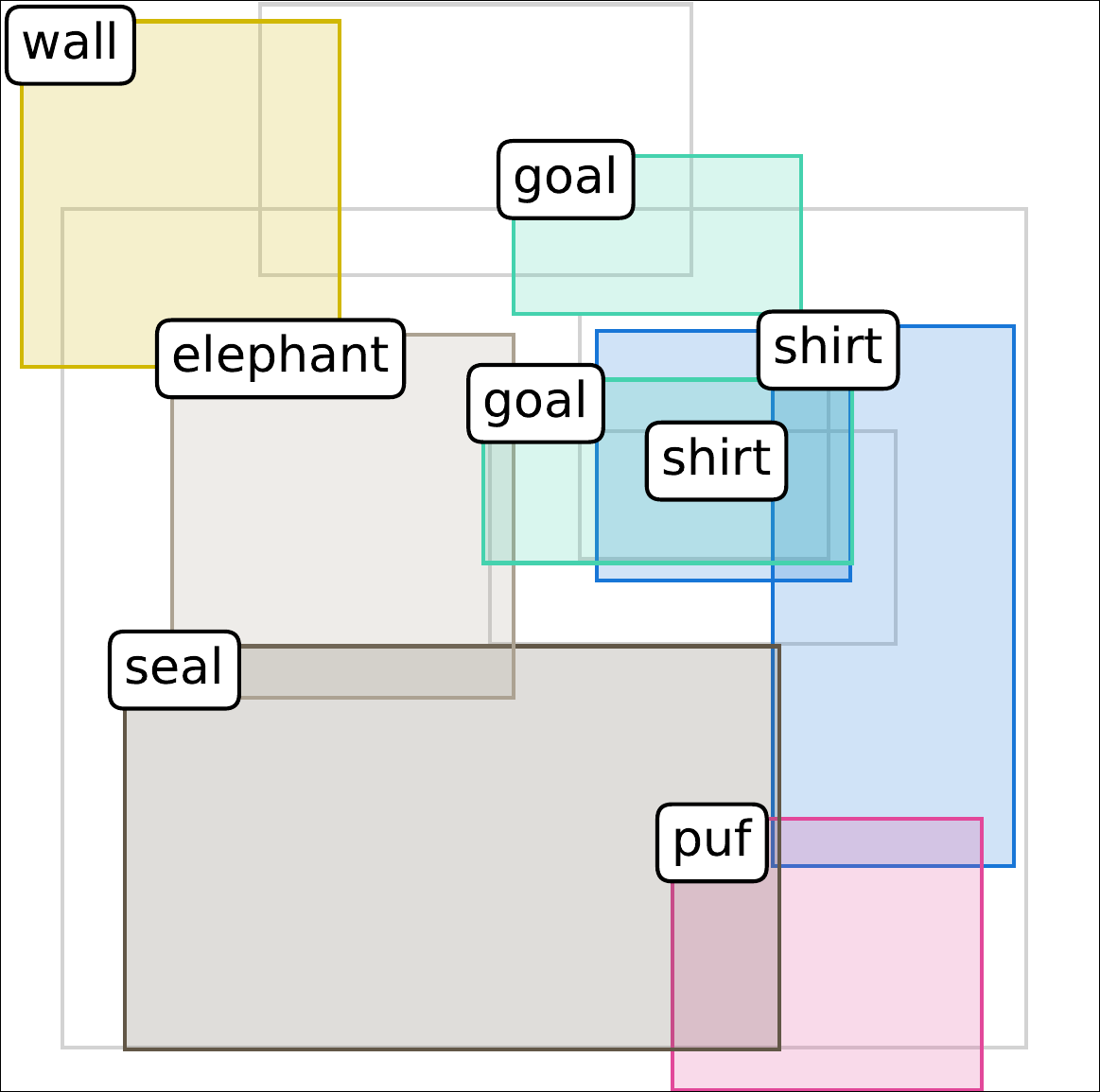} &
        \includegraphics[width=0.3\textwidth]{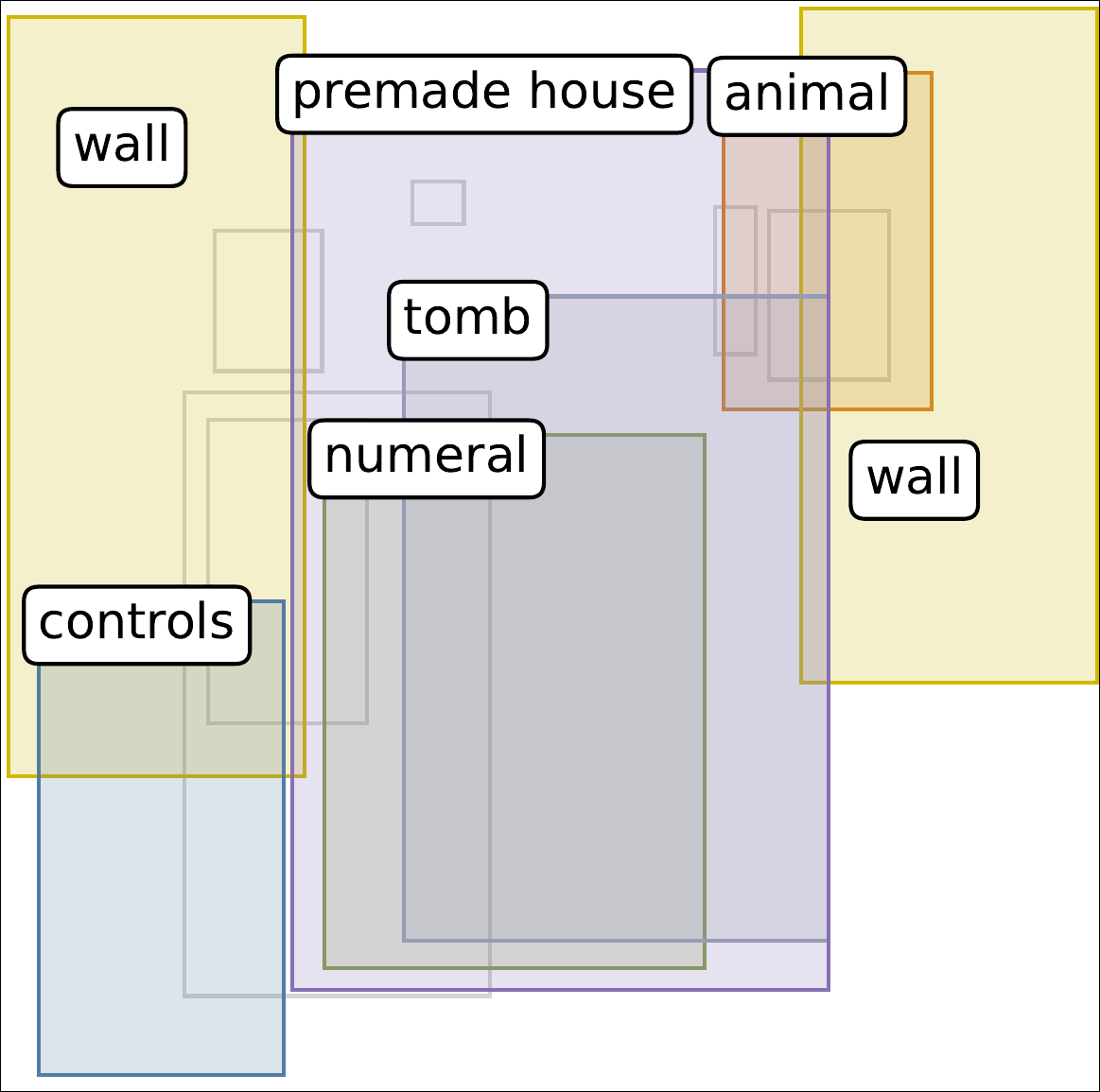} &
        \includegraphics[width=0.3\textwidth]{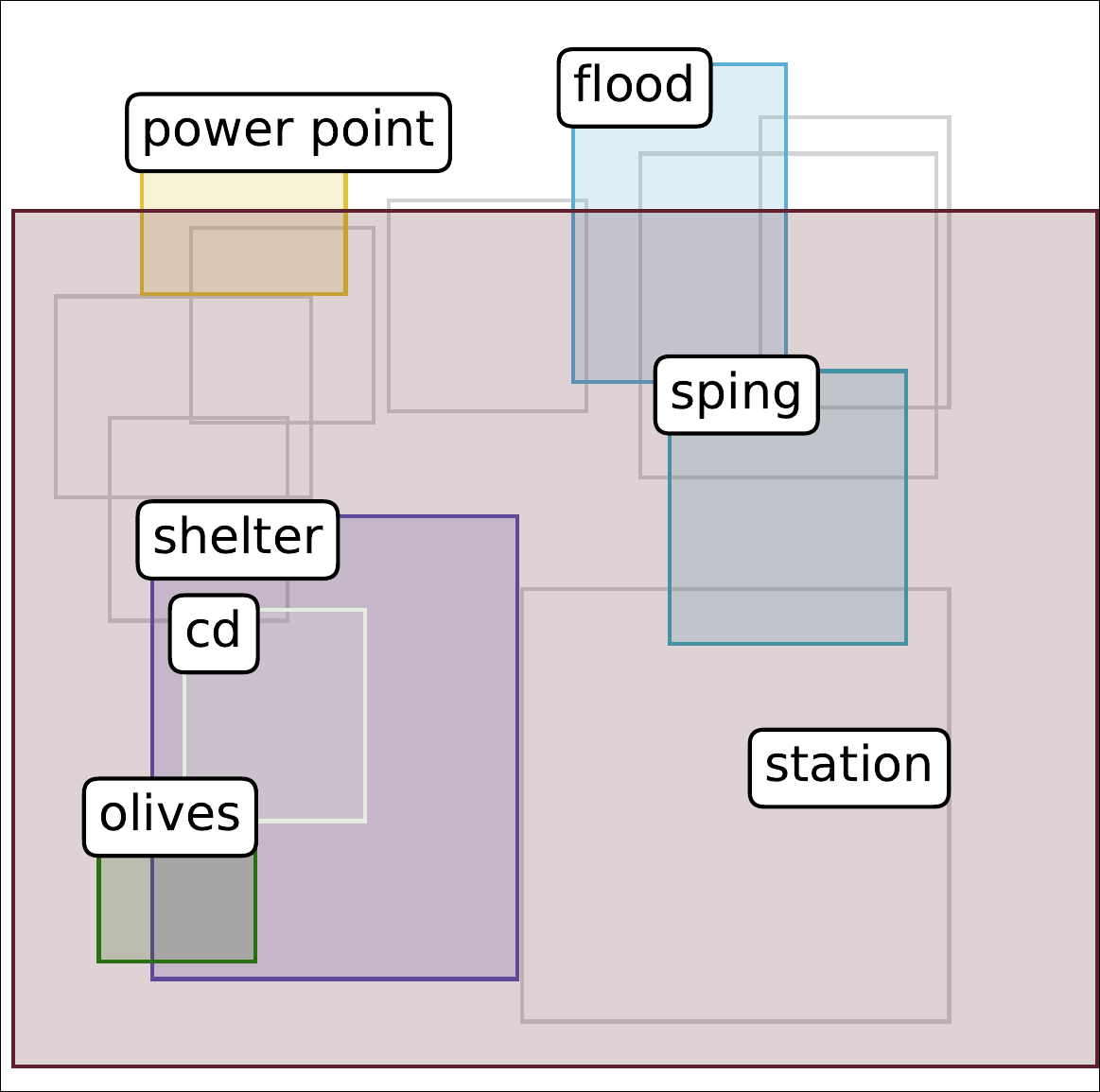} \\
    \end{tabular}
    \caption{Our Model with DDIM instead of Rectified Flow - Street. The bounding box labels match poorly to the desired scene, and the resulting images appear to be implausible.}
    \label{fig:diffusion_examples}
\end{figure*}

We show our statistics in ~\cref{tab:diffusion_ablation_stats}, and some visual examples from the model in ~\cref{fig:diffusion_examples}. We speculate that the straighter transit paths of samples rectified flow ~\cite{liu2022flowstraightfastlearning} increases the model's ability to effectively learn high dimensional data like the PCA-reduced CLIP embeddings.

\begin{table*}[ht]
\footnotesize
\newrobustcmd\B{\DeclareFontSeriesDefault[rm]{bf}{b}\bfseries}  
\def\Uline#1{#1\llap{\uline{\phantom{#1}}}}

\sisetup{detect-weight=true,
         mode=text,
         table-format=2.2,   
         add-integer-zero=false,
         table-space-text-post={*},
         table-align-text-post=false
         }

    \centering
        \begin {tabular}{
        l
        S
        S
        S
        S
        S
        S
        S
        S
        S
        S
        S
        c   
        }
        \toprule
        \toprule
        {\textbf{Model}} & {\textbf{FID} ($\downarrow$)} & {\textbf{KID} ($10^{-2}$)($\downarrow$)} & {\textbf{CMMD} ($\downarrow$)} & {$\objectNumeracyScore(\downarrow)$} & 
        {$\firstOrderPositionalLikelihood (\uparrow)$} &
        {$\secondOrderPositionalLikelihood (\uparrow)$} &    
        {$\positionalVarianceScore(\uparrow)$}
        \\
        
        \midrule

        \textbf{Ours (DDIM)} & 0.95 & 8.60 & 1.77 &  7.89 & 4.33 & 0.01 & \B 239\\
        \midrule
        \textbf{Ours (Rectified Flow)} & \B 0.17 & \B 0.27 & \B 0.03  & \B 1.91 & \B 4.76 & \B 2.03 & 187  \\
        \bottomrule
        \bottomrule
        
        \end{tabular}
    \caption{Generated image metrics, and our generated layout numerical metrics applied on our model architecture with DDIM or rectified flow. Our model performs better on everything except positional variance $\positionalVarianceScore$, but this is at the cost of the layouts being largely nonsense (see ~\cref{fig:diffusion_examples}) }
    \label{tab:diffusion_ablation_stats}
\end{table*}

\clearpage
\section{Additional Images and Layouts}\label{sec:bonus_images}
\begin{figure*}[ht]
    \centering
    \begin{tabular}{c c c}
        \includegraphics[width=0.3\textwidth]{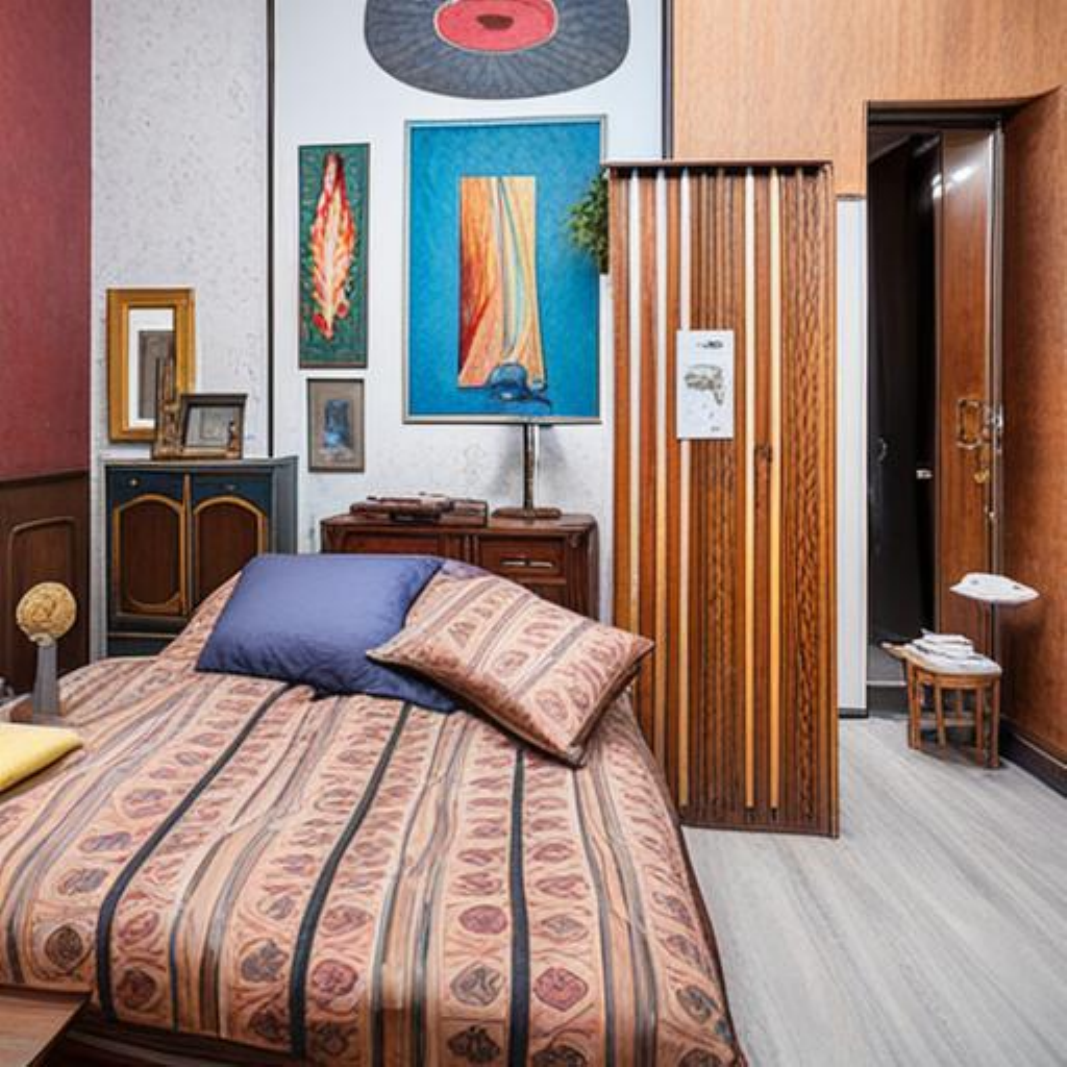} &
        \includegraphics[width=0.3\textwidth]{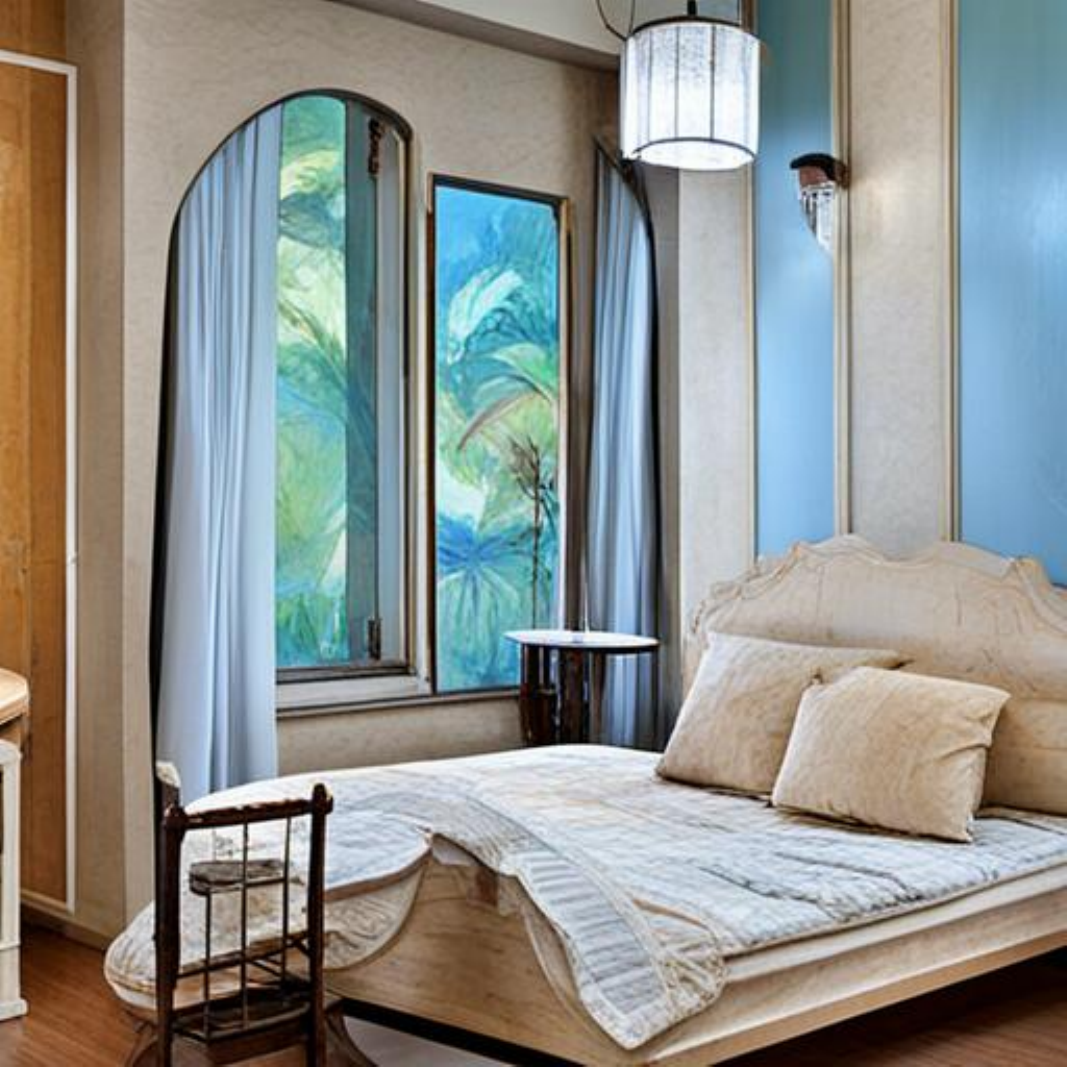} &
        \includegraphics[width=0.3\textwidth]{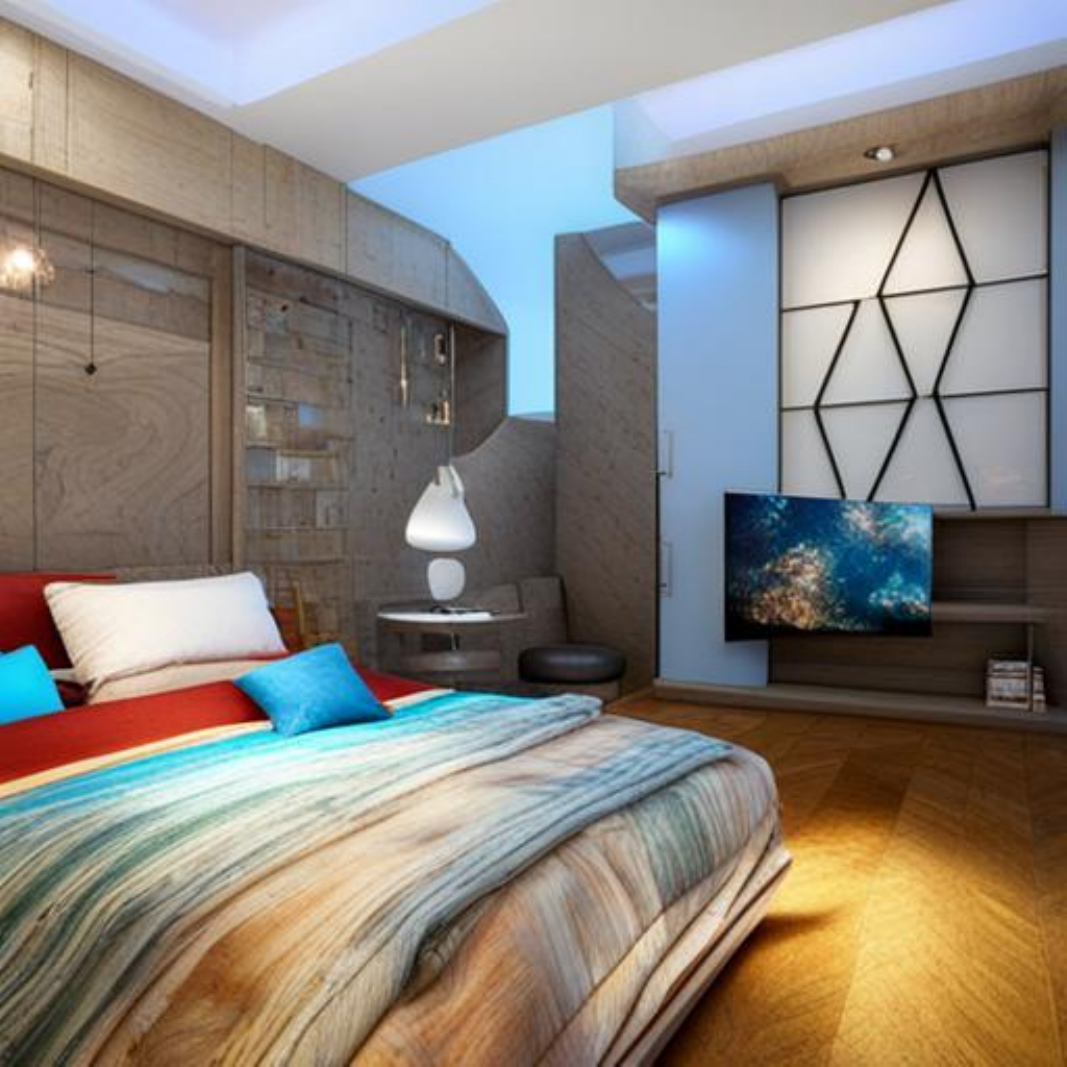} \\
        
        \includegraphics[width=0.3\textwidth]{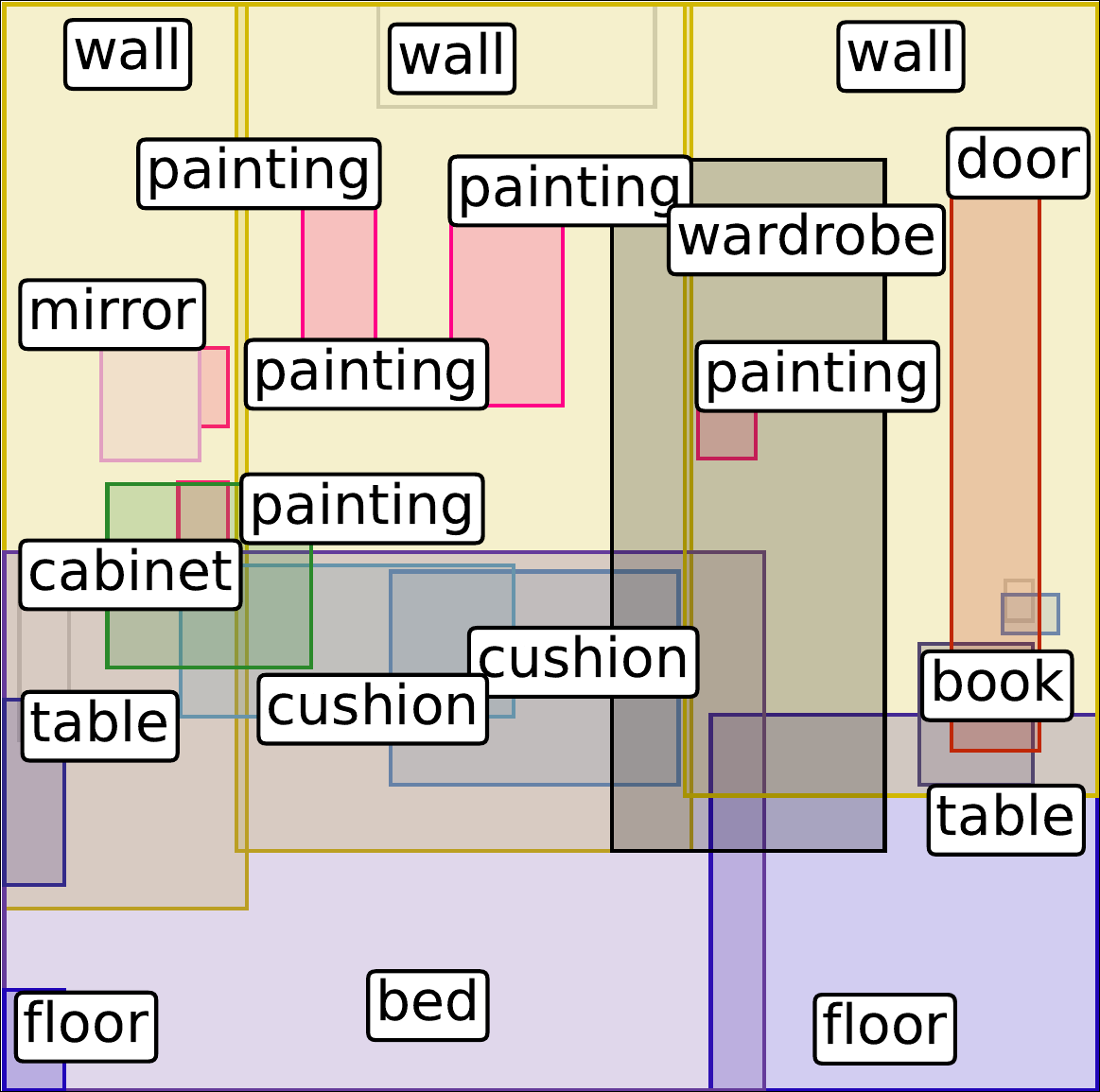} &
        \includegraphics[width=0.3\textwidth]{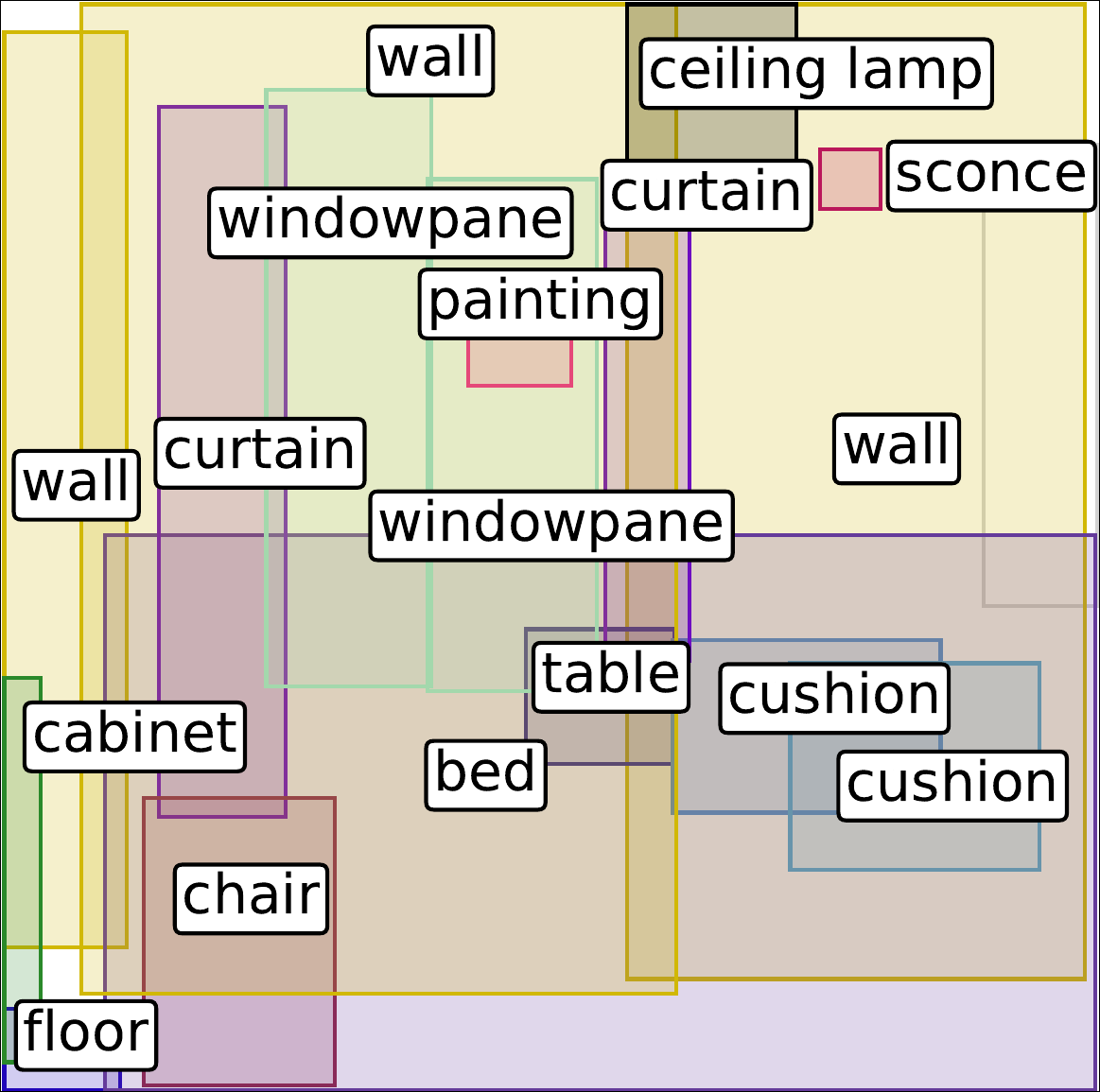} &
        \includegraphics[width=0.3\textwidth]{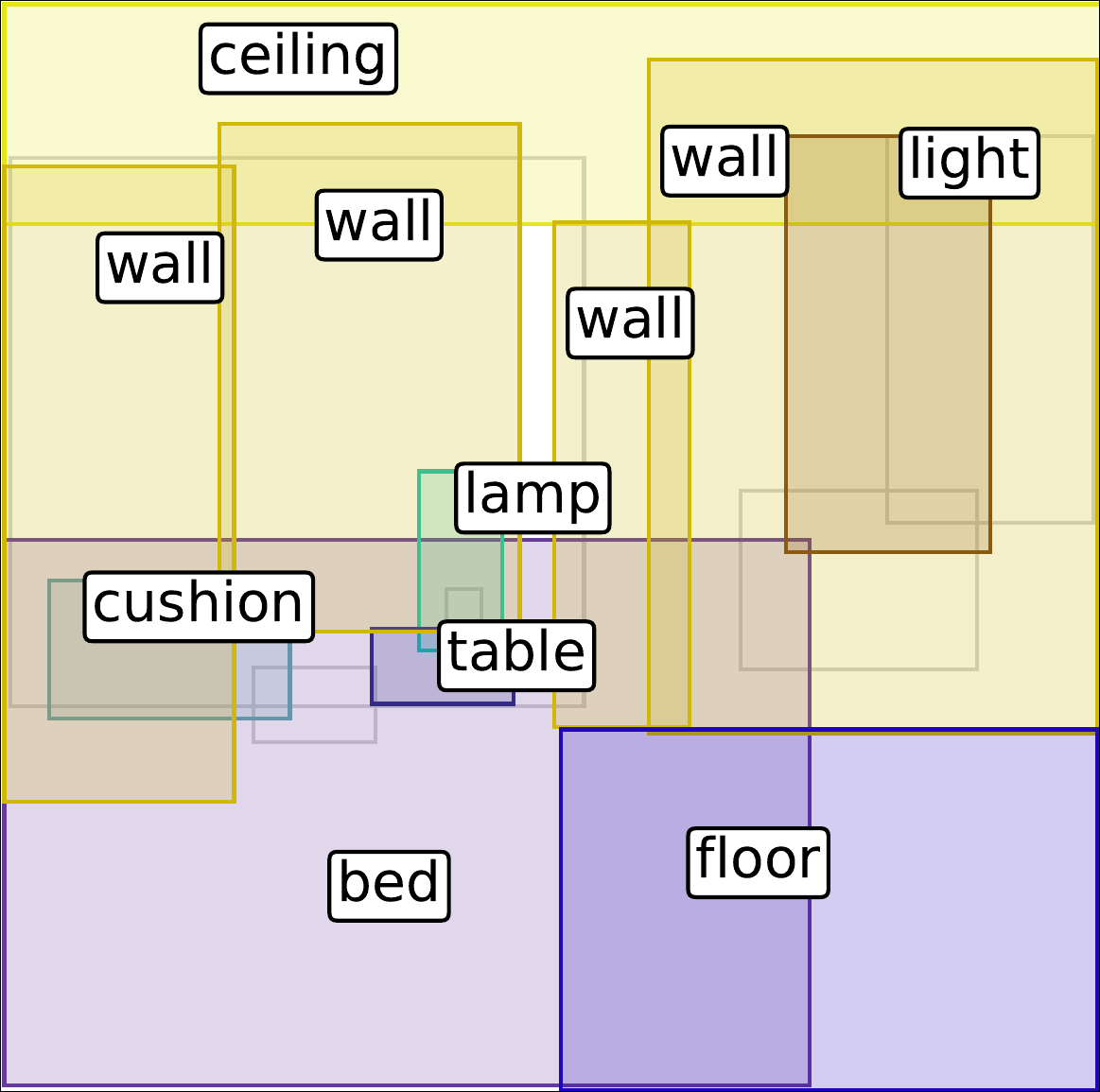} \\
    \end{tabular}
    \caption{Ours - Bedroom.}
    \label{fig:ours_bedroom}
\end{figure*}

\begin{figure*}[ht]
    \centering
    \begin{tabular}{c c c}
        \includegraphics[width=0.3\textwidth]{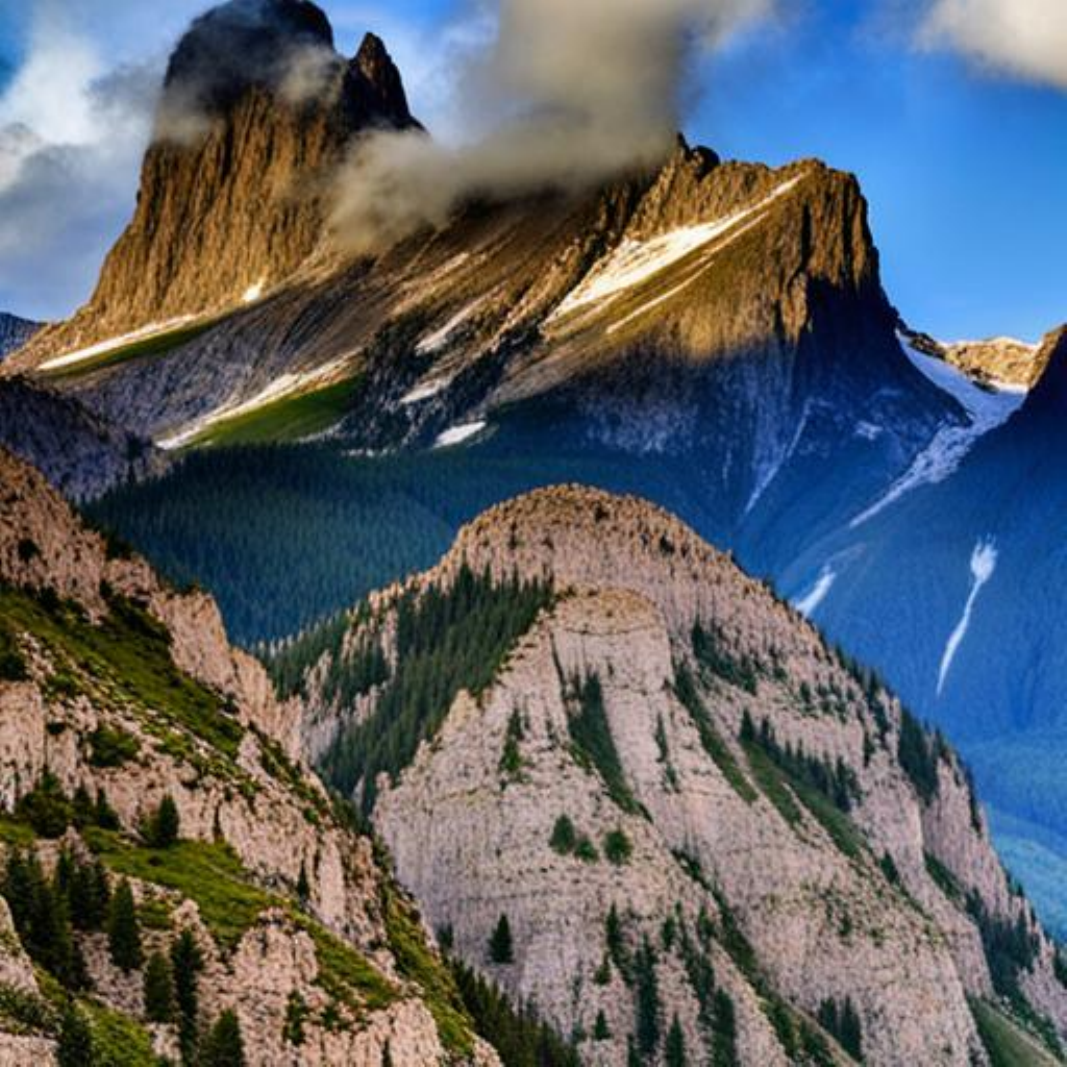} &
        \includegraphics[width=0.3\textwidth]{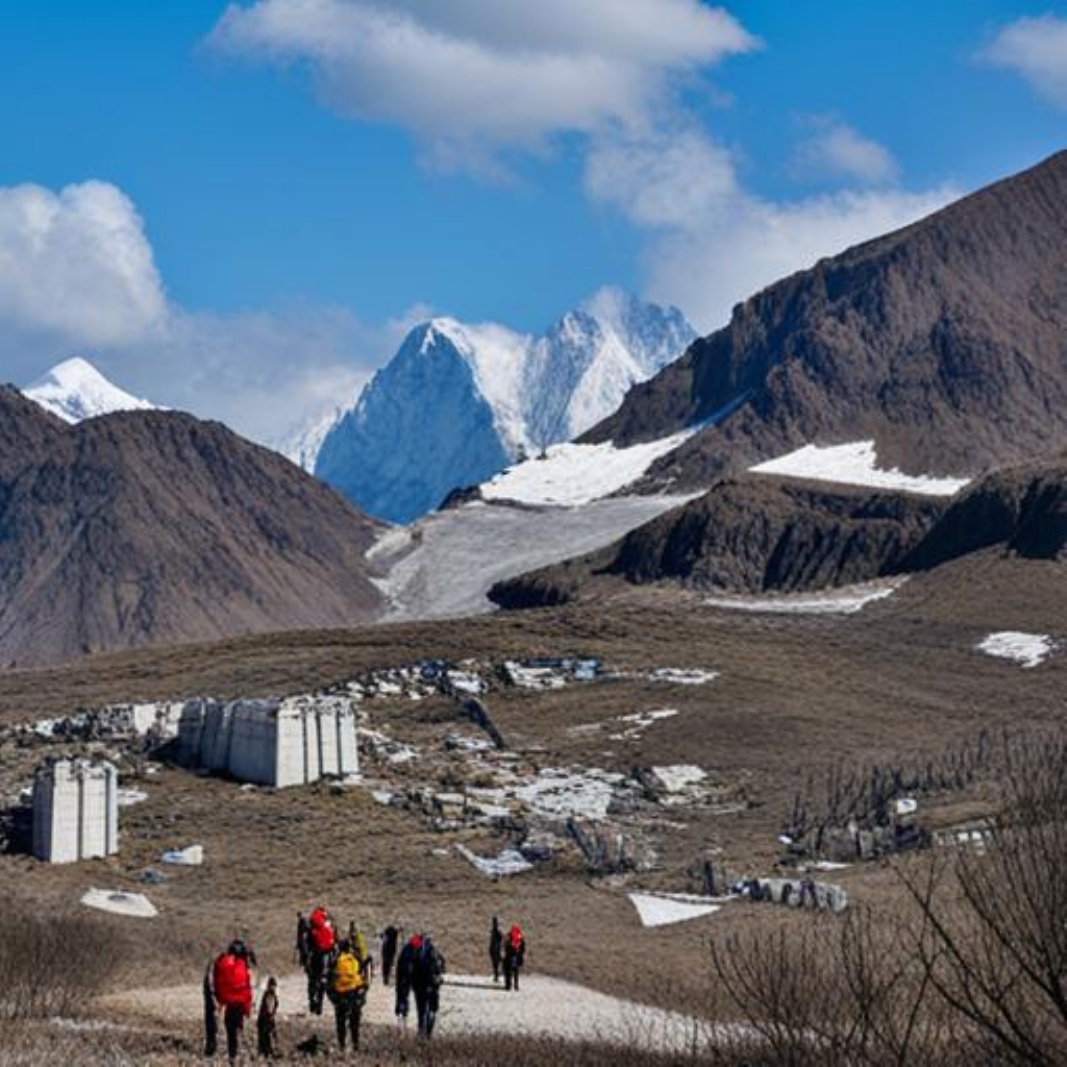} &
        \includegraphics[width=0.3\textwidth]{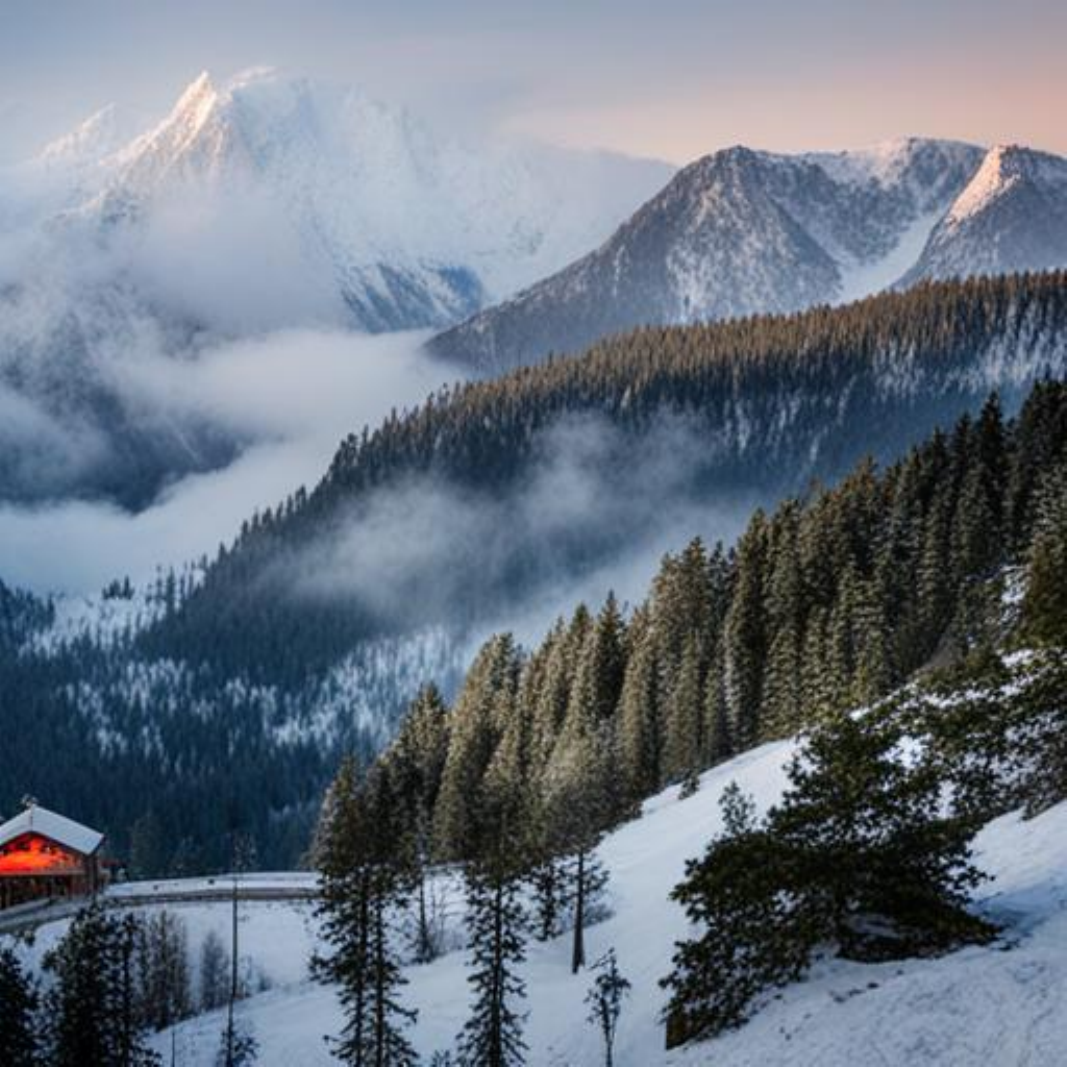} \\
        
        \includegraphics[width=0.3\textwidth]{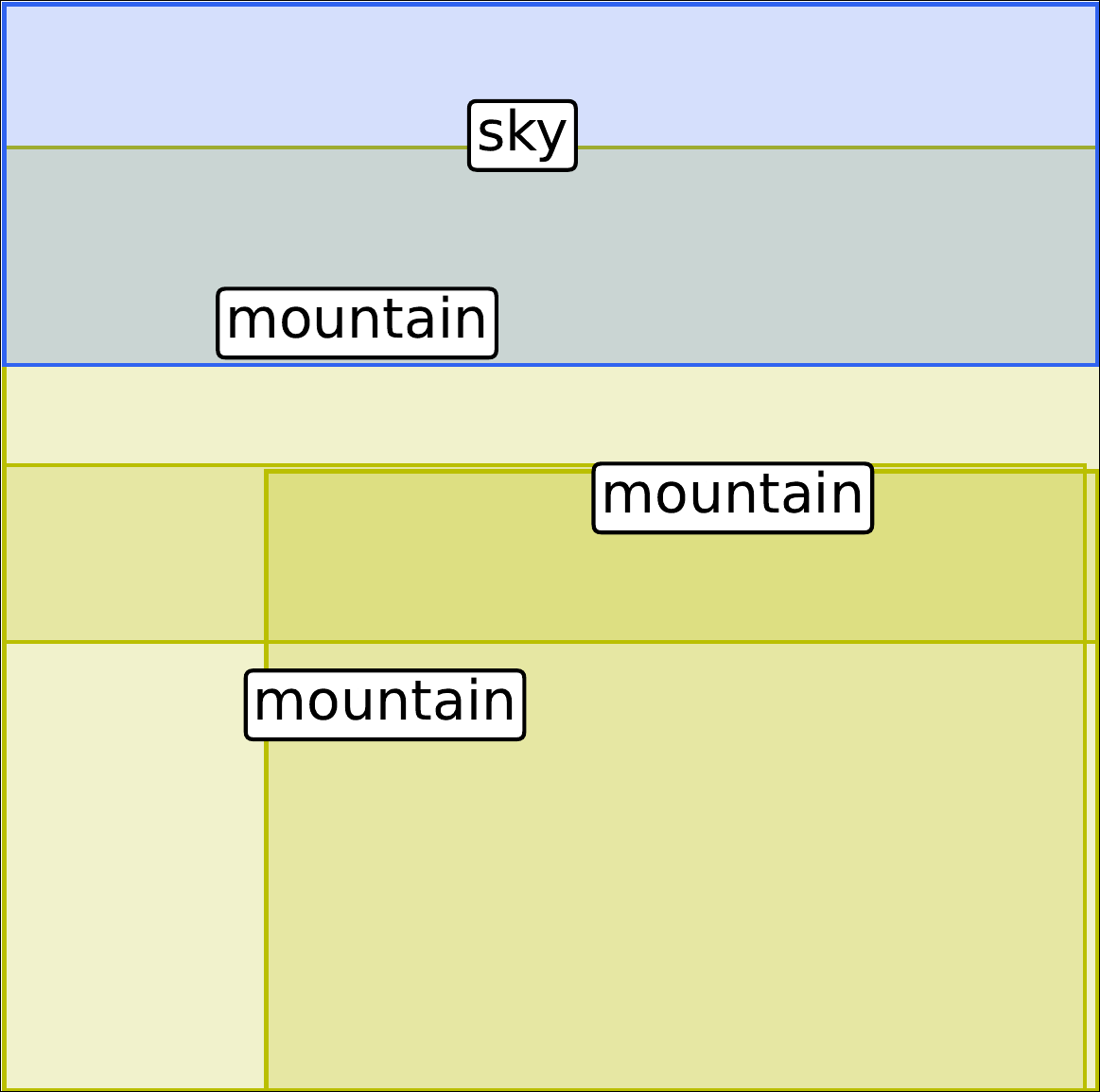} &
        \includegraphics[width=0.3\textwidth]{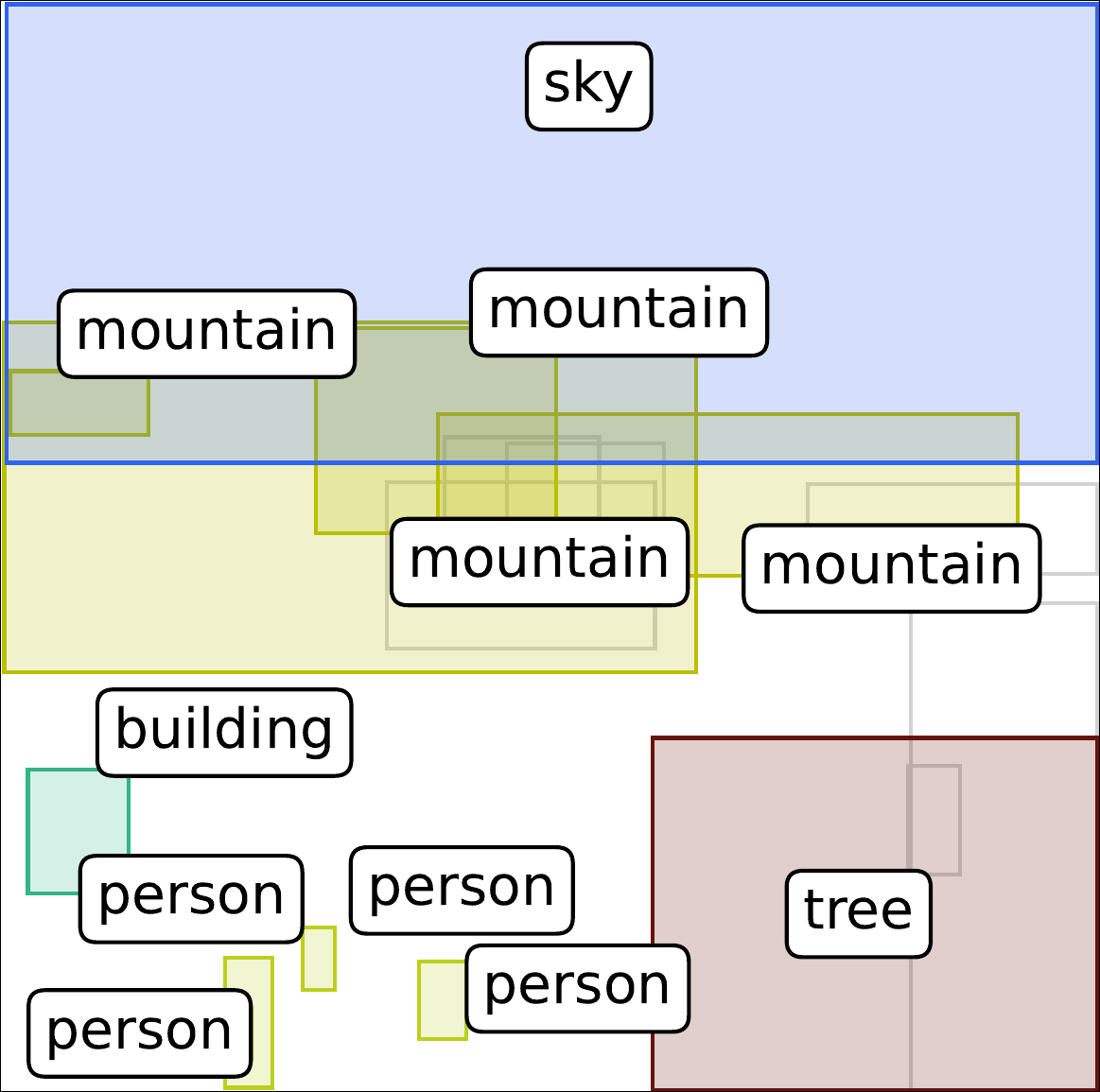} &
        \includegraphics[width=0.3\textwidth]{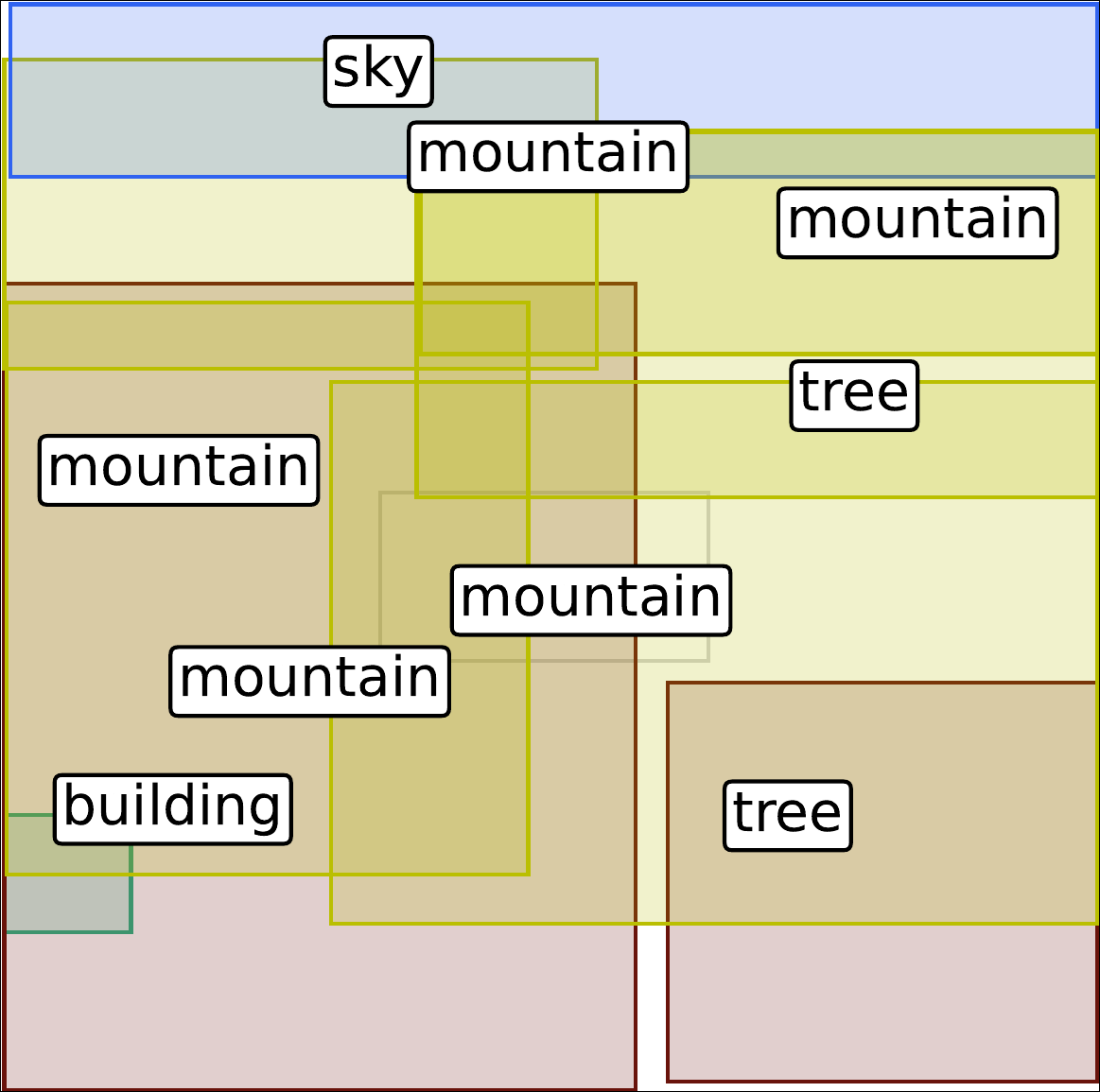} \\
    \end{tabular}
    \caption{Ours - Mountain.}
    \label{fig:ours_mountain}
\end{figure*}

\begin{figure*}[ht]
    \centering
    \begin{tabular}{c c c}
        \includegraphics[width=0.3\textwidth]{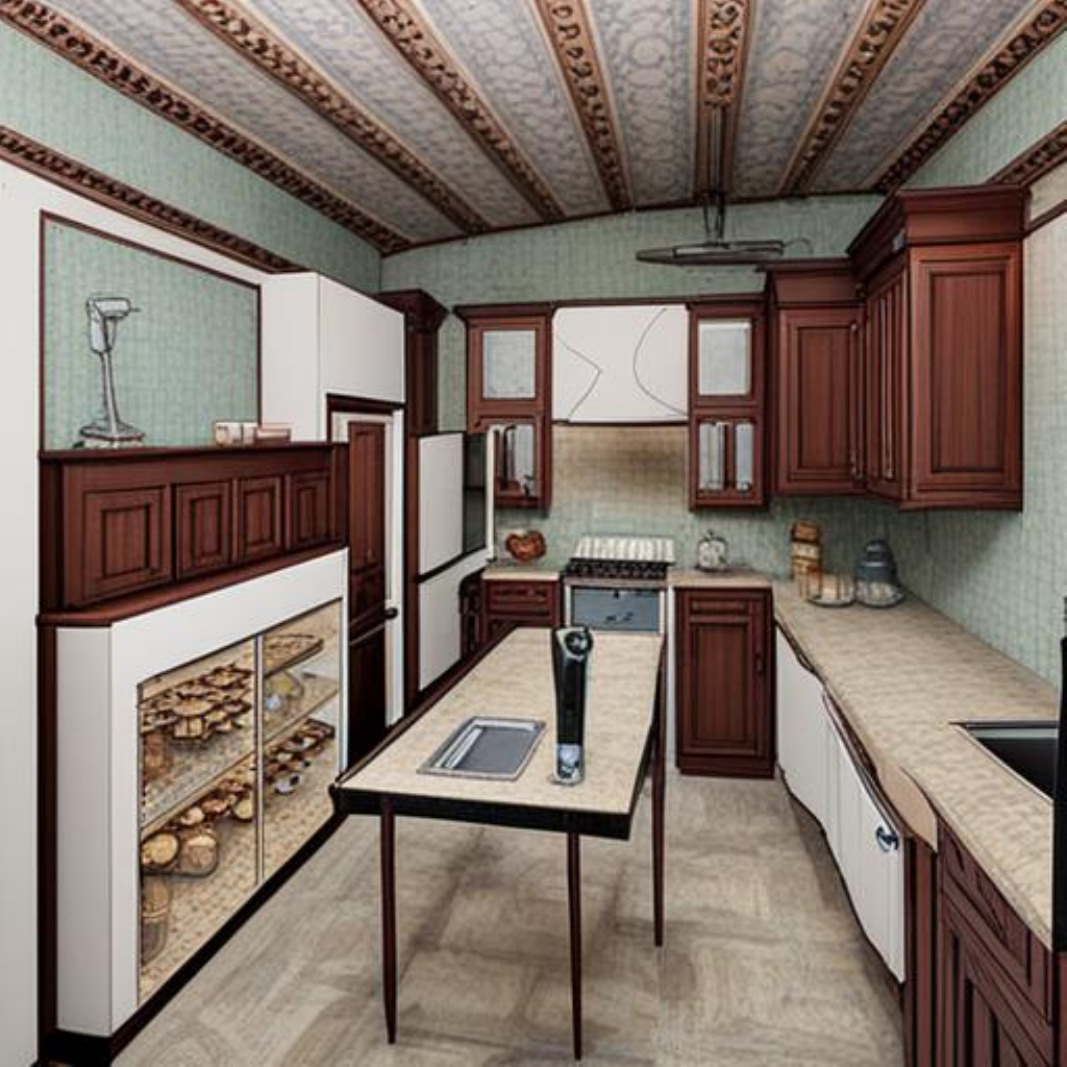} &
        \includegraphics[width=0.3\textwidth]{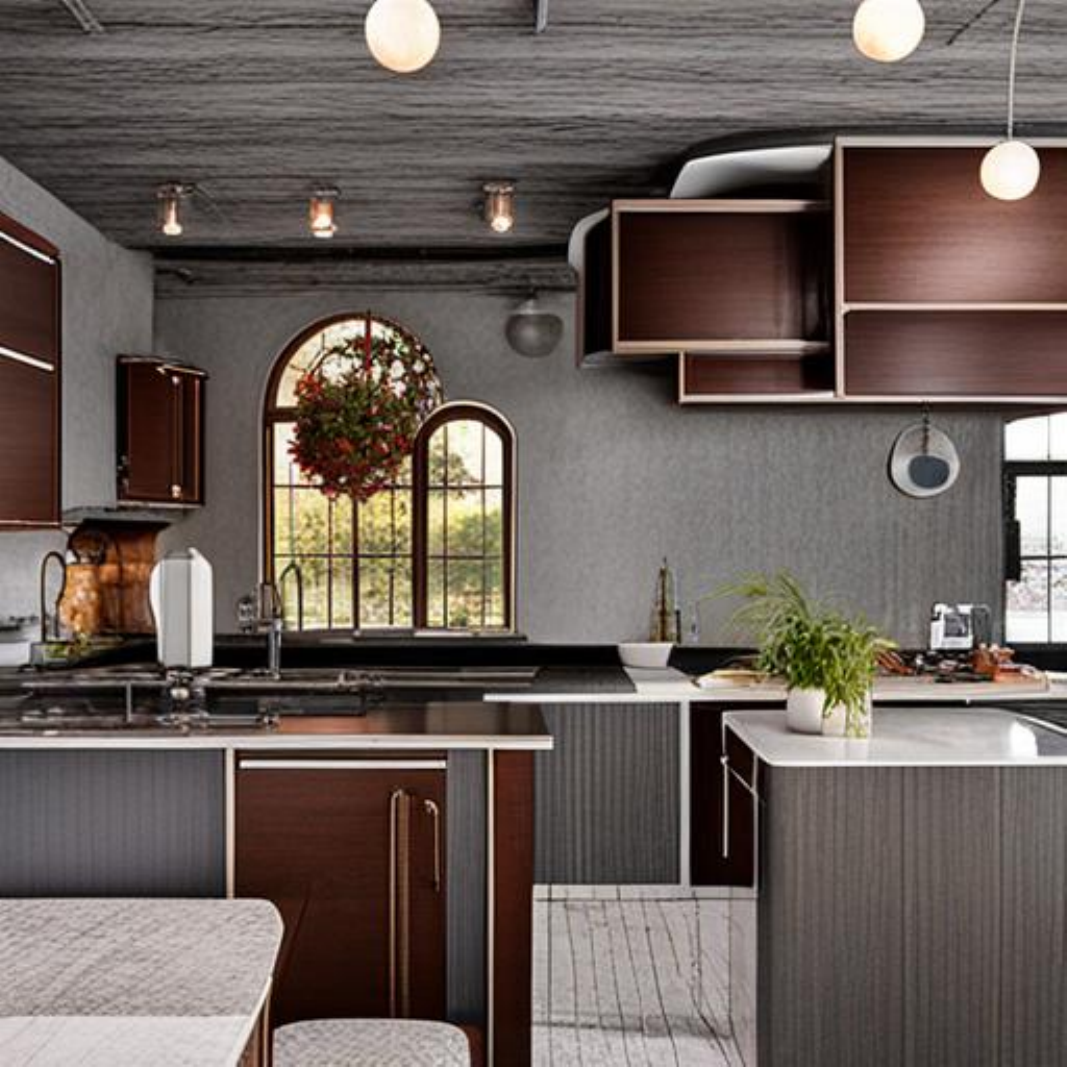} &
        \includegraphics[width=0.3\textwidth]{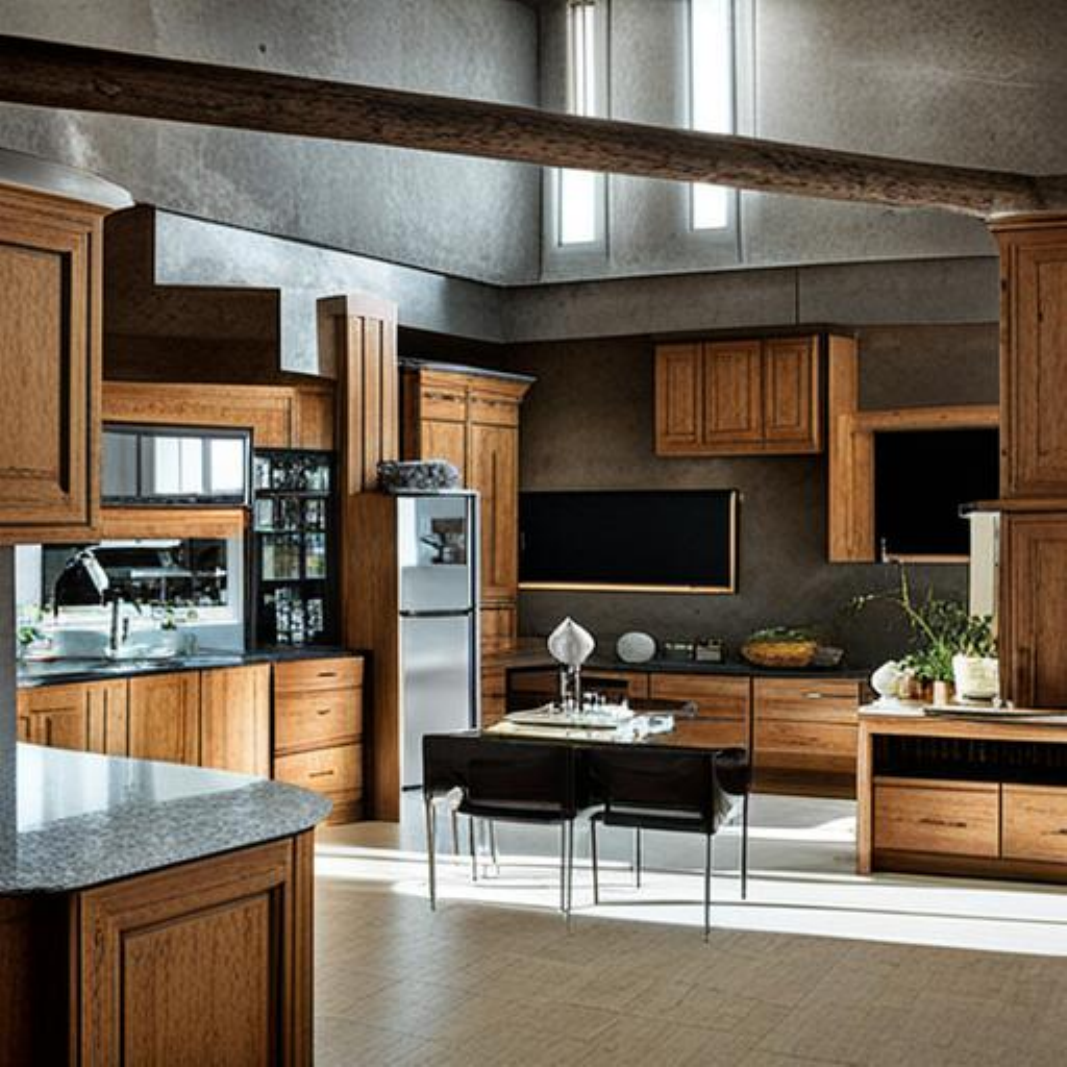} \\
        
        \includegraphics[width=0.3\textwidth]{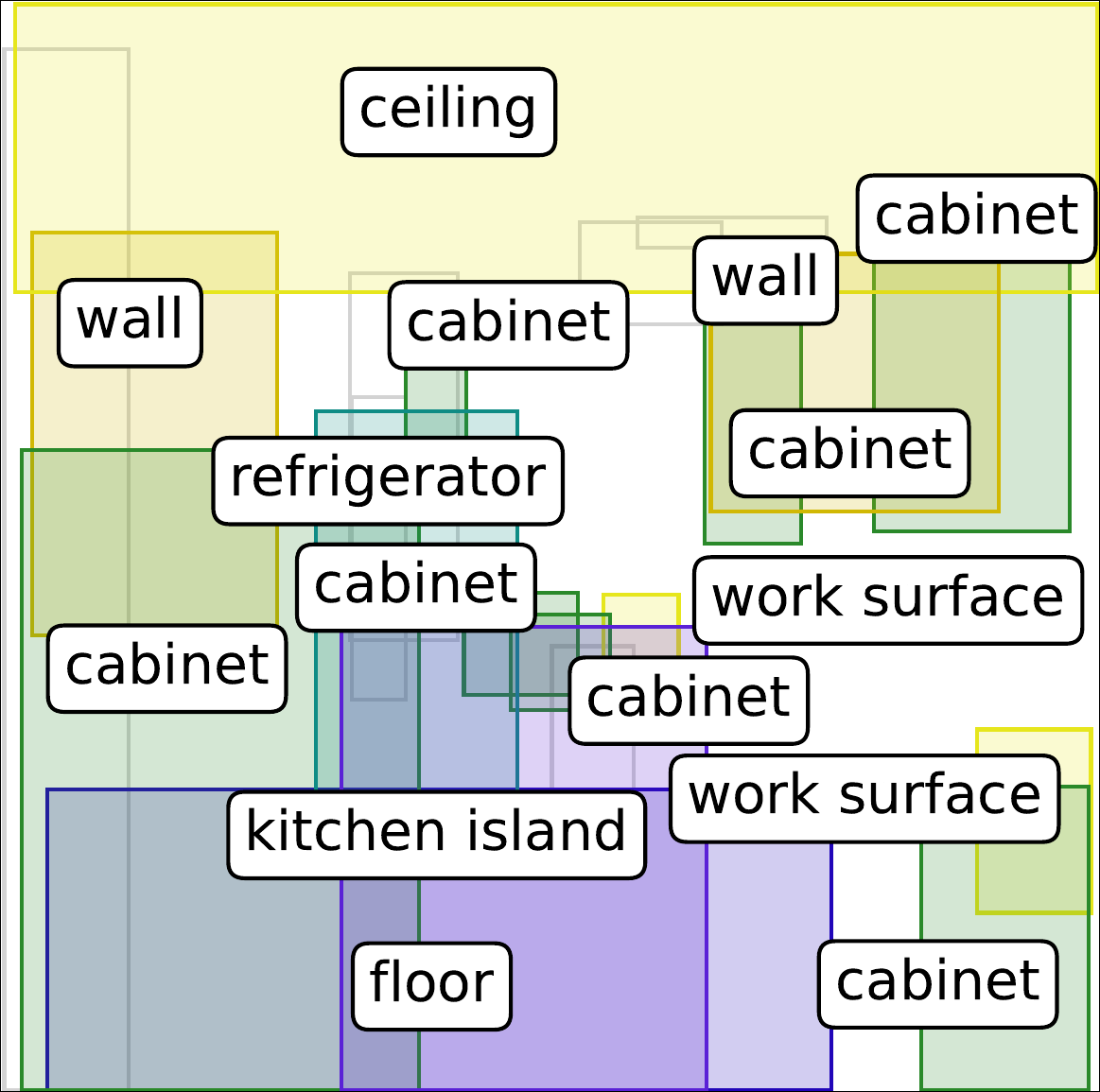} &
        \includegraphics[width=0.3\textwidth]{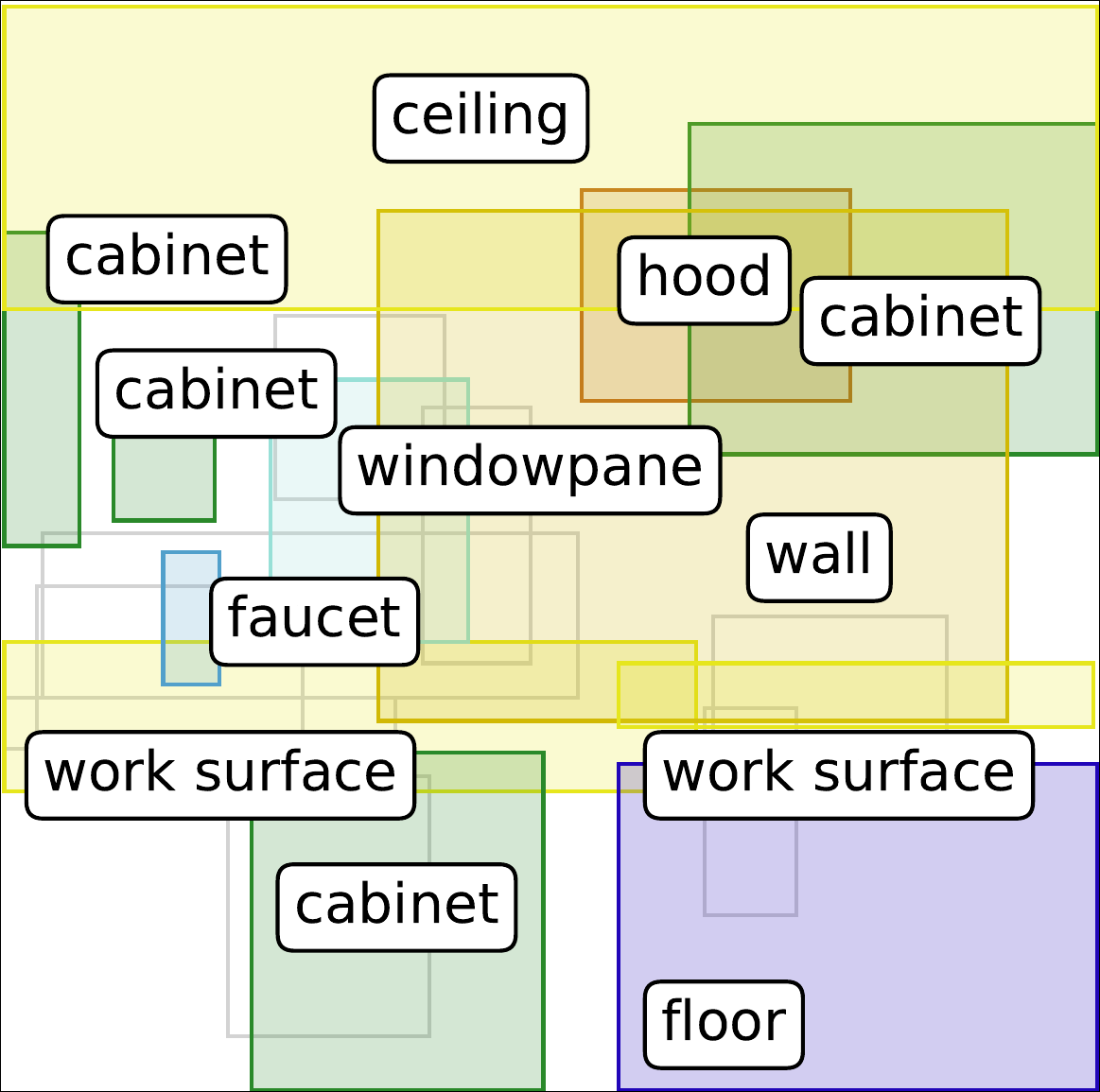} &
        \includegraphics[width=0.3\textwidth]{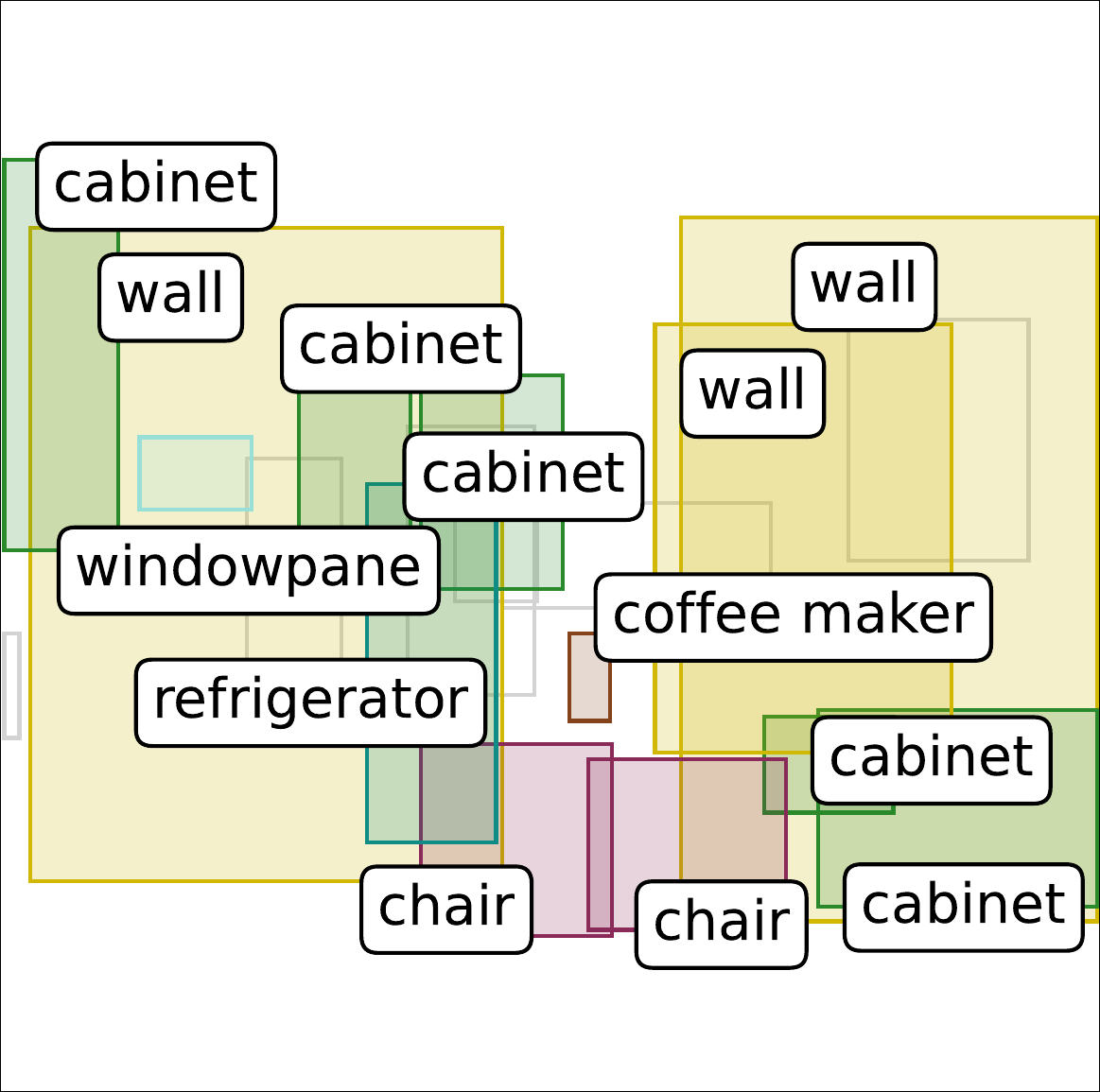} \\
    \end{tabular}
    \caption{Ours - Kitchen.}
    \label{fig:ours_kitchen}
\end{figure*}

Here we present additional examples of our model's generated layouts, and conditionally generated images, for the prompts \emph{bedroom} (\cref{fig:ours_bedroom}), \emph{mountain} (\cref{fig:ours_mountain}), and \emph{kitchen} (\cref{fig:ours_kitchen}).

\clearpage
\section{Print-ready Main Results Diagram} 
\label{sec:high_res}
For readers who prefer the document on paper, we include our visual results diagram from ~\cref{fig:visual_results} at a size where the annotations are large enough to be printed clearly. The annotations can also be zoomed into on our main document.

\begin{figure*}[ht]
    \centering
    \begin{tabular}{c c c}
        \includegraphics[width=0.3\textwidth]{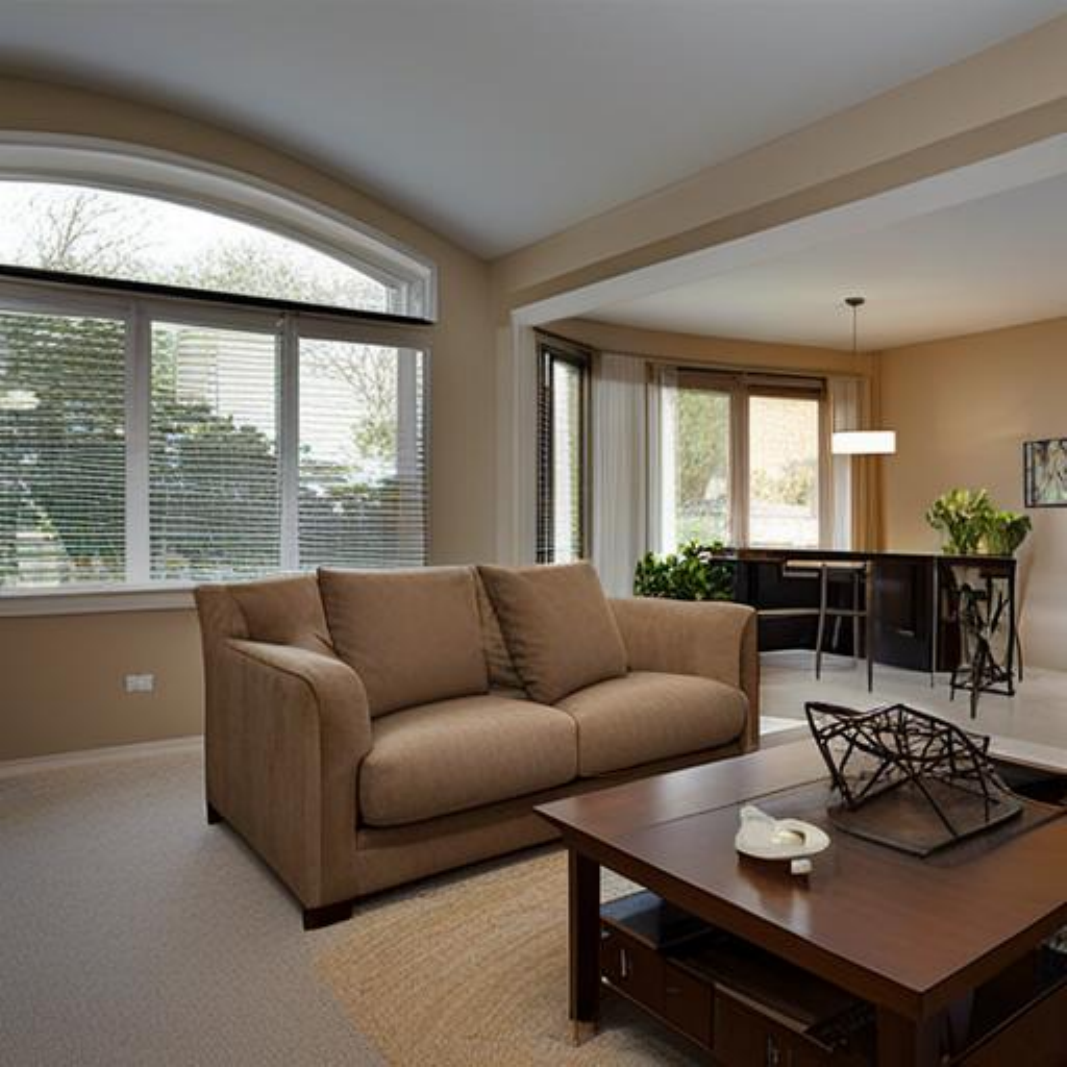} &
        \includegraphics[width=0.3\textwidth]{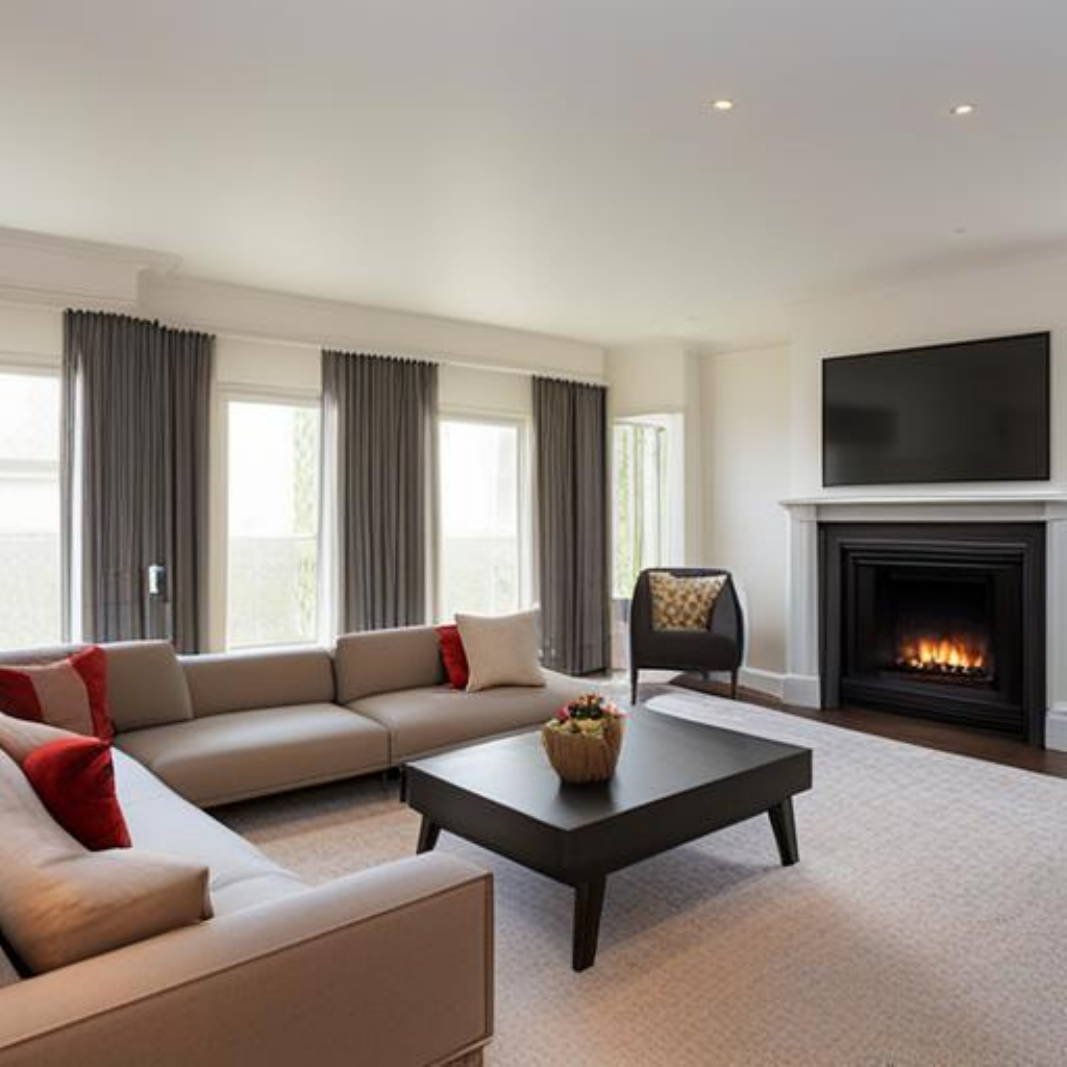} &
        \includegraphics[width=0.3\textwidth]{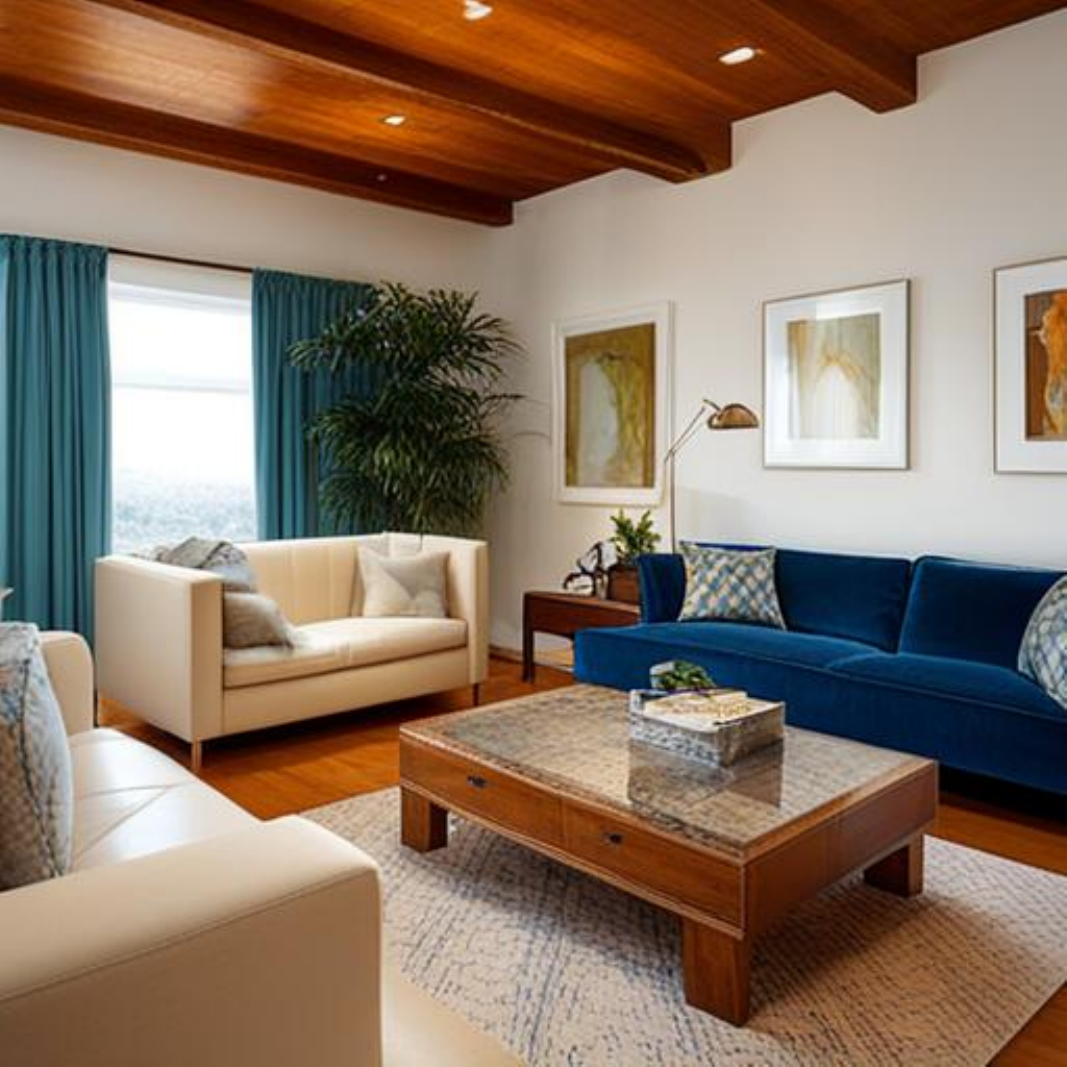} \\
        
    \end{tabular}
    \caption{No Layout - Living Room.}
    \label{fig:no_layout_living_room}
\end{figure*}

\begin{figure*}[ht]
    \centering
    \begin{tabular}{c c c}
        \includegraphics[width=0.3\textwidth]{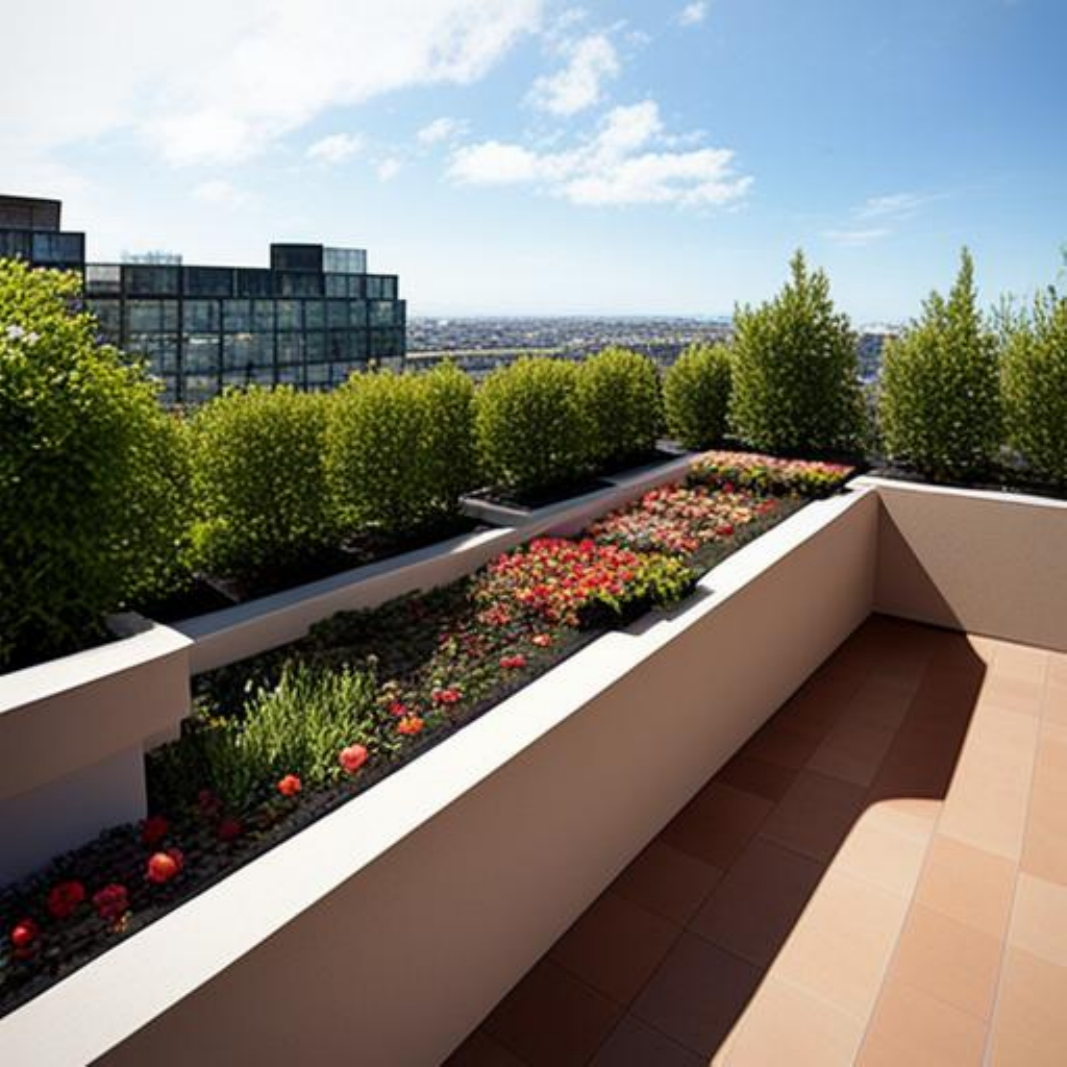} &
        \includegraphics[width=0.3\textwidth]{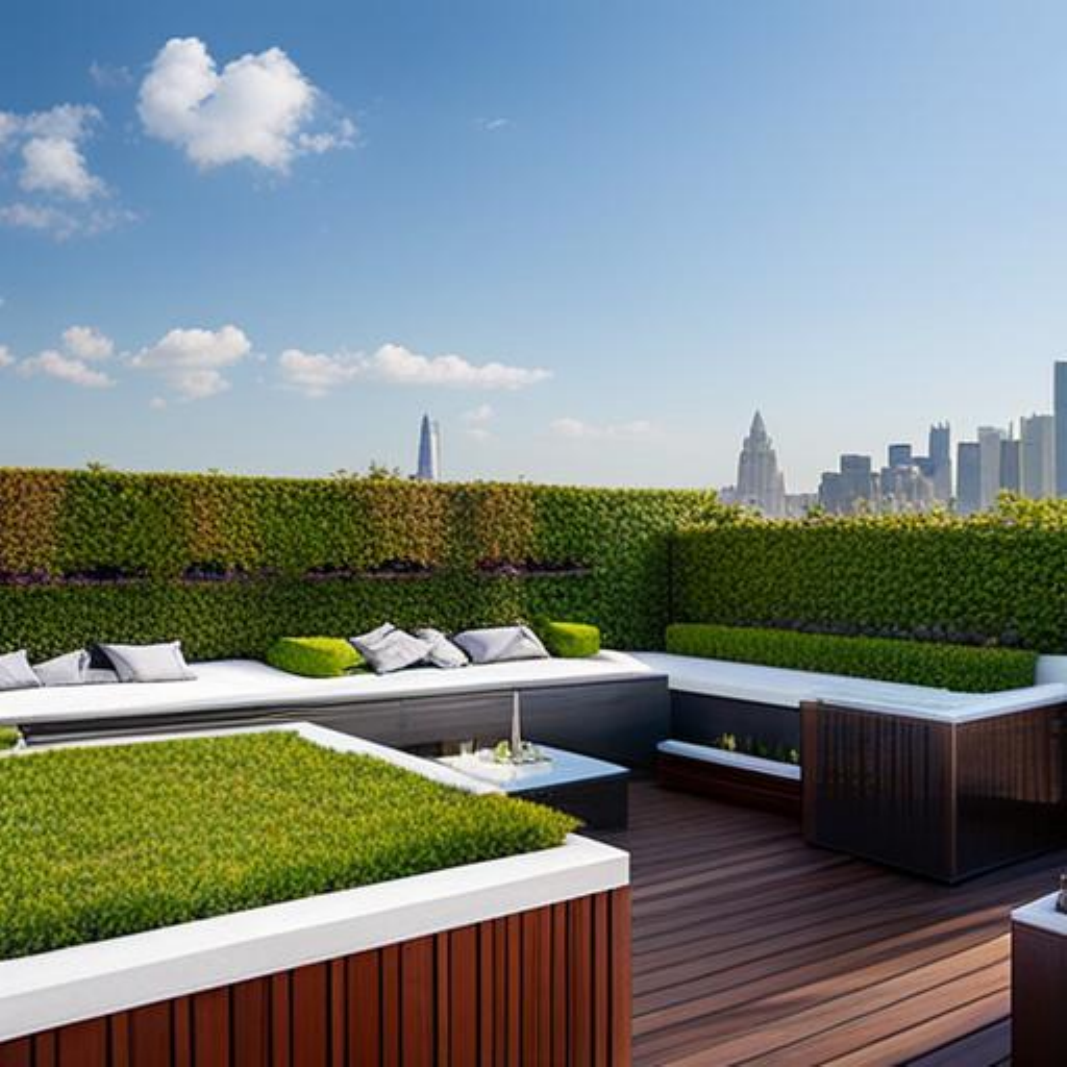} &
        \includegraphics[width=0.3\textwidth]{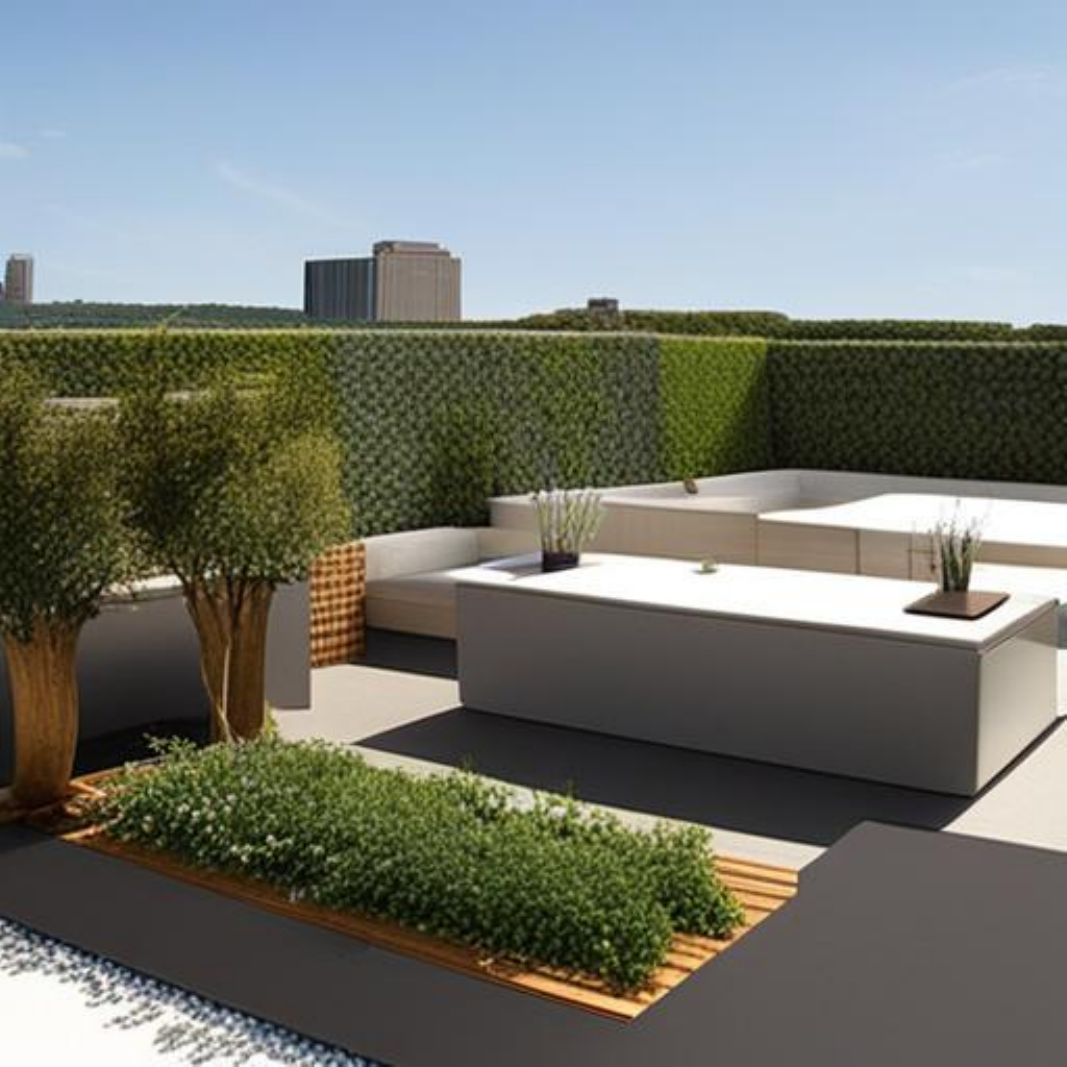} \\
        
    \end{tabular}
    \caption{No Layout - Roof Top}
    \label{fig:no_layout_roof_top}
\end{figure*}

\begin{figure*}[ht]
    \centering
    \begin{tabular}{c c c}
        \includegraphics[width=0.3\textwidth]{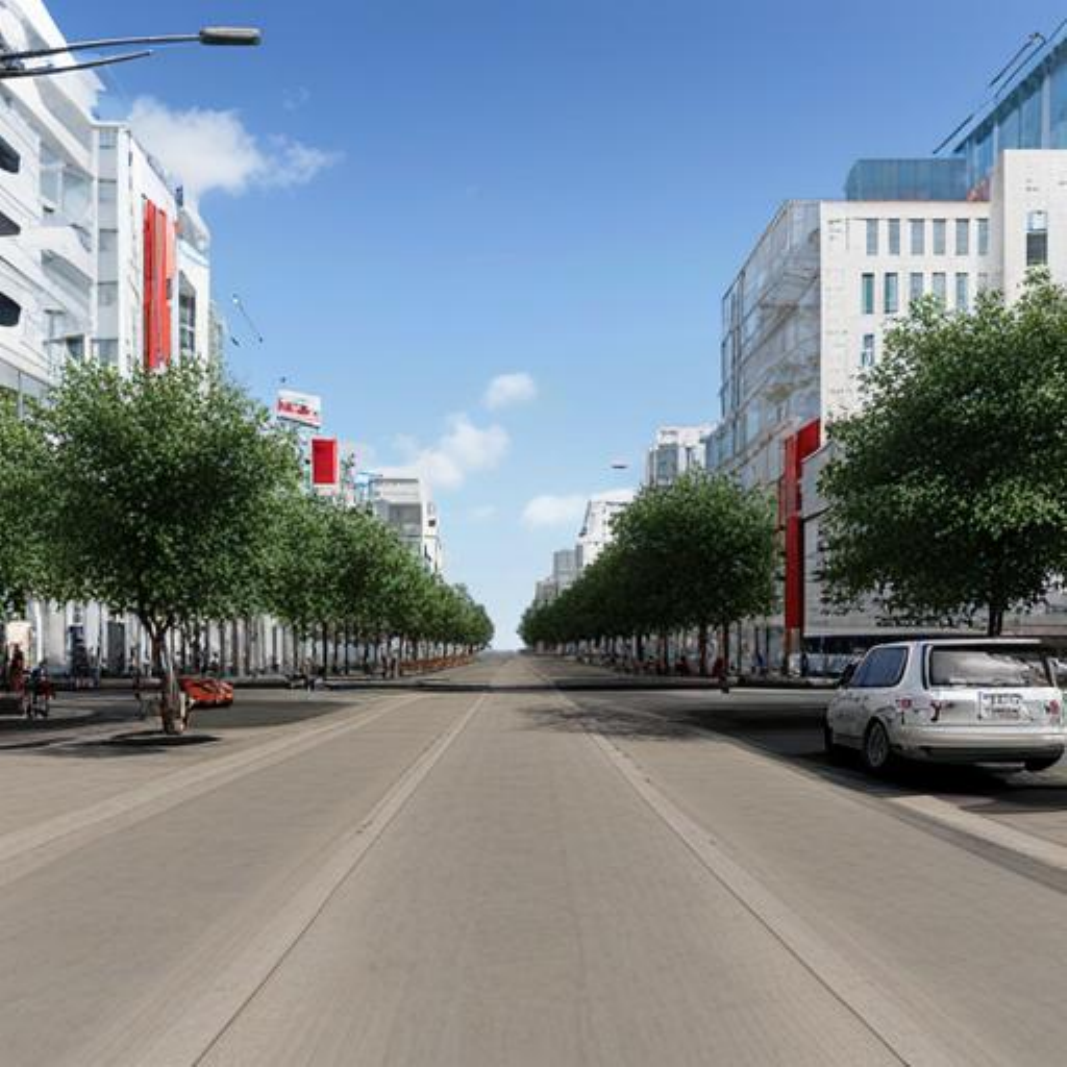} &
        \includegraphics[width=0.3\textwidth]{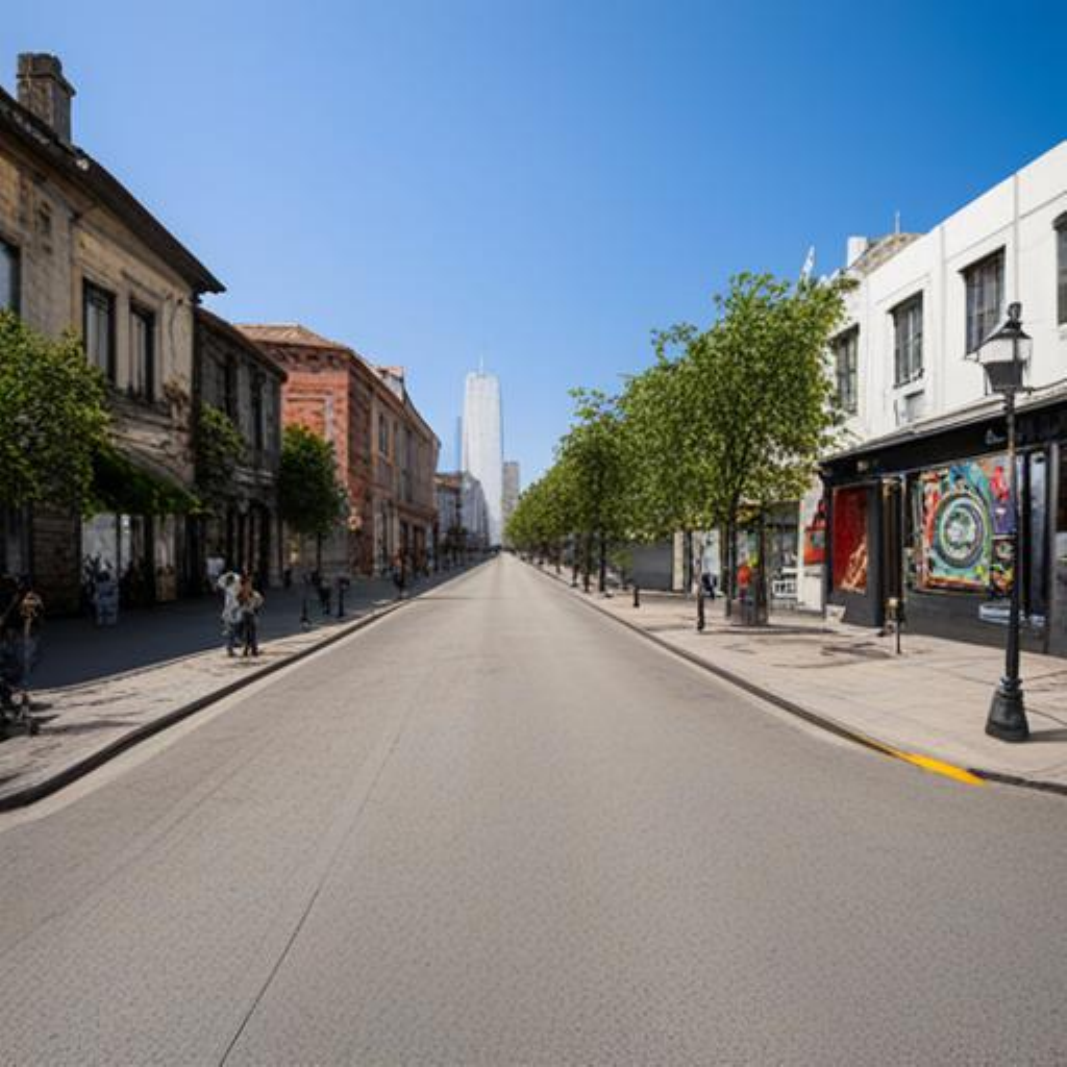} &
        \includegraphics[width=0.3\textwidth]{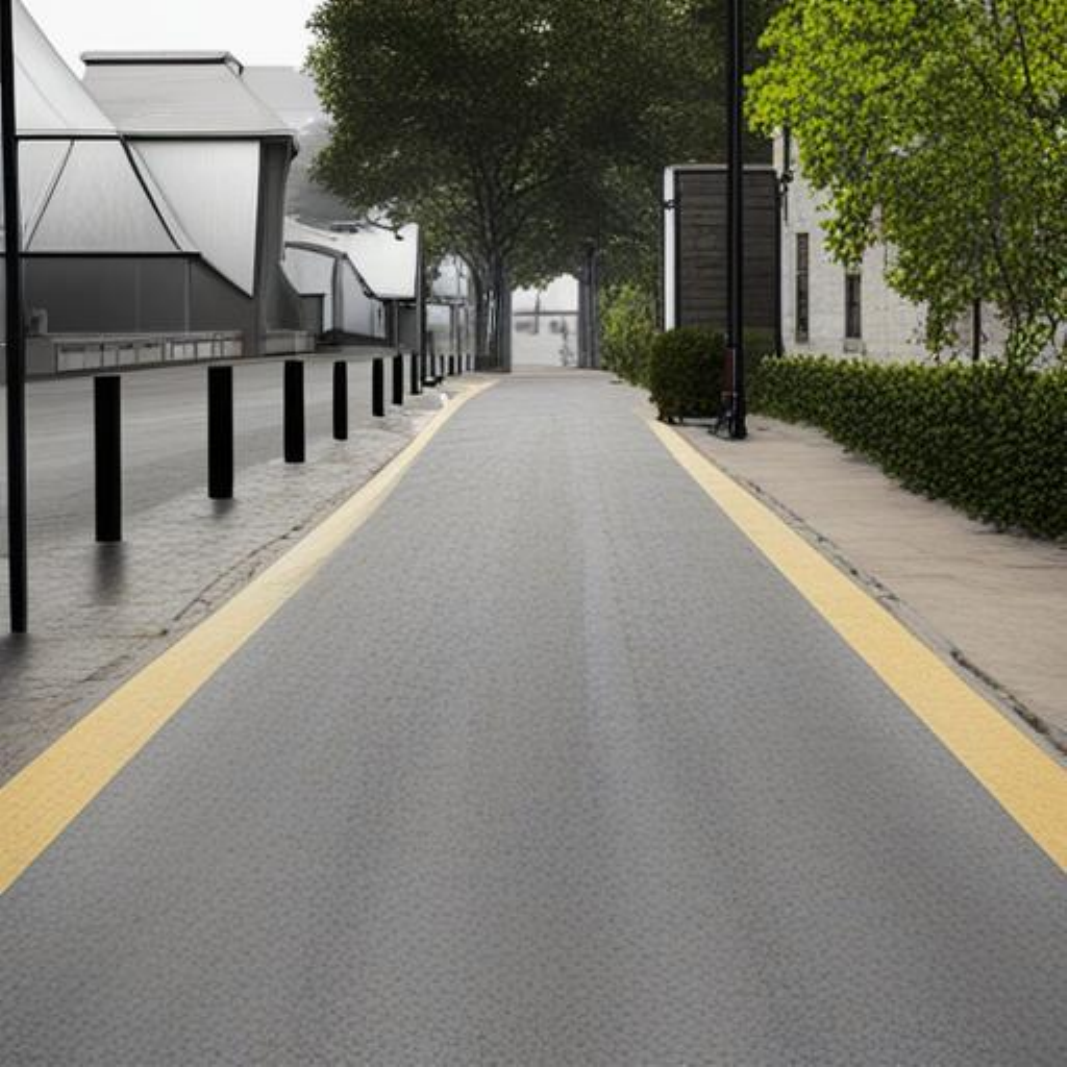} \\
        
    \end{tabular}
    \caption{No Layout - Street}
    \label{fig:no_layout_street}
\end{figure*}

\begin{figure*}[ht]
    \centering
    \begin{tabular}{c c c}
        \includegraphics[width=0.3\textwidth]{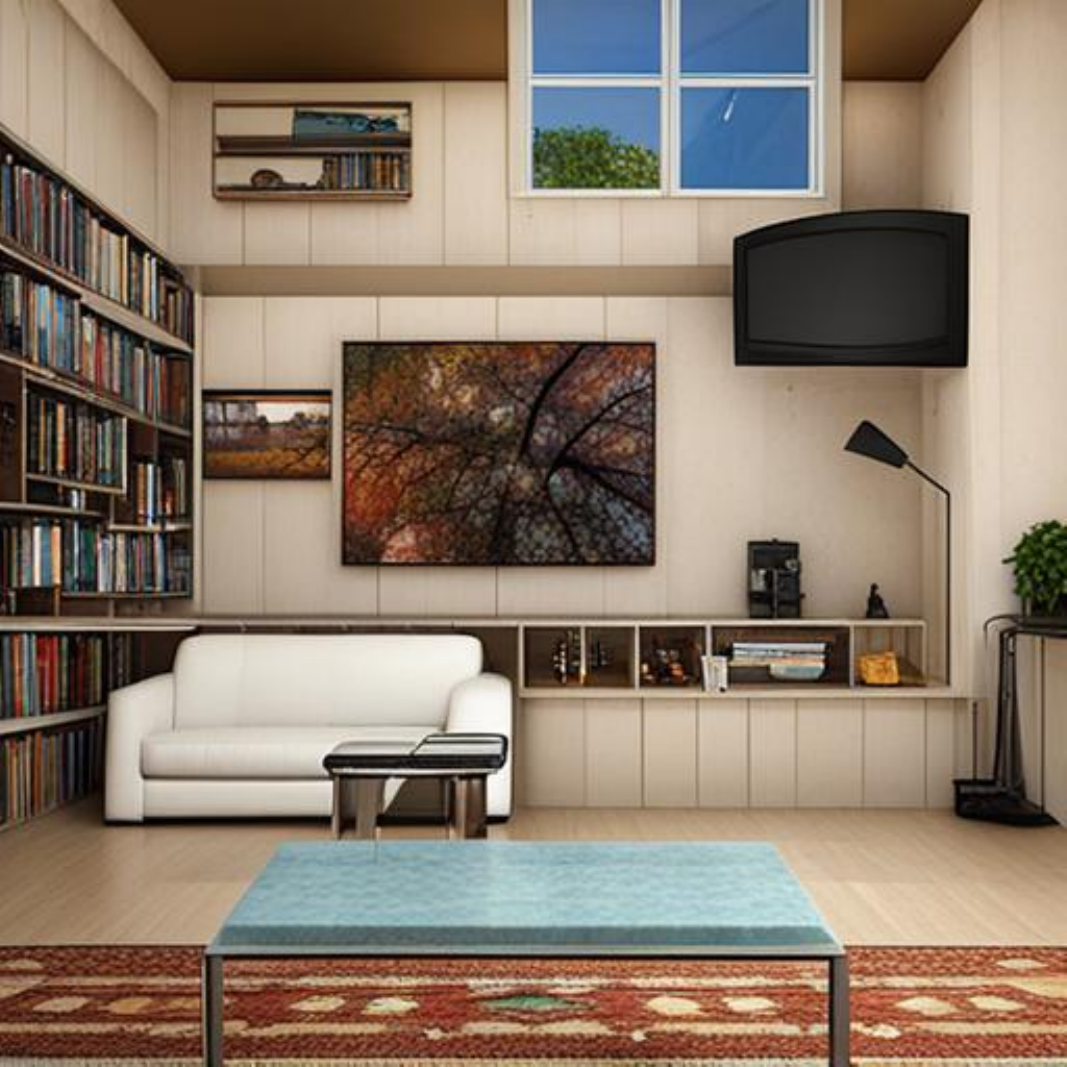} &
        \includegraphics[width=0.3\textwidth]{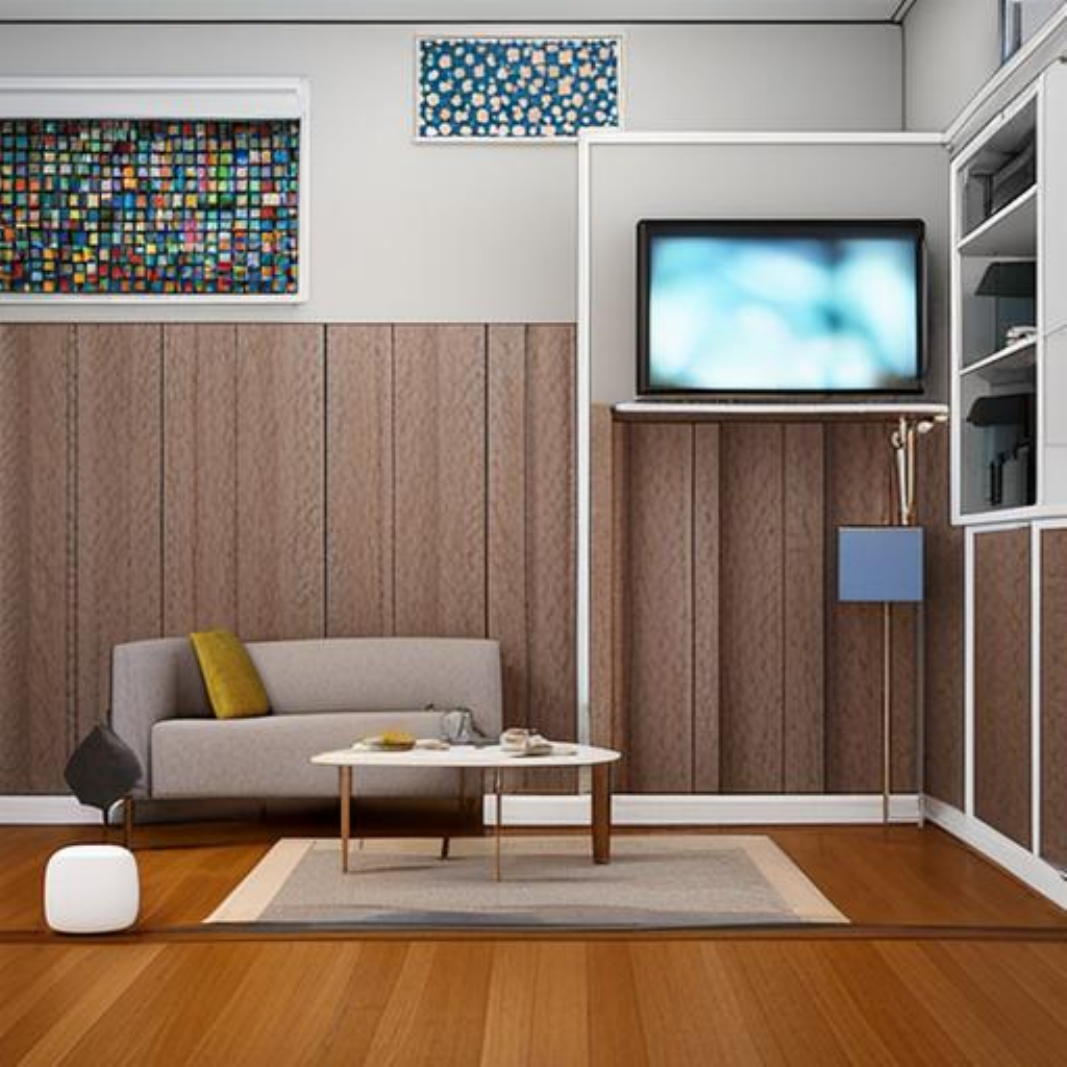} &
        \includegraphics[width=0.3\textwidth]{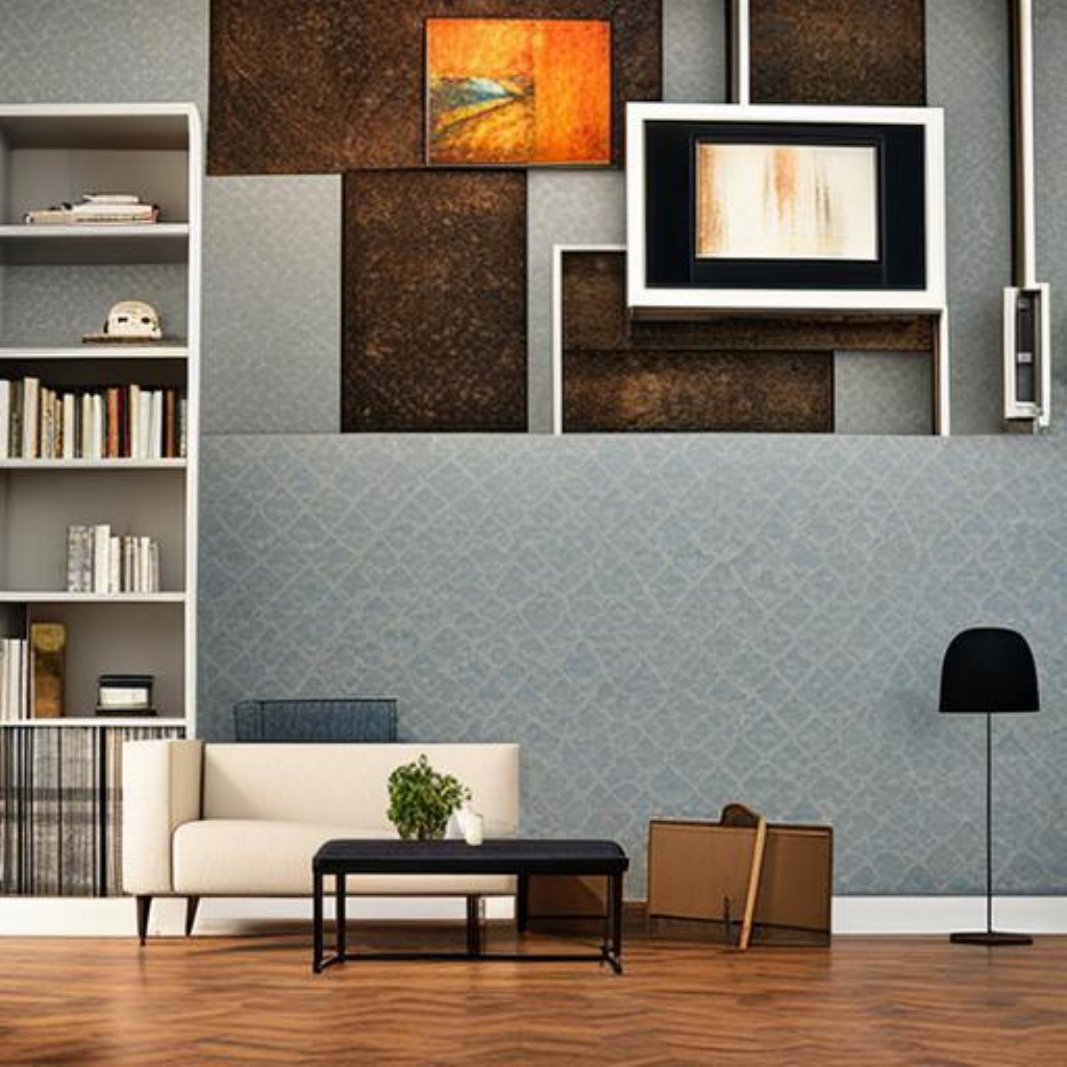} \\
        
        \includegraphics[width=0.3\textwidth]{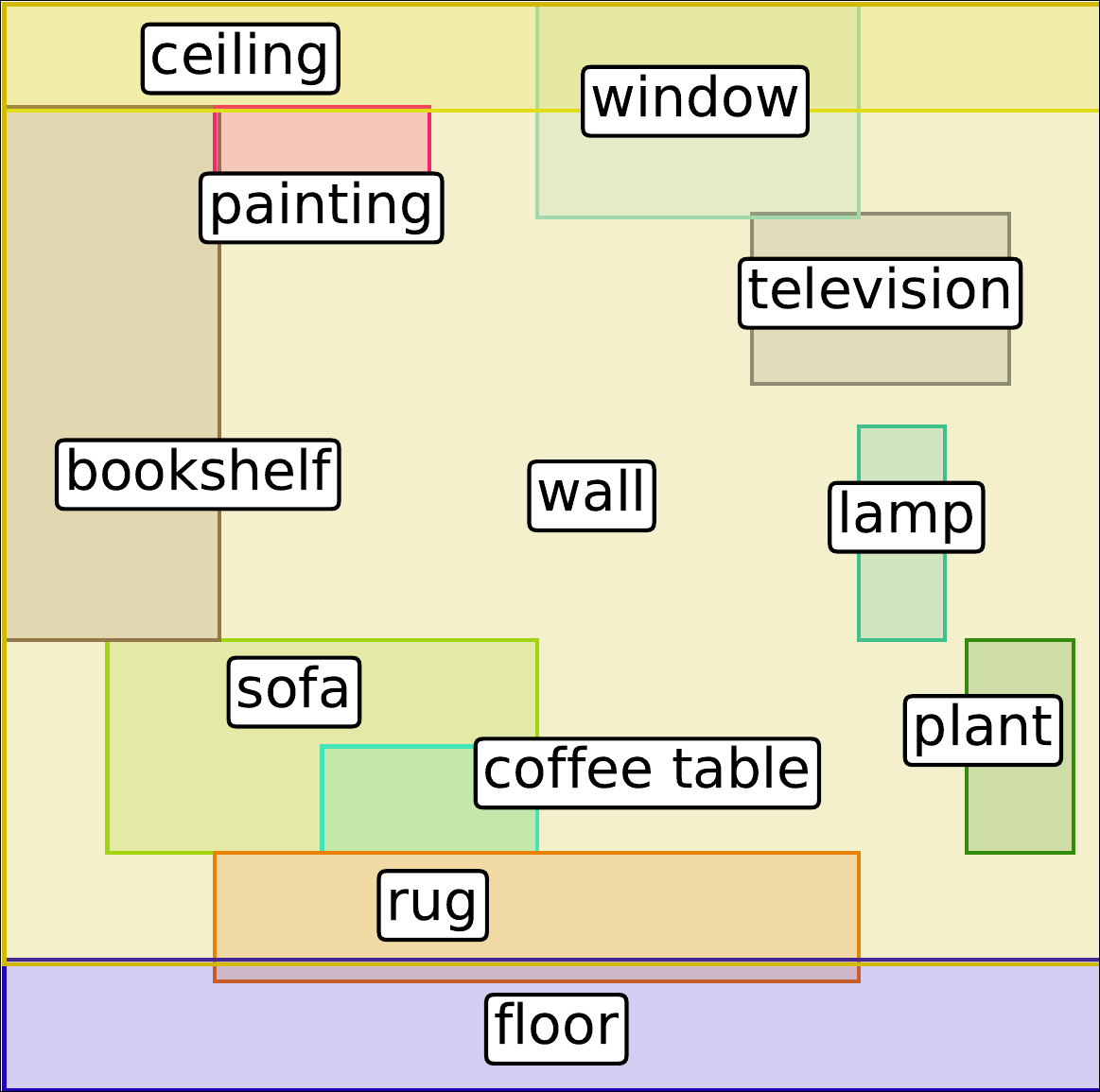} &
        \includegraphics[width=0.3\textwidth]{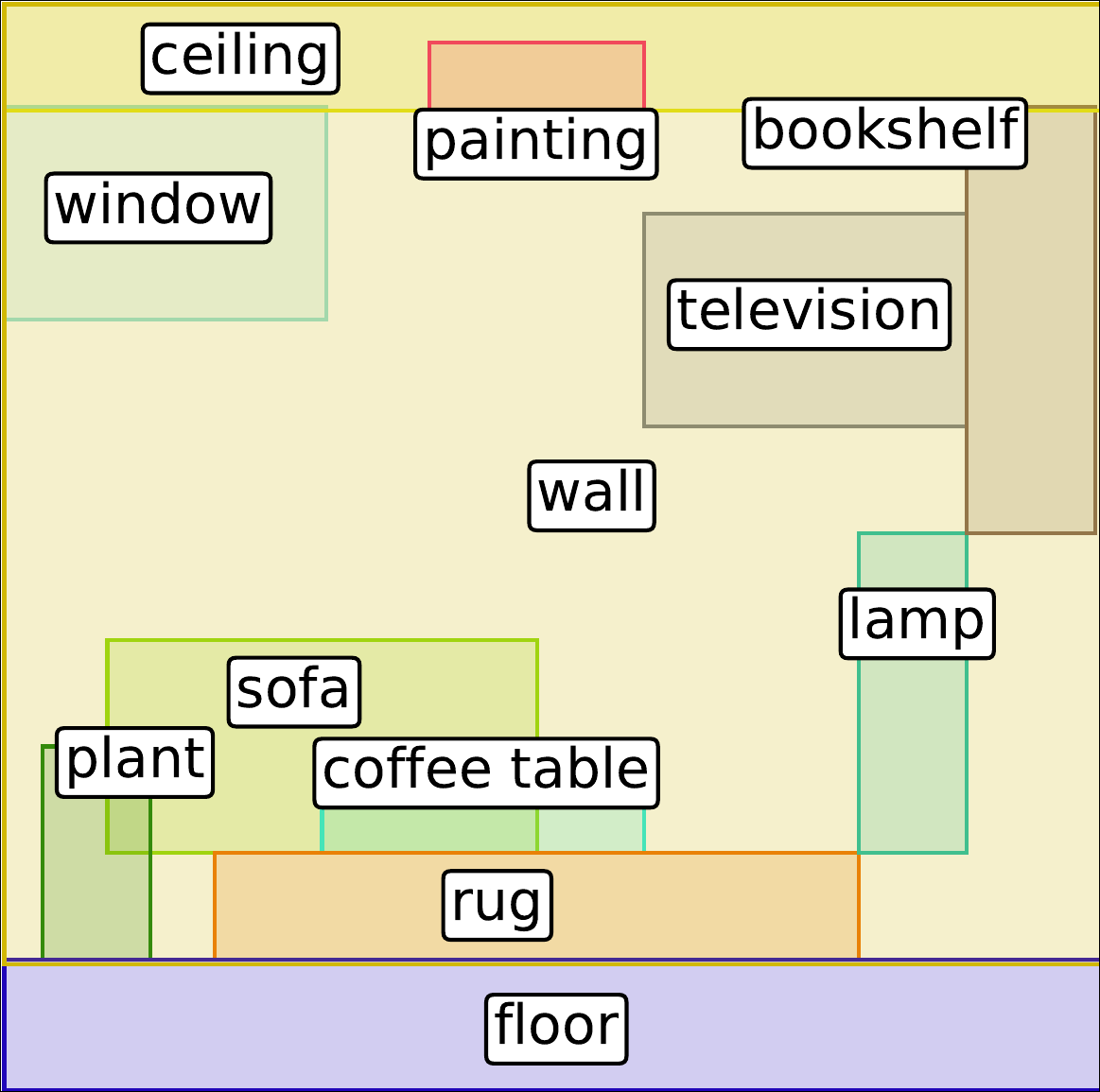} &
        \includegraphics[width=0.3\textwidth]{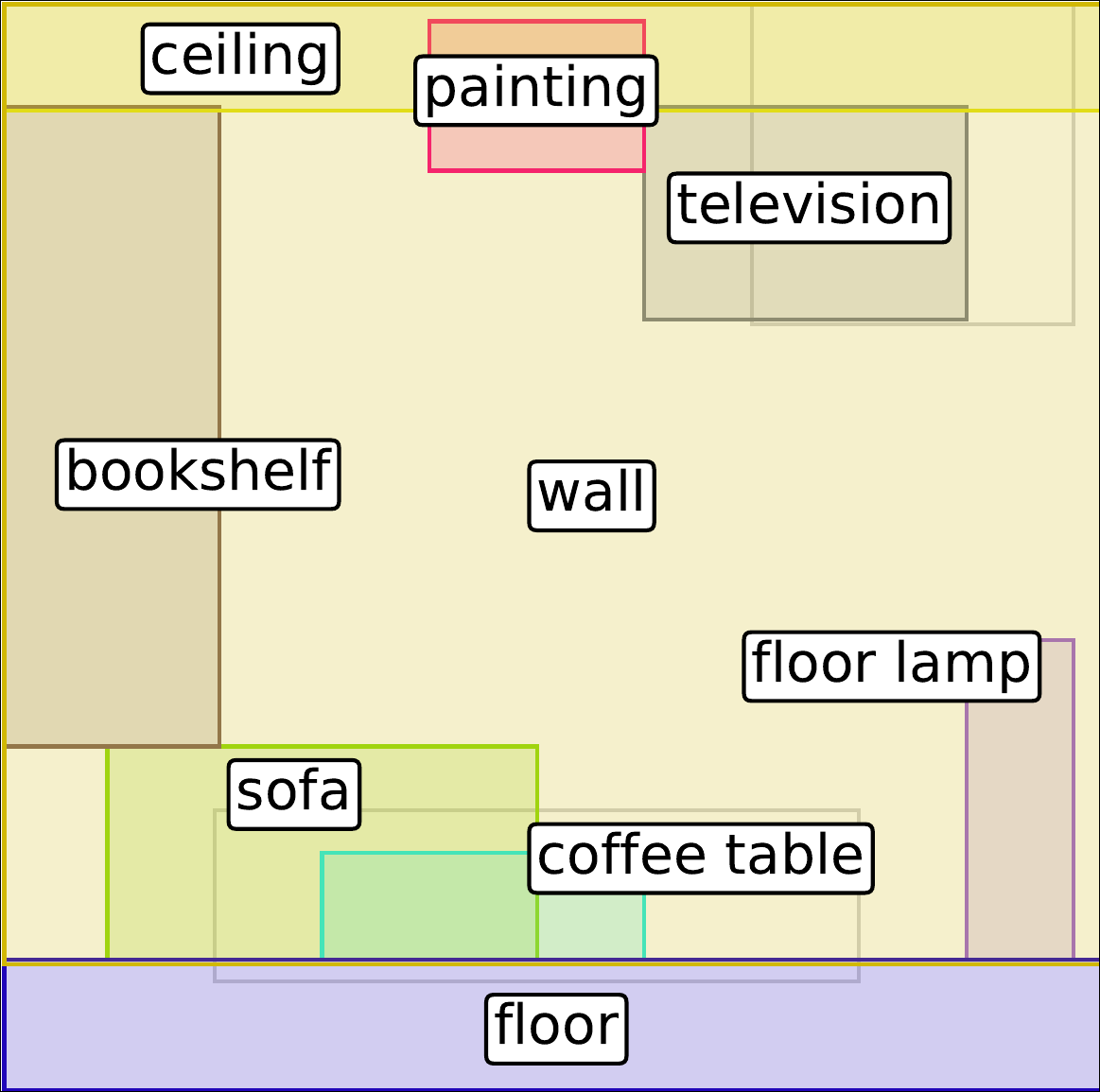} \\
    \end{tabular}
    \caption{GPT4o - Living Room.}
    \label{fig:gpt4o_living_room}
\end{figure*}
    
\begin{figure*}[ht]
    \centering
    \begin{tabular}{c c c} 
        \includegraphics[width=0.3\textwidth]{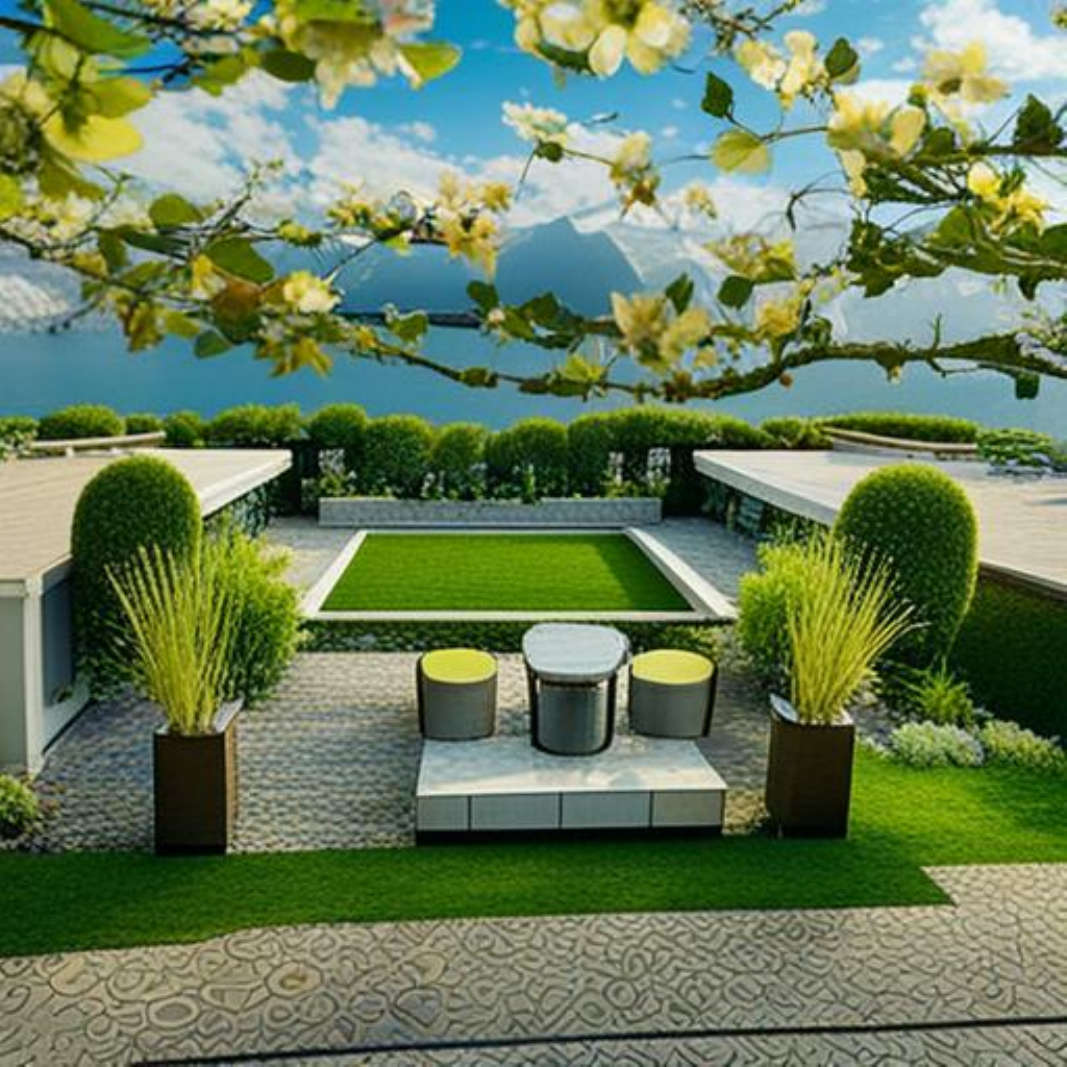} &
        \includegraphics[width=0.3\textwidth]{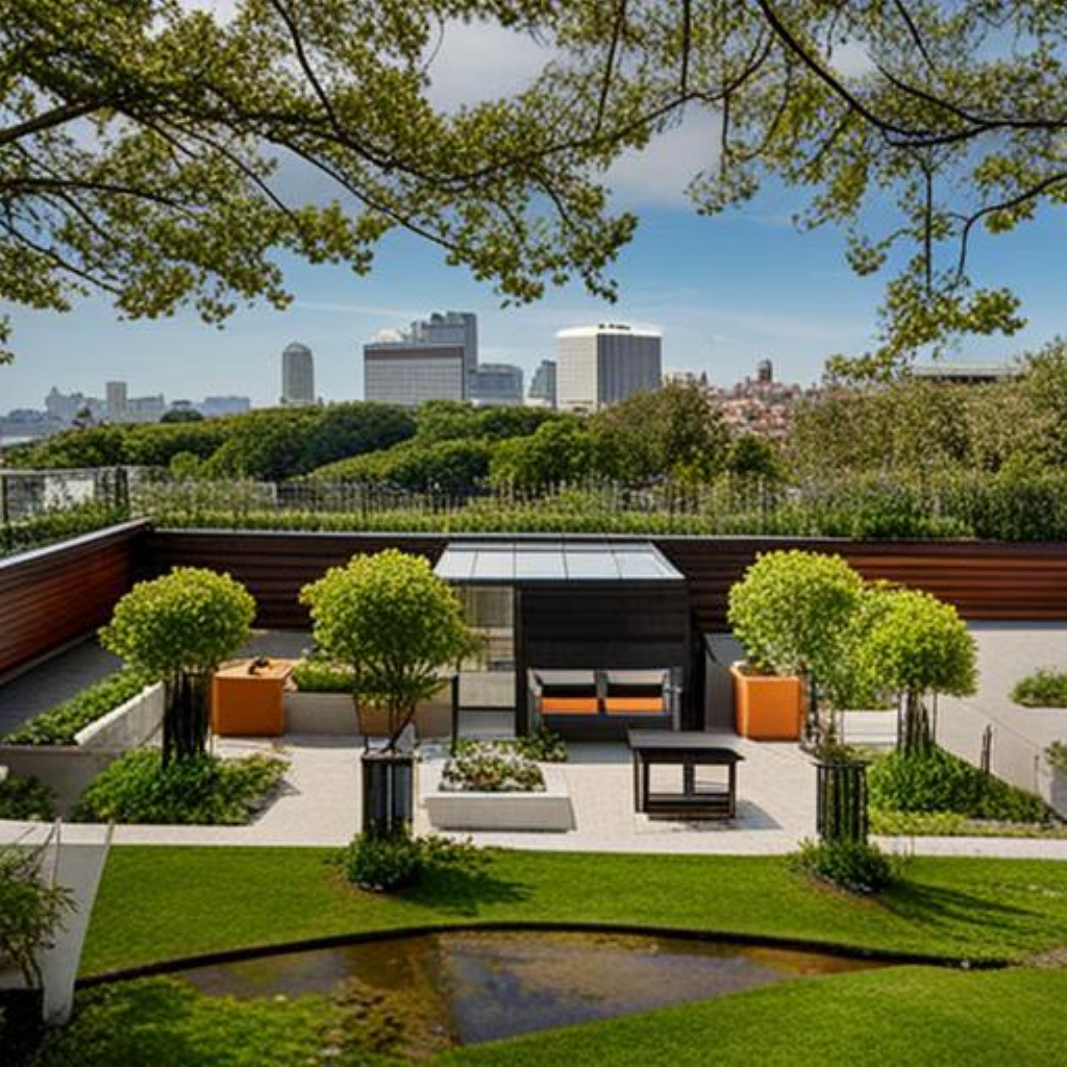} &
        \includegraphics[width=0.3\textwidth]{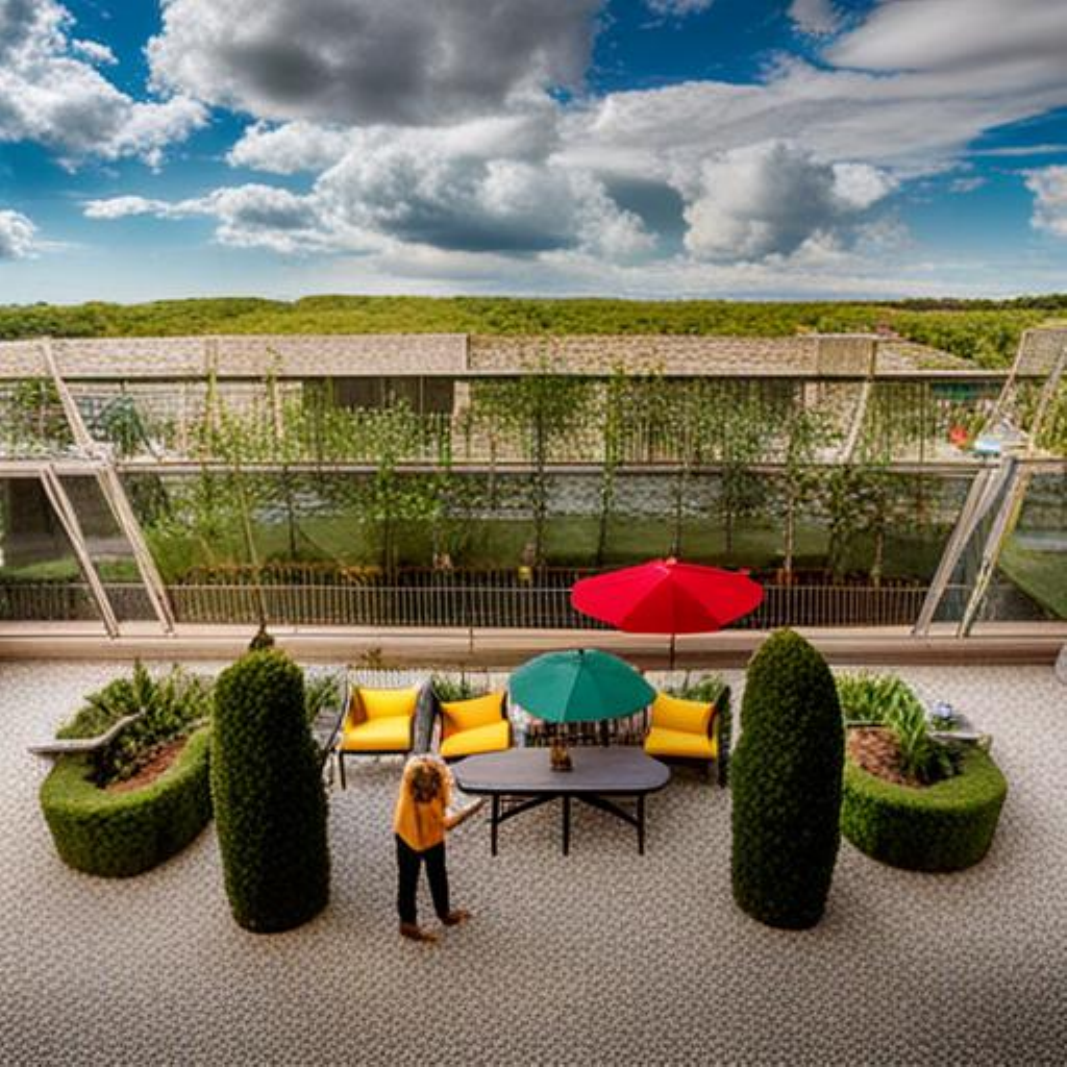} \\
        
        \includegraphics[width=0.3\textwidth]{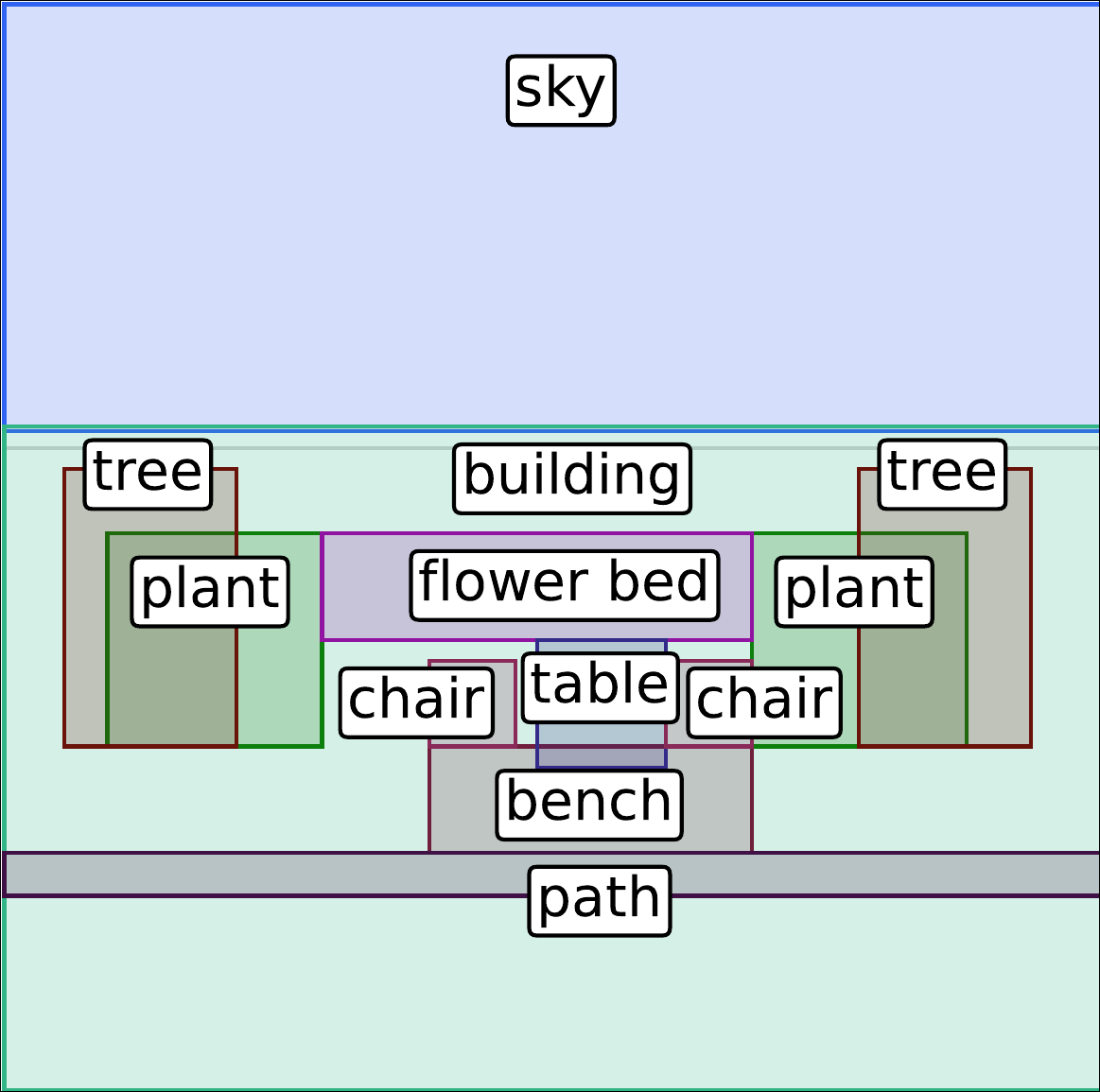} &
        \includegraphics[width=0.3\textwidth]{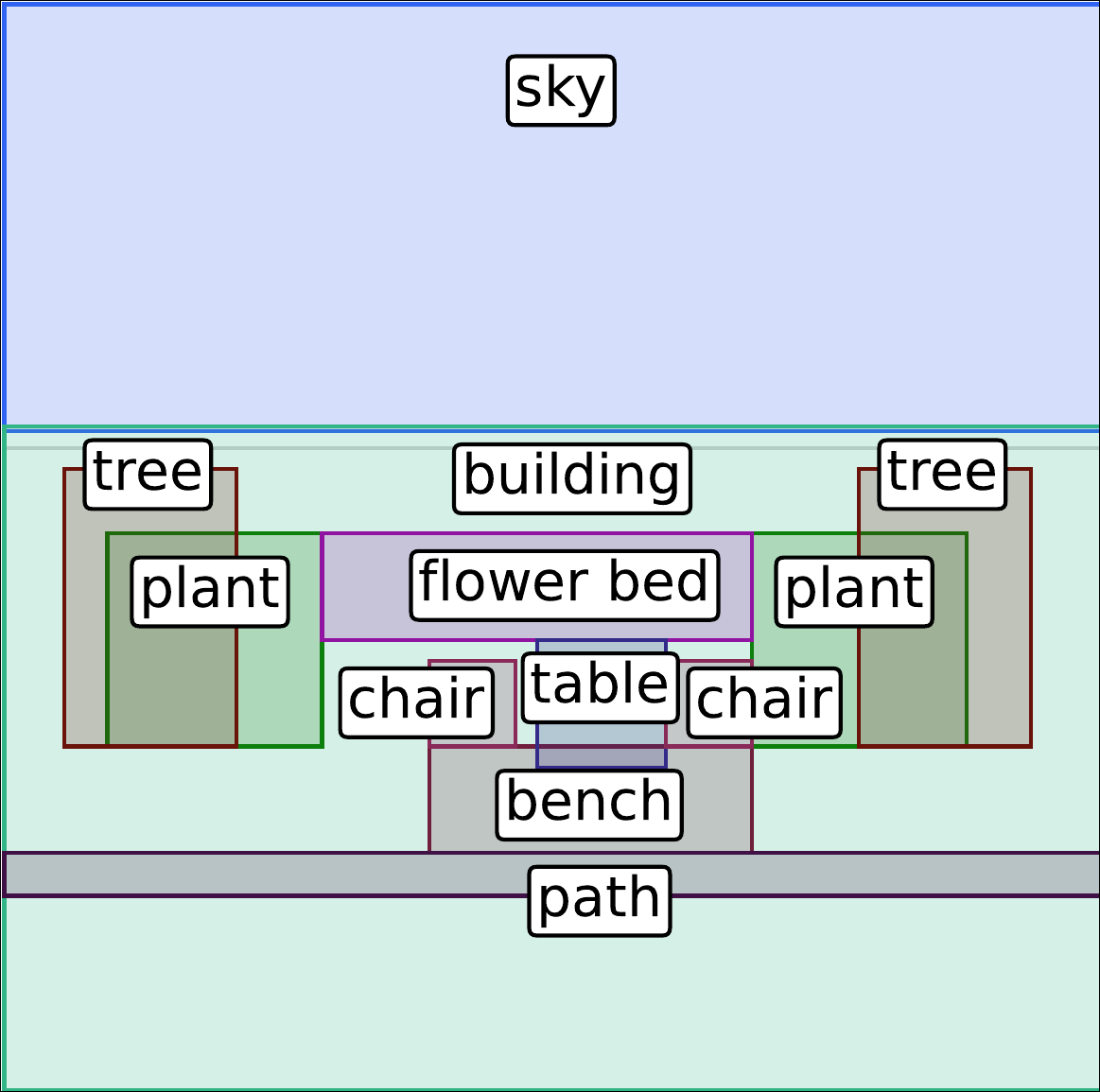} &
        \includegraphics[width=0.3\textwidth]{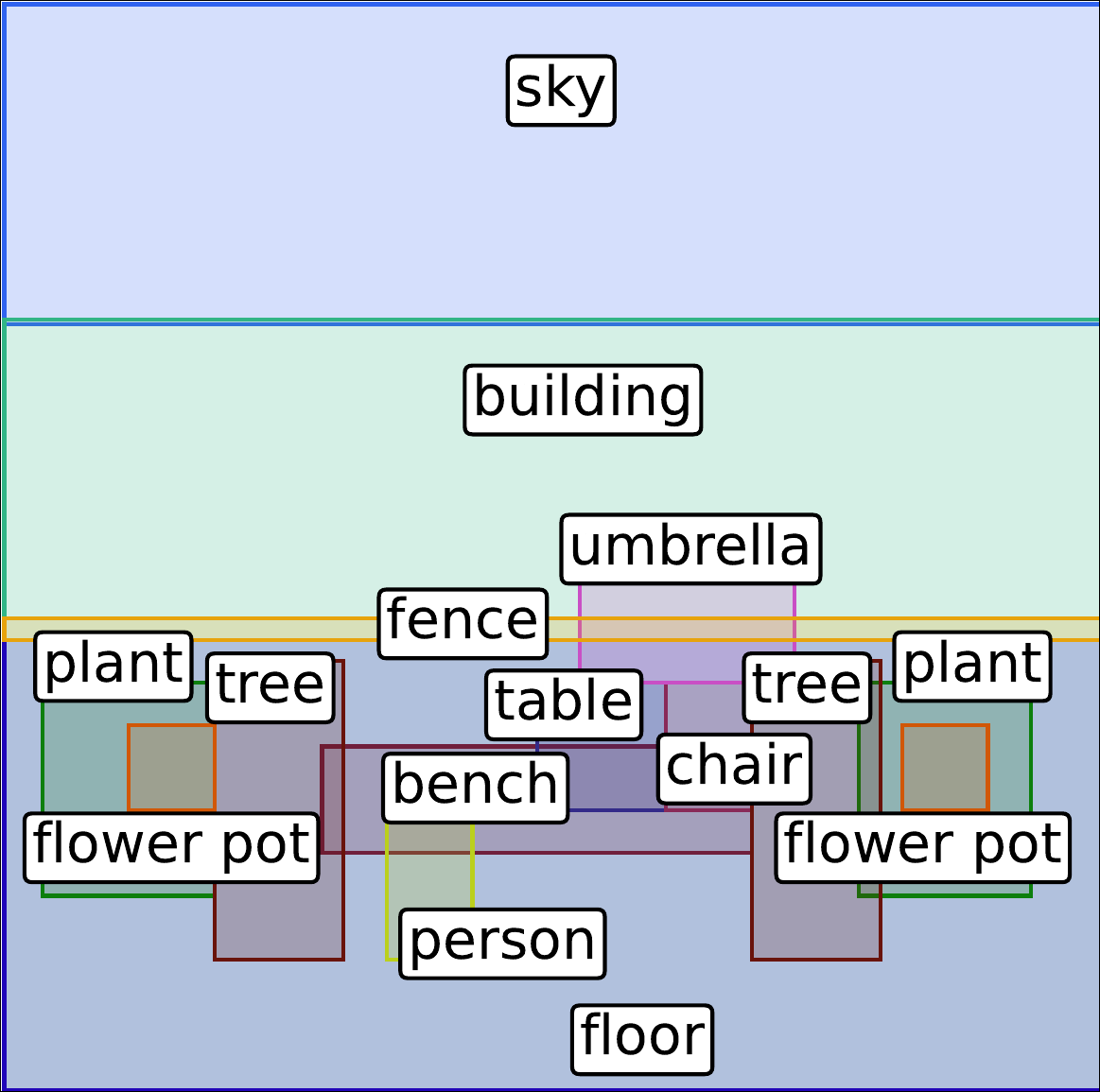} \\
    \end{tabular}
    \caption{GPT4o - Roof Top.}
    \label{fig:gpt4o_rooftop}
\end{figure*}
    
\begin{figure*}[ht]
    \centering
    \begin{tabular}{c c c}
        \includegraphics[width=0.3\textwidth]{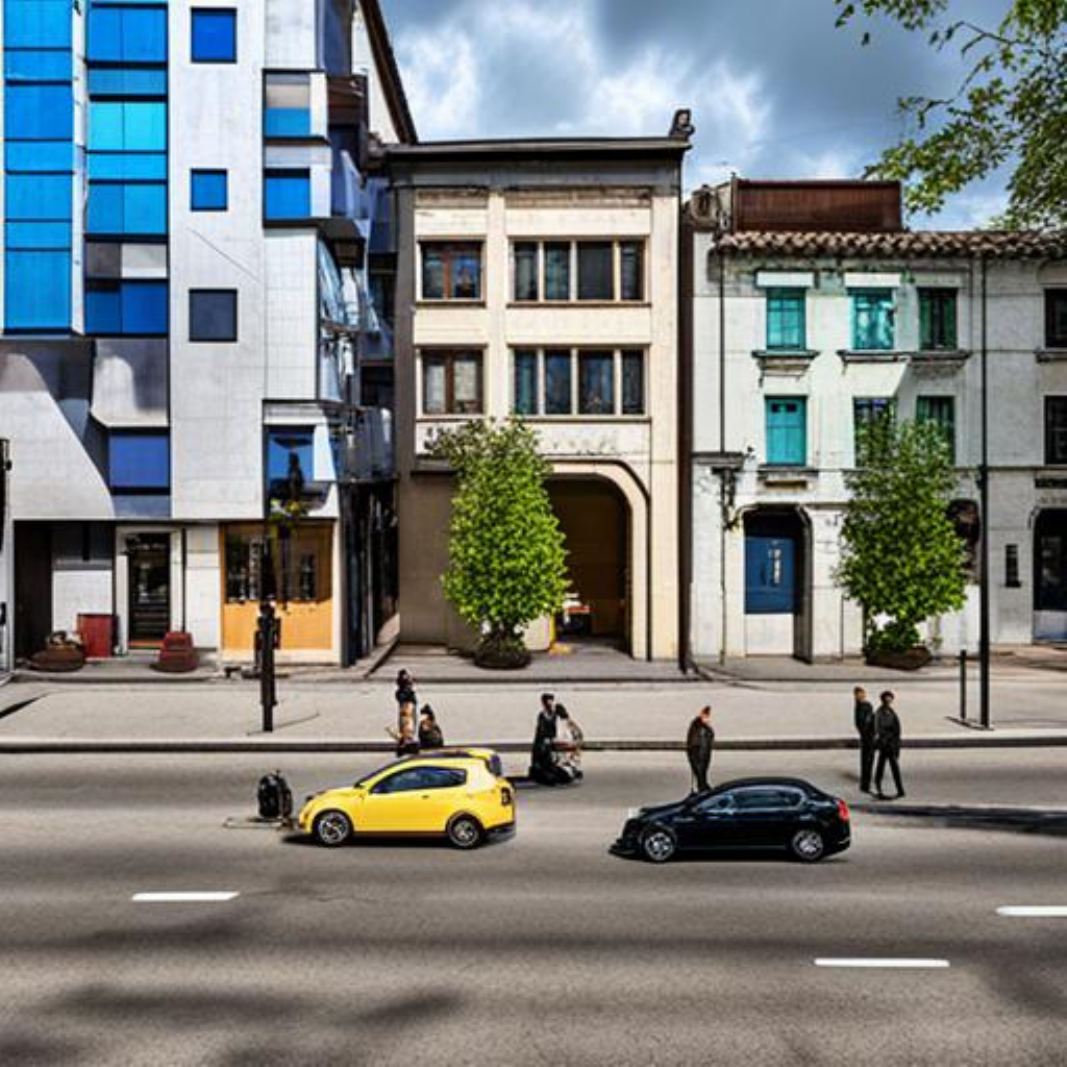} &
        \includegraphics[width=0.3\textwidth]{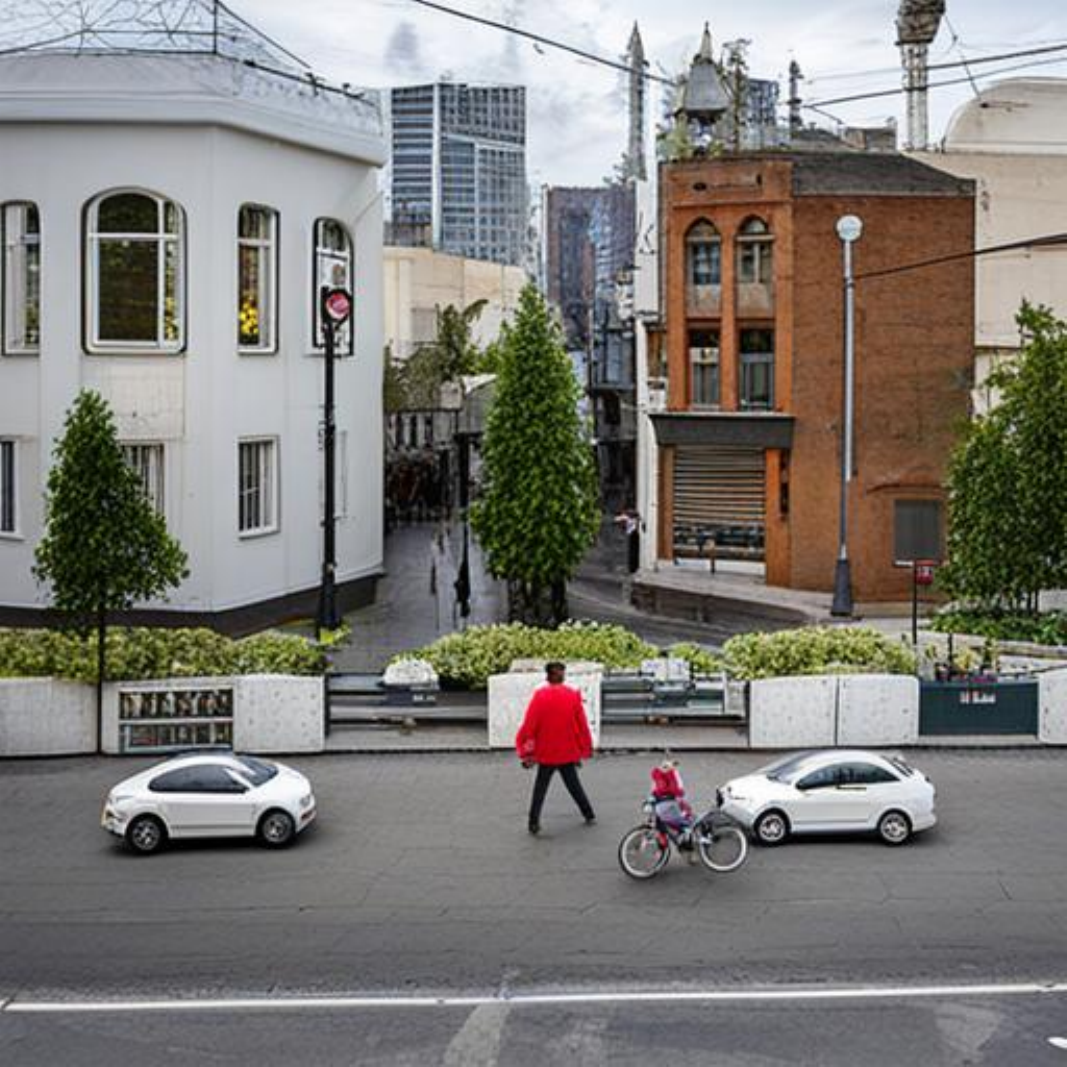} &
        \includegraphics[width=0.3\textwidth]{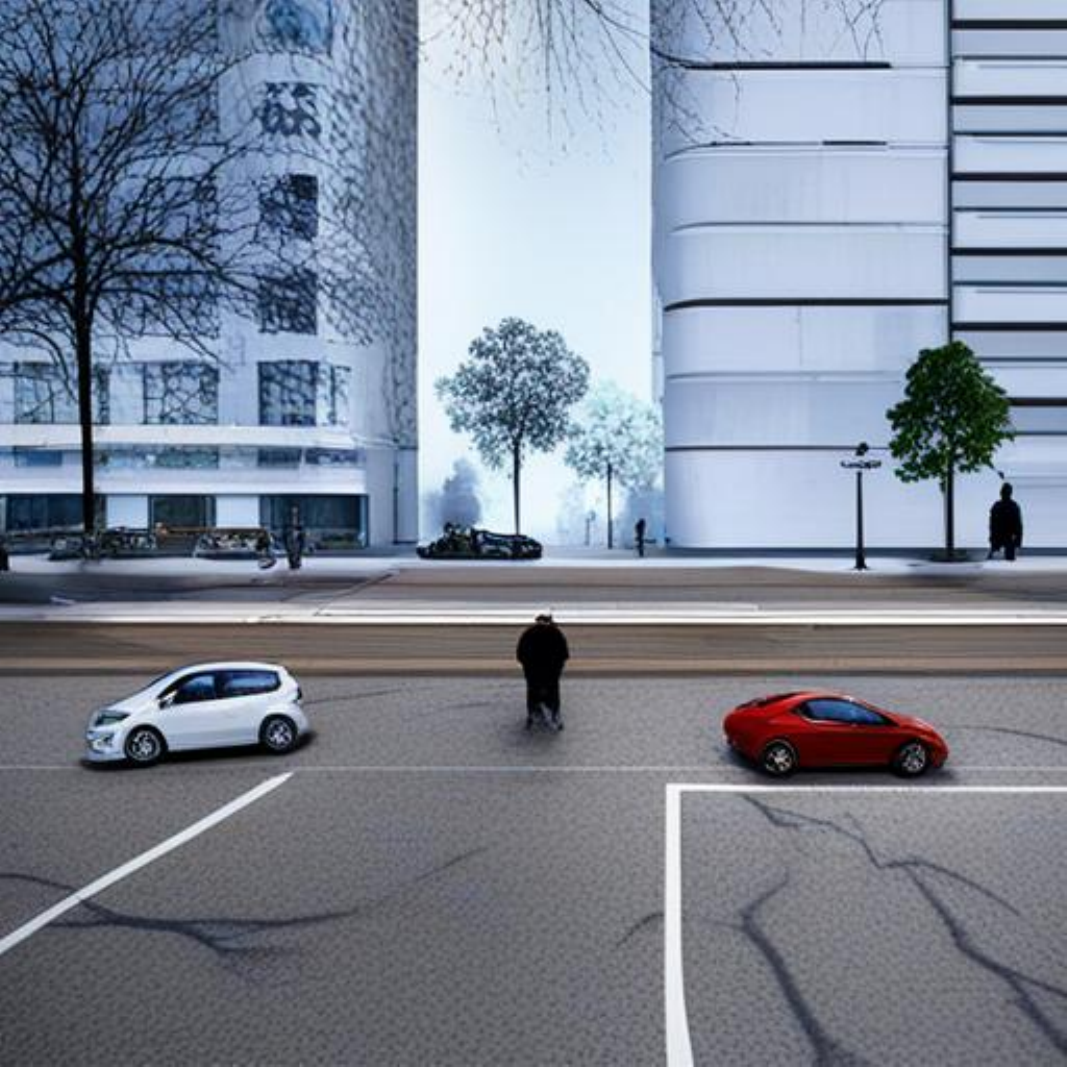} \\
        
        \includegraphics[width=0.3\textwidth]{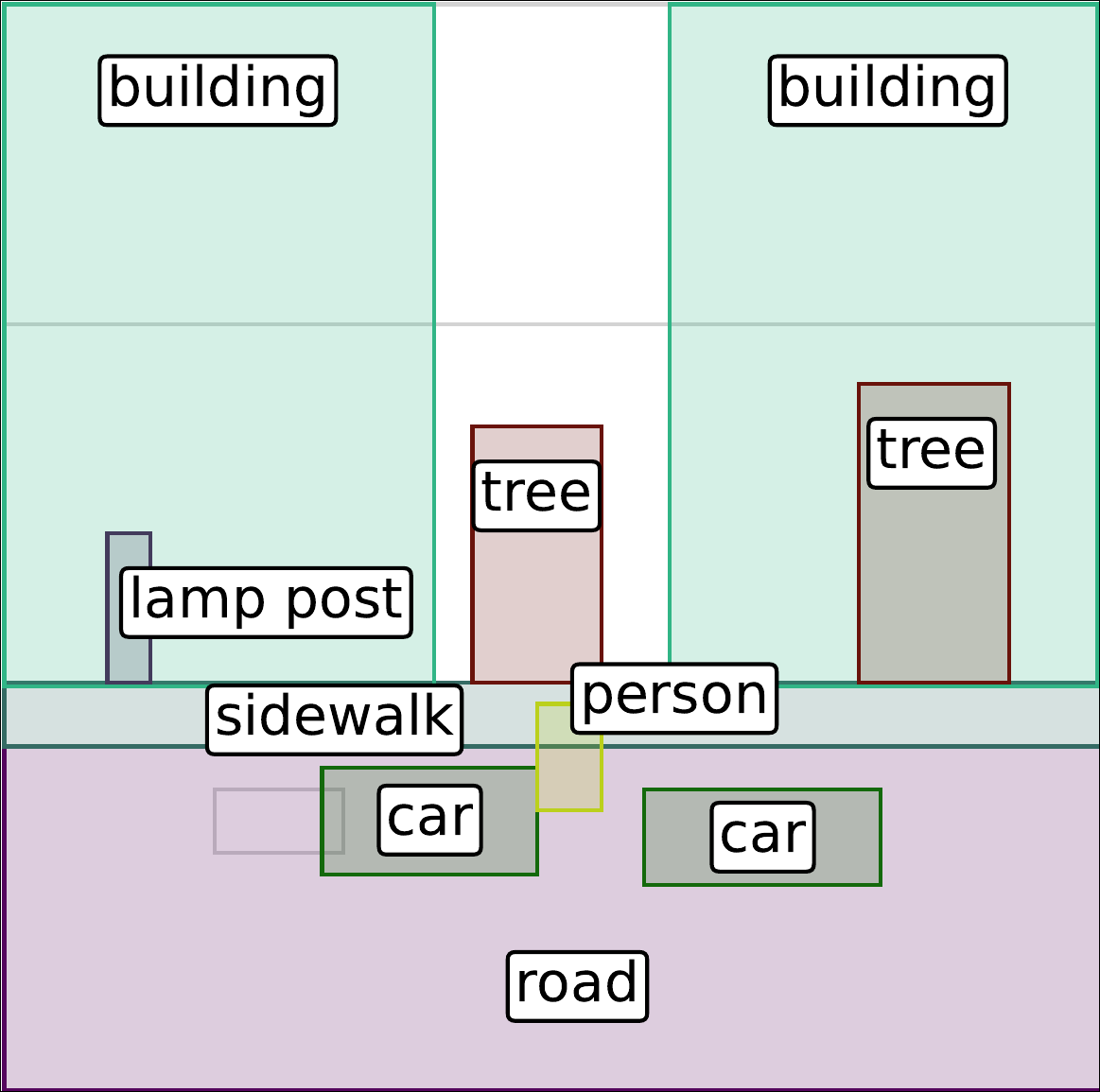} &
        \includegraphics[width=0.3\textwidth]{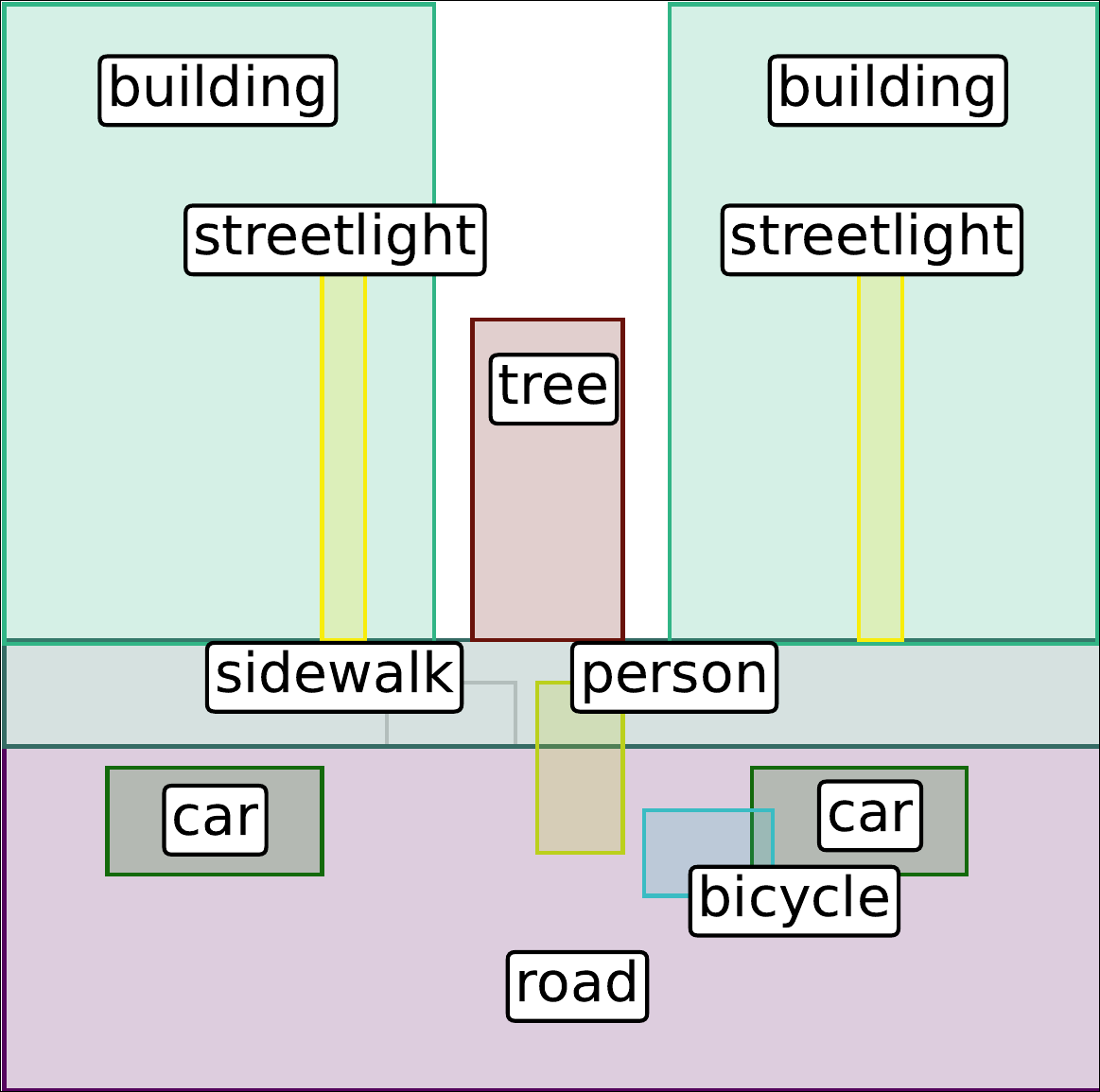} &
        \includegraphics[width=0.3\textwidth]{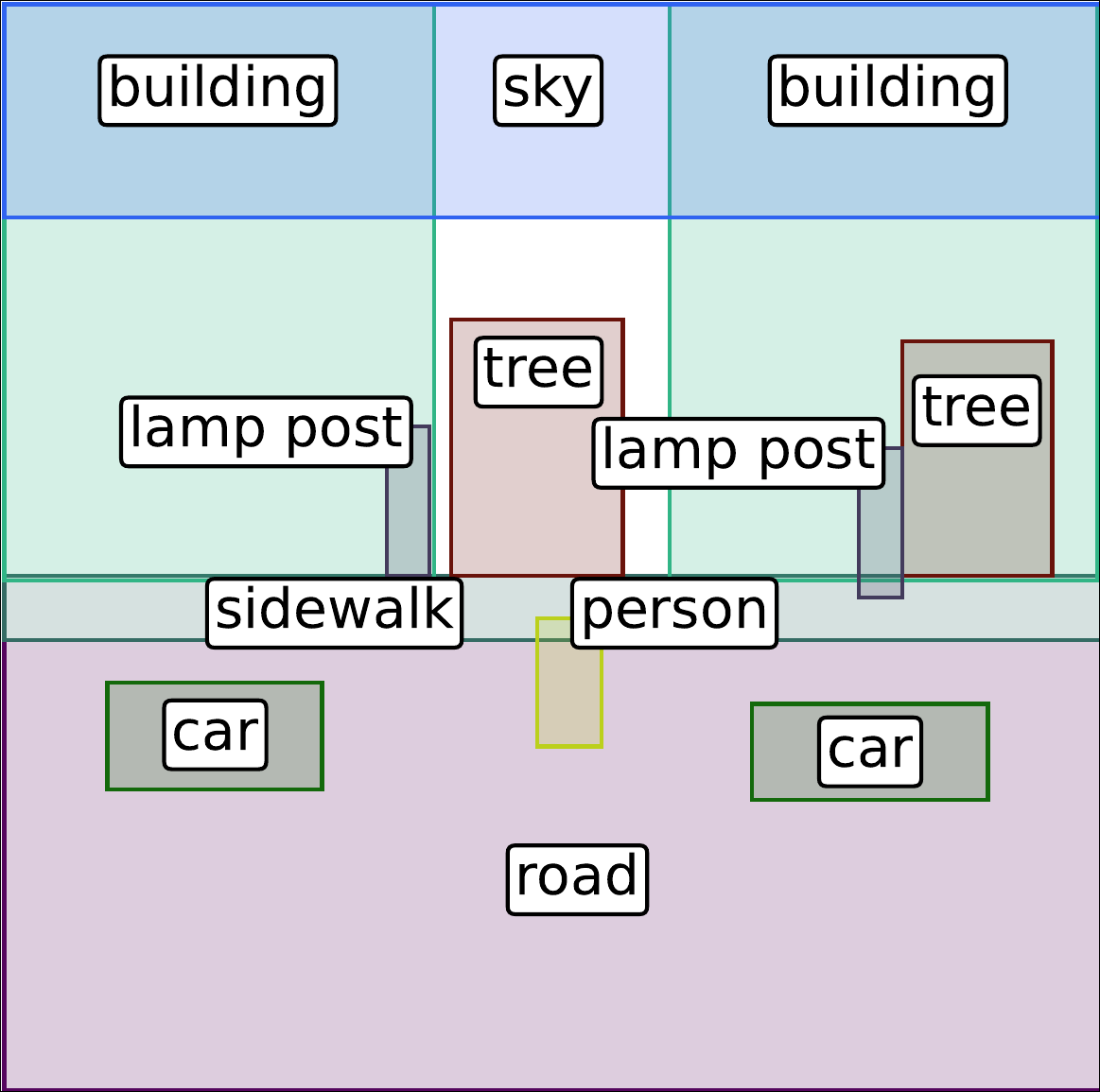} \\
    \end{tabular}
    \caption{GPT4o - Street.}
    \label{fig:gpt4o_street}
\end{figure*}

\begin{figure*}[ht]
    \centering
    \begin{tabular}{c c c}
        \includegraphics[width=0.3\textwidth]{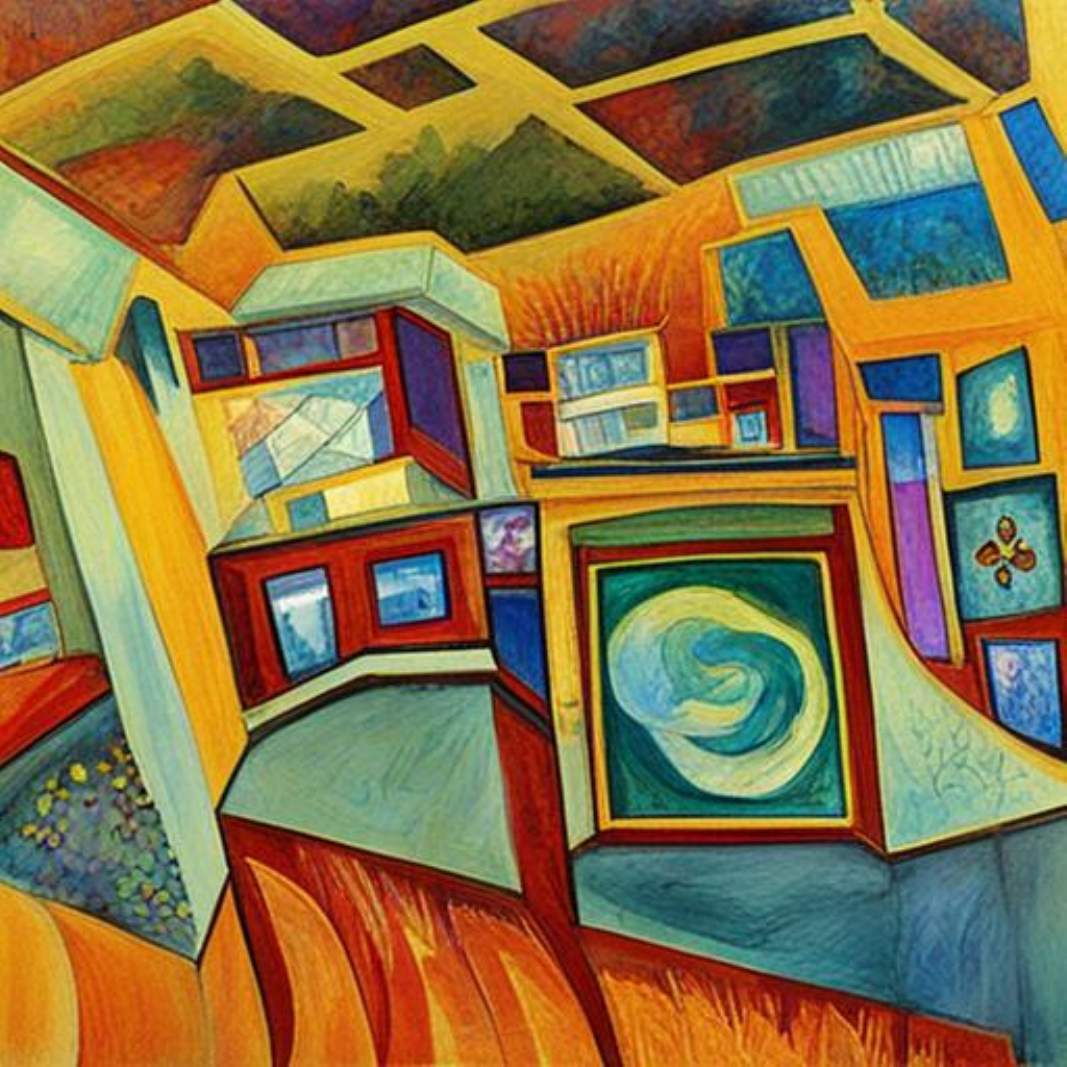} &
        \includegraphics[width=0.3\textwidth]{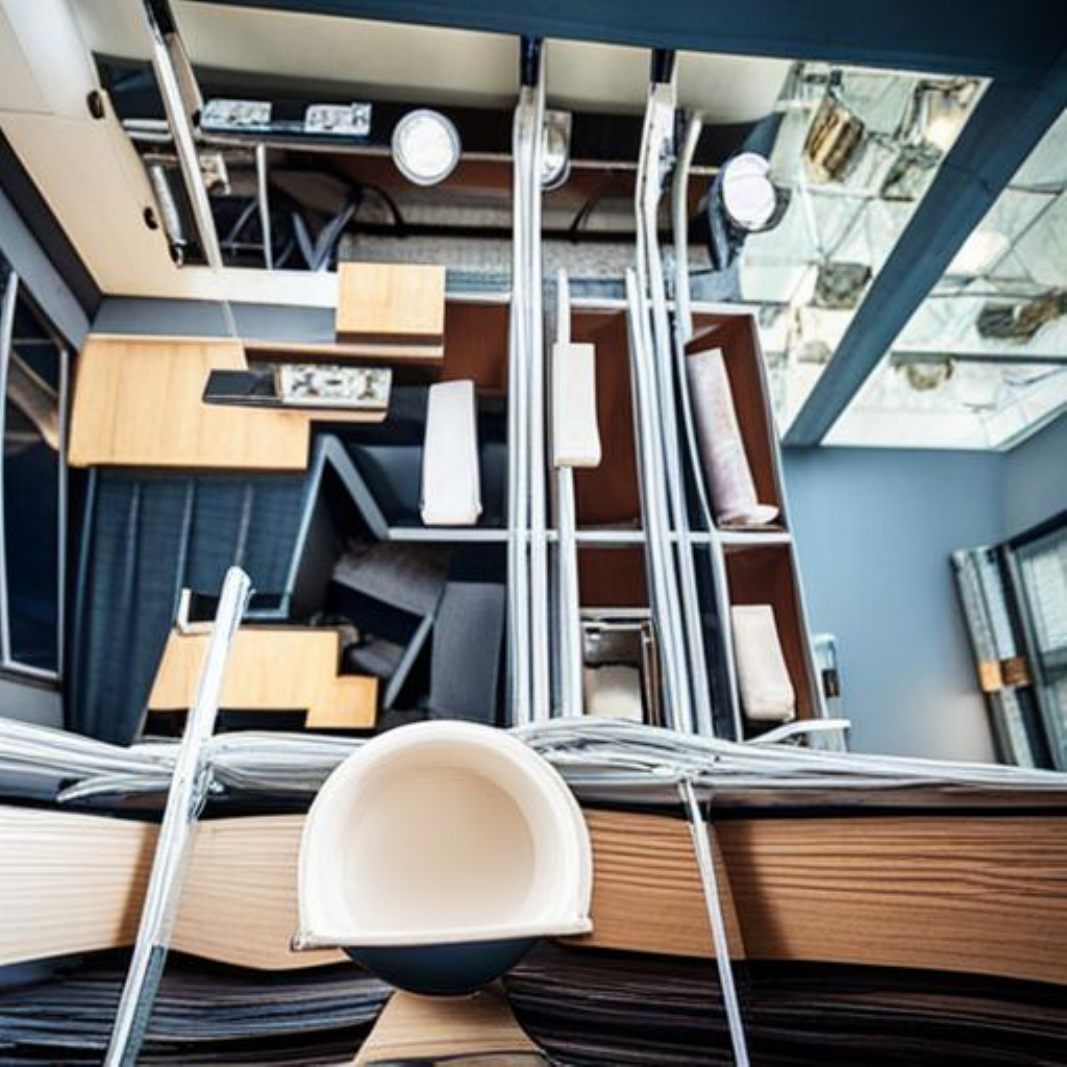} &
        \includegraphics[width=0.3\textwidth]{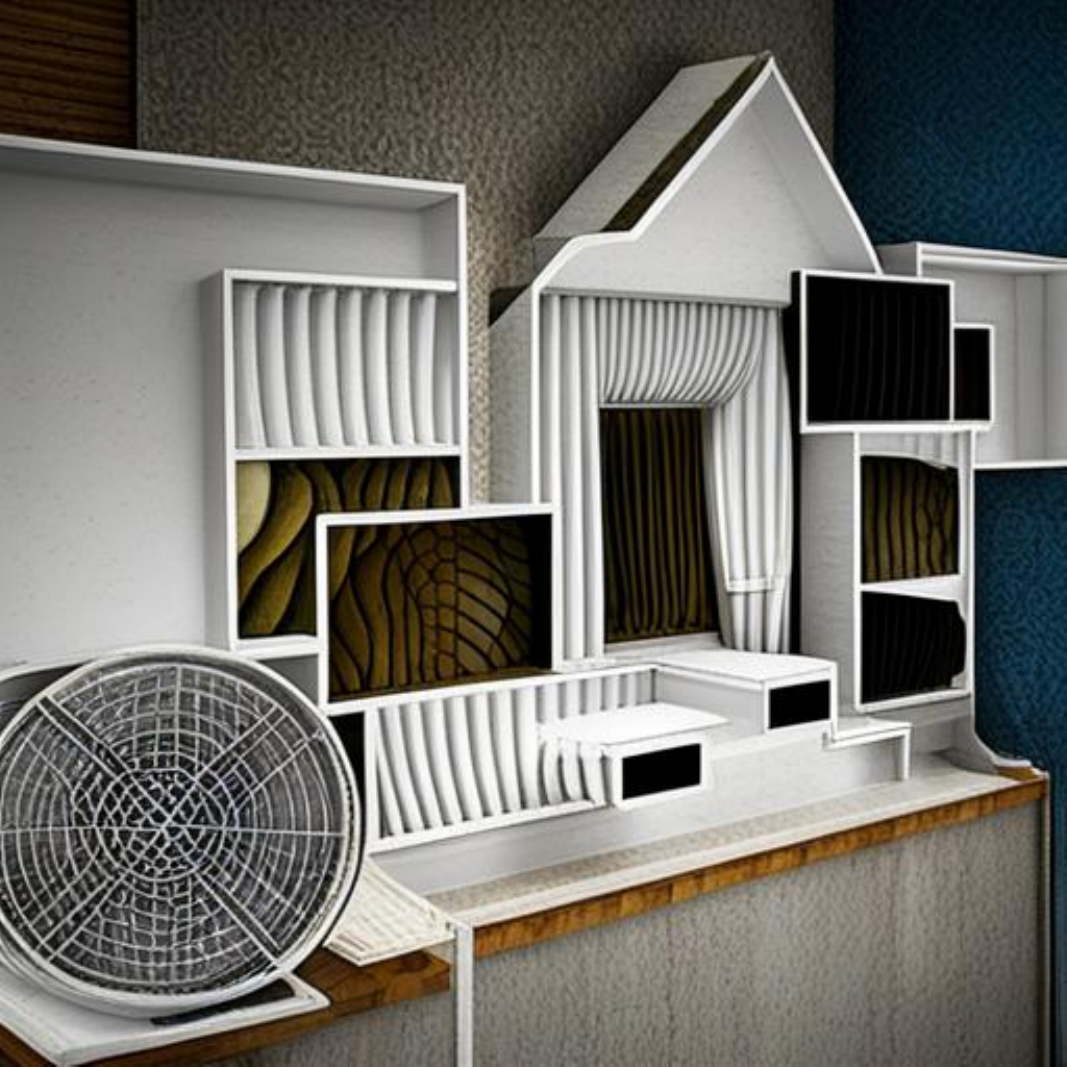} \\
        
        \includegraphics[width=0.3\textwidth]{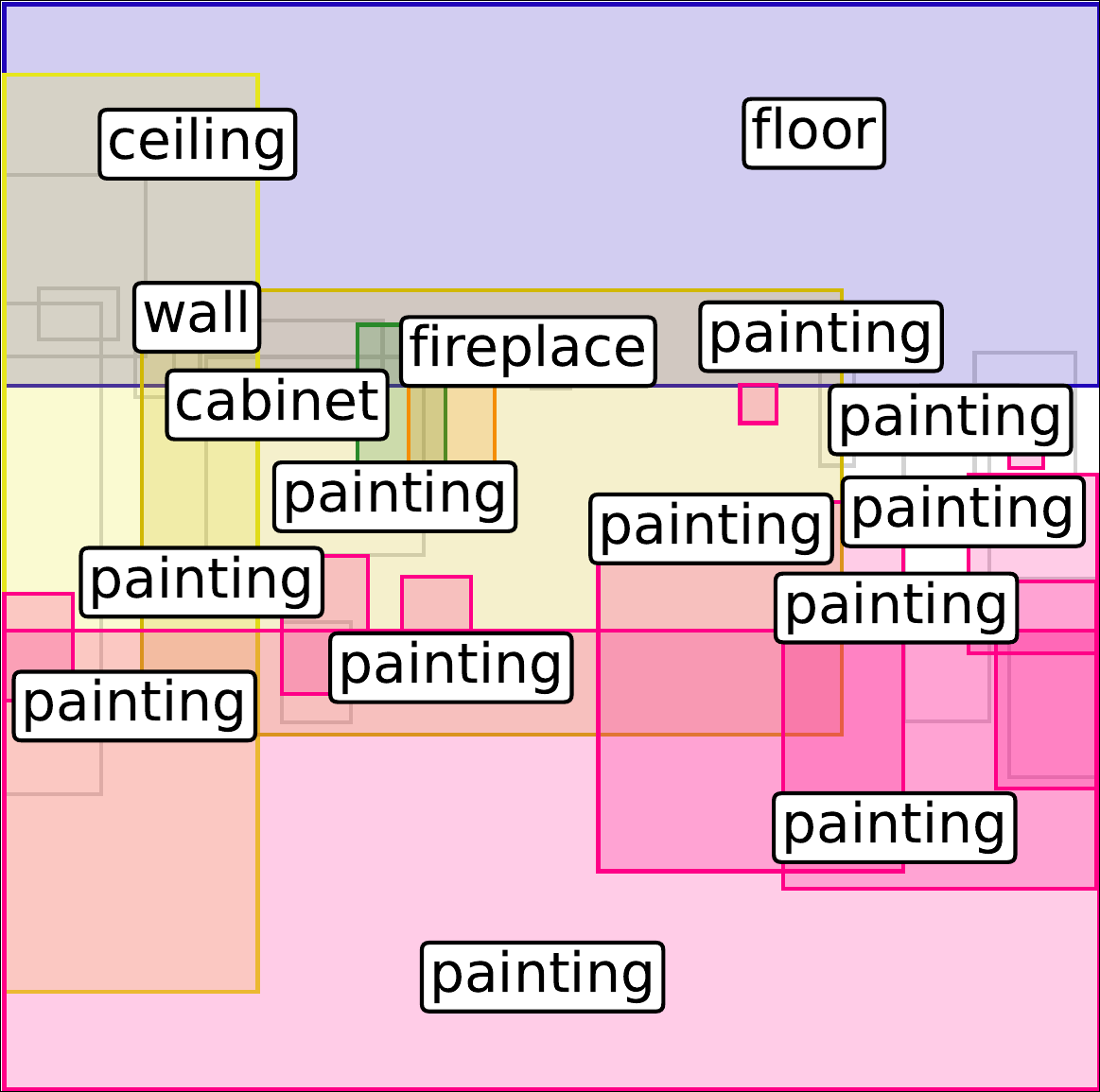} &
        \includegraphics[width=0.3\textwidth]{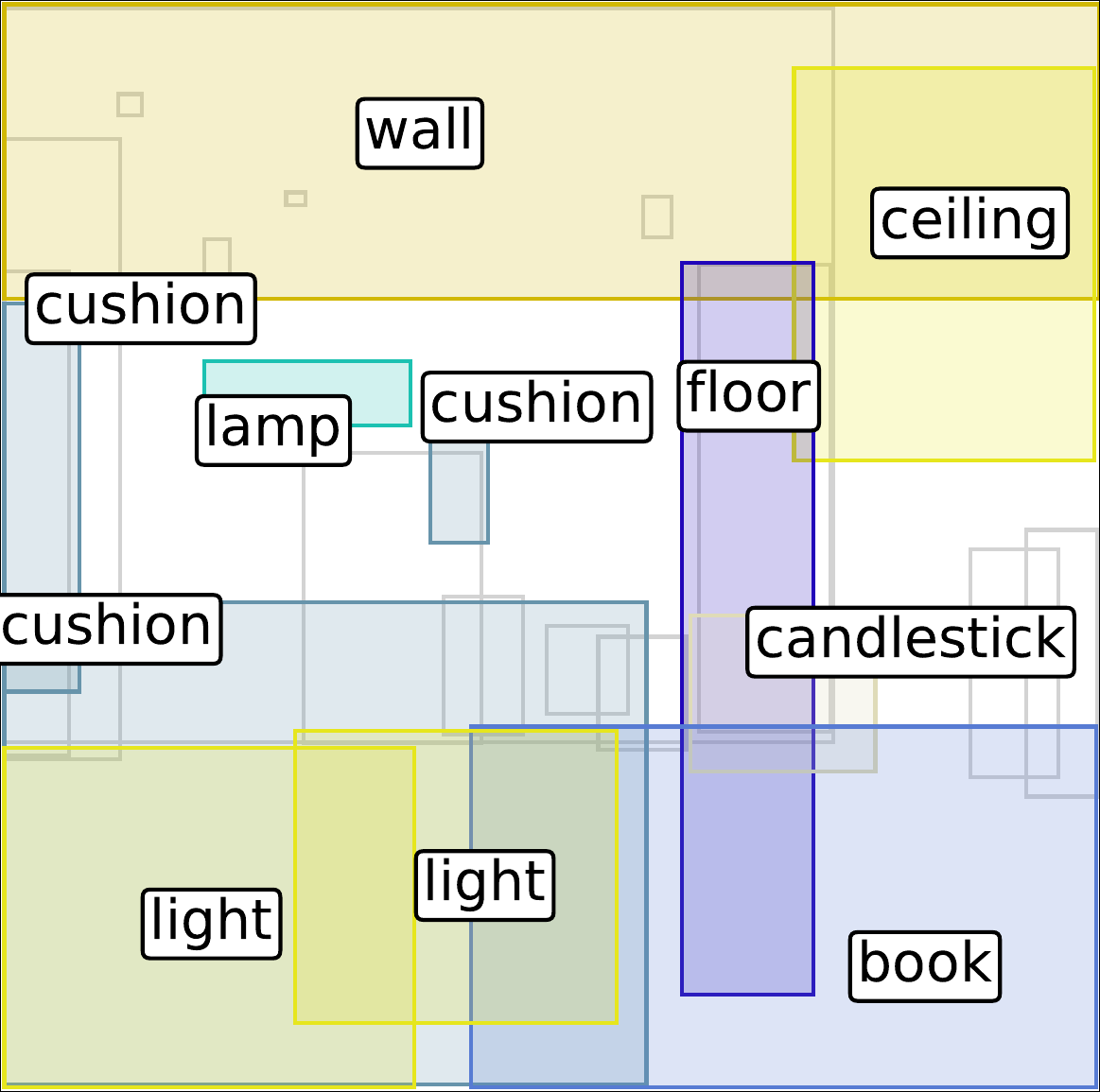} &
        \includegraphics[width=0.3\textwidth]{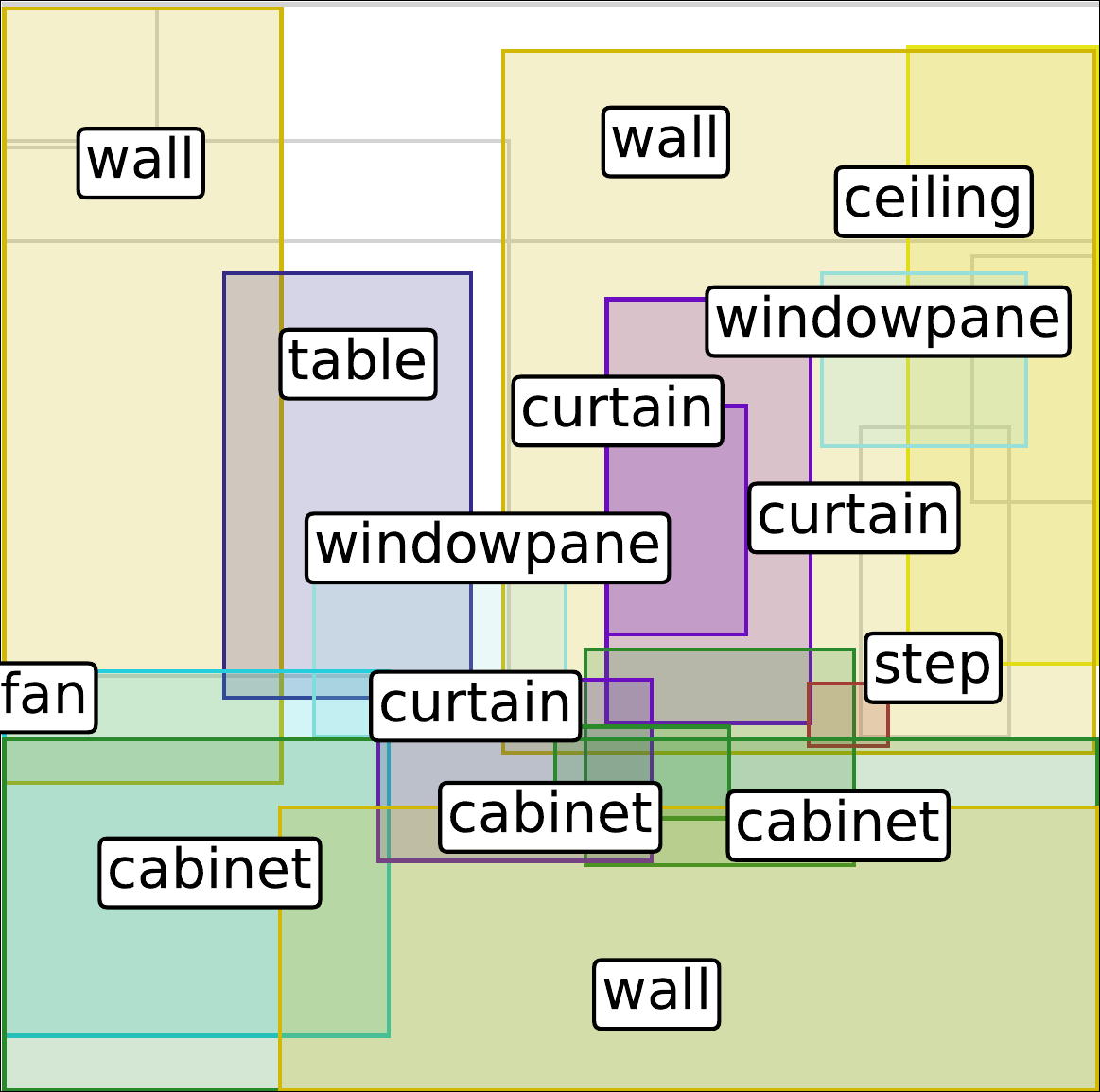} \\
    \end{tabular}
    \caption{Layout Transformer - Living Room.}
    \label{fig:lt_living_room}
\end{figure*}

\begin{figure*}[ht]
    \centering
    \begin{tabular}{c c c}
        \includegraphics[width=0.3\textwidth]{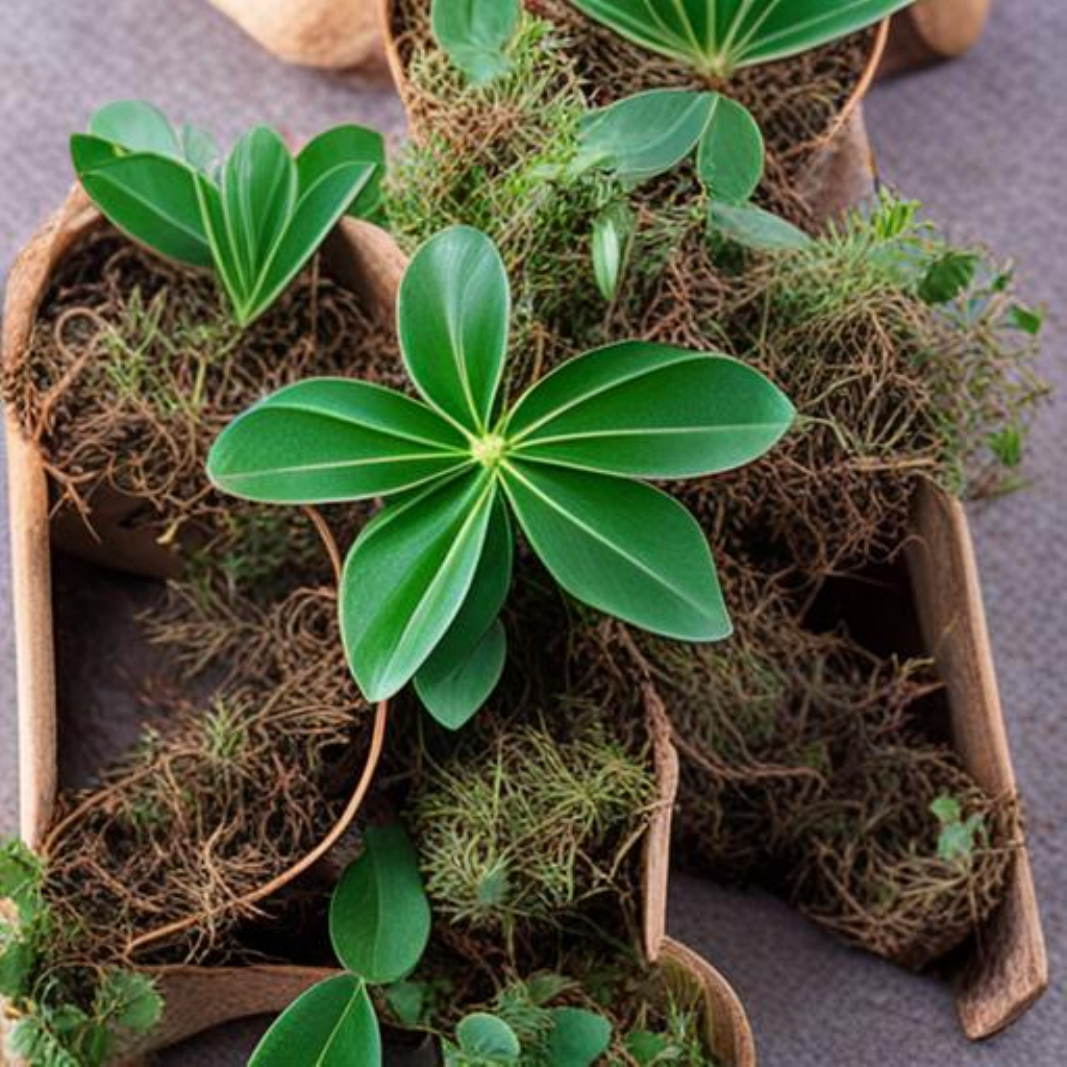} &
        \includegraphics[width=0.3\textwidth]{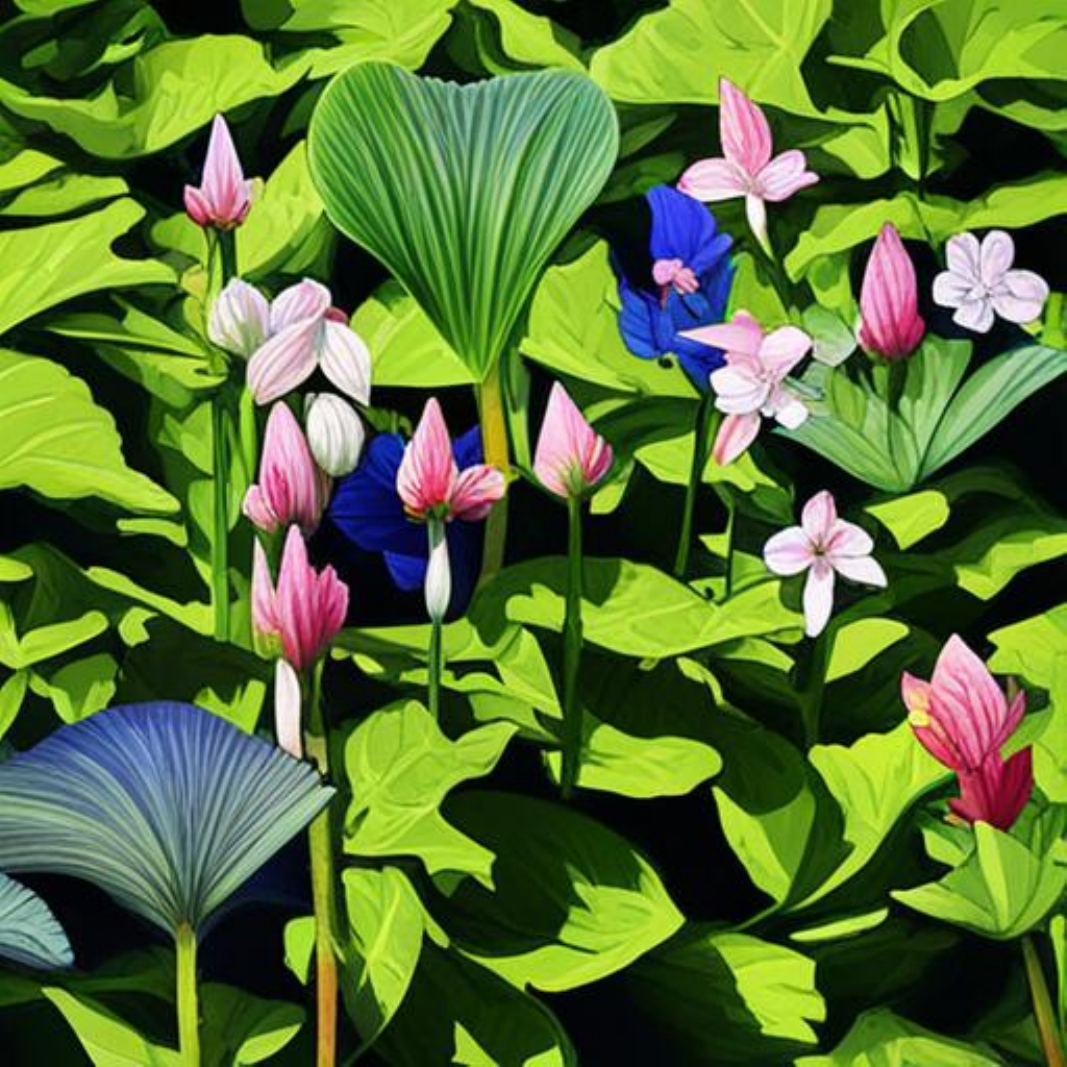} &
        \includegraphics[width=0.3\textwidth]{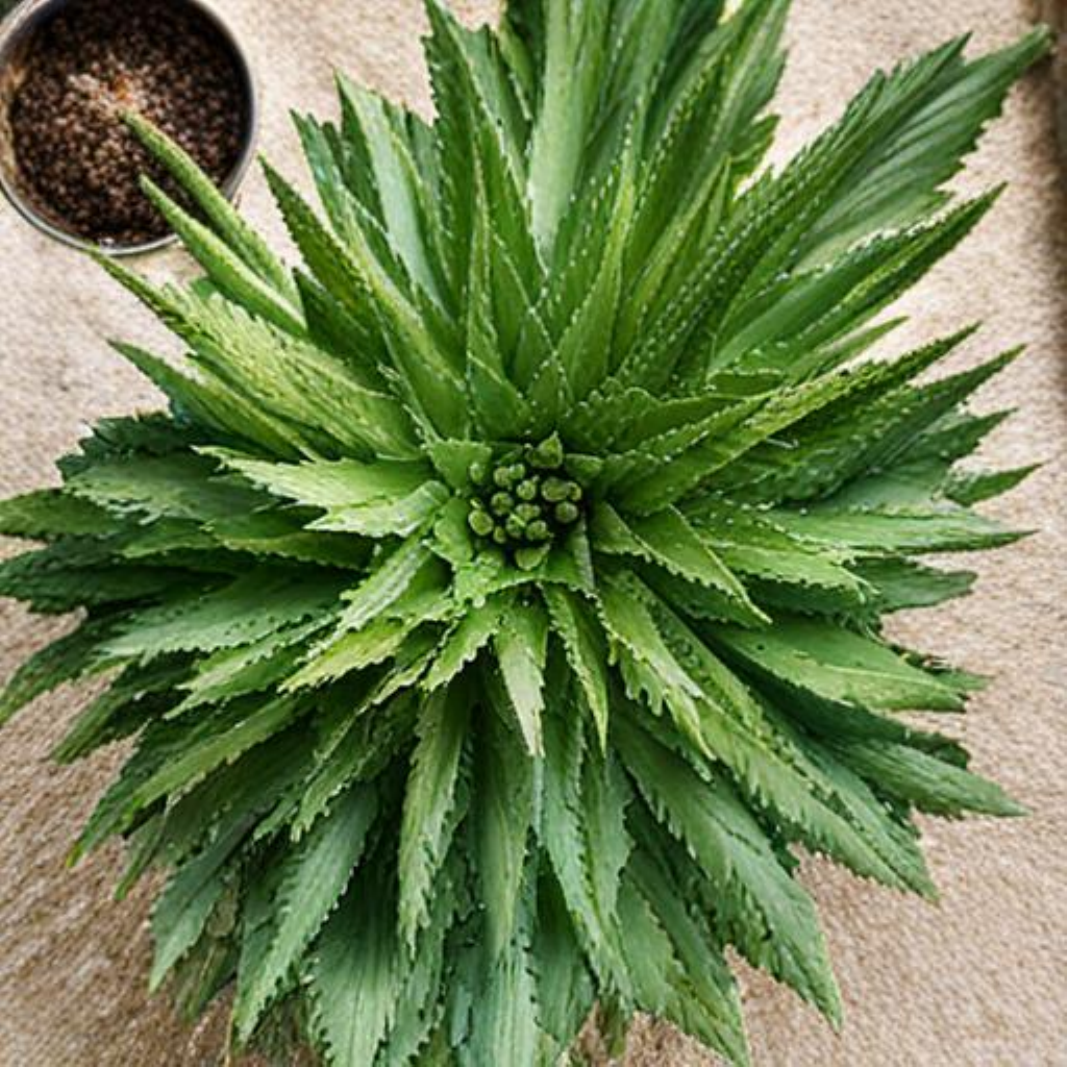} \\
        
        \includegraphics[width=0.3\textwidth]{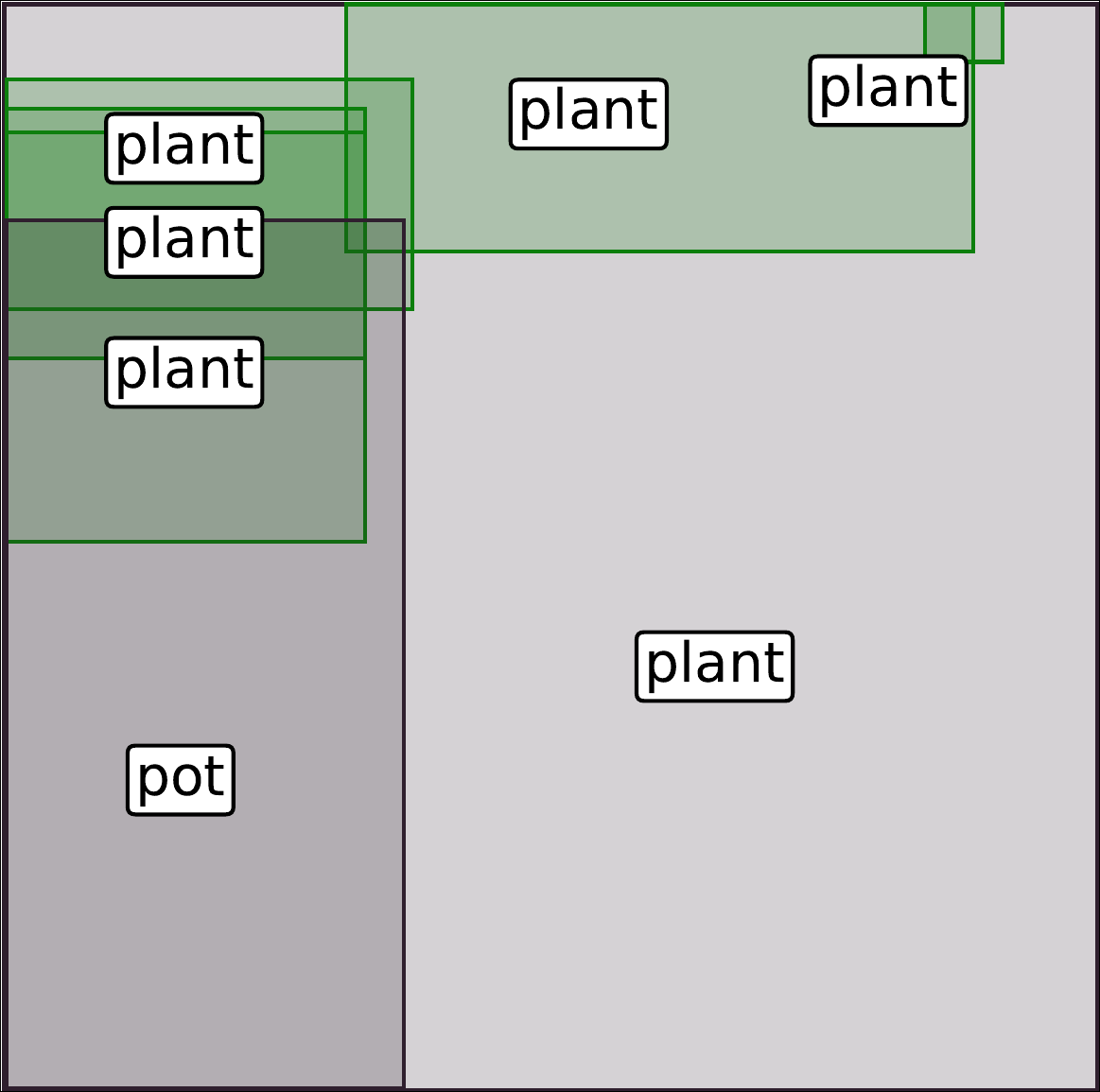} &
        \includegraphics[width=0.3\textwidth]{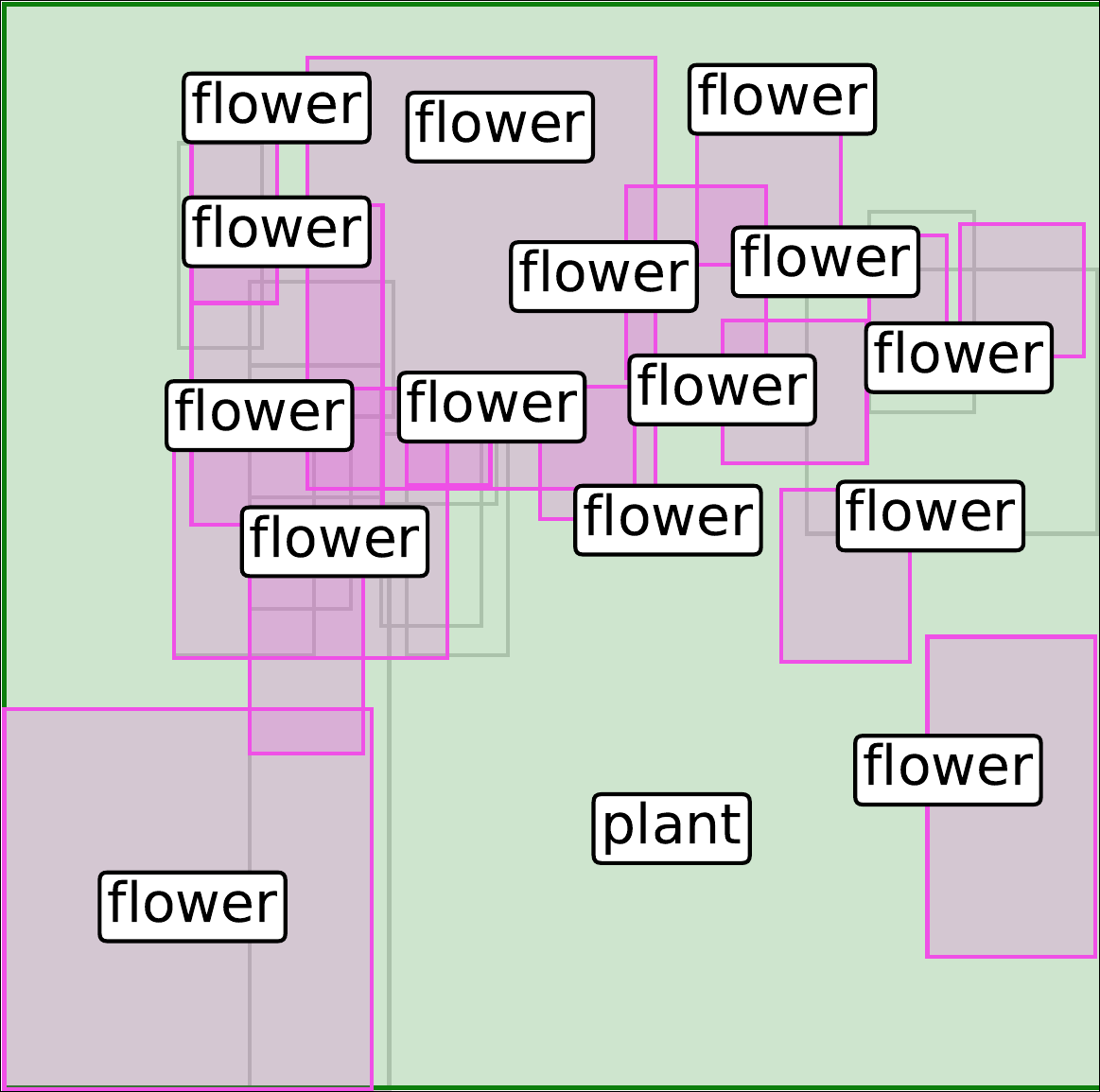} &
        \includegraphics[width=0.3\textwidth]{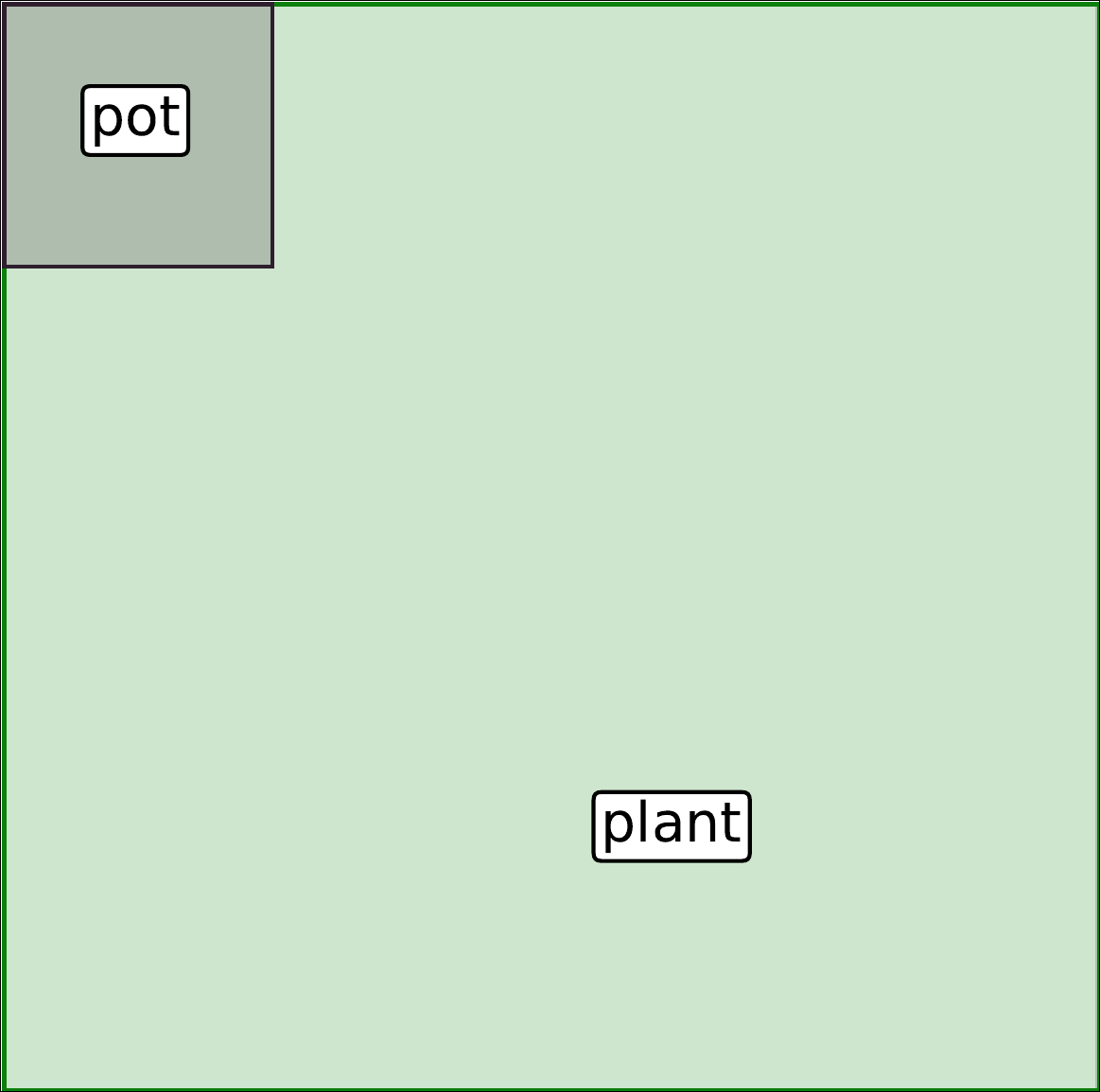} \\
    \end{tabular}
    \caption{Layout Transformer - Roof Top.}
    \label{fig:lt_rooftop}
\end{figure*}

\begin{figure*}[ht]
    \centering
    \begin{tabular}{c c c}
        \includegraphics[width=0.3\textwidth]{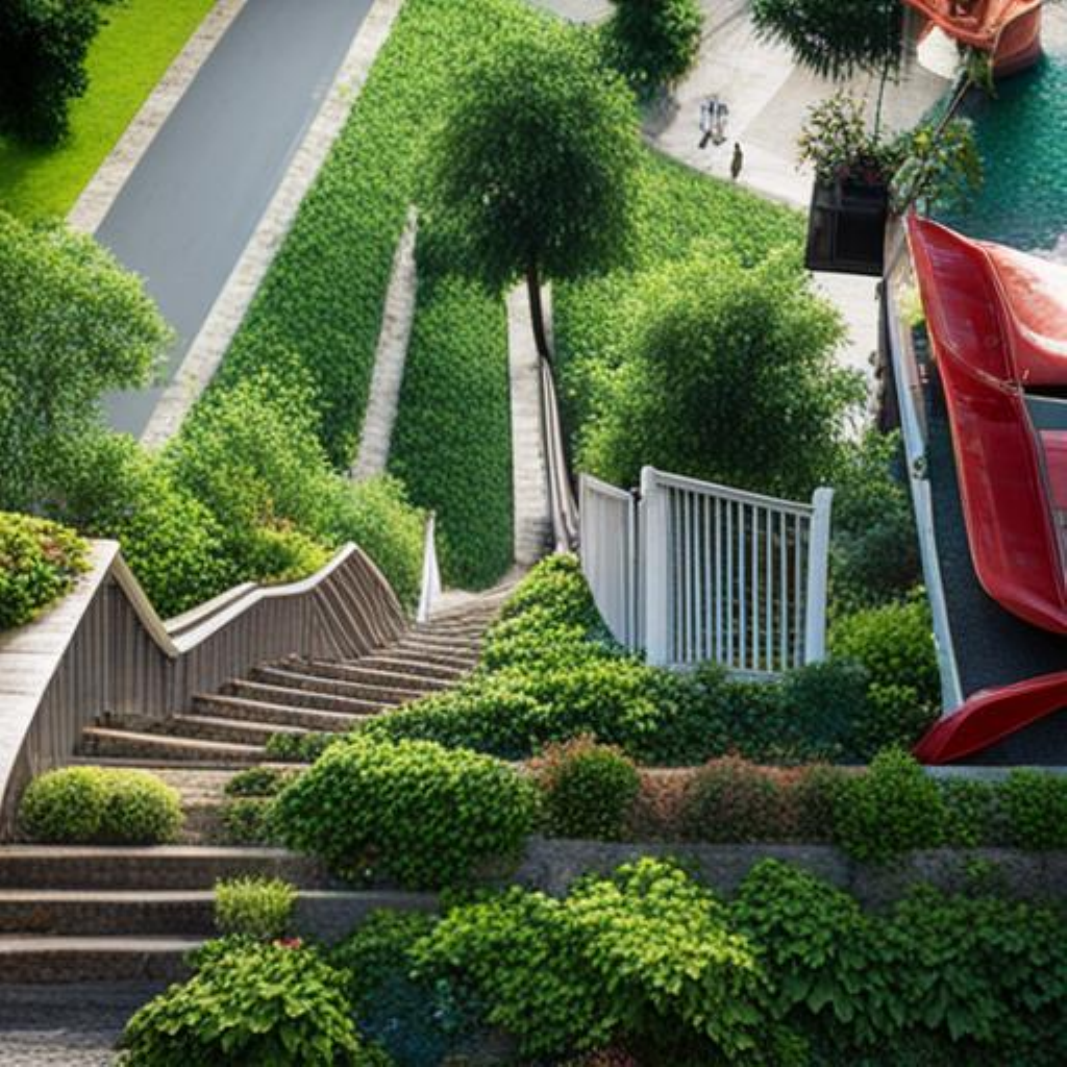} &
        \includegraphics[width=0.3\textwidth]{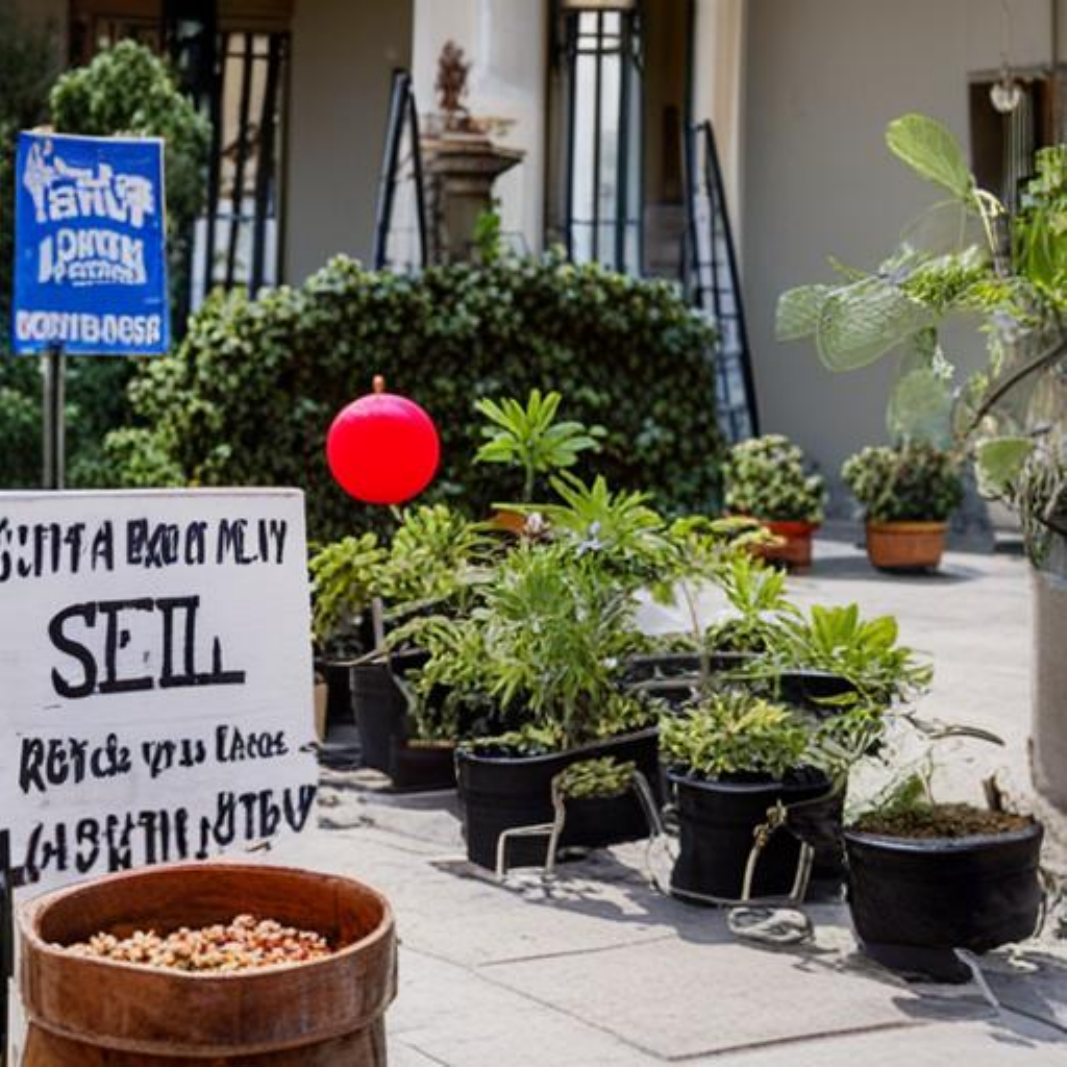} &
        \includegraphics[width=0.3\textwidth]{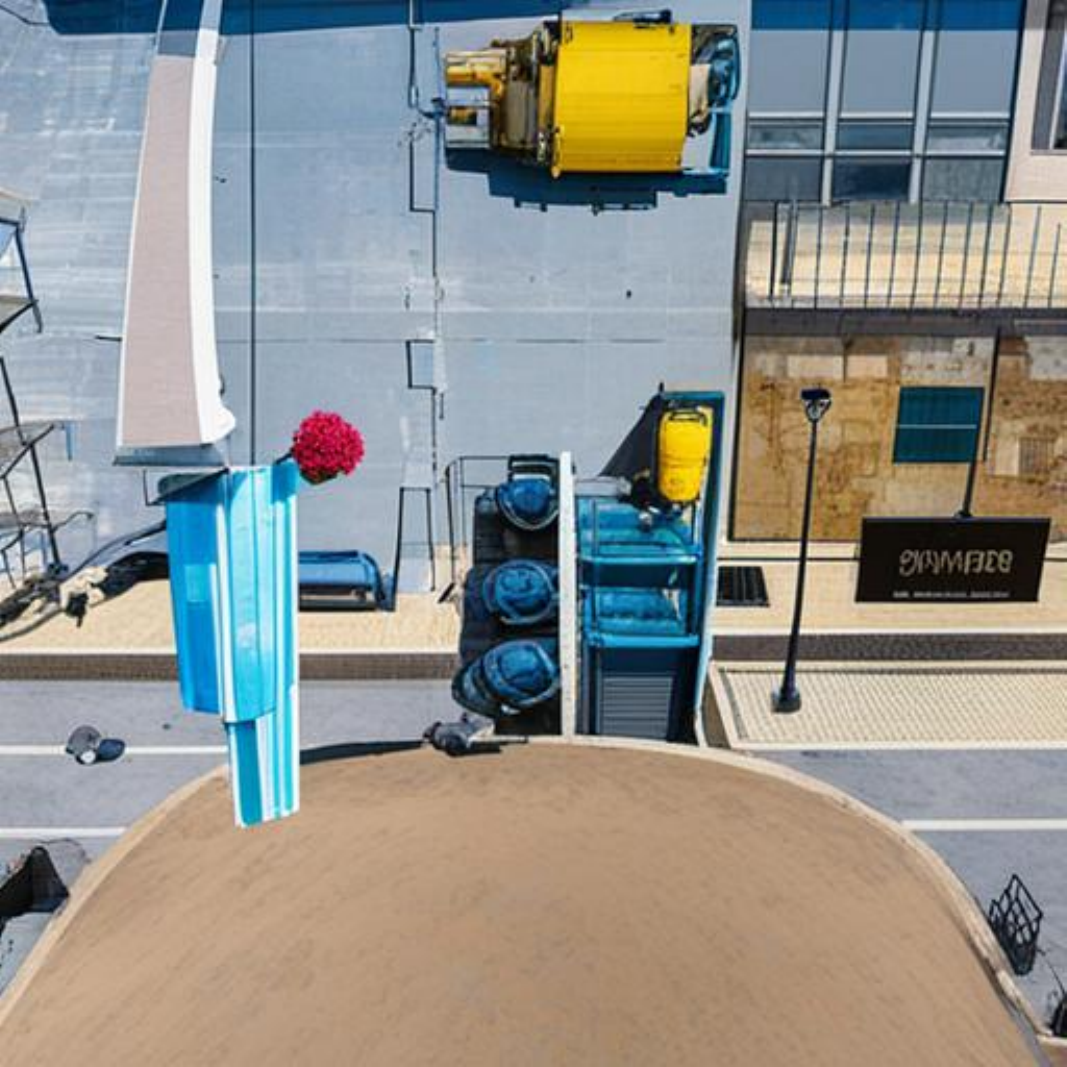} \\
        
        \includegraphics[width=0.3\textwidth]{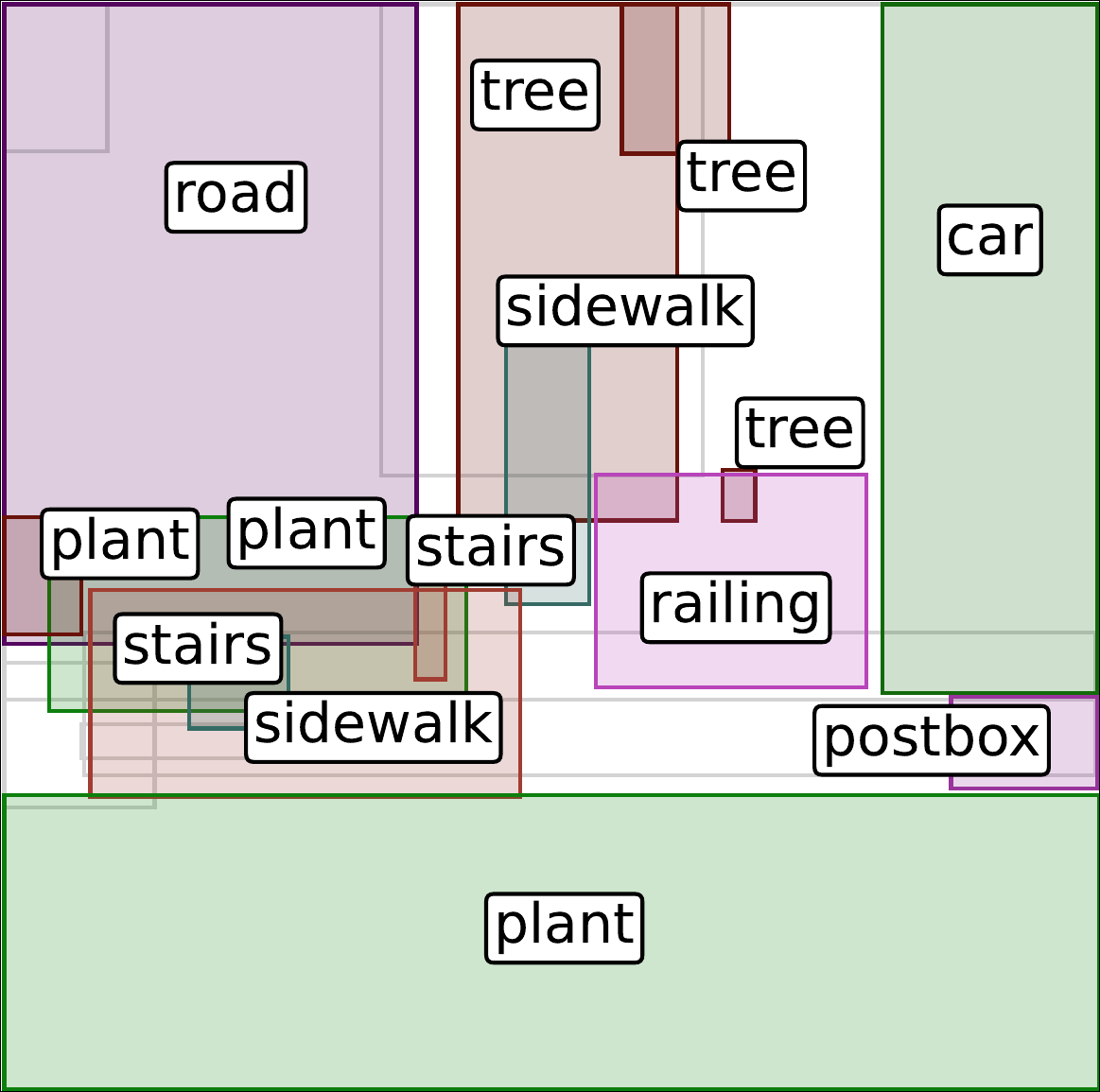} &
        \includegraphics[width=0.3\textwidth]{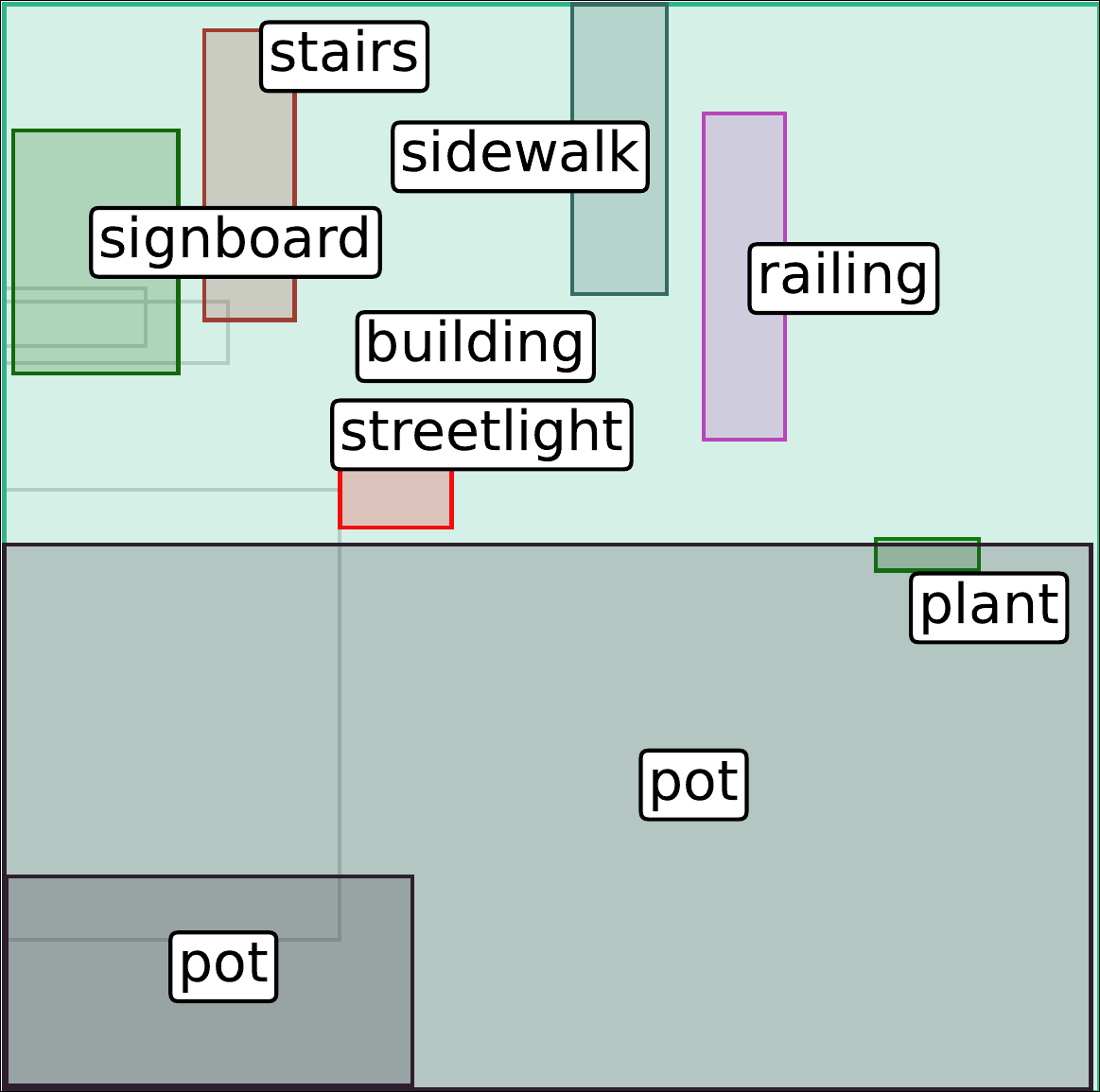} &
        \includegraphics[width=0.3\textwidth]{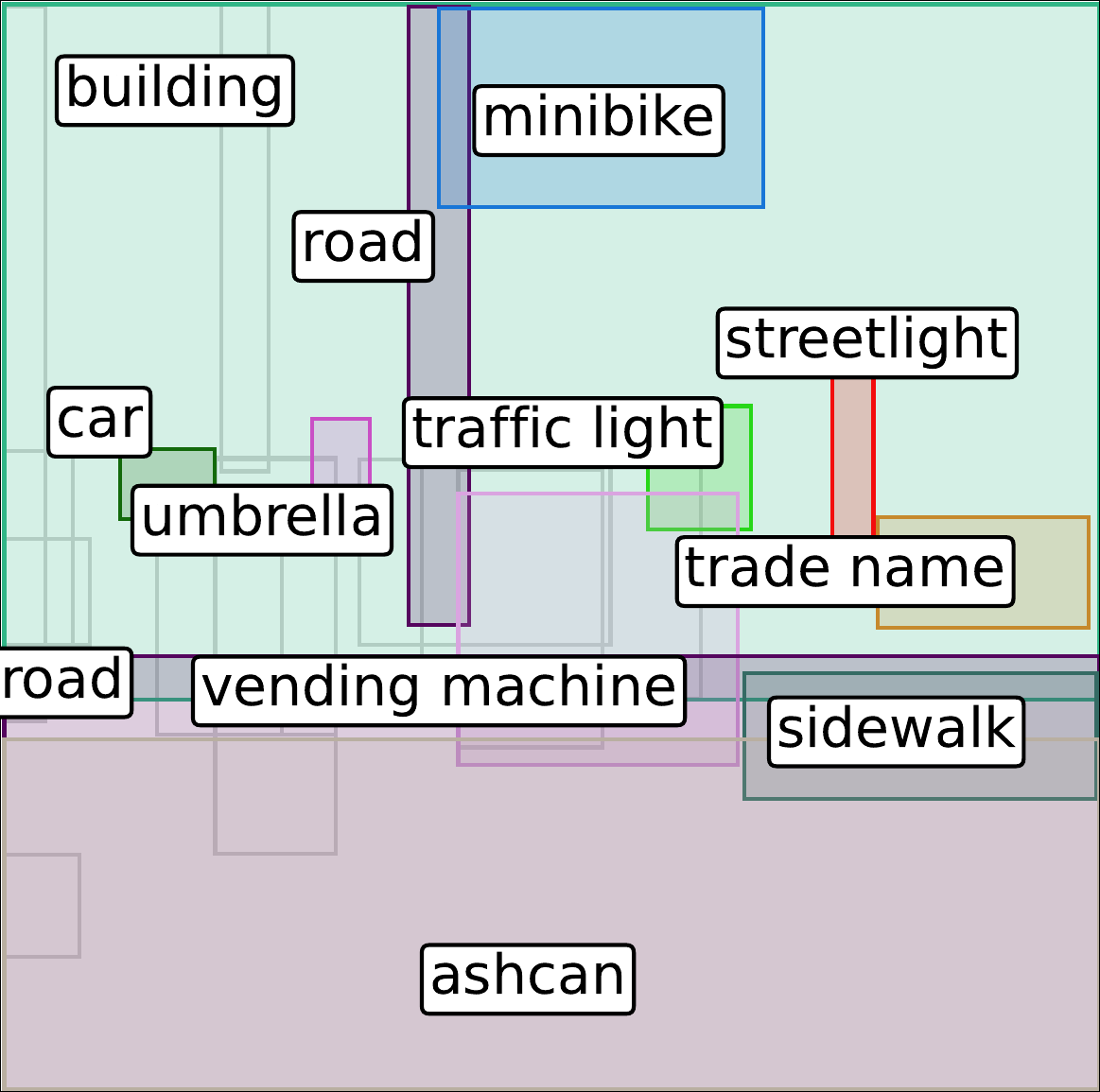} \\
    \end{tabular}
    \caption{Layout Transformer - Street.}
    \label{fig:layout_transformer_street}
\end{figure*}
    
\begin{figure*}[ht]
    \centering
    \begin{tabular}{c c c}
        \includegraphics[width=0.3\textwidth]{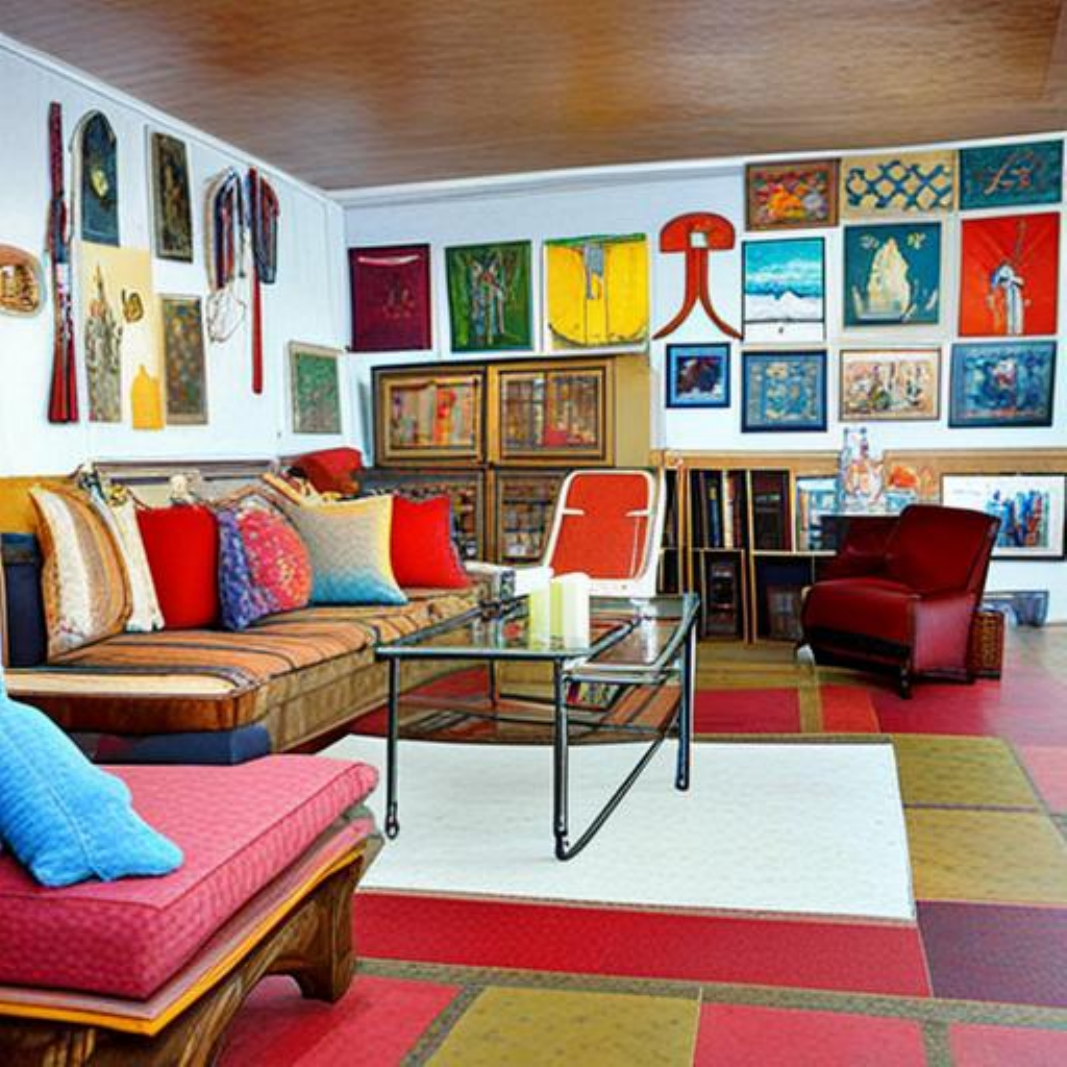} &
        \includegraphics[width=0.3\textwidth]{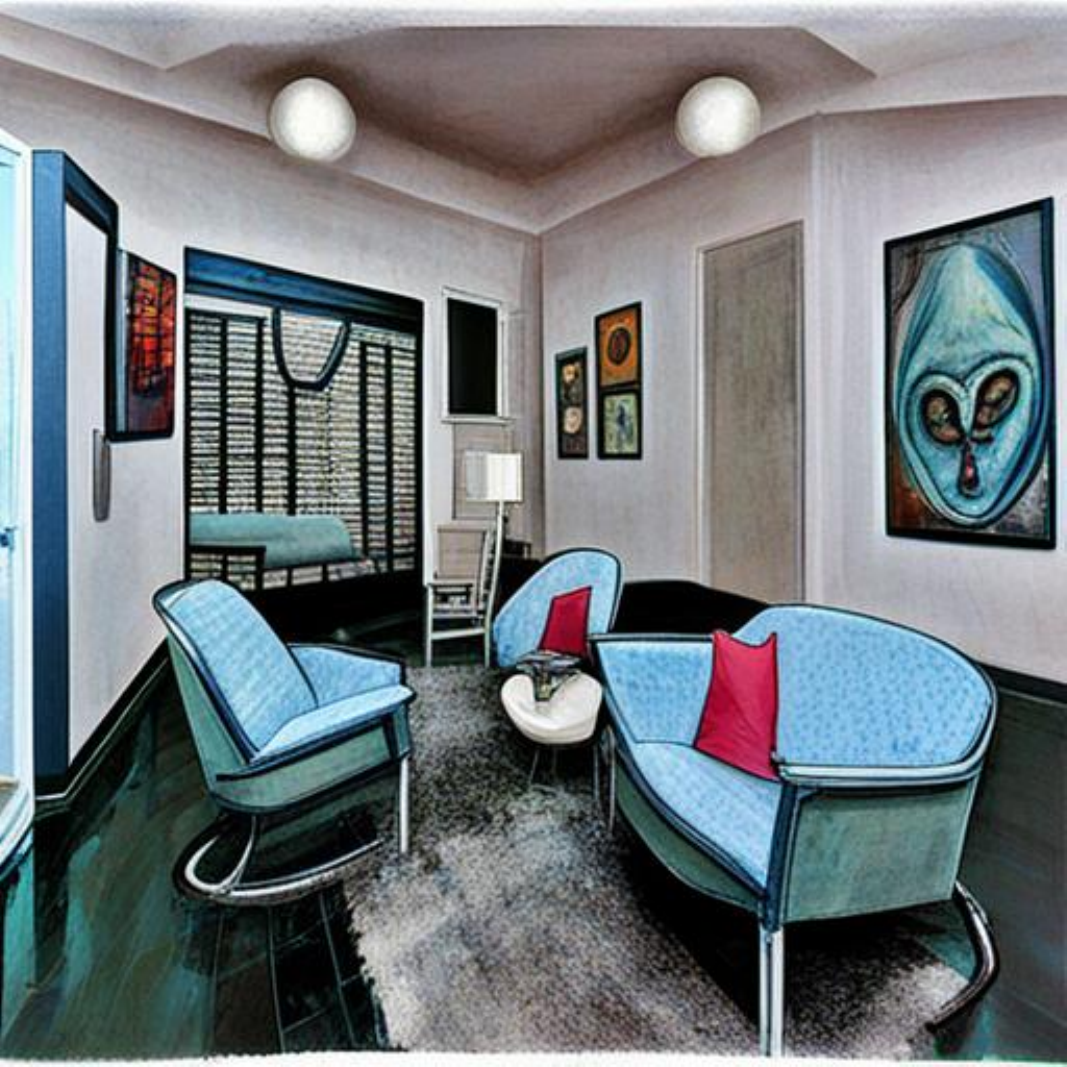} &
        \includegraphics[width=0.3\textwidth]{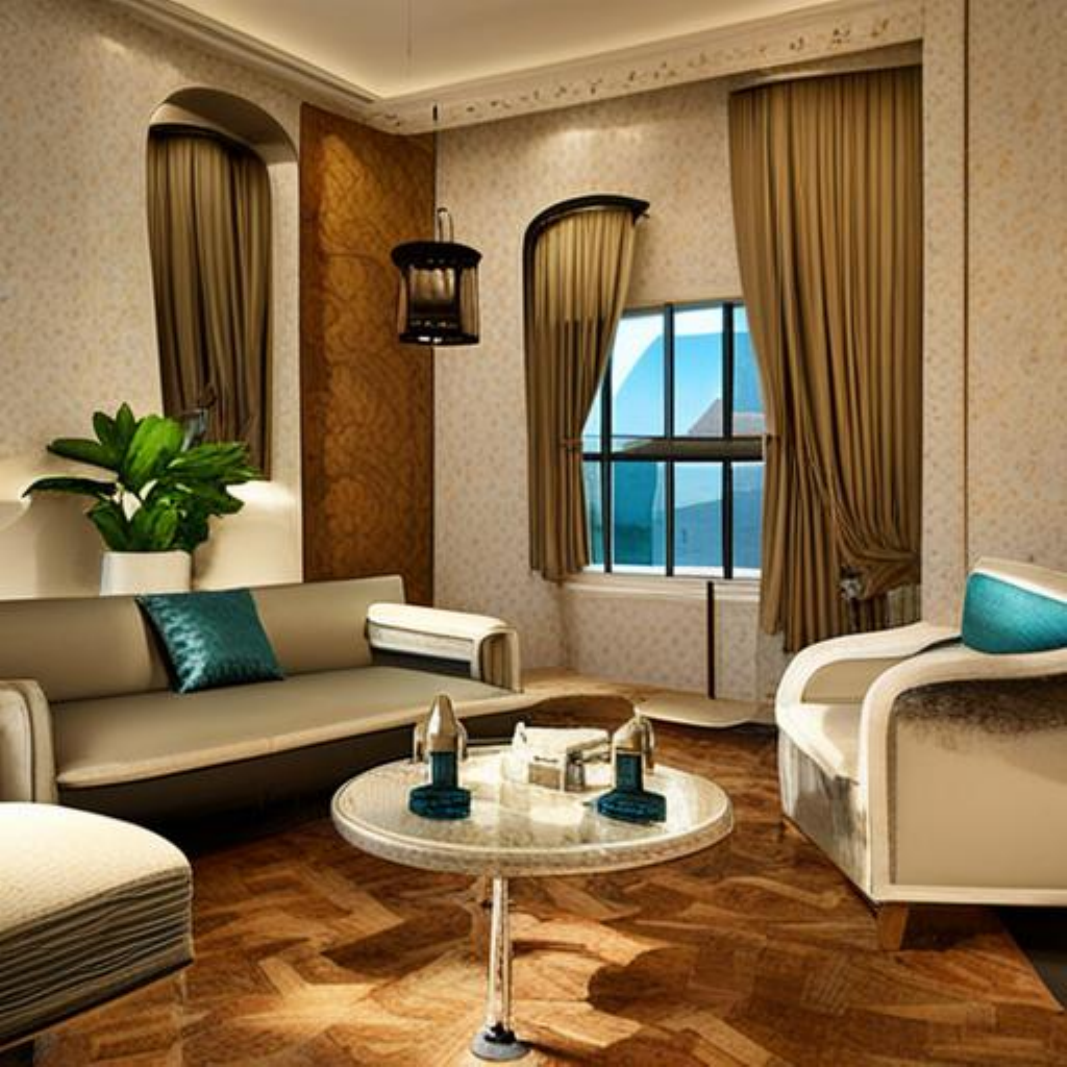} \\
        
        \includegraphics[width=0.3\textwidth]{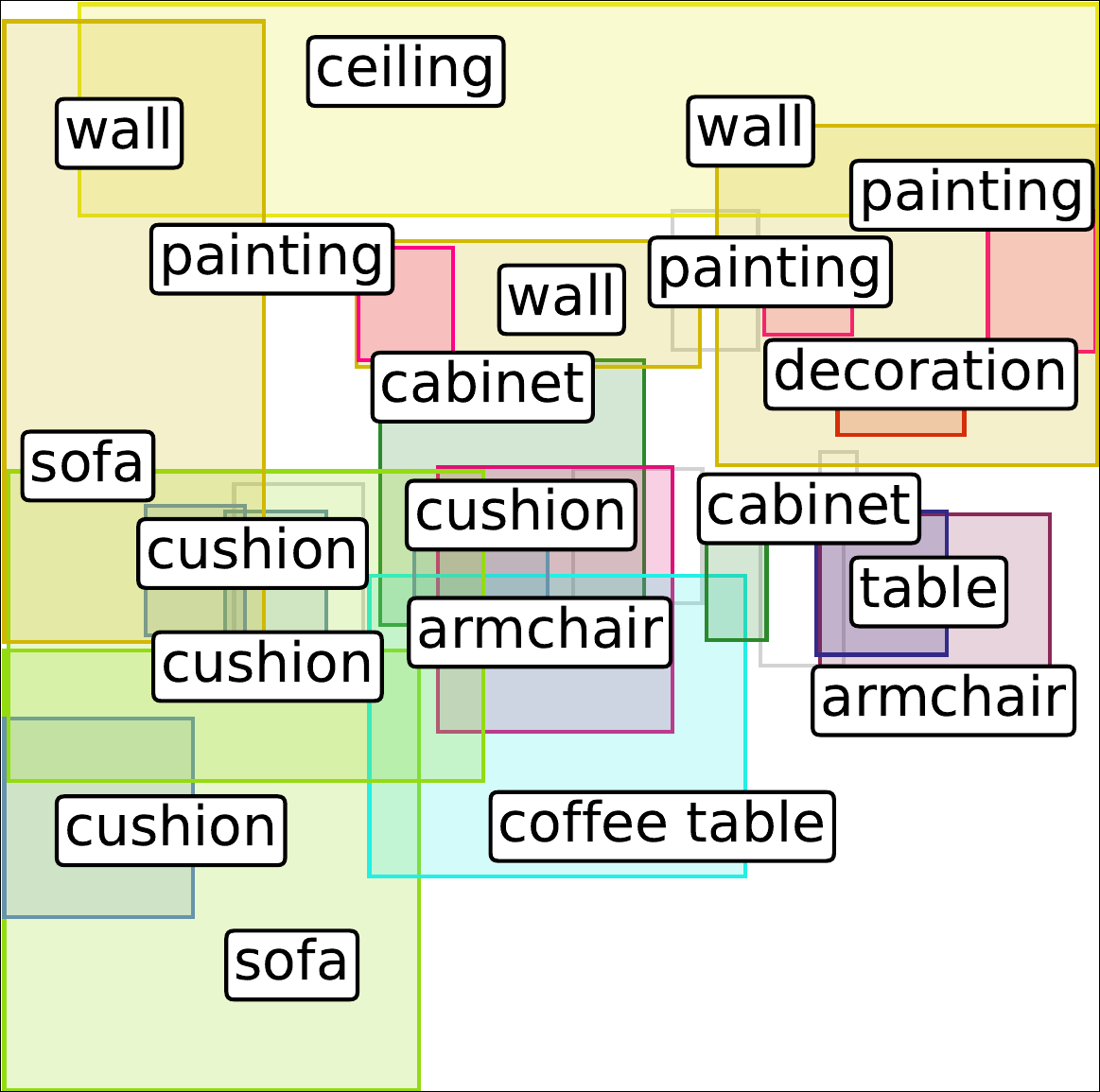} &
        \includegraphics[width=0.3\textwidth]{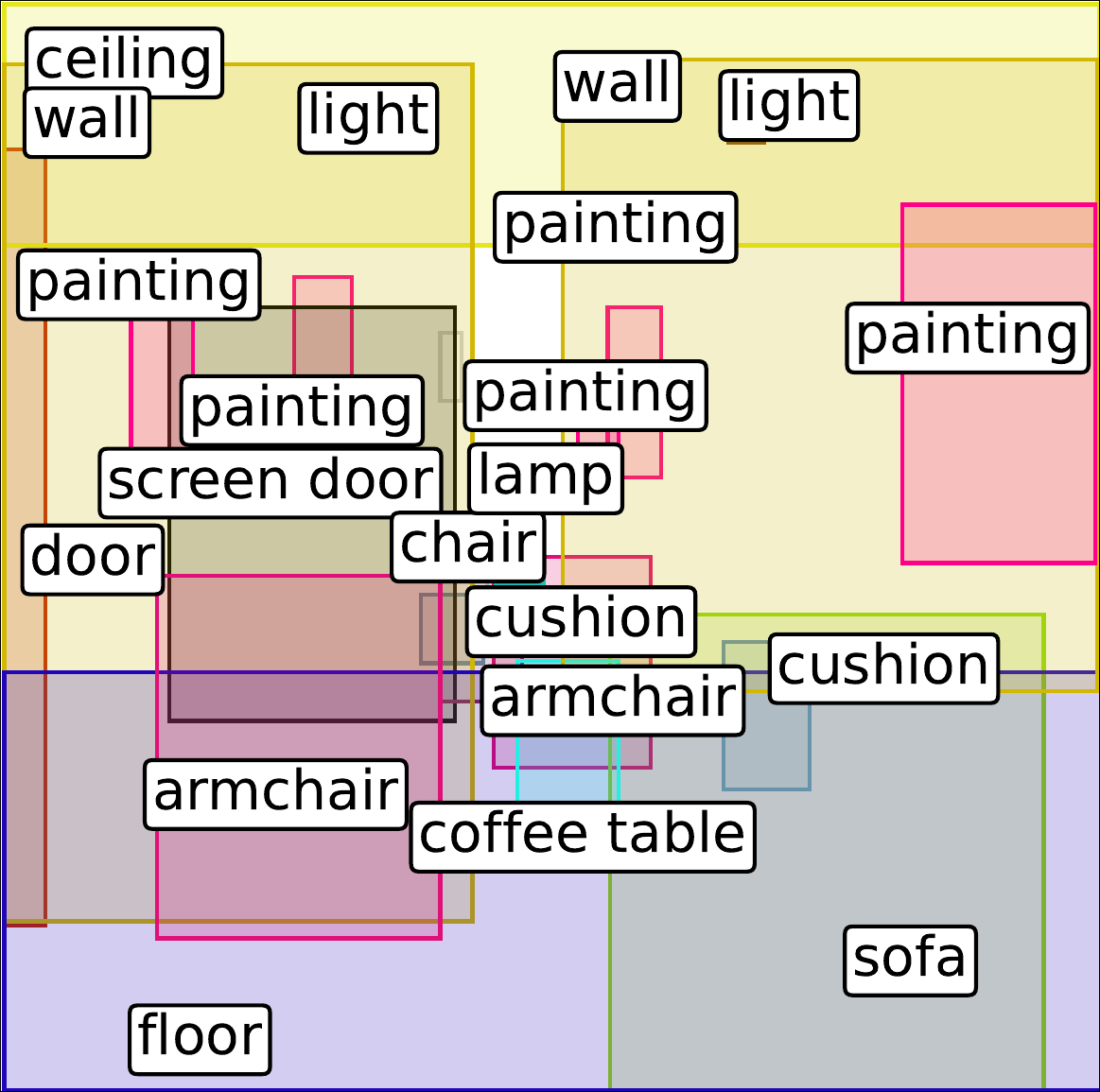} &
        \includegraphics[width=0.3\textwidth]{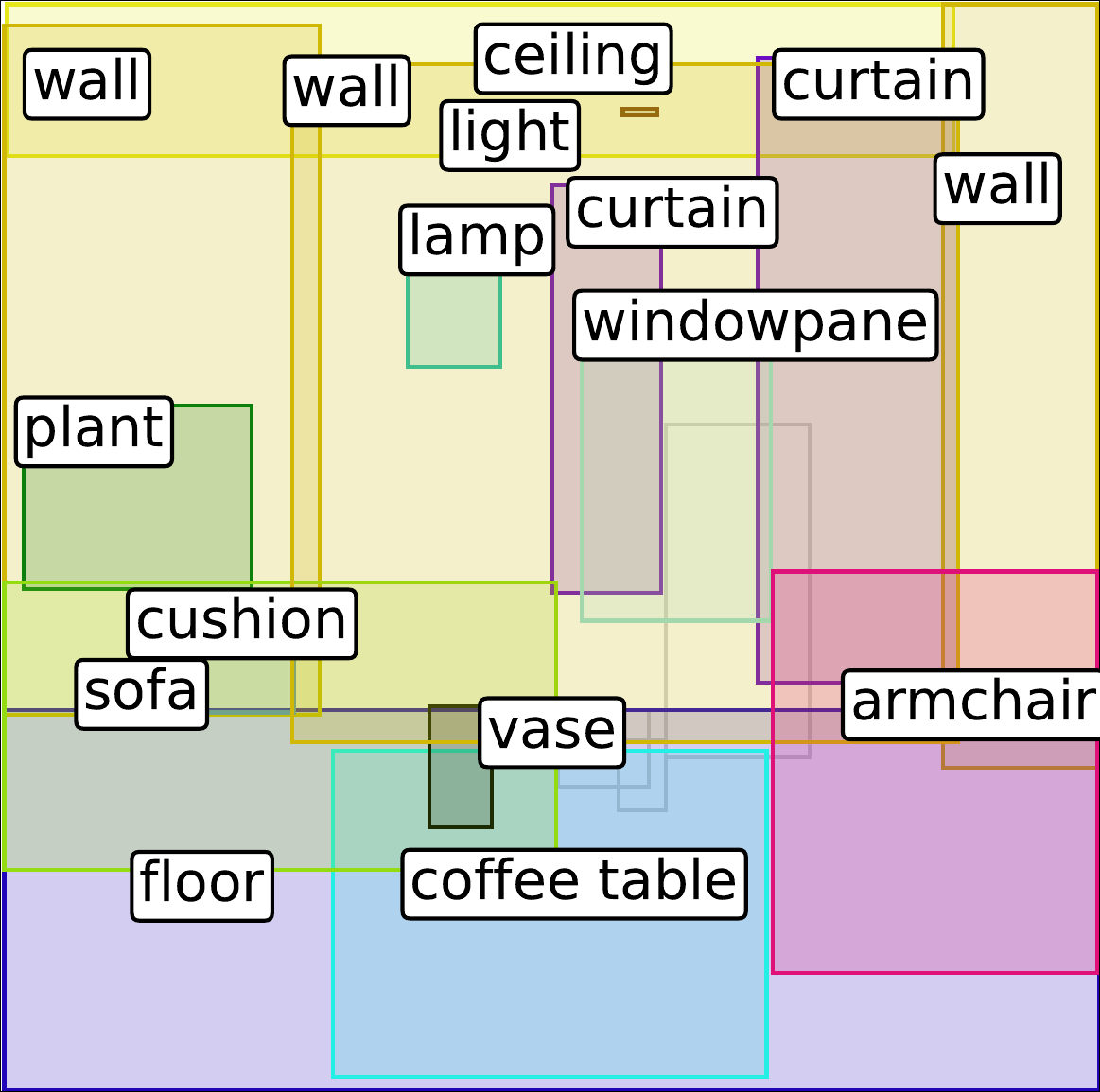} \\
    \end{tabular}
    \caption{Ours - Living Room.}
    \label{fig:ours_living_room}
\end{figure*}

\begin{figure*}[ht]
    \centering
    \begin{tabular}{c c c}
        \includegraphics[width=0.3\textwidth]{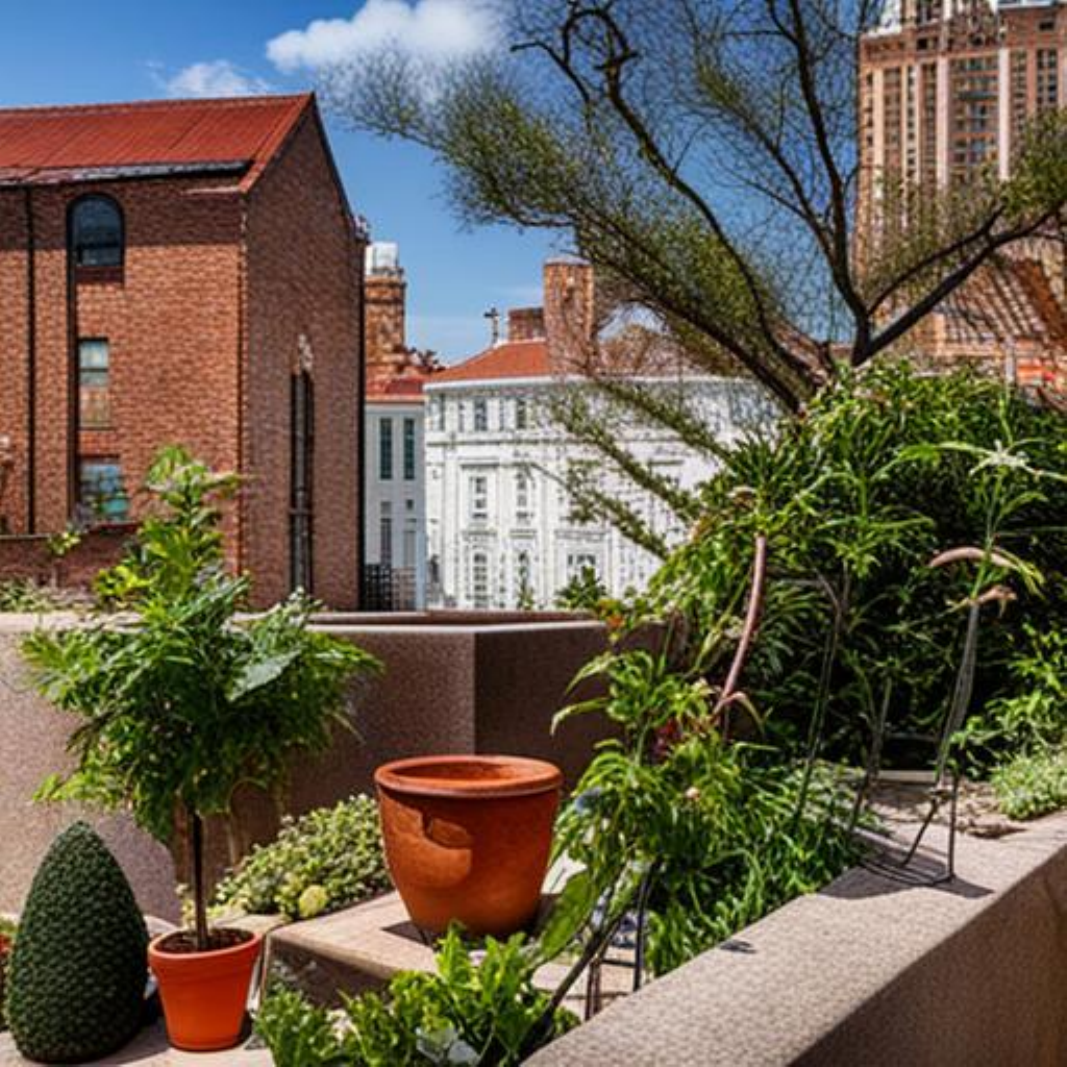} &
        \includegraphics[width=0.3\textwidth]{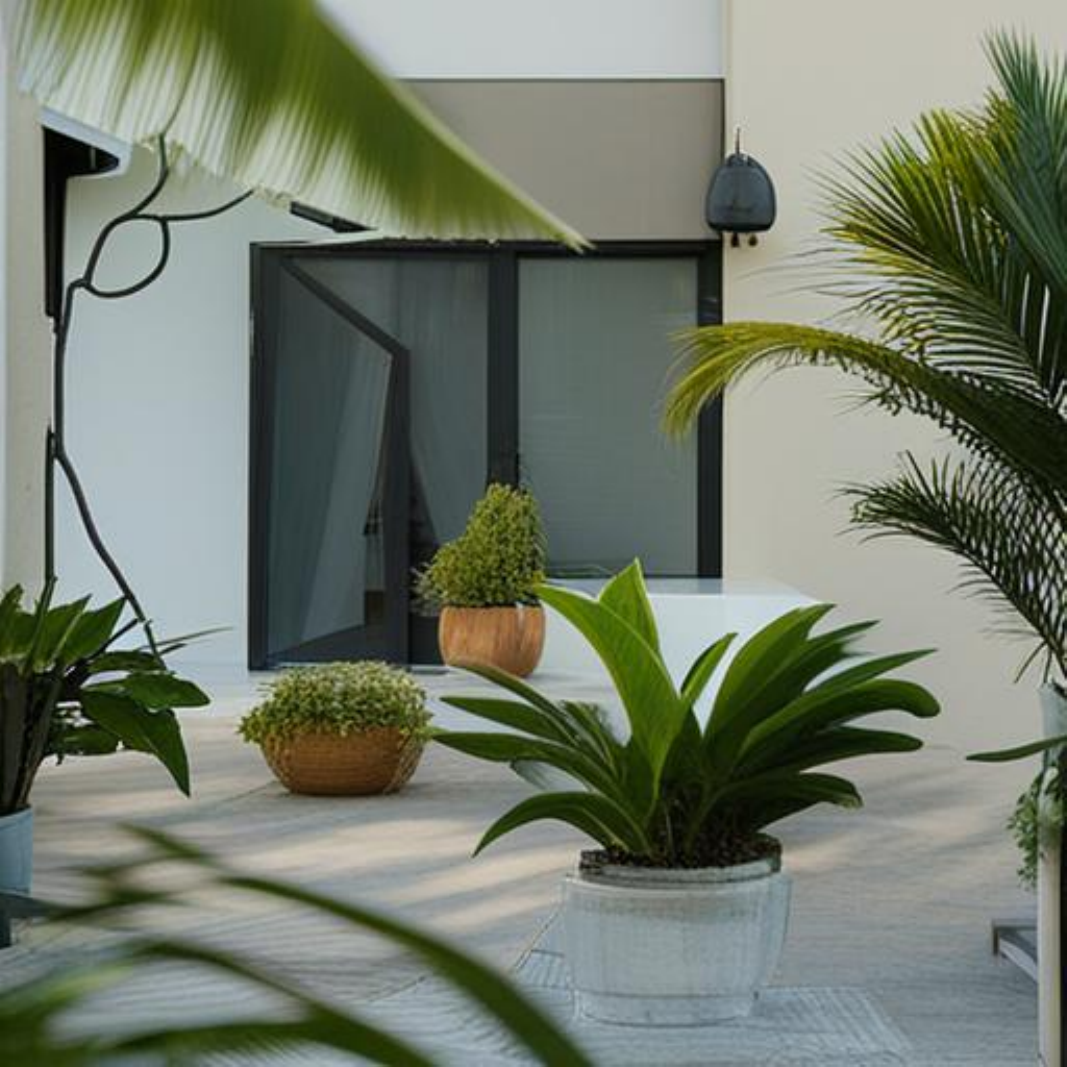} &
        \includegraphics[width=0.3\textwidth]{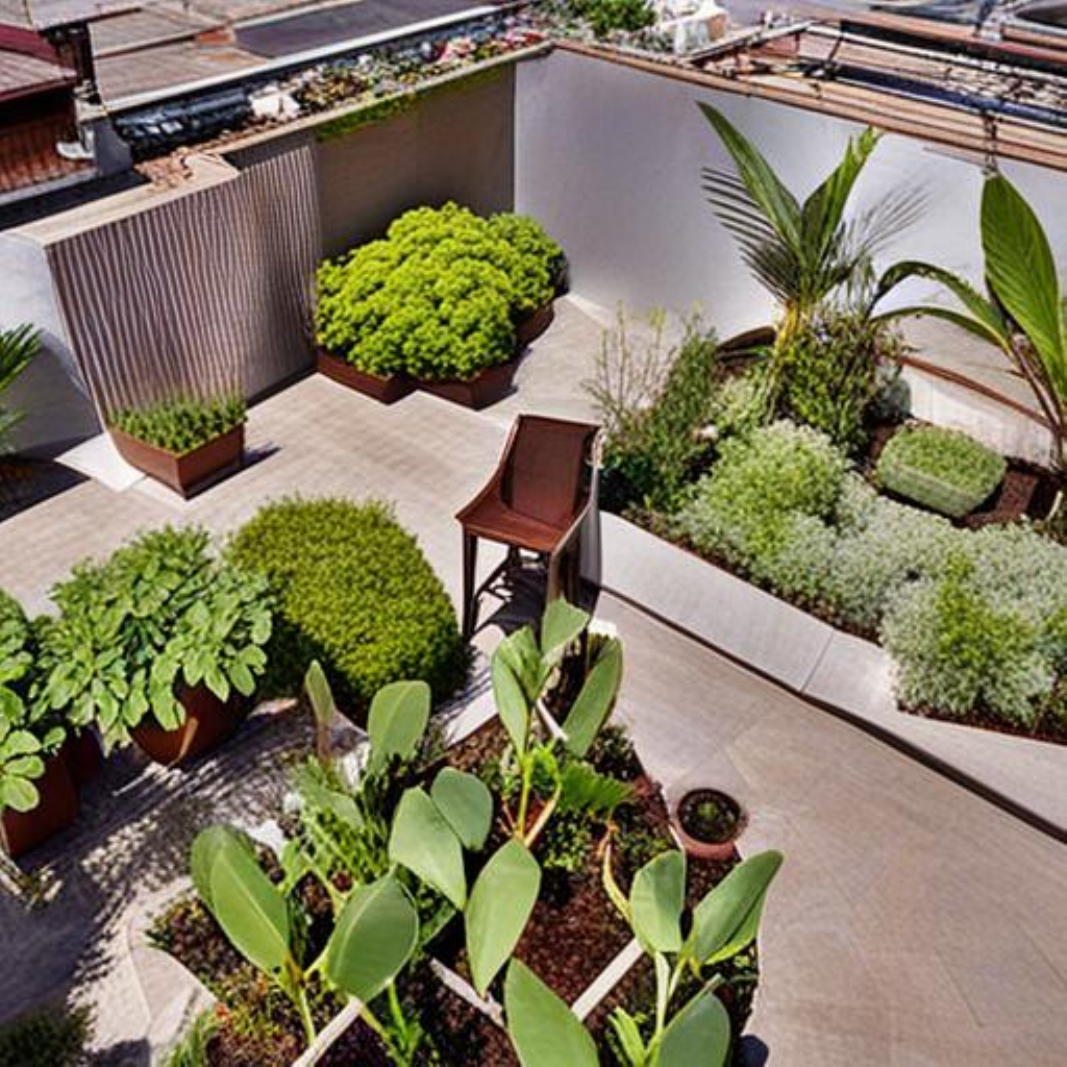} \\
        
        \includegraphics[width=0.3\textwidth]{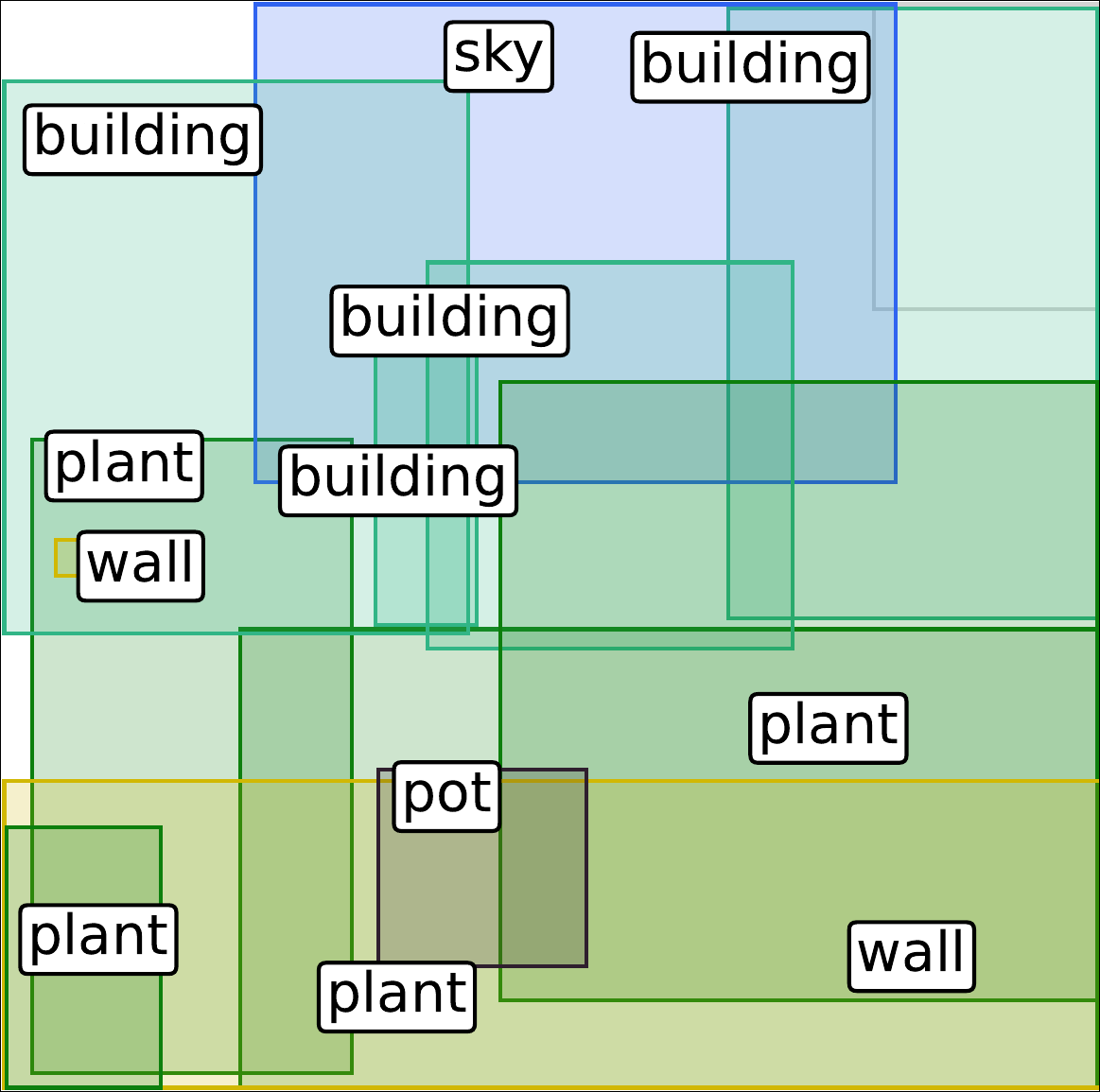} &
        \includegraphics[width=0.3\textwidth]{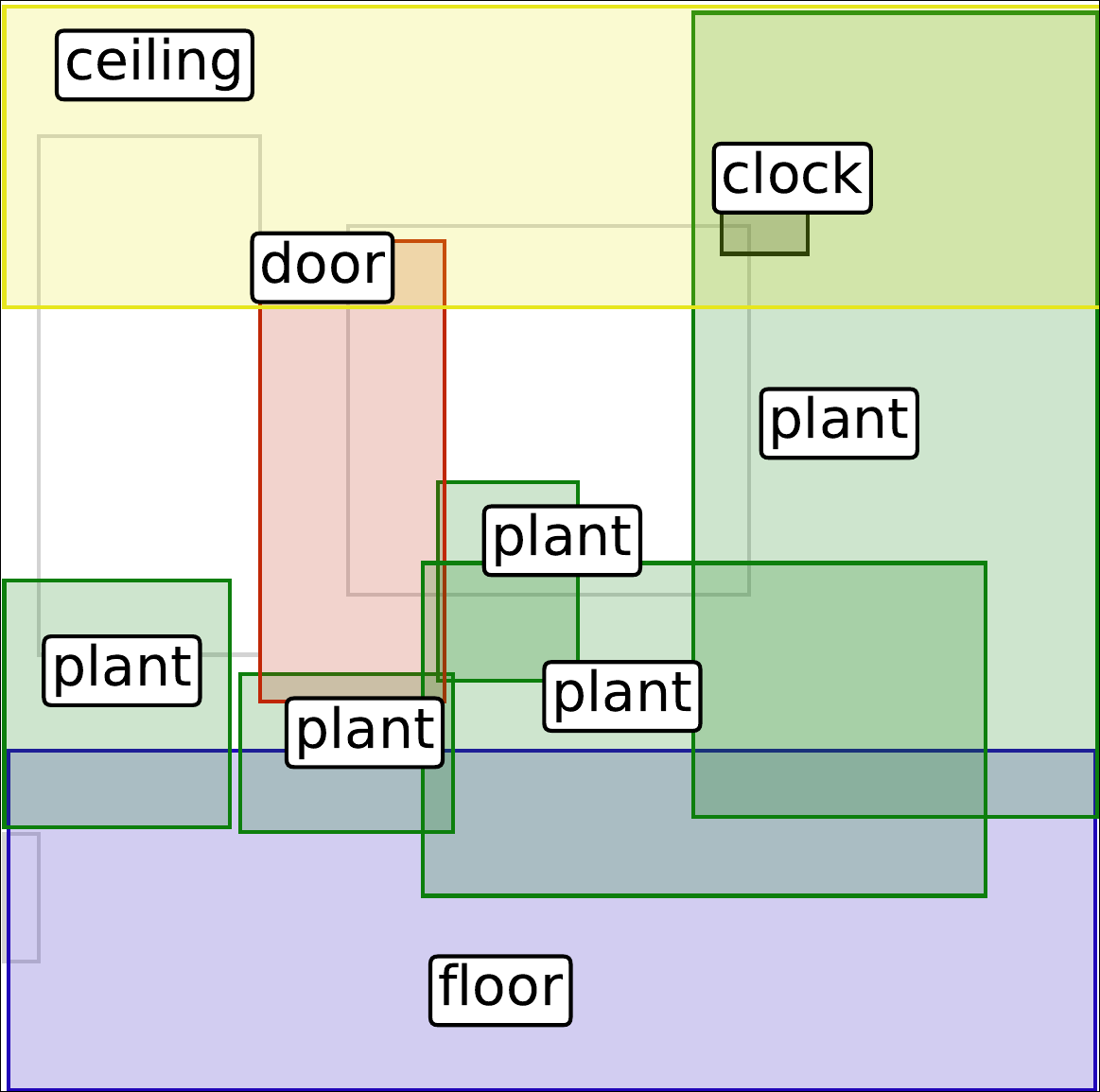} &
        \includegraphics[width=0.3\textwidth]{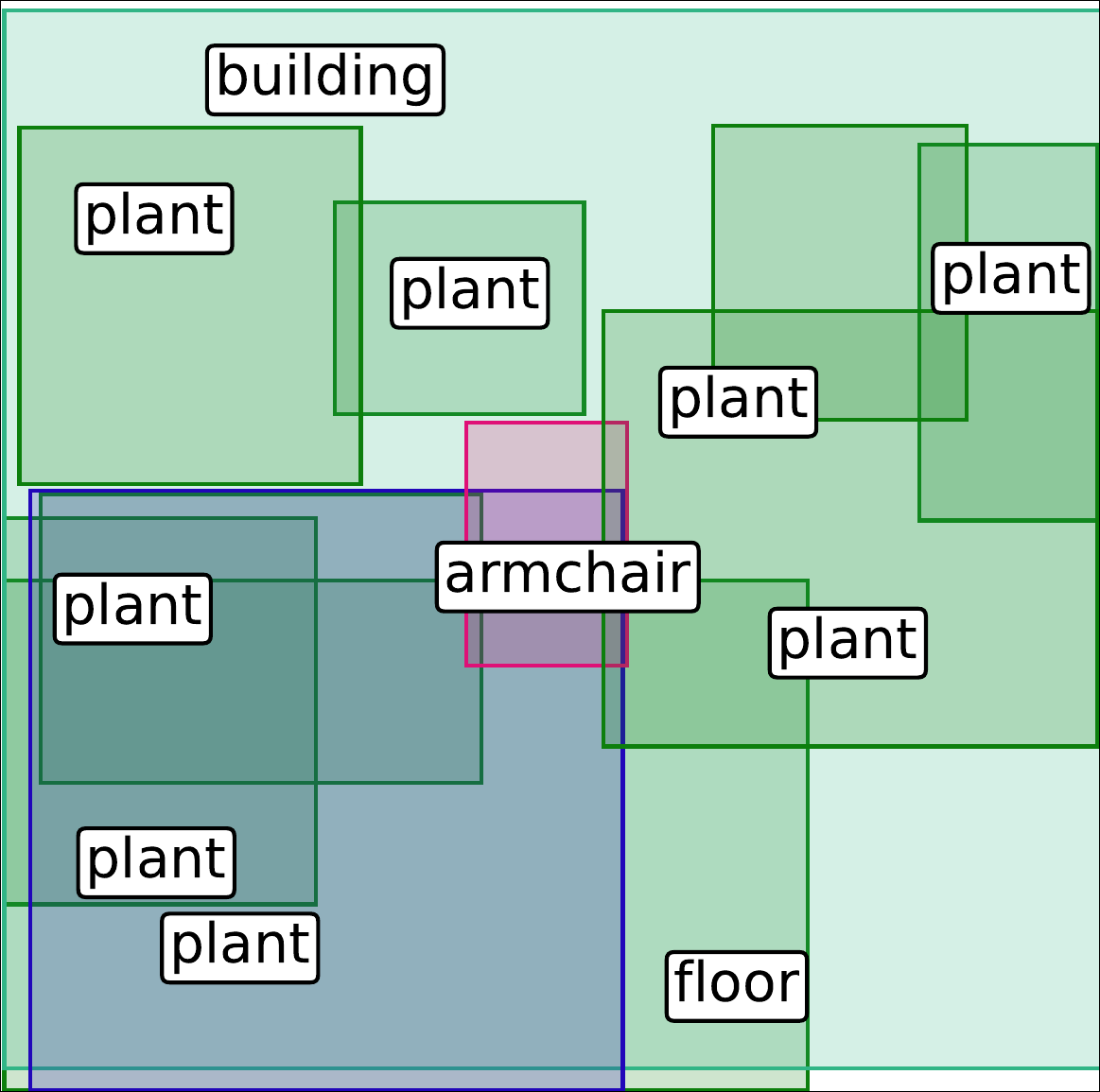} \\
    \end{tabular}
    \caption{Ours - Roof Top.}
    \label{fig:ours_rooftop}
\end{figure*}

\begin{figure*}[ht]
    \centering
    \begin{tabular}{c c c}
        \includegraphics[width=0.3\textwidth]{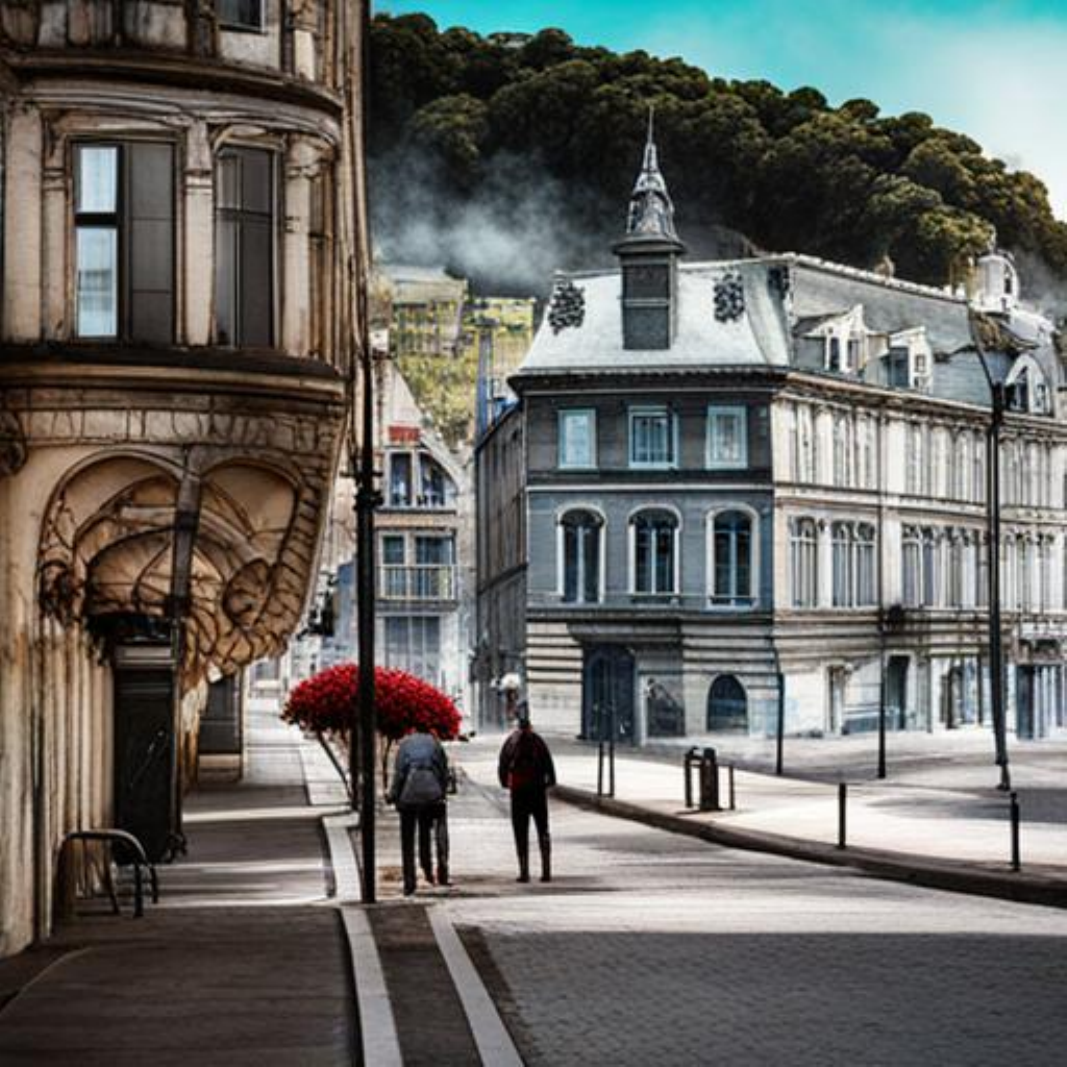} &
        \includegraphics[width=0.3\textwidth]{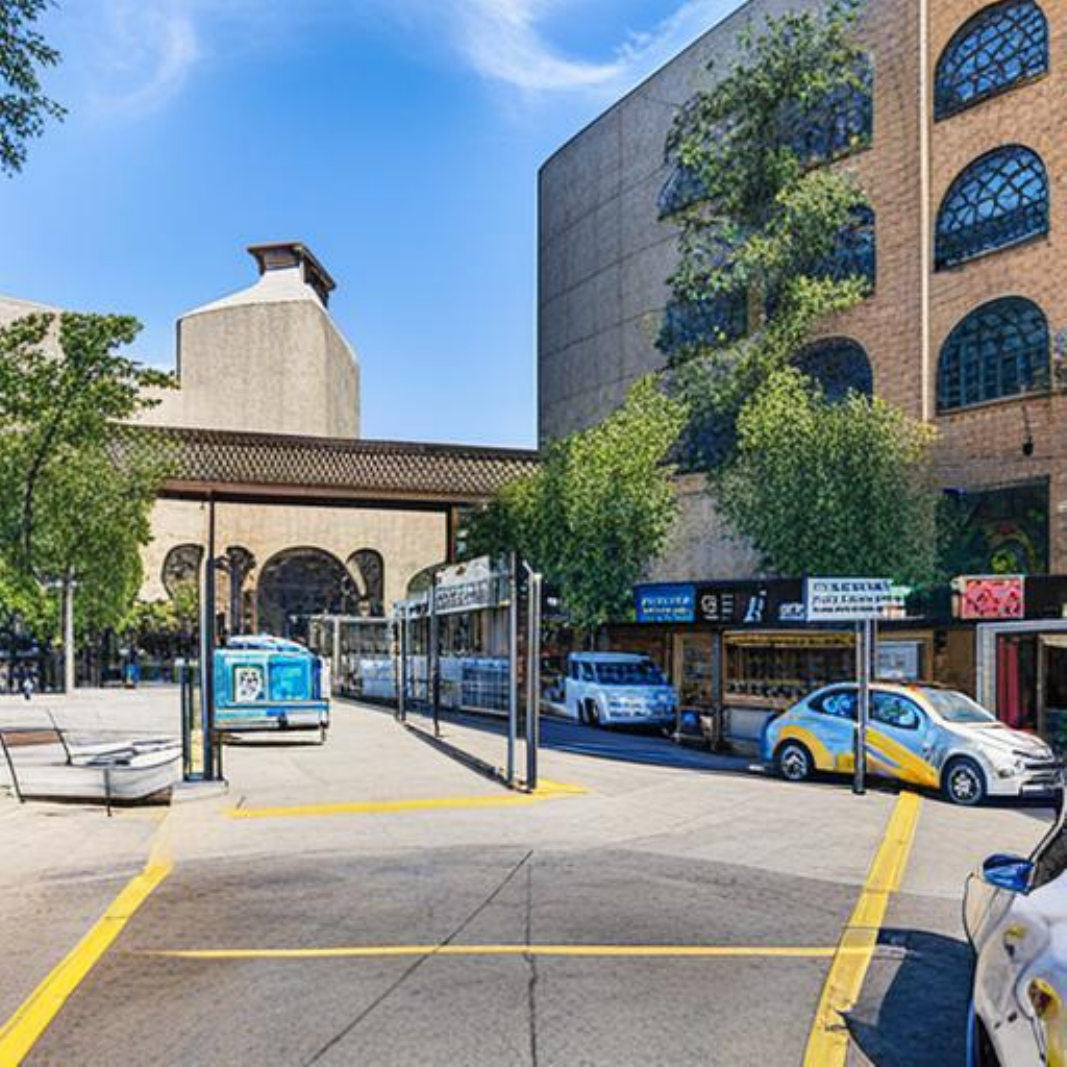} &
        \includegraphics[width=0.3\textwidth]{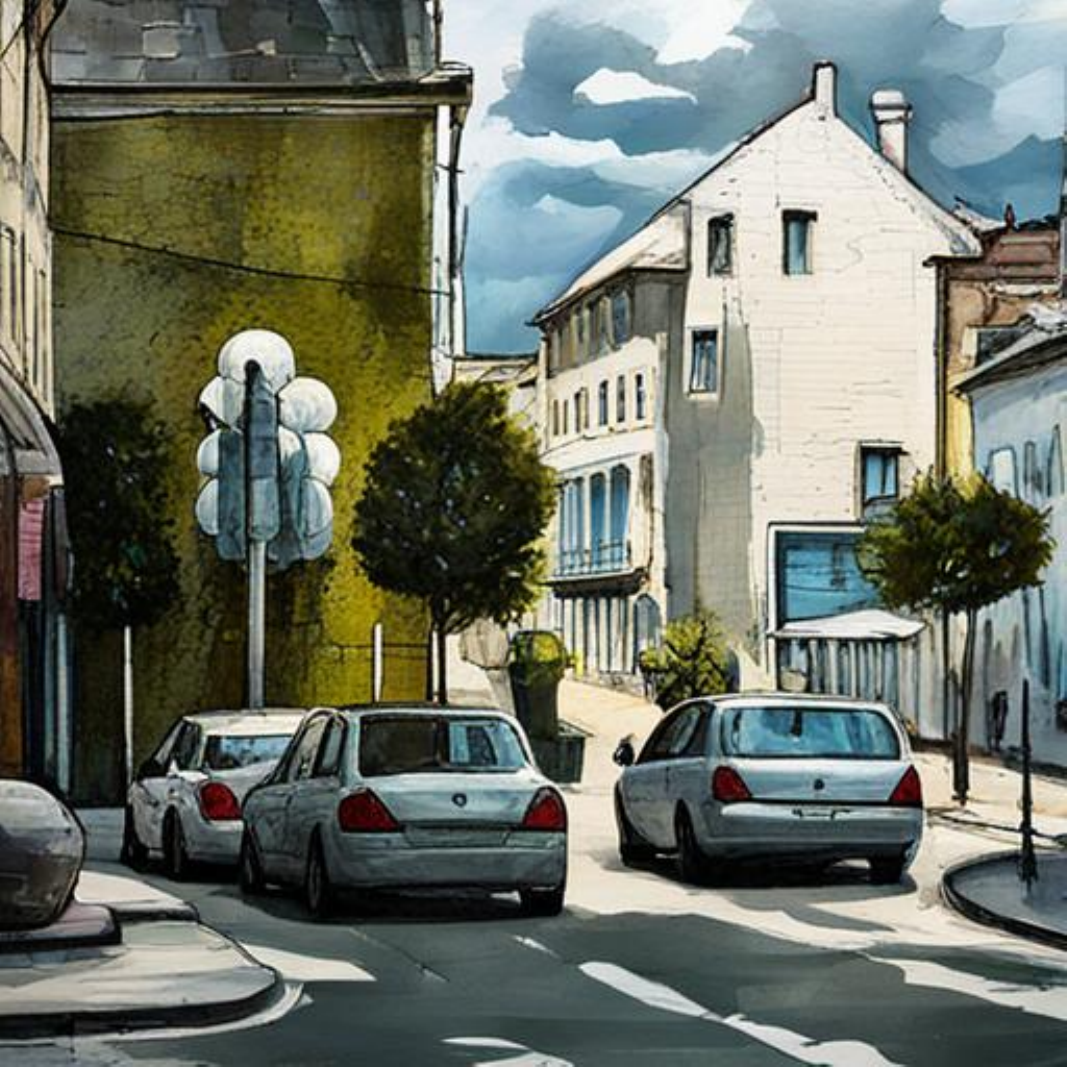} \\
        
        \includegraphics[width=0.3\textwidth]{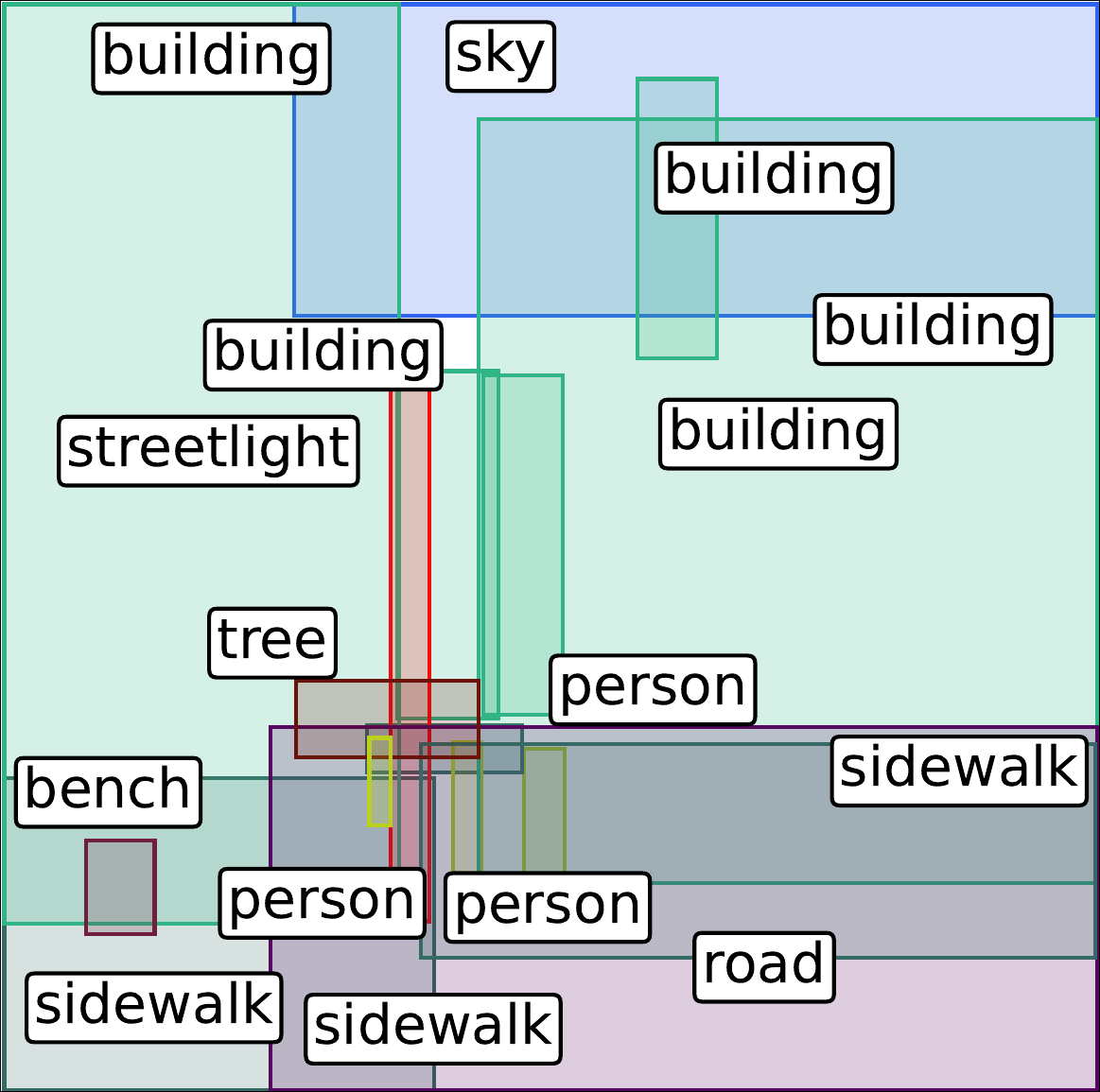} &
        \includegraphics[width=0.3\textwidth]{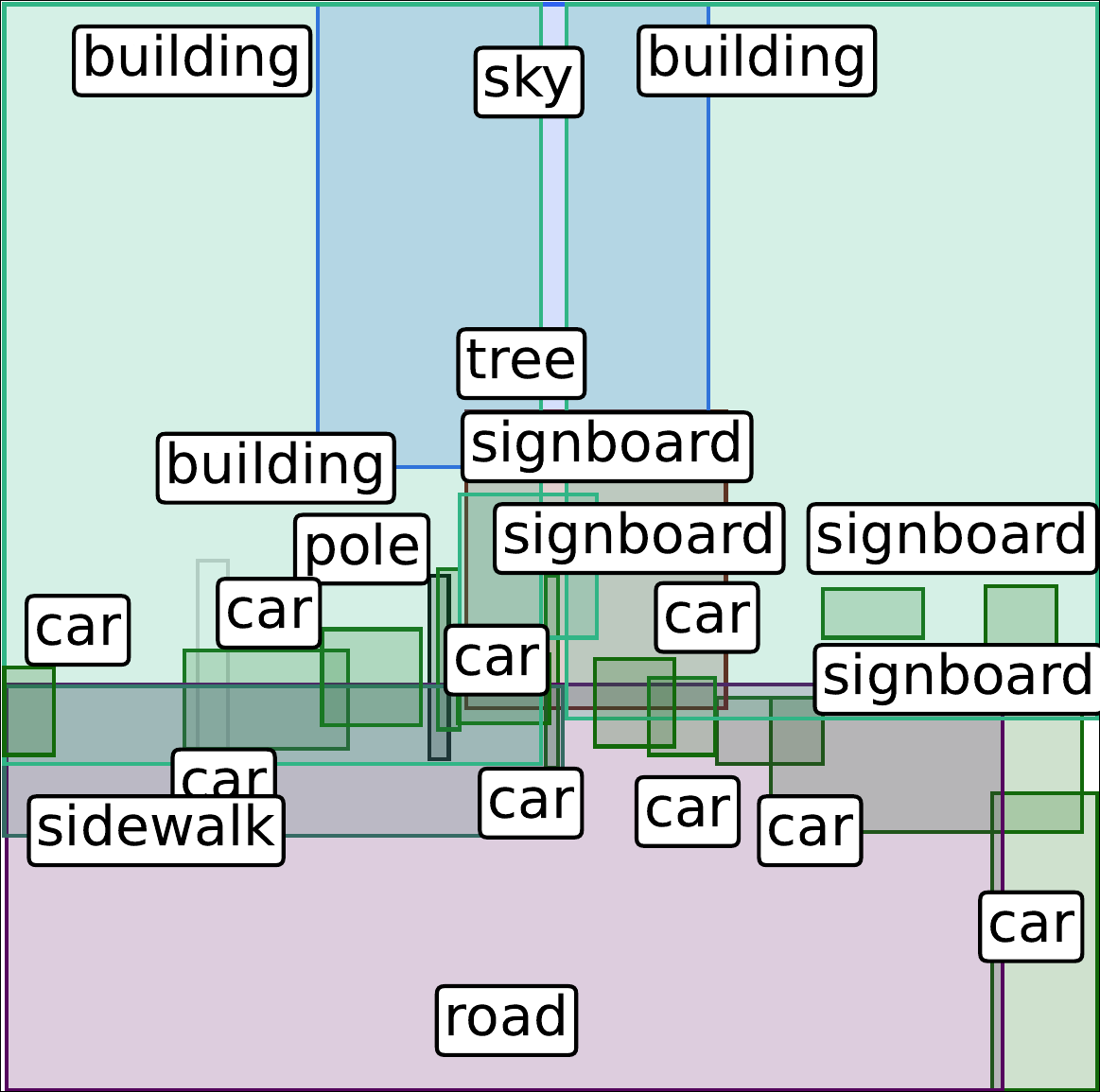} &
        \includegraphics[width=0.3\textwidth]{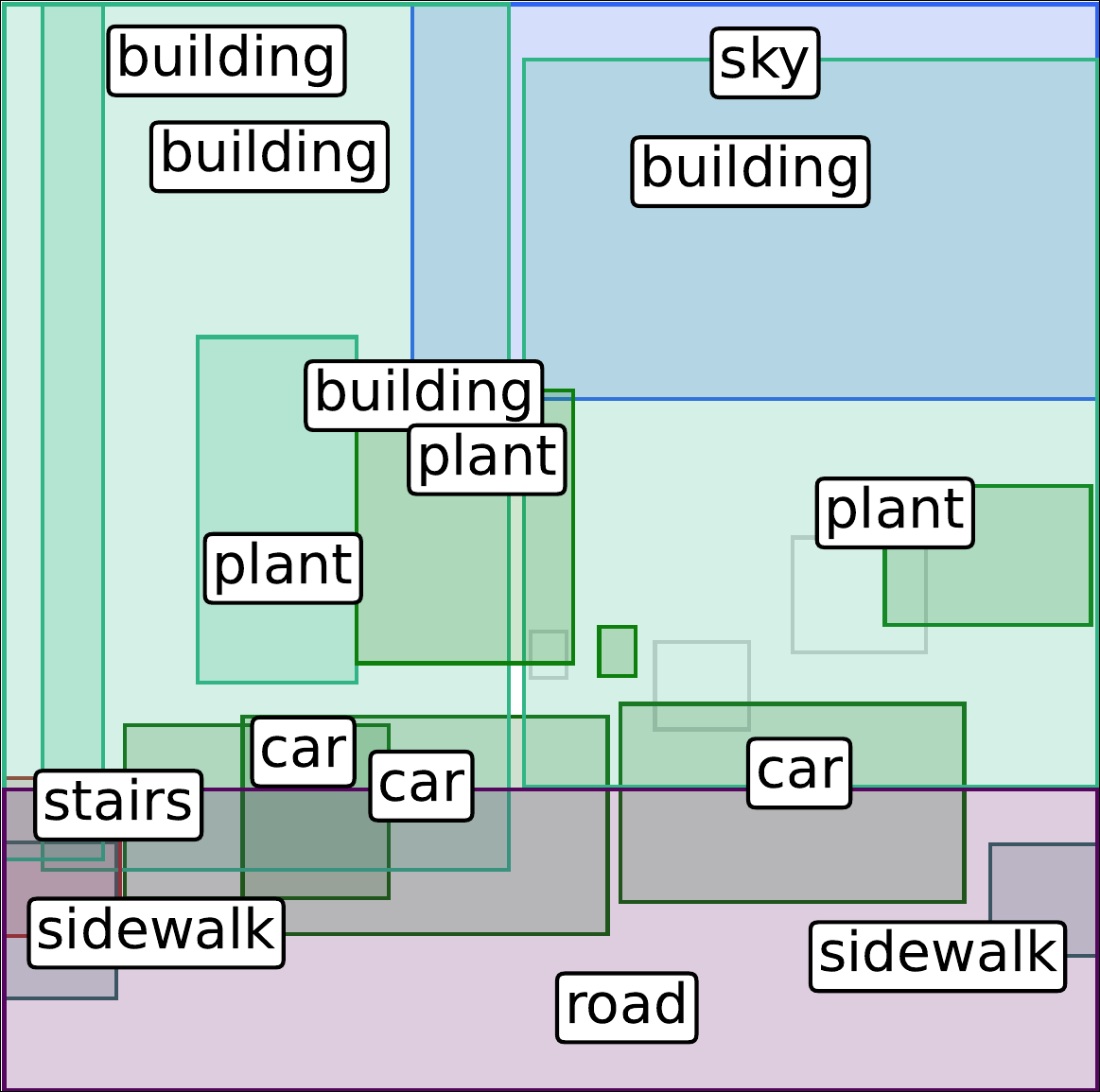} \\
    \end{tabular}
    \caption{Ours - Street.}
    \label{fig:ours_street}
\end{figure*}

\clearpage
\clearpage
\clearpage
\clearpage
\clearpage
\clearpage
\clearpage
\clearpage

\end{document}